\documentclass[10pt]{article} %
\usepackage[preprint]{tmlr_stylefiles/tmlr}

\renewcommand{\headrulewidth}{0pt}

\usepackage{amsmath,amsfonts,bm}

\def\eqref#1{equation~\ref{#1}}

\def\1{\bm{1}}

\DeclareMathAlphabet{\mathsfit}{\encodingdefault}{\sfdefault}{m}{sl}
\SetMathAlphabet{\mathsfit}{bold}{\encodingdefault}{\sfdefault}{bx}{n}

\newcommand{\E}{\mathbb{E}}

\usepackage{graphicx}
\usepackage{xcolor}
\usepackage{xspace}
\usepackage{booktabs}
\usepackage{makecell}
\usepackage{ragged2e}
\usepackage{multirow}
\usepackage{caption}

\usepackage{algorithm}
\usepackage{algpseudocode}

\usepackage{amsmath}
\usepackage{amssymb}
\usepackage{mathtools}
\usepackage{bm}
\usepackage{textgreek}
\allowdisplaybreaks %
\usepackage{twoopt}

\usepackage{natbib}
\usepackage[pagebackref=true]{hyperref}
\usepackage{url}
\renewcommand*\backref[1]{\ifx#1\relax \else (Cited on #1) \fi}
\hypersetup{
    plainpages=false,
    colorlinks,
    linkcolor={red!50!black},
    citecolor={blue!50!black},
    urlcolor={blue!80!black}
}

\title{CFMI: Flow Matching for Missing Data Imputation}

\author{\name Vaidotas Simkus \email vaidotas.simkus@ed.ac.uk\\
        \name Michael U.\ Gutmann \email michael.gutmann@ed.ac.uk \\
        \addr School of Informatics\\
        University of Edinburgh}

\definecolor{revisioncolor}{RGB}{175, 2, 2}

\def\month{}  %
\def\year{} %
\def\openreview{} %

\usepackage{cleveref}
\crefformat{equation}{eq.\ (#2#1#3)}
\crefrangeformat{equation}{eqs.\ (#3#1#4) to (#5#2#6)}
\crefmultiformat{equation}
  {eqs.\ (#2#1#3)}
  { \andor\  (#2#1#3)}
  {, (#2#1#3)}
  { \andor\ (#2#1#3)}
\crefrangemultiformat{equation}
  {eqs.\ (#3#1#4) to (#5#2#6)}
  { and (#3#1#4) to (#5#2#6)}
  {, (#3#1#4) to (#5#2#6)}
  { and (#3#1#4) to (#5#2#6)}
\newcommand{\andor}{and}

\newcommand{\obs}{\mathrm{o}}
\newcommand{\mis}{\mathrm{m}}
\newcommand{\x}{{\bm{x}}}
\newcommand{\xm}{\x_\mis}
\newcommand{\xo}{\x_\obs}

\newcommand{\m}{\bm{m}}
\newcommand{\s}{\bm{s}}

\newcommand{\xtarg}{\x_{\text{t}}}
\newcommand{\xcond}{\x_{\text{c}}}

\newcommand{\starg}{\s_{\text{t}}}
\newcommand{\scond}{\s_{\text{c}}}

\usepackage{twoopt}
\newcommand{\thetab}{{\bm{\theta}}}
\newcommand{\pt}{{p_\thetab}}
\newcommand{\ptt}[1][t]{{p_\thetab^{#1}}}
\newcommandtwoopt{\vtt}[2][\cdot][t]{{v_\thetab(#1; #2)}}
\newcommandtwoopt{\phitt}[2][\cdot][t]{{\phi_\thetab(#1; #2)}}
\newcommandtwoopt{\invphitt}[2][\cdot][t]{{\phi^{\text{-}1}_\thetab(#1; #2)}}
\newcommand{\ps}{{p^*}}

\newcommand{\qtt}[1][t]{{q^{#1}}}
\newcommandtwoopt{\utt}[2][\cdot][t]{{u(#1; #2)}}

\newcommand{\zp}[1]{\widetilde{#1}}

\usepackage{suffix}

\newcommand{\Expect}[2]{\E_{#1} \left[ #2 \right]}
\WithSuffix\newcommand\Expect*[2]{\E_{#1} #2 }

\DeclarePairedDelimiterX{\infdivx}[2]{(}{)}{%
  #1\;\delimsize|\delimsize|\;#2%
}

\newcommand\norm[1]{\left\lVert#1\right\rVert}

\newcommand{\Lc}{\mathcal{L}}

\newcommand{\Xc}{\mathcal{X}}

\newcommand{\LcCFM}{\Lc_{\text{CFM}}}
\newcommand{\LcCFMI}{\Lc_{\text{CFMI}}}

\newcommand{\dif}{\mathop{}\!\mathrm{d}}

\usepackage{xspace}
\newcommand{\ie}{i.e.\@\xspace}
\newcommand{\eg}{e.g.\@\xspace}

\begin{document}

\maketitle

\begin{abstract}
We introduce conditional flow matching for imputation (CFMI), a new general-purpose method to impute missing data. The method combines continuous normalising flows, flow-matching, and shared conditional modelling to deal with intractabilities of traditional multiple imputation. Our comparison with nine classical and state-of-the-art imputation methods on 24 small to moderate-dimensional tabular data sets shows that CFMI matches or outperforms both traditional and modern techniques across a wide range of metrics. Applying the method to zero-shot imputation of time-series data, we find that it matches the accuracy of a related diffusion-based method while outperforming it in terms of computational efficiency. Overall, CFMI performs at least as well as traditional methods on lower-dimensional data while remaining scalable to high-dimensional settings, matching or exceeding the performance of other deep learning-based approaches, making it a go-to imputation method for a wide range of data types and dimensionalities.
\end{abstract}

\section{Introduction}

Missing data is a persistent challenge that hinders the full potential of machine learning in critical domains such as healthcare and the sciences.
Data in these domains are often incomplete due to issues like participant dropout in longitudinal studies or the malfunction of measurement devices.
Addressing this issue is essential to enable robust analyses and downstream applications.

A fundamental statistical procedure to address this issue is multiple imputation \citep{rubinMultipleImputationNonresponse1987,littleStatisticalAnalysisMissing2020}, which aims to sample (or impute) the missing data with draws from the corresponding conditional data distribution.
While modern, deep learning-based imputation methods \citep[\eg][]{yoonGAINMissingData2018,matteiMIWAEDeepGenerative2019,miaoGenerativeSemisupervisedLearning2021,tashiroCSDIConditionalScorebased2021} excel in high-dimensional settings, they frequently underperform compared to traditional imputation techniques \citep[\eg][]{vanbuurenMultivariateImputationChained2000,stekhovenMissForestNonparametricMissing2012} when applied to small- or moderate-dimensional data.
This highlights the need to look for a more generalisable approach that can deliver robust performance across diverse data types and dimensionalities.

\begin{figure}[tb]
  \centering
  \includegraphics[width=0.8\linewidth]{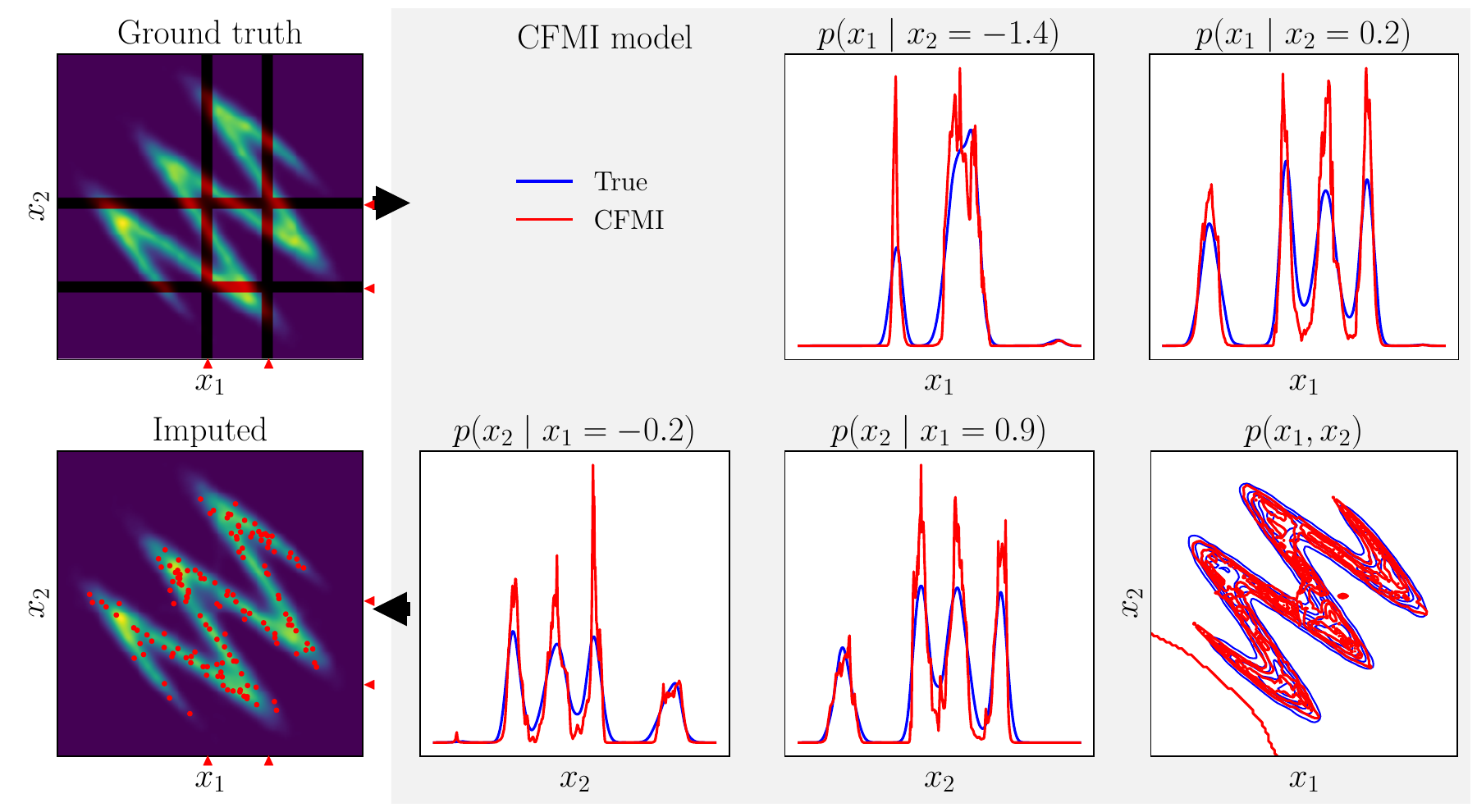}
  \caption{\emph{Imputation using CFMI on synthetic 2D data.} 
  The left-most column shows the kernel density estimate (KDE) of the ground truth distribution (top) and the KDE of the imputed data distribution (bottom). 
  The red-tinted rows and columns in top-left illustrate that some data-points may be missing one or both (not visualized) of the dimensions, and thus imputation requires sampling the corresponding (conditional) data distribution.  
  CFMI learns a model of all conditional distributions that are needed for imputation (illustrated in the grey box on the right, corresponding to the red-tinted rows and columns). 
  This model is then used to impute the data by sampling the appropriate conditional distribution for each incomplete data point. 
  A sample of imputations is shown as red dots in the bottom-left, and the KDE of the imputed data closely matches that of the ground truth distribution. 
  Additional synthetic-data examples are in \cref{apx:fig:toy_cfmi_visualizations} of the appendix.}
  \label{fig:toy_ground_truth_vs_cfmi_visualization}
\end{figure}

Recently, advances in diffusion \citep{sohl-dicksteinDeepUnsupervisedLearning2015,hoDenoisingDiffusionProbabilistic2020} and continuous normalising flow \citep[CNF,][]{chenNeuralOrdinaryDifferential2019,lipmanFlowMatchingGenerative2023} literature has demonstrated remarkable flexibility, robustness, and generalisability across various domains. 
Building on these developments, we propose an imputation method based on CNFs trained via flow-matching \citep{lipmanFlowMatchingGenerative2023,tongImprovingGeneralizingFlowbased2023}. 
To facilitate efficient learning and imputation, the model is specified conditionally, enabling all required conditional distributions for imputation to be captured by a single shared model, drawing on insights from prior work  \citep{ivanovVariationalAutoencoderArbitrary2019,liFlowModelsArbitrary2020,tashiroCSDIConditionalScorebased2021,straussArbitraryConditionalDistributions2021}. 
Our approach, CFMI (\cref{fig:toy_ground_truth_vs_cfmi_visualization}), combines the strengths of CNFs, flow-matching, and shared conditional modelling to achieve imputation performance on par with traditional methods for lower-dimensional data while maintaining scalability to high-dimensional settings, matching and outperforming the capabilities of other deep learning-based methods. 

We summarise our main contributions as follows:
\begin{itemize}
    \item In \cref{sec:cfmi}, we propose conditional flow matching for imputation (CFMI), a conditional CNF model trained via flow matching for missing data imputation.
    \item In \cref{sec:eval-synthetic-2d-data}, we evaluate the ability of the CNF model to capture all necessary conditional distributions for missing data imputation on a synthetic data set, where ground truth conditionals can be computed.
    \item In \cref{sec:eval-uci}, we thoroughly evaluate the method's imputation effectiveness on 24 small-to-moderate dimensional tabular UCI data sets, matching or outperforming the performance of traditional and modern techniques across a wide range of metrics.
    \item In \cref{sec:eval-time-series}, we also evaluate the method's performance for zero-shot imputation of held-out time-series data, matching the accuracy of a related state-of-the-art diffusion-based method while outperforming it in terms of computational efficiency. 
\end{itemize}
These contributions represent a step toward a general-purpose imputation framework that provides robust, accurate, and scalable performance across a wide variety of data types and dimensionalities.

\section{Background}

\subsection{Missing data imputation}
\label{sec:background-missing-data-imputation}

Let $\xo^{(i)}$ and $\xm^{(i)}$ be the observed and missing parts of the $i$-th data-point $\x^{(i)} \in \Xc^D$ drawn from an unknown distribution $\ps(\x)$. 
The missingness pattern for $\x^{(i)}$ is indicated by a binary mask $\m^{(i)} \in \{0,1\}^D$, which follows a generally unknown missingness mechanism $\ps(\m \mid \x^{(i)})$. Formally, the observed and missing data are thus defined as
\begin{align}
    \xo^{(i)} \overset{\text{def}}{=} \x^{(i)}[\m^{(i)}] \quad \text{and} \quad \xm^{(i)} \overset{\text{def}}{=} \x^{(i)}[1 - \m^{(i)}], \label{eq:def-obs-miss-variable}
\end{align}
where there are up to $2^D$ possible missingness patterns, and hence combinations of $\xo$ and $\xm$. To simplify the notation, we omit the sample index $i$ in the remainder of this discussion.

Using the common assumption that the data are missing-at-random \citep[MAR, \eg][Section~1.3]{littleStatisticalAnalysisMissing2020}, where the probability of missingness depends only on the observed values and not on the missing values, %
we would like to sample $\ps(\xm \mid \xo^{(i)})$, such that the imputations follow the true distribution of the missing variables \citep[\eg][Section~2.2.6]{vanbuurenFlexibleImputationMissing2018}.
However, as the ground truth distribution $\ps(\x)$ is generally unknown approximations are needed.

Given an incomplete data set $\{\xo^{(i)}, \m^{(i)}\}_{i=1}^N$, one way to obtain imputations is to approximate the unknown joint distribution $\ps(\x)$ with a probabilistic model $\pt(\x)$ \citep{dempsterMaximumLikelihoodIncomplete1977}, and then perform probabilistic inference to sample from $\pt(\xm \mid \xo)$. 
However, estimating modern jointly-specified models $\pt(\x)$ directly from incomplete data is generally challenging \citep[][]{simkusVariationalGibbsInference2023}. Moreover, performing probabilistic inference to impute the incomplete data-points can be computationally demanding.

\subsection{Missing data imputation with conditional models}
\label{sec:background-mi-via-conditional-models}

A practical alternative to learning a \emph{joint} model $\pt(\x)$ is to model the \emph{conditional} distributions $\pt(\xm \mid \xo)$ directly. 
However, since the number of missingness patterns grows exponentially with the dimensionality $D$ (\ie, up to $2^D$ patterns), learning a separate model for each pattern is computationally infeasible.

To address similar challenges, several works have proposed modelling all conditional distributions using a single model \citep{douglasUniversalMarginalizerAmortized2017,ivanovVariationalAutoencoderArbitrary2019,liFlowModelsArbitrary2020,tashiroCSDIConditionalScorebased2021,straussArbitraryConditionalDistributions2021}. 
Such models can be interpreted as learning a set of conditional distributions, $\{\pt(\xm \mid \xo) \mid \xo = \x[\m], \xm = \x[1-\m], \forall \m \in \{0, 1\}^D\}$, with shared parameters $\thetab$ across all missingness patterns.
This conditional modelling approach enables efficient imputation by directly sampling from the appropriate conditional distribution for any given missingness pattern.

When fully-observed training data are available, arbitrarily-conditional models are typically trained by randomly splitting each data-point $\x$ into target $\xtarg$ and conditioning $\xcond$ parts according to some heuristic splitting strategy \citep[see \eg][Section~4.3]{tashiroCSDIConditionalScorebased2021}.
The training objective is then to approximate the true conditional distribution, such that $\pt(\xtarg \mid \xcond) \approx \ps(\xtarg \mid \xcond)$ for any $\xtarg$ and $\xcond$. 
When the training data are incomplete, the target and conditioning parts are similarly obtained by partitioning the observed parts $\xo$ of the incomplete data-points. 

An important caveat to note is that when the models are trained on incomplete data, these imputation methods rely on the (often implicit) assumption that a model trained to approximate $\pt(\xtarg \mid \xcond)$, where $(\xtarg, \xcond) = \xo$, can generalise to the conditional distribution of missing variables $\pt(\xm \mid \xo)$, where $(\xo, \xm) = \x$. 
Nevertheless, the conditional modelling approach remains appealing due to its simplicity, scalability, and empirical success.%

\subsection{Continuous normalising flows via flow matching}

A continuous normalising flow \citep[CNF,][]{chenNeuralOrdinaryDifferential2019} is a time-dependent probabilistic model $\ptt(\x)$, also called a probability path, where $t \in [0, 1]$ is a continuous time index. At $t=0$, $\ptt[0](\x)$ is a simple base distribution, while at $t=1$, $\ptt[1](\x)$ approximates the target distribution. 
The probability path is generated by a time-dependent vector field $\vtt: \mathbb{R}^D \rightarrow \mathbb{R}^D$, which yields an invertible flow $\phitt: \mathbb{R}^D \rightarrow \mathbb{R}^D$:
\begin{align}
    \phitt[\x^0] = \x^0 + \int_{0}^{t} \vtt[\x^{\tau}][\tau] \dif \tau \overset{\text{def}}{=} \x^t, \label{eq:flow}
\end{align}
where $\phitt[\x^0]$ maps a sample $\x^0$ from the base distribution $\ptt[0](\x)$ to a sample $\x^t \sim \ptt(\x)$ along the probability path. 
The target distribution $\ptt[1](\x)$ is determined using the change-of-variables formula \citep[\eg][Section~2.8.3]{murphyProbabilisticMachineLearning2021}:
\begin{align}
    \ptt[1](\x) = \ptt[0](\invphitt[\x][1]) \det \left[ \frac{\partial \invphitt[\x][1]}{\partial \x} \right], \label{eq:cnf-density}
\end{align}
where $\invphitt[\cdot][1]$ is the inverse of the flow at $t=1$. Calculating this density involves solving for the inverse of the integral in \cref{eq:flow} and computing the determinant of the Jacobian, both of which can be computationally demanding. 
As a result, maximum-likelihood estimation (MLE) for CNFs may be resource-intensive and potentially unstable. 

Recently, inspired by the connection between CNFs and diffusion models \citep{sohl-dicksteinDeepUnsupervisedLearning2015,songDenoisingDiffusionImplicit2022}, \citet{lipmanFlowMatchingGenerative2023} proposed an alternative method for fitting CNFs that avoids computationally expensive simulation.
This method, called flow matching, constructs a fixed target probability path $\qtt(\x)$ along with a corresponding (conditional) target vector field $\utt[\x \mid \x^1]$. 
The CNF is then fitted by minimising the following conditional flow matching objective:
\begin{align}
    \LcCFM(\thetab) = \Expect*{q(t) q(\x) \qtt(\x^t \mid \x)}{\norm{\vtt[\x^t] - \utt[\x^t \mid \x]}_2^2},
\end{align}
where $q(t)$ is a distribution over the time indices, $q(\x)$ is the empirical data distribution formed by samples from the ground truth $\ps(\x)$, and $\qtt(\x^t \mid \x)$ is the conditional probability path such that $\qtt(\x^t) = \Expect{q(\x)}{\qtt(\x^t \mid \x)}$.
The flow matching objective, unlike MLE, does not need computing the density in \cref{eq:cnf-density}, making the training process significantly more scalable and stable.

Yet, using the joint model $\ptt[1](\x)$ for missing data imputation is challenging since fitting a joint CNF model from incomplete data is difficult \citep[\eg][]{simkusVariationalGibbsInference2023} and conditional sampling of the density in \cref{eq:cnf-density} is computationally demanding. 
Instead, as discussed in \cref{sec:background-mi-via-conditional-models}, we will use a conditional modelling approach.

\section{Conditional normalising flows for missing data imputation via flow matching}
\label{sec:cfmi}

We propose conditional flow matching for imputation (CFMI), a conditional continuous normalising flow (CNF) model trained via flow matching for missing data imputation. 
Our approach leverages flow matching to enable both fast training and accurate model fitting. 
Building on the motivation discussed in \cref{sec:background-mi-via-conditional-models}, CFMI learns a shared model that captures all conditional distributions needed for imputation, making it particularly effective for high-dimensional data sets. 
Through extensive evaluation on both synthetic, real-world tabular, and time-series data (see \cref{sec:evaluation}), we demonstrate that CFMI achieves substantial improvements in both computational efficiency and imputation accuracy compared to existing methods.

\subsection{Conditional CNF model for missing data imputation}
\label{sec:conditional-cfn-miss}

Continuous normalising flows, like diffusion models, are well-known for their highly-flexible modelling capabilities despite being conceptually relatively simple. 
In general, a CNF operates by taking as input a $D$-dimensional sample $\x_t$ from the learnt probability path $\ptt(\x)$ (or the target path $\qtt(\x)$ during training), along with (an encoding of) the time index $t$, and predicts a $D$-dimensional vector field $\vtt[\x_t]$.
The vector-field can be parametrised using virtually any function approximator, such as neural networks \citep{lipmanFlowMatchingGenerative2023} or random forests \citep{jolicoeur-martineauGeneratingImputingTabular2024}.
This allows CNFs to adapt to a wide range of tasks while maintaining simplicity and flexibility in their architectural choices.
When using a neural network for parametrisation, the CNF framework further enables the modelling of multiple conditional distributions within a single model.

For conditional modelling, the CNF model takes additional inputs, namely the observed conditioning variables $\xo$, alongside the sample from the probability path and the time index. 
To accommodate arbitrary configurations of missing (target) and observed (conditioning) variables, the model maintains a fixed output dimensionality of $D$. 
Similarly, the dimensionalities of the probability-path sample and the conditioning variables are also padded to $D$ dimensions before being passed as inputs to the model.
Specifically, let $\xo$ and $\xm$ represent the observed and missing variables, respectively, and $\m$ denote the missingness pattern.
Then, $\zp{\xo}$ and $\zp{\xm}$ are zero-padded versions of $\xo$ and $\xm$ such that the padded vectors have $D$ dimensions and the indices are respected, that is, $\zp{\xo}$ is zero where $\m$ is zero, and $\zp{\xm}$ is zero where $1-\m$ is zero.
The conditional density $\pt(\xm \mid \xo)$ of the CNF can then be expressed via the vector field $\vtt[\zp{\xm^\tau}][\zp{\xo}, \tau][1-\m]$:
\begin{align}
    \pt(\xm \mid \xo) &= \ptt[0](\invphitt[\xm][\xo, 1]) \det \left[ \frac{\partial \invphitt[\xm][\xo,1]}{\partial \xm} \right] \label{eq:cnf-incomplete-density}
    \\
    \intertext{with}
    \phitt[\xm^0][\xo, t] &= \xm^0 + \int_{0}^{t} \vtt[\zp{\xm^\tau}][\zp{\xo}, \tau][1-\m] \dif \tau 
    \label{eq:incomplete-flow}
\end{align}
where $\ptt[0](\cdot)$ is an unconditional base distribution, $[1-\m]$ denotes indexing the outputs of the model using the mask as in \cref{eq:def-obs-miss-variable} and $\zp{\xm^\tau}$ is a zero-padded sample from the (conditional) probability path $\ptt[\tau](\xm^\tau \mid \xo)$.
This specification ensures that the same model can seamlessly handle any missingness pattern $\m$.

While the vector field can be parametrised using any function approximator, the choice of a network architecture can significantly impact the performance, and the optimal choice often depends on the specific data set. 
For the evaluations in this paper, we used both residual fully-connected network and transformer architectures, grounding the choice on the target task. 
Additionally, while incorporating the binary missingness mask $\m$ as an input is theoretically unnecessary in the MAR case (\cref{sec:background-missing-data-imputation}), we include it as an auxiliary input to the model $\vtt[\zp{\xm^t}][\zp{\xo}, \m, t][1-\m]$. 
This generally improves the model's ability to effectively learn the required conditional distributions by providing additional information to the neural network that specifies which conditional distribution is being estimated. 

\subsection{Training CNF for imputation from incomplete data via flow matching}

To perform missing data imputation, we first train a conditional CNF model, defined by the conditional vector field $\vtt[\zp{\xm^t}][\zp{\xo}, \m, t][1-\m]$ as introduced in \cref{sec:conditional-cfn-miss}.
However, since the true missing data $\xm$ are typically unavailable, we follow prior works \citep{ivanovVariationalAutoencoderArbitrary2019,liFlowModelsArbitrary2020,tashiroCSDIConditionalScorebased2021} and train the model by partitioning the observed part $\xo$ of each potentially incomplete data point $\x$.%

\paragraph{Partitioning observed data.} Given an incomplete data point $\xo$ and its missingness mask $\m$, we randomly partition the available variables into two subsets: conditioning variables $\xcond$ and target variables $\xtarg$. 
This partitioning is determined by a stochastic splitting strategy $q(\starg, \scond \mid \m)$, where $\starg$ and $\scond$ are binary masks indicating the target and conditioning variables, respectively.
These splitting masks are constrained such that elements corresponding to missing variables ($m_j=0$) remain zero in both $s_{\text{targ}, j}$ and $s_{\text{cond}, j}$ for $\forall j \in [1, \ldots, D]$, and $1 - \scond[\m] = \starg[\m]$.
Thus, missing dimensions cannot be selected as target or conditioning variables.
Using these masks, the target and conditioning variables are defined as follows:
\begin{align}
     \xcond \overset{\text{def}}{=} \xo[\scond[\m]] \quad \text{and} \quad \xtarg \overset{\text{def}}{=} \xo[\starg[\m]]. \label{eq:def-targ-cond-variable}
\end{align}
We formalise the joint distribution of $\xtarg$, $\xcond$, $\starg$, and $\scond$ as:
\begin{align}
    q(\xtarg, \xcond, \starg, \scond) &= \iint q(\xo, \m) q(\starg, \scond \mid \m) 
    \delta(\xtarg - \xo[\starg[\m]]) 
    \delta(\xcond - \xo[\scond[\m]]) 
    \dif \xo \dif \m,
\end{align}
where $q(\xo, \m)$ is the empirical distribution of incomplete data-points and missingness masks, and $\delta(\cdot)$ is the Dirac delta, ensuring that the split sample $(\xtarg, \xcond)$ is derived from $\xo$ using the masks $\starg$ and $\scond$. 

The choice of the splitting strategy $q(\starg, \scond \mid \m)$ introduces some flexibility. 
An effective strategy should closely align with the true missingness mechanism to help the model prioritise learning the most important conditional distributions. 
In this work, we use two simple heuristic strategies based on the data type:
\begin{itemize}
    \item \textbf{Random strategy:} For synthetic and tabular data sets in \cref{sec:eval-synthetic-2d-data,sec:eval-uci}, we split the observed dimensions randomly into conditioning and target variables. This ensures a diverse set of patterns for training the model without introducing systematic biases.
    \item \textbf{Random-Historical strategy:} For time-series data in \cref{sec:eval-time-series}, we adopt the heuristic approach from \citet{tashiroCSDIConditionalScorebased2021}. 
    This strategy uses a mixture of the random strategy above and a heuristic strategy that uses a missingness pattern from another training data point and determines the target and conditioning variables by intersecting the two patterns. 
    The heuristic component of this strategy aims to mirror real-world missingness structures, such as the consecutive sequences of missing values often observed in time-series data, whereas the random component of this strategy attemps to ensure that all conditional distributions are learnt.  
    While we apply this strategy primarily to time-series data, it is versatile and can also be applied to other data types.
\end{itemize}
For more examples of splitting strategies see \citet[][Section~4.3]{tashiroCSDIConditionalScorebased2021}.

\paragraph{Training objective.} The conditional flow matching training objective is then defined as:
\begin{align}
    &\LcCFMI(\thetab) = \E_{q(t) q(\xtarg, \xcond, \starg, \scond) \qtt(\xtarg^t \mid \xtarg, \xcond)} \left[ \norm{\starg}_0^{-1} 
    \norm{
    \vtt[\zp{\xtarg^t}][\zp{\xcond}, \scond, t][\starg]
    - \utt[\xtarg^t \mid \xtarg][\xcond, t]
    }_2^2
    \right].
    \label{eq:cfmi-objective}
\end{align}
Here, $q(t)$ is the distribution over the time index $t$, and $\qtt[t](\xtarg^t \mid \xtarg, \xcond)$ represents the fixed conditional probability path for the target variables $\xtarg$ conditioned on $\xcond$, and $\zp{\xtarg^t}$ and $\zp{\xcond}$ are the zero-padded versions of $\xtarg^t$ and $\xcond$ such that $\zp{\xtarg^t}$ is zero where $\starg$ is zero and $\zp{\xcond}$ is zero where $\scond$ is zero.
The additional scaling factor $\norm{\starg}_0^{-1}$, normalises the objective by the number of target variables, ensuring comparable magnitudes for different target variable sets. 
This normalisation balances the learning of the shared model for all conditional distributions, regardless of the number of target variables. 
Importantly, the scaling term does not change the optimum of the flow matching objective.
As a result, similar to the original flow-matching objective, this loss function minimises the L2 norm between the predicted vector field  $\vtt[\cdot][\zp{\xcond}, \scond, t]$ and the true conditional vector field $\utt[\cdot][\xcond, t]$, ensuring that the learned CNF can faithfully model the conditional distributions needed for imputation.

For the probability path $\qtt(\xtarg^t \mid \xtarg, \xcond)$ we adopt the independent coupling scheme from \citet{tongImprovingGeneralizingFlowbased2023}, where $\qtt(\xtarg^t \mid \xtarg, \xcond) = \mathcal{N}(\xtarg^t \mid t \xtarg + (1-t) \xtarg^0, \sigma^2)$ and $\utt[\xtarg^t \mid \xtarg][\xcond, t] = \xtarg - \xtarg^0$, where $\xtarg$ is a sample from the empirical distribution and $\xtarg^0$ is a sample from the base distribution which is assumed to be standard normal. 
We leave the evaluation of the effect of the other probability paths from the flow-matching literature \citep{lipmanFlowMatchingGenerative2023,tongImprovingGeneralizingFlowbased2023,gatDiscreteFlowMatching2024} to missing data imputation as future work. 

\subsection{Imputation of missing data using a trained model}

Once the conditional CNF model has been trained to approximate the conditional distributions using \cref{eq:cfmi-objective}, it can be used to impute missing data. 
The data to be imputed may be the incomplete training data itself, or a separate test data set.
The imputation process involves solving the flow equation in \cref{eq:incomplete-flow} using the trained vector-field model $\vtt[\zp{\xm^t}][\zp{\xo}, \m, t][1-\m]$. 
While a variety of numerical ODE solvers can be used for this purpose, we observe that a simple Euler integrator performs effectively across all experiments in our evaluation.

It is important to highlight a potential generalisation gap between the training and imputation phases. 
During training, the model learns to predict the conditional distribution of the target variables $\xtarg$, given the conditioning variables $\xcond$. These subsets are derived by splitting \emph{incomplete} training data points $\xo$ into target and conditioning variables based on a splitting strategy $q(\starg, \scond \mid \m)$.

In contrast, during imputation, the task typically involves predicting the missing values $\xm$ from the observed incomplete data points $\xo$. 
This difference between the two stages means that the average dimensionality of the target and conditioning variables is generally higher during imputation phase than during training, which can potentially affect the performance of the model.
Despite this potential gap, our evaluations demonstrate that the trained model generalises well and achieves accurate missing data imputation. 
This robustness highlights the model’s capacity to adapt from the training setup to the imputation task.

\section{Evaluation}
\label{sec:evaluation}

In this section, we evaluate the proposed CFMI method for missing data imputation. 
The evaluation begins with synthetic 2D data sets, where we examine the method's capacity to approximate complex conditional distributions and demonstrate its flexibility. 
Next, CFMI is extensively evaluated on 24 tabular data sets using a wide range of metrics, providing insights into its effectiveness in imputing incomplete training data. 
Finally, the method is applied to time-series data in a zero-shot imputation setting, highlighting its generalisation capabilities across diverse domains.

Throughout these experiments, CFMI is consistently compared to the score-based diffusion method CSDI \citep{tashiroCSDIConditionalScorebased2021} to assess its relative performance, since flow matching and diffusion models are related \citep{albergoStochasticInterpolantsUnifying2023}. 
Details of our implementation of CSDI are provided in \cref{apx:csdi-implementation}.
The code to reproduce our method and the following results is available at \url{https://github.com/vsimkus/cfmi}.

\subsection{Learning conditionals from synthetic 2D data}
\label{sec:eval-synthetic-2d-data}

As discussed in \cref{sec:background-mi-via-conditional-models}, accurately imputing missing data requires estimating up to $2^D-1$ conditional distributions $\pt(\xm \mid \xo)$, one for each pattern of missingness.
In this section, we evaluate the performance of CFMI for missing data imputation on synthetic 2D data sets with known distributions from \citet{wenliangLearningDeepKernels2019}. This setting allows reference (conditional) distributions to be estimated accurately via numerical integration, providing a robust baseline for comparison.
For the evaluation, samples from the true distribution were made 50\% missing uniformly at random before training the CFMI model. Further details about the experimental setup are provided in \cref{apx:exp-details-synth-2d-data}.

For the cosine data set, \cref{fig:toy_ground_truth_vs_cfmi_visualization} demonstrates CFMI's ability to accurately capture all three (conditional) distributions: $p(x_1 \mid x_2)$, $p(x_2 \mid x_1)$, and $p(x_1,x_2)$. This is achieved even when the model is trained directly on incomplete data.
Further results on four additional synthetic datasets are provided in \cref{apx:fig:toy_cfmi_visualizations}, where CFMI consistently captures the necessary conditional distributions for imputation. 
These results highlight the flexibility and effectiveness of the model for various data distributions.

Interestingly, the CFMI distributions exhibit a slight mode-seeking trait, reminiscent of variational inference with the reverse KL objective \citep[\eg][Section~6.2.6]{murphyProbabilisticMachineLearning2021}. 
This behaviour likely arises from the flow-matching objective, which does not directly correspond to maximum-likelihood estimation (MLE) \citep{albergoStochasticInterpolantsUnifying2023,daxFlowMatchingScalable2023} that typically has a mass-covering trait instead.
While flow matching does not generally correspond to MLE, \citet{daxFlowMatchingScalable2023} demonstrated that if we assume some strong regularity conditions on the vector field model, then flow matching can upper-bound MLE, which would have mass-covering properties.  
However, our findings contrast with theirs, potentially due to differences in the target probability paths used. 
Specifically, while we use the independent coupling method from \citet{tongImprovingGeneralizingFlowbased2023}, they used the optimal transport coupling from \citet{lipmanFlowMatchingGenerative2023}, which may explain the discrepancy.
Despite this mode-seeking tendency, kernel density estimates of the imputed datasets (\cref{fig:toy_ground_truth_vs_cfmi_visualization,apx:fig:toy_cfmi_visualizations}) reveal that the imputed distributions with CFMI closely match the ground truth distributions. This suggests that the mode-seeking behaviour is inconsequential, provided the degree of missingness remains moderate.

In \cref{apx:fig:toy_cfmi_vs_csdi_with_diff_budgets,apx:fig:toy_cfmi_vs_csdi_with_diff_budgets_diff}, we compare CFMI with the diffusion-based alternative, CSDI \citep{tashiroCSDIConditionalScorebased2021}. 
CFMI not only achieves comparable imputation accuracy but also trains faster. 
These findings highlight the potential advantages of flow-matching for scalable and efficient missing data imputation.
In the following sections, we extend our investigation to realistic tabular and time-series data sets, evaluating CFMI’s performance in practical scenarios.

\subsection{Imputing incomplete tabular training data sets}
\label{sec:eval-uci}
\begin{figure}[tb]%
\begin{minipage}{\textwidth}%
\centering%
\captionof{table}{Average rank of each method for all metrics (mean and standard error) under 25\% MCAR missingness. Lowest means for each metric are bold and underlined; values within one standard error of the lowest are bolded.}%
\resizebox{1.\linewidth}{!}{\begin{tabular}{@{}lccccccccccc@{}}\toprule
Method             & W2             & RMSE           & CRPS           & Energy MMD     & Gaussian MMD   & Laplacian MMD  & 1-AUROC        & Pred. CRPS     & Reg. PB        & Reg. 1-CR      & Reg. AW       \\
\midrule
Mean               & $9.67 (0.51)$  & $8.96 (0.58)$  & $10.75 (0.11)$  & $10.54 (0.12)$  & $9.46 (0.26)$  & $10.67 (0.12)$  & $9.53 (0.61)$  & $10.83 (0.08)$  & $7.43 (1.46)$  & $6.86 (1.81)$  & $\bm{3.00 (1.41)}$\\
OT imputer         & $5.00 (0.43)$  & $4.04 (0.46)$  & $7.33 (0.37)$  & $6.92 (0.30)$  & $6.79 (0.43)$  & $6.92 (0.30)$  & $4.47 (0.58)$  & $7.25 (0.34)$  & $4.57 (0.90)$  & $5.71 (1.15)$  & $3.86 (1.35)$\\
SoftImpute         & $5.17 (0.64)$  & $3.33 (0.53)$  & $8.00 (0.57)$  & $8.12 (0.39)$  & $5.88 (0.78)$  & $8.21 (0.38)$  & $8.63 (0.29)$  & $8.12 (0.55)$  & $6.29 (1.02)$  & $4.57 (1.46)$  & $8.43 (1.21)$\\
MICE               & $7.21 (0.63)$  & $8.00 (0.60)$  & $4.38 (0.70)$  & $4.46 (0.61)$  & $4.08 (0.48)$  & $4.46 (0.61)$  & $7.32 (0.64)$  & $4.04 (0.69)$  & $5.00 (1.76)$  & $5.57 (1.74)$  & $7.43 (1.72)$\\
MissForest         & $\underline{\bm{1.08 (0.26)}}$  & $\underline{\bm{0.75 (0.20)}}$  & $3.62 (0.40)$  & $3.42 (0.51)$  & $3.04 (0.49)$  & $3.42 (0.51)$  & $5.37 (0.78)$  & $\underline{\bm{3.25 (0.37)}}$  & $\bm{4.00 (1.02)}$  & $\underline{\bm{3.14 (1.03)}}$  & $8.14 (1.22)$\\
ForestDiffusion    & $7.04 (0.51)$  & $8.00 (0.41)$  & $4.58 (0.57)$  & $5.38 (0.48)$  & $7.38 (0.53)$  & $5.25 (0.47)$  & $2.74 (0.79)$  & $4.33 (0.58)$  & $4.71 (1.29)$  & $5.86 (1.32)$  & $\underline{\bm{2.86 (0.77)}}$\\
GAIN               & $9.04 (0.25)$  & $8.50 (0.32)$  & $8.58 (0.39)$  & $9.75 (0.23)$  & $10.33 (0.19)$  & $9.67 (0.21)$  & $6.68 (0.89)$  & $8.54 (0.40)$  & $8.86 (0.86)$  & $8.29 (0.92)$  & $4.86 (0.94)$\\
\bottomrule
\end{tabular}}%
\label{tab:avg-rank-mcar25}%
\vspace{0.5cm} %
\centering%
\includegraphics[width=0.45\linewidth]{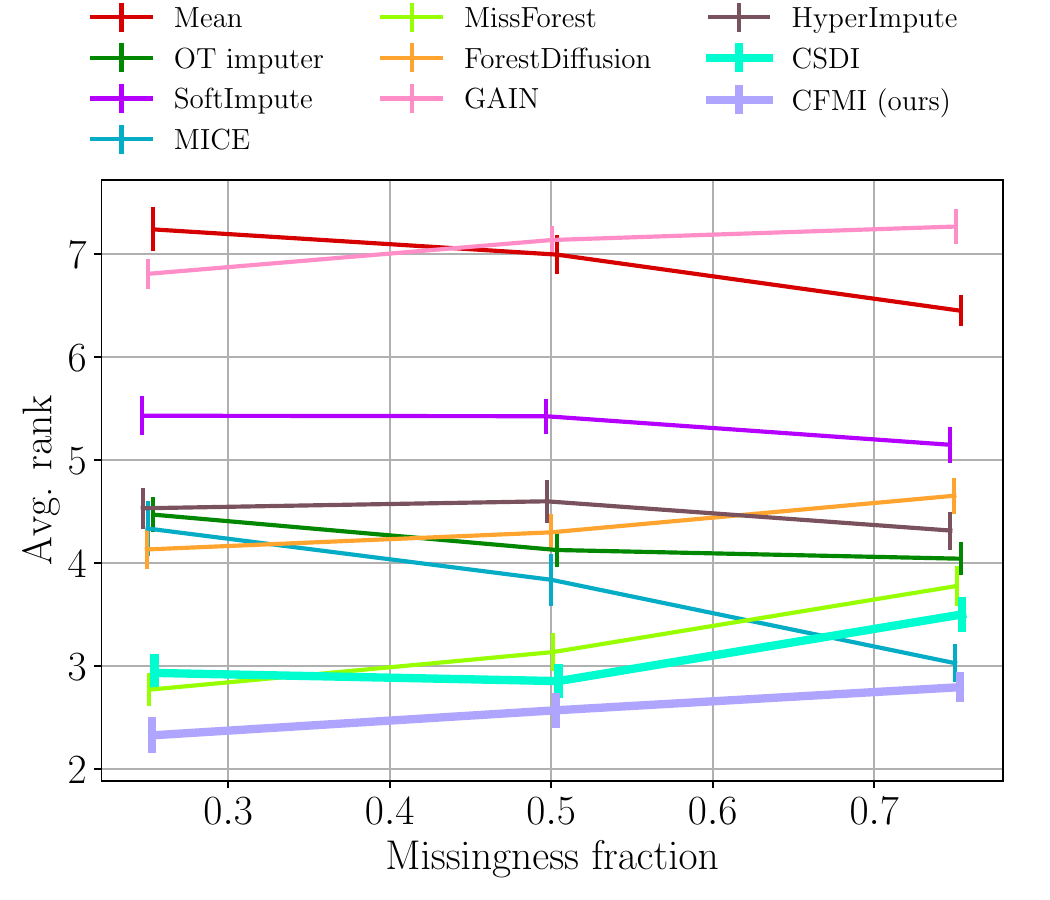}%
\includegraphics[width=0.45\linewidth]{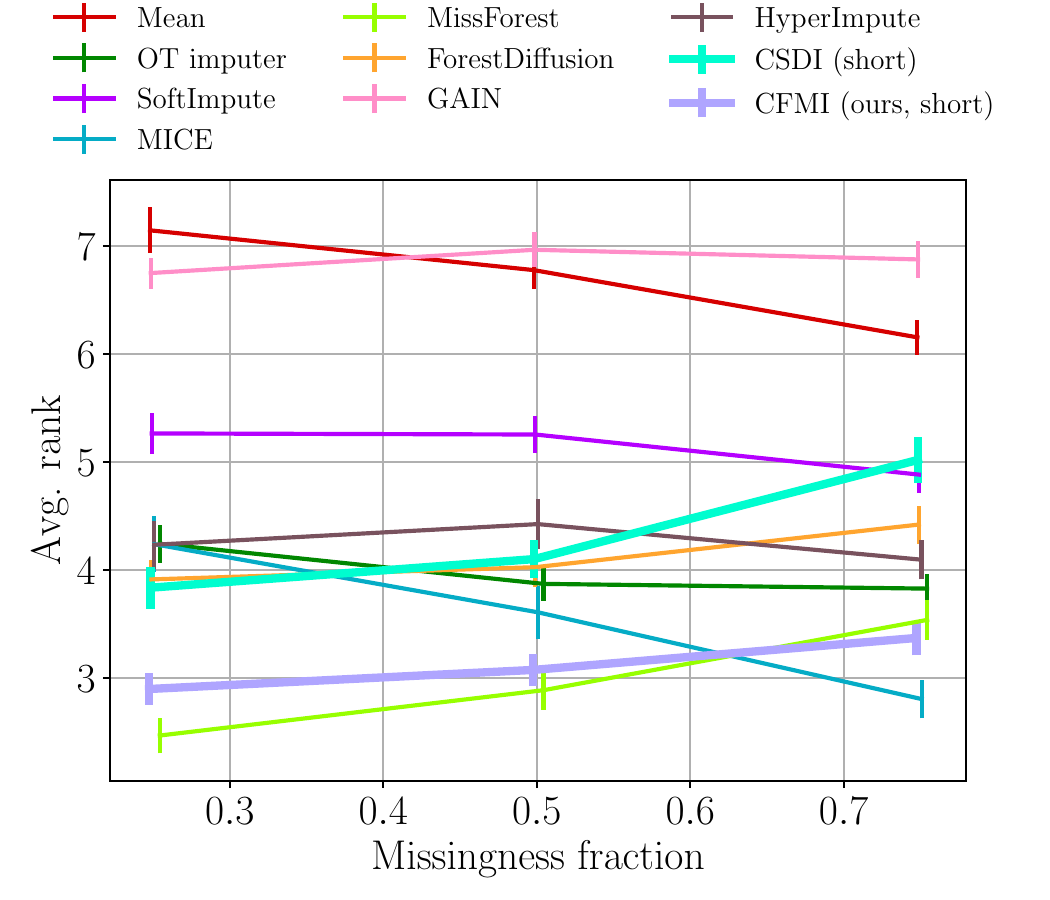}%
\captionof{figure}{\emph{Average rank over all metrics and data sets versus MCAR missingness fraction.} Left: long run CFMI and CSDI. Right: short run CFMI and CSDI. CFMI remains competitive even at very small training budgets, whereas CSDI's performance drops significantly.}%
\label{fig:uci_average_allmetric_rank_vs_missfrac}%
\end{minipage}%
\end{figure}%

In this section, we evaluate CFMI on 24 tabular data sets from the UCI repository \citep{duaUCIMachineLearning2017}, detailed in \cref{apx:exp-datasets-uci}. The classical imputation scenario is considered, where the goal is to impute an incomplete 
data set. 
As before, for this evaluation, the complete UCI data were first made 25/50/75\% missing-completely-at-random (MCAR) and 25\% missing-at-random (MAR) \citep{schoutenGeneratingMissingValues2018}.
Imputation was then performed using various methods based solely on the observed data. The true missing values were used for assessing imputation accuracy, with evaluation metrics outlined in \cref{apx:exp-details-uci-eval-metrics}.

We compare CFMI against a range of traditional \citep{vanbuurenMultivariateImputationChained2000,stekhovenMissForestNonparametricMissing2012,hastieMatrixCompletionLowRank2015} and modern \citep{yoonGAINMissingData2018,muzellecMissingDataImputation2020,jarrettHyperImputeGeneralizedIterative2022,jolicoeur-martineauGeneratingImputingTabular2024} imputation methods, as well as, the related score-based diffusion method \citep[CSDI,][]{tashiroCSDIConditionalScorebased2021}, comparing the proposed method to nine other methods in total.
For CFMI and CSDI, we use the same neural network architecture, and evaluate with two different training budgets: 5000 and 75000 training steps, where the conservative training budget often imputes faster than the traditional methods (including the training time), whereas the more generous training budget generally takes longer but often achieves better overall imputation results. 
Further details of the experimental setup, including hyperparameters, are provided in \cref{apx:exp-hyperparam-uci}.

\Cref{tab:avg-rank-mcar25} shows the per-metric ranks of each method averaged over the 24 data sets for the 25\% MCAR data.\footnote{See \cref{apx:uci-avg-rank-tables} for more MCAR and MAR results.} 
Firstly, we observe that the traditional missForest method performs as well as, or better than, the other traditional and modern baselines. 
Despite being a deterministic imputation method, missForest interestingly often outperforms the probabilistic baselines, suggesting that conditional distributions of missing data in these settings may have low entropy.
This trend reverses with increasing missingness levels, as seen in \cref{apx:uci-avg-rank-tables}, where missForest's performance declines and is eventually outperformed by the less-flexible but probabilistic imputation baseline, MICE, likely due to growing entropy in the missing data distribution. This is further highlighted in 
\cref{fig:uci_average_allmetric_rank_vs_missfrac}.

The proposed CFMI method achieves a comparable or better performance than missForest across most metrics, except for Wasserstein-2 and RMSE, where missForest holds a slight advantage but CFMI still outperforms the other methods.
Notably, CFMI outperforms missForest at higher missingness levels, as shown in \cref{apx:uci-avg-rank-tables}, maintaining robust performance as the degree of missingness increases. 
\Cref{fig:uci_average_allmetric_rank_vs_missfrac}
further summarises the results by showing the average rank over all data sets and all metrics against the missingness fraction, which further highlights that the proposed method outperforms all other approaches. 

Furthermore, CFMI consistently outperforms CSDI across missingness levels and training budgets, as illustrated in 
\cref{fig:uci_average_allmetric_rank_vs_missfrac}.
While CFMI remains competitive even with small training budgets, CSDI’s performance degrades significantly under similar conditions 
(\cref{fig:uci_average_allmetric_rank_vs_missfrac}, right).
This aligns with prior findings that flow-matching methods train faster than diffusion-based ones \citep{lipmanFlowMatchingGenerative2023}.
Additionally, we plot the imputation metrics versus training budget in \cref{apx:uci-metrics-vs-num-train-steps}, which further confirms this observation on a more granular level.

Overall, in this section, we have established that CFMI is a robust imputation method for a diverse range of tabular data sets.
Importantly, CFMI performs comparably or surpasses the best traditional method, missForest, even in low- to moderate-dimensional scenarios, where deep methods often struggle, and maintains strong performance across varying missingness levels and training budgets, demonstrating its scalability and efficiency.

To aid future meta-analyses, we provide supplementary results in \cref{apx:uci-results}: tables and box plots with raw metrics for each data set (\cref{apx:uci-raw-tables,apx:uci-raw-boxplots}), average relative performance of methods for each metric (\cref{apx:uci-relperformance}), and average rank versus missingness fraction for individual metrics (\cref{apx:uci-avgrank-vs-missingness}).

\begin{figure*}[tbp]
  \centering
  \includegraphics[width=0.95\linewidth]{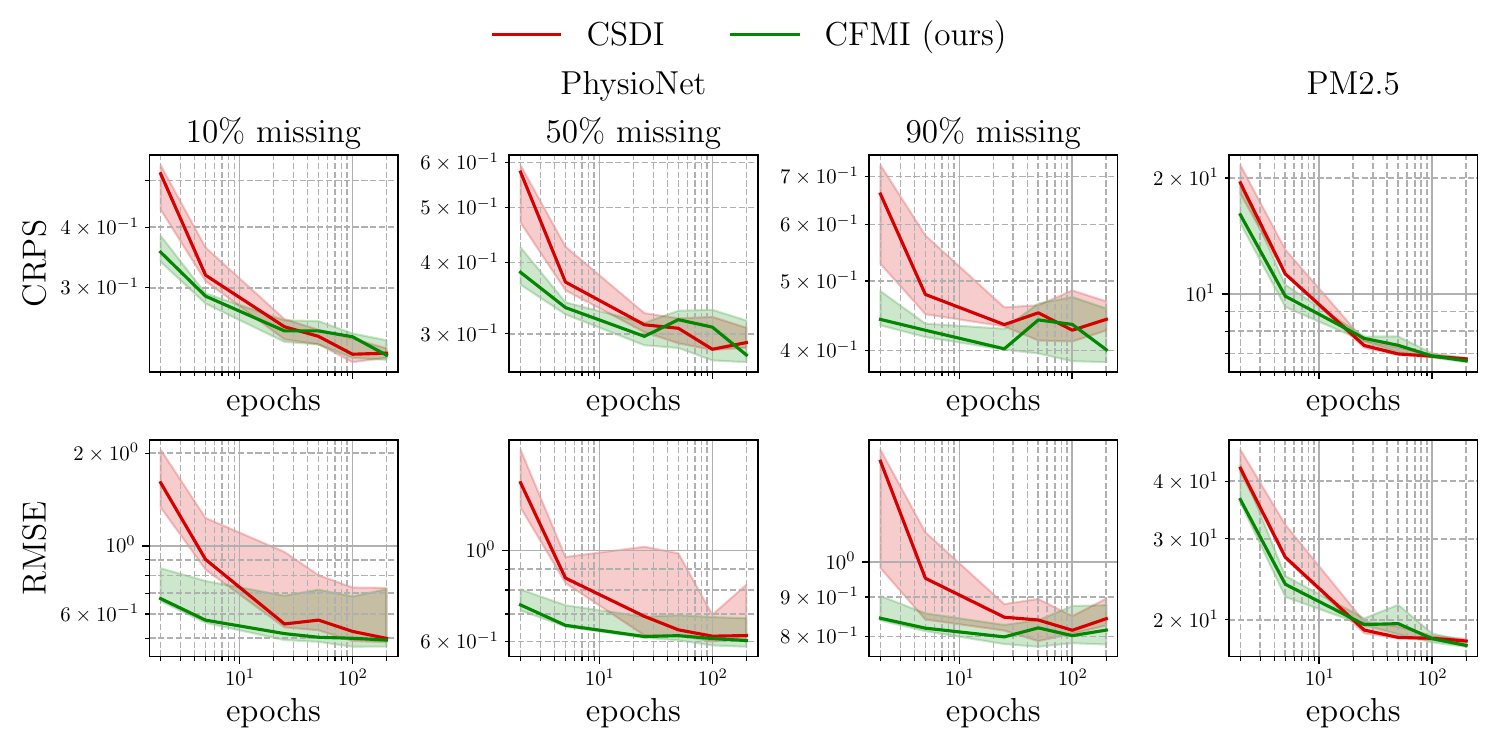}
  \caption{CRPS and RMSE versus number of training epochs. Solid lines represent the median imputation performance over 5 runs, with shaded areas denoting 95\% confidence intervals.}
  \label{fig:timeseries-crps-and-rmse-vs-epochs}
\end{figure*}

\subsection{Imputing held-out time-series data}
\label{sec:eval-time-series}

In this section, we extend our evaluation of CFMI to time-series data imputation. 
Unlike the previous section, we consider the zero-shot imputation scenario, where a model trained on one incomplete data set is used to impute a held-out incomplete data set.

Following \citet{tashiroCSDIConditionalScorebased2021}, we evaluate on two time-series data sets: PhysioNet \citep{silvaPredictingInHospitalMortality2012}, a clinical time series with 35 variables sampled hourly for 48 hours from an intensive care unit, and PM2.5 \citep{yiSTMVLFillingMissing2016}, an air quality data set from 36 weather stations for 12 months.
The PhysioNet data has 80\% inherent missingness, and for evaluation, we further mask 10/50/90\% of the observed data uniformly at random as true missing values. 
The PM2.5 data set contains 13\% missing values not-at-random (MNAR), with artificial true missing values provided for evaluation. 
Further experimental details and evaluation metrics are outlined in \cref{apx:exp-hyperparam-timeseries,apx:exp-details-timeseries-eval-metrics}.

We here focus on a comparison to the score-based diffusion method, CSDI, which has previously achieved state-of-the-art performance on these data sets. 
In \cref{fig:timeseries-crps-and-rmse-vs-epochs} we plot the 
performance of the methods in terms of CRPS and RMSE against the number of training epochs. 
As before, we observe that CFMI train significantly faster than CSDI, often achieving comparable or better performance in less time than CSDI. 
This advantage is particularly evident on the RMSE metric, where CFMI exceeds CSDI by orders of magnitude in terms of training speed. 
In addition, we provide the final metric values and comparisons with additional methods in \cref{apx:timeseries-raw-tables}, as well as results for additional metrics in  \cref{apx:timeseries-metrics-vs-num-train-epochs}, reinforcing CFMI's robust performance across various metrics. 

Overall, these results further demonstrate the effectiveness of CFMI for missing data imputation, extending the benefits beyond tabular data to time-series data. 
Notably, CFMI excels in zero-shot imputation of time-series data, effectively generalising to held-out data, and trains faster than the previous state-of-the-art diffusion-based method while achieving comparable accuracy, making it overall more efficient.

\section{Conclusion}

Despite many advances in machine learning, missing data remains a significant challenge, limiting its effectiveness in key areas of application. 
Missing data imputation is a common solution; deep learning methods often excel in high-dimensional settings, whereas traditional approaches remain more effective in low- to moderate-dimensional cases. 
Additionally, methods for tabular and time-series data---both crucial in scientific domains---are often not interchangeable, highlighting the need for a more versatile approach.

In this paper, we proposed a conditional continuous normalising flow (CNF) model for missing data imputation, efficiently trained via flow matching. 
The model builds on prior work by specifying a shared conditional model for all patterns of missingness, while leveraging the strengths of CNFs and flow matching for modelling capability and training efficiency.
Our extensive evaluations demonstrated strong performance across both tabular (low- to moderate-dimensional) and time-series (high-dimensional) datasets, achieving comparable or better results than traditional or diffusion-based methods with faster training.

A scalable and general-purpose imputation method, like CFMI, that effectively handles incomplete data across modalities and dimensionalities is fundamental for building foundation models in scientific domains. 
We anticipate that future efforts to develop foundation models in these crucial domains will greatly benefit from the generality, scalability, and efficiency of imputation methods such as CFMI.

\vskip 0.2in
\bibliography{references}
\bibliographystyle{icml2024}

\clearpage

\onecolumn %
\appendix

\section{Experiment details}

\subsection{CSDI implementation details}
\label{apx:csdi-implementation}

Our implementation of CSDI closely follows the original approach by \citet{tashiroCSDIConditionalScorebased2021}, with a few modifications outlined below. 

The primary difference lies in how we normalise the training objective. 
In our implementation, the objective is normalised by the number of target variables for each input, $\norm{\starg}_0^{-1}$, equivalent to \cref{eq:cfmi-objective}. 
In contrast, the original implementation normalises the mini-batch loss by the total number of target variables across the entire mini-batch.
Importantly, this per-data-point normalisation does not alter the optimum of the objective. 
We observe that the per-data-point normalisation typically improves the method’s performance by better balancing the learning of conditional distributions across different patterns of missingness.

Additionally, we identified and remedied an inconsistency in the original paper. 
Specifically, Figure~2 of the paper suggests that the noisy target inputs to the denoising network are set to zero everywhere except where the splitting mask, $\starg$ in our paper, assumes the targets are observed.
Conversely, Figure~5 implies that inputs correspond to $1-\scond$ are set to zero, with the remaining non-zero inputs being either noisy targets \emph{or} pure noise in dimensions that are neither in $\scond$ nor in $\starg$. This is because $1-\scond$ does not correspond to $\starg$ when training on incomplete data.
In the released code by the original authors, the latter approach is implemented.
While this discrepancy appears to have minimal impact on model performance, we adopted the approach described in Figure~2 of the paper. This decision ensures architectural consistency between CFMI and CSDI. 

\subsection{Synthetic 2D data sets}
\label{apx:exp-details-synth-2d-data}

The experiments in \cref{sec:eval-synthetic-2d-data} are based on five synthetic data distributions from \citet{wenliangLearningDeepKernels2019}. 
To generate the training data set, we sampled 20K samples from each distribution and then masked 50\% of the data values as missing uniformly at random (MCAR). 

For CFMI we have used the I-CFM form of flow-matching from \citet{tongImprovingGeneralizingFlowbased2023} with $\sigma=0$, using the \texttt{torchcfm} package. 
And for CSDI, following \citet{tashiroCSDIConditionalScorebased2021} we have used a quadratic noise scheduler with a total of 50 steps and a minimum noise level of $10^{-4}$ and a maximum of $0.5$. 

For the vector field and denoising models, we have used a residual network with 4 residual blocks, 256 hidden dimensions, and ReLU activations. 
Similar to the existing literature, the neural network takes a sinusoidal time embedding, a zero-padded observed data vector, a zero-padded target (missing) variable vector, and a binary missingness mask vector as input, and predicts the vector field. 
To optimise the model parameters we have used Adam optimiser \citep{kingmaAdamMethodStochastic2014} with a learning rate of $10^{-3}$ and gradient norm clipping with a maximum magnitude of 2 for a total of 1M gradient steps. 

\subsection{Tabular UCI data sets}
\label{apx:exp-details-uci}

\subsubsection{Datasets}
\label{apx:exp-datasets-uci}

We use the same UCI \citep{duaUCIMachineLearning2017} data sets as \citet{muzellecMissingDataImputation2020}, with the addition of the banknote dataset \citep{lohwegBanknoteAuthentication2012}, for a total of 24 diverse data sets.

\begin{table}[H]
\caption{UCI datasets}
\vspace{0.1em}
\label{tab:uci-datasets}
\centering
\begin{tabular}{@{}lcccl@{}}
\toprule
 Dataset & Reference & Num.\ samples &   Num.\ dimensions ($D$) & Outcome \\
\midrule
        airfoil self noise & \citep{brooksAirfoilSelfNoise1989} &   1503 &   5 & continuous \\
        banknote authentication & \citep{lohwegBanknoteAuthentication2012} & 1272 & 4 & binary \\
          blood transfusion & \citep{yehBloodTransfusionService2008} &    748 &   4 & binary\\
  breast cancer diagnostic & \citep{wolbergBreastCancerWisconsin1993} &    569 &  30 & binary \\
                  california housing & \citep{paceSparseSpatialAutoregressions1997} &  20640 &   8 & continuous \\
     climate model crashes & \citep{lucasClimateModelSimulation2013} &    540 &  18 & binary \\
       concrete compression & \citep{yehConcreteCompressiveStrength1998} &   1030 &   7 &continuous \\
             concrete slump & \citep{yehConcreteSlumpTest2007} &    103 &   7 & continuous \\
 connectionist bench sonar & \citep{sejnowskiConnectionistBenchSonar1988} &    208 &  60 &  binary \\
 connectionist bench vowel & \citep{deterdingConnectionistBenchVowel1988} &    990 &  10 &  binary \\
                       ecoli & \citep{nakaiEcoli1996} &    336 &   7  & categorical \\
                       glass & \citep{germanGlassIdentification1987} &    214 &   9 & categorical \\
                  ionosphere & \citep{sigillitoIonosphere1989} &    351 &  34 & binary \\
                        iris & \citep{fisherIris1936} &    150 &   4 & categorical \\
                      libras & \citep{diasLibrasMovement2009} &    360 &  90  & categorical \\
                  parkinsons & \citep{littleParkinsons2007} &    195 &  23  & binary \\
             planning relax & \citep{bhattPlanningRelax2004} &    182 &  12  & binary \\
        qsar biodegradation & \citep{mansouriQSARBiodegradation2013} &   1055 &  41  & binary \\
                       seeds & \citep{charytanowiczSeeds2010} &    210 &   7    & categorical \\
                        wine & \citep{aeberhardWine1992} &    178 &  13    & categorical \\
          wine quality red & \citep{cortezWineQuality2009} &   1599 &  10  & integer \\
        wine quality white & \citep{cortezWineQuality2009} &   4898 &  11 & integer \\
        yacht hydrodynamics & \citep{gerritsmaYachtHydrodynamics1981} &    308 &   6 & continuous \\
                       yeast & \citep{nakaiYeast1991} &   1484 &   8 & categorical \\
\bottomrule
\end{tabular}
\end{table}

For each of the 24 data sets, 3 different incomplete data sets were generated using MCAR and MAR mechanisms. 
Then, for each of the incomplete data sets 5 imputations were generated using each of the imputation methods outlined in the next section.
These imputations were then evaluated using a range of metrics outlined in \cref{apx:exp-details-uci-eval-metrics}.

\subsubsection{Imputation method details}
\label{apx:exp-hyperparam-uci}

Here we outline the details of the imputation methods used in \cref{sec:eval-uci}. 
We start with the baselines, which closely follow the defaults from \citet{muzellecMissingDataImputation2020}:
\begin{itemize}
    \item \textbf{OT imputer \citep{muzellecMissingDataImputation2020}.} We use the default implementation provided by the original authors with 3000 optimisation steps. 
    \item \textbf{SoftImpute \citep{hastieMatrixCompletionLowRank2015}.} We use the implementation provided by \citet{muzellecMissingDataImputation2020} with 3000 optimisation steps. 
    \item \textbf{MICE \citep{vanbuurenMultivariateImputationChained2000}.} We use the \texttt{IterativeImputer} implementation with \texttt{BayesianRidge} model from \texttt{scikit-learn} \citep{pedregosaScikitlearnMachineLearning2011} with 2000 iterations, each iteration consisting of a full scan of all dimensions.
    \item \textbf{MissForest \citep{stekhovenMissForestNonparametricMissing2012}.} We use the \texttt{IterativeImputer} implementation with \texttt{RandomForestRegressor} model from \texttt{scikit-learn} \citep{pedregosaScikitlearnMachineLearning2011} with 200 iterations, each iteration consisting of a full scan of all dimensions.
    \item \textbf{ForestDiffusion \citep{jolicoeur-martineauGeneratingImputingTabular2024}.} We use the \texttt{ForestDiffusion} package released by the authors with the default hyperparameters, using the \texttt{RePaint} method \citep{lugmayrRePaintInpaintingUsing2022} for imputation.
    \item \textbf{GAIN \citep{yoonGAINMissingData2018}.} We use the  implementation provided in the \texttt{hyperimpute} \citep{jarrettHyperImputeGeneralizedIterative2022} package with 25000 iterations and default hyperparameters.
    \item \textbf{HyperImpute \citep{jarrettHyperImputeGeneralizedIterative2022}.} We use the \texttt{hyperimpute} package released by the authors with 60 iterations, each iteration consisting of a full scan of all dimensions.
\end{itemize}

For CFMI we have used the I-CFM form of flow-matching from \citet{tongImprovingGeneralizingFlowbased2023} with $\sigma=0$, using the \texttt{torchcfm} package. 
And for CSDI, following \citet{tashiroCSDIConditionalScorebased2021} we have used a quadratic noise scheduler with a total of 50 steps and a minimum noise level of $10^{-4}$ and a maximum of $0.5$. 

For the vector field and denoising models, we have used a residual network with 4 residual blocks, 256 hidden dimensions, and ReLU activations. 
Similar to the existing literature, the neural network takes a sinusoidal time embedding, a zero-padded observed data vector, a zero-padded target (missing) variable vector, and a binary missingness mask vector as input, and predicts the vector field. 
To optimise the model parameters we have used AdamW optimiser \citep{loshchilovDecoupledWeightDecay2017} with a learning rate of $10^{-3}$ and a cosine learning rate schedule, batch size of 64, weight decay of $10^{-5}$, and gradient norm clipping with a maximum magnitude of 2 for a total of 5000 to 75000 gradient steps.

For imputation using a trained CFMI model we have approximated the flow equation in \cref{eq:incomplete-flow} using 100 steps of Euler integration.

\subsubsection{Evaluation metrics}
\label{apx:exp-details-uci-eval-metrics}

Here we summarise the main evaluation metrics used for the tabular data imputation in \cref{sec:eval-uci}:
\begin{itemize}
    \item \textbf{Wasserstein-2 distance (W2).} The Wasserstein-2 distance quantifies the discrepancy between the ground-truth complete training dataset and an imputed dataset. This metric is computed using the POT Python package \citep{flamaryPOTPythonOptimal2021}.\footnote{\url{https://github.com/PythonOT/POT}} We show the average W2 distance of 5 imputed data sets.
    \item \textbf{Average RMSE.} The root-mean-squared-error (RMSE) is computed between the imputed missing values and their ground-truth counterparts, averaged across multiple imputations. While this metric does not capture sample diversity and achieves an optimum when the imputation method predicts the conditional mean of the missing values \citep{vanbuurenFlexibleImputationMissing2018}, it remains widely used and is included for comparison.
    \item \textbf{Continuous ranked probability score (CRPS).} CRPS \citep{mathesonScoringRulesContinuous1976} evaluates the compatibility of a predicted probability distribution (or its samples) with true observed data. It generalises the pinball (or quantile) loss across all quantiles. CRPS is computed independently for each dimension of the imputed data, and the scores are averaged. We use the \texttt{properscoring} Python package to calculate CRPS.\footnote{\url{https://github.com/properscoring/properscoring}}
    \item \textbf{Maximum mean discrepancy (MMD).} 
    MMD \citep{grettonKernelTwoSampleTest2012} is a sample-based metric measuring discrepancies between two distributions. It is zero if and only if the distributions are identical. In our evaluation, we use kernelised MMD with Laplacian, Gaussian, and energy-distance kernels to characterize the distance between samples. MMD scores are computed using the \texttt{geomloss} Python package \citep{feydyInterpolatingOptimalTransport2019}.\footnote{\url{https://www.kernel-operations.io/geomloss/}}
    \item \textbf{Area Under the Receiver Operating Characteristic Curve (AUROC).} 
    For datasets intended for classification tasks, we evaluate AUROC by training a classifier on the imputed data. Higher-quality imputations lead to higher classification performance and thus better AUROC scores. For consistency with other metrics, we report 1-AUROC, where smaller values indicate better performance.
    \item \textbf{Prediction CRPS.} 
    To address potential biases from averaging CRPS across dimensions, we compute a one-dimensional CRPS focused on the target variable of each dataset. Specifically, a linear (or logistic) regression model is first fitted to the ground-truth data, and CRPS is calculated between predicted distributions for the ground-truth and imputed datasets. Low CRPS values indicate high accuracy in the imputed distributions.
    \item \textbf{Downstream statistical inference accuracy.} 
    Imputation is often used for statistical inference of downstream model parameters \citep[][Section~2.5.2]{vanbuurenFlexibleImputationMissing2018}. For UCI datasets, where downstream tasks typically involve regression, we evaluate the accuracy of imputations using three metrics: percent-bias (PB), coverage rate (CR), and average width (AW).
    \begin{itemize}
        \item \textbf{Percent-bias (PB):} Measures the bias in model parameters estimated from imputed data relative to ground-truth data.
        \item \textbf{Coverage rate (CR):} Assesses the proportion of confidence intervals for model parameters that include the ground-truth values.
        \item \textbf{Average width (AW):} Reflects the statistical efficiency of the imputation method. An ideal AW is small but not so small that the coverage rate falls below a nominal level \citep[][Section~2.5.2]{vanbuurenFlexibleImputationMissing2018}.
    \end{itemize}
\end{itemize}

Using these diverse metrics, we provide a comprehensive evaluation of the accuracy of the imputed data for various downstream tasks.

\subsection{Time series data sets}
\label{apx:exp-details-timeseries}

\subsubsection{Imputation method details}
\label{apx:exp-hyperparam-timeseries}

For CFMI we have used the I-CFM form of flow-matching from \citet{tongImprovingGeneralizingFlowbased2023} with $\sigma=0$, using the \texttt{torchcfm} package. 
And for CSDI, following \citet{tashiroCSDIConditionalScorebased2021} we have used a quadratic noise scheduler with a total of 50 steps and a minimum noise level of $10^{-4}$ and a maximum of $0.5$. 

For the vector field and denoising models we use the same transformer architecture from \citet[][Section~5]{tashiroCSDIConditionalScorebased2021}, which is based on DiffWave by \citet{kongDiffWaveVersatileDiffusion2021}.
The transformer uses 4 \texttt{TransformerEncoder} layers from PyTorch \citep{paszkePyTorchImperativeStyle2019}, which is composed of a multi-head attention layer with 8 heads, fully-connected layers and layer normalisation.
To encode the time-step of the time series we use 128-dimensional temporal embeddings \citep[][Appendix~E.1]{tashiroCSDIConditionalScorebased2021}.
Each feature is similarly encoded using a learnable 16-dimensional encoding. 
And to optimise the model parameters we use AdamW optimiser \citep{loshchilovDecoupledWeightDecay2017} with a learning rate of $10^{-3}$ and a multi-step learning rate schedule, batch size of 16, weight decay of $10^{-6}$, and gradient norm clipping with a maximum magnitude of 2 for a total of 200 epochs steps.

For imputation using a trained CFMI model we have approximated the flow equation in \cref{eq:incomplete-flow} using 100 steps of Euler integration.

\subsubsection{Evaluation metrics}
\label{apx:exp-details-timeseries-eval-metrics}

Here we summarise the main evaluation metrics used for the time-series imputation in \cref{sec:eval-time-series}:
\begin{itemize}
    \item \textbf{Continuous ranked probability score (CRPS).} 
    CRPS \citep{mathesonScoringRulesContinuous1976} is a proper scoring metric for one-dimensional distributions, assessing the compatibility of a predicted distribution (or its samples) with true observed data. It generalises the pinball (or quantile) loss over a continuum of quantiles. CRPS is computed independently for each dimension, and the results are averaged to produce a single score. We calculate CRPS using the \texttt{properscoring} Python package.\footnote{\url{https://github.com/properscoring/properscoring}}
    \item \textbf{Magnitude-normalised CRPS.} 
    In \citet{tashiroCSDIConditionalScorebased2021}, to assess the accuracy of probabilistic imputation, the authors use a related metric. Rather than averaging the CRPS over different dimensions, the authors sum the CRPS scores of all dimensions and then normalise the sum-score by the magnitude of the data set. While the metric is similar to CRPS, we observe it to exhibit slightly different properties to the regular CRPS.
    \item \textbf{Mean absolute error (MAE).} 
    To compute MAE, we first derive the median imputation by taking the element-wise median of imputations for each data point. The absolute error between these median imputations and the ground truth values is then averaged across the dataset. This metric achieves its optimum when the median imputation corresponds to the median of the conditional distribution for the missing variables.
    \item \textbf{Root mean squared error (RMSE).} 
    RMSE is calculated by first obtaining the mean imputation for each data point as the mean of imputations. The squared error between these mean imputations and the ground truth values is then averaged across the dataset, with the square root of the result providing the RMSE. The metric is optimal when the mean imputation matches the mean of the conditional distribution of the missing variables.
\end{itemize}

Using these metrics we evaluate the both distributional and point-estimate accuracy of imputations in time-series data.

\newpage
\section{Additional figures}

\subsection{Synthetic 2D data}

\begin{figure}[H]
  \centering
  \includegraphics[width=0.495\linewidth]{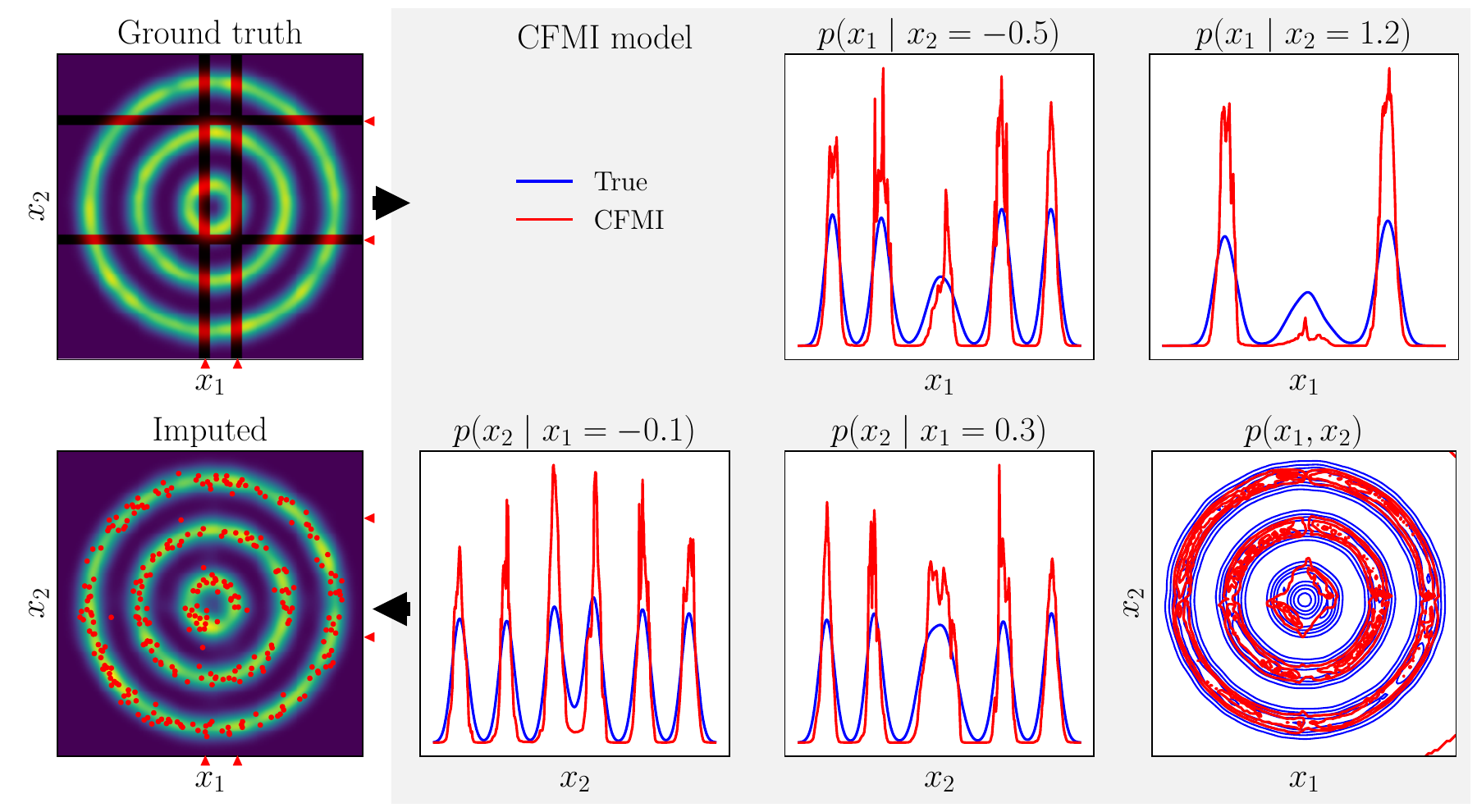}
  \includegraphics[width=0.495\linewidth]{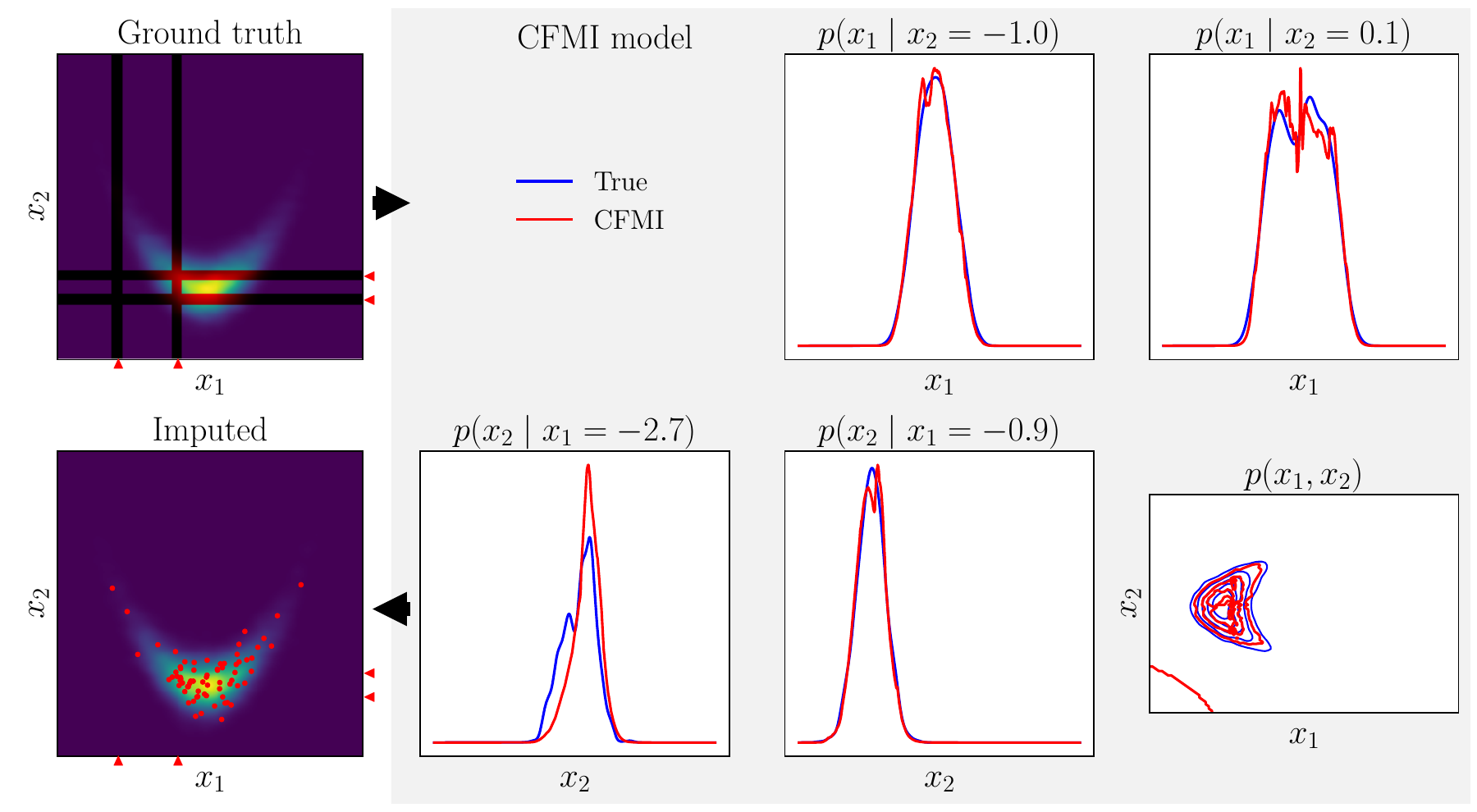}
  \includegraphics[width=0.495\linewidth]{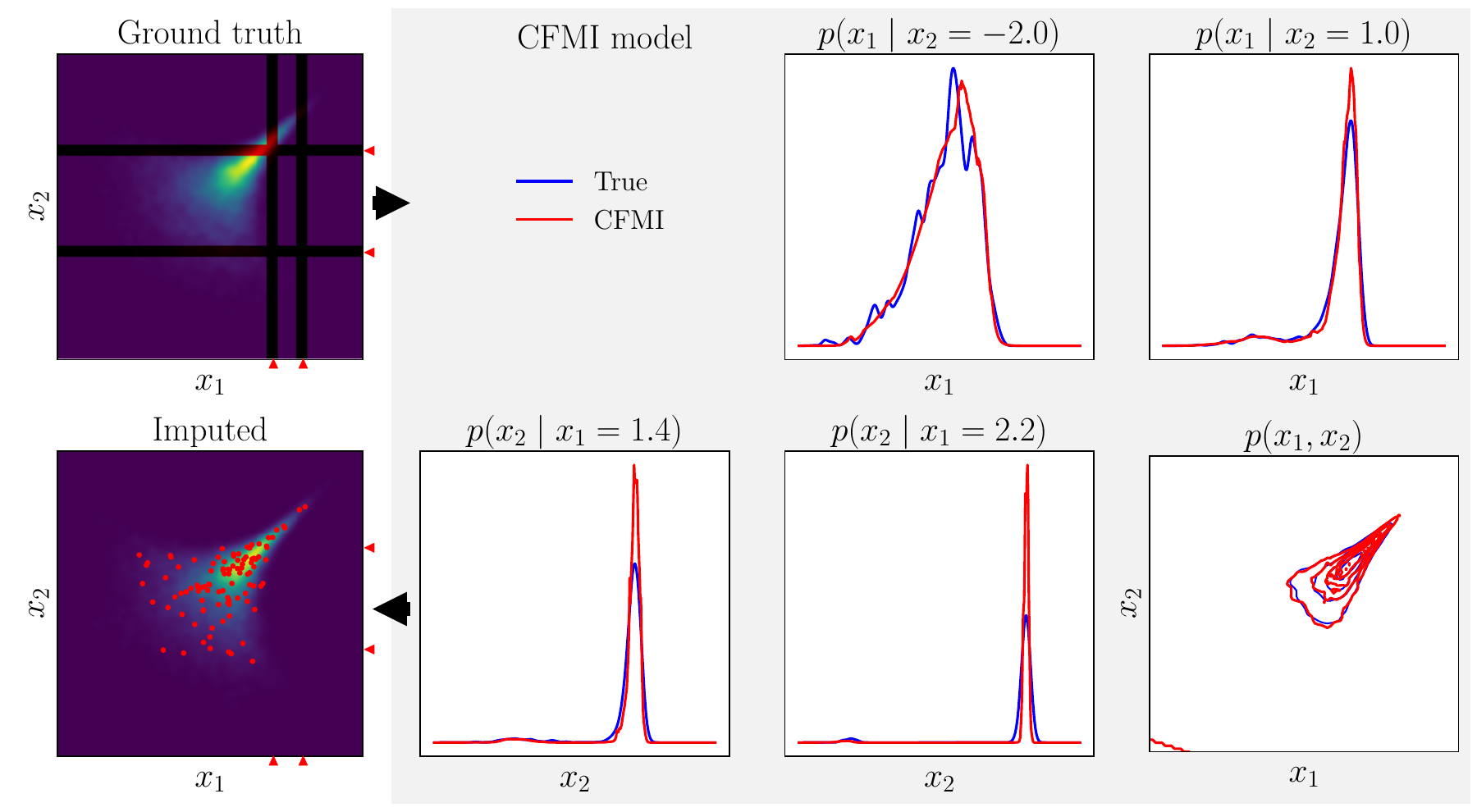}
  \includegraphics[width=0.495\linewidth]{figures/toy/toy_cosine_cfmi_visualization.pdf}
  \includegraphics[width=0.495\linewidth]{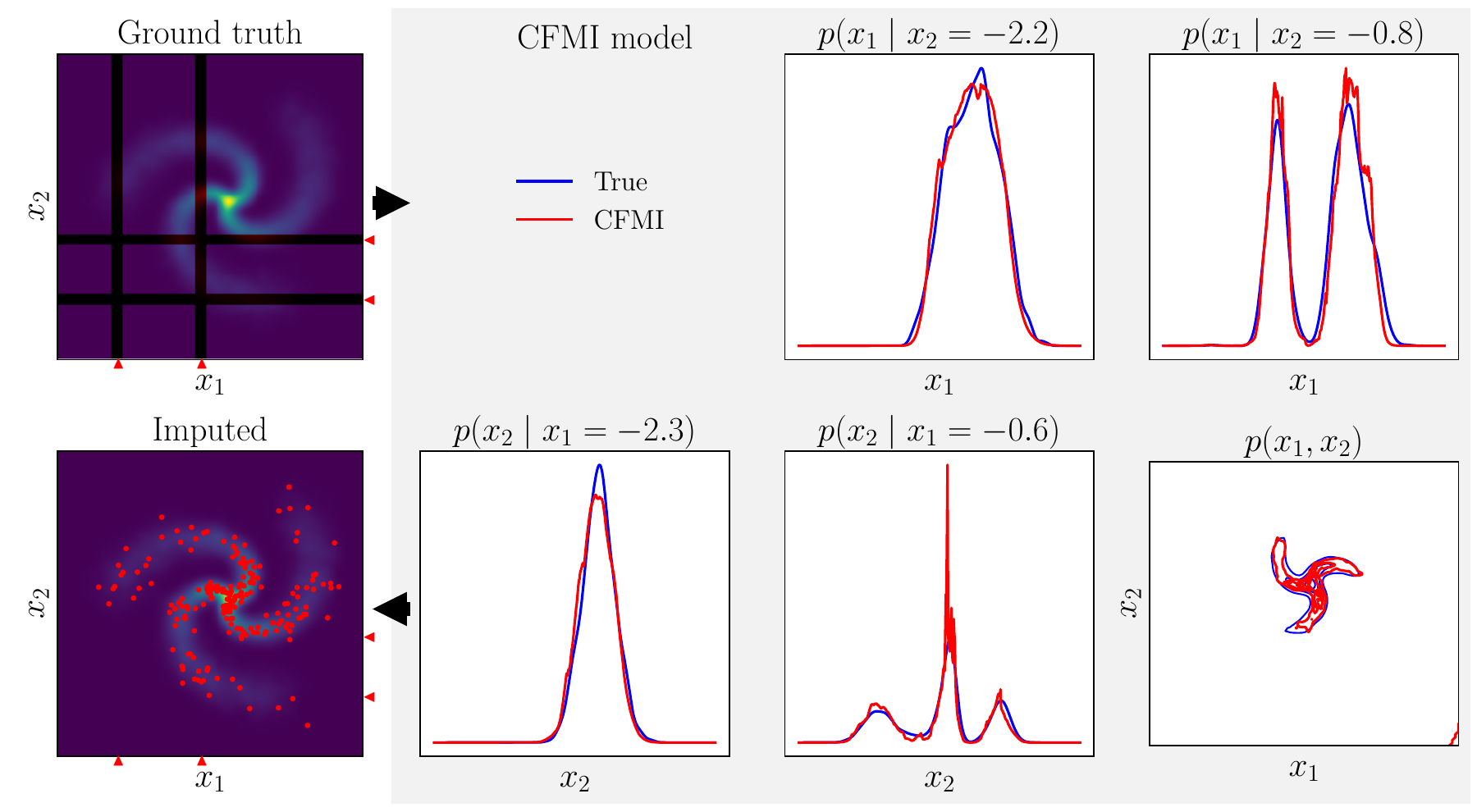}
  \caption{\emph{Imputation using CFMI on five synthetic 2D data sets.} 
  In each figure, the left-most column shows the kernel density estimate (KDE) of the ground truth distribution (top) and the KDE of the imputed data distribution (bottom). 
  The red-tinted rows and columns illustrate that some data-points may be missing one (or both) of the dimensions, and thus imputation requires sampling the corresponding (conditional) data distribution.  
  CFMI learns a model of all conditional distributions that are needed for imputation (illustrated in the grey box on the right, corresponding to the red-tinted rows and columns).
  This model is then used to impute the data by sampling the appropriate conditional distribution for each incomplete data point. 
  A sample of imputations is shown as red dots in the bottom-left, and the KDE of the imputed data closely matches that of the ground truth distribution.}
  \label{apx:fig:toy_cfmi_visualizations}
\end{figure}

\begin{figure}[H]
  \centering
  \includegraphics[width=0.6\linewidth]{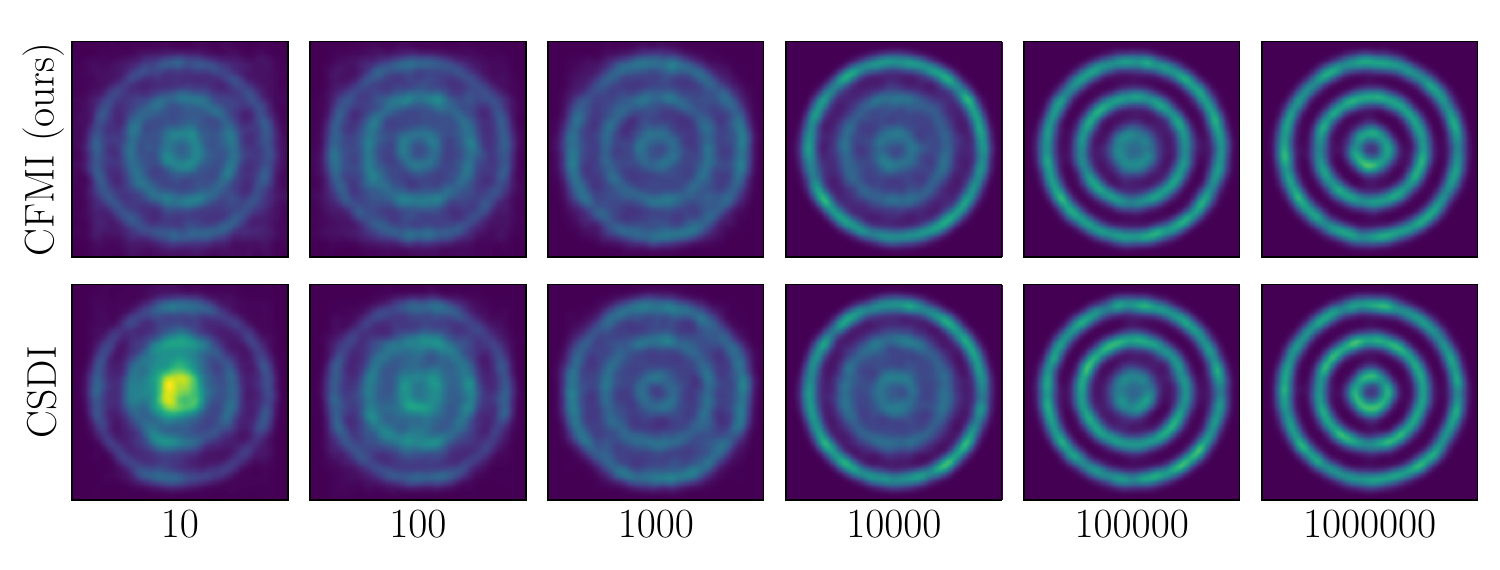}
  \includegraphics[width=0.6\linewidth]{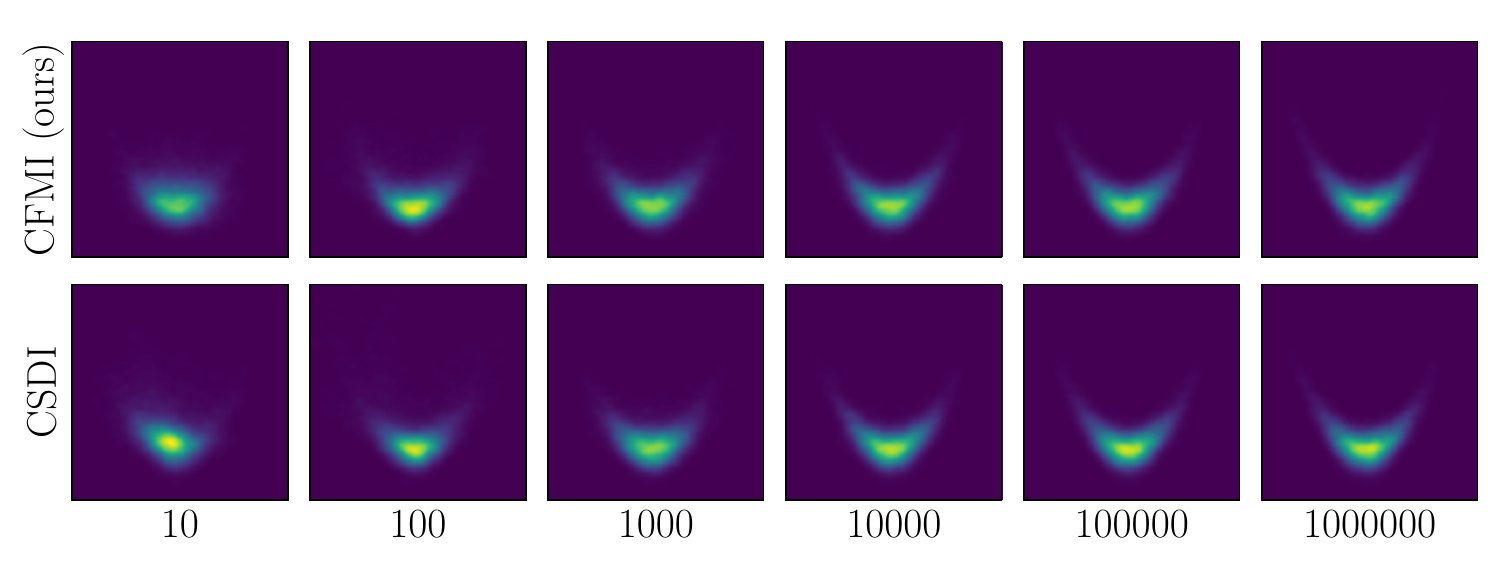}
  \includegraphics[width=0.6\linewidth]{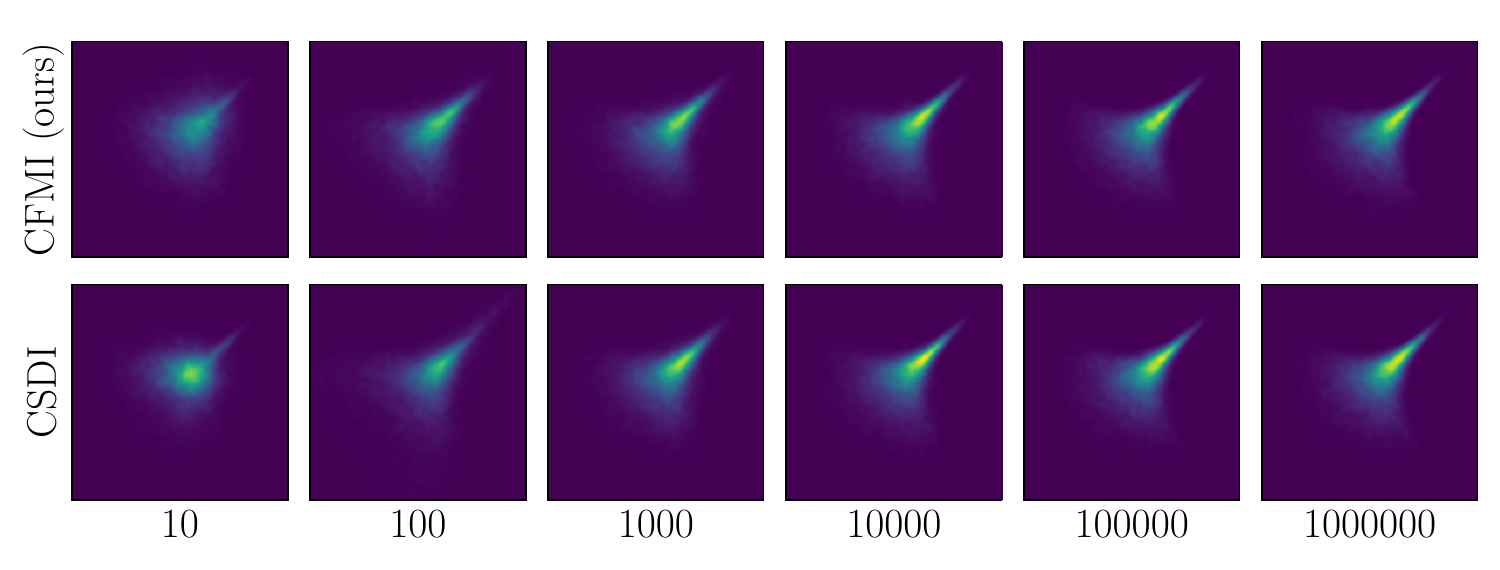}
  \includegraphics[width=0.6\linewidth]{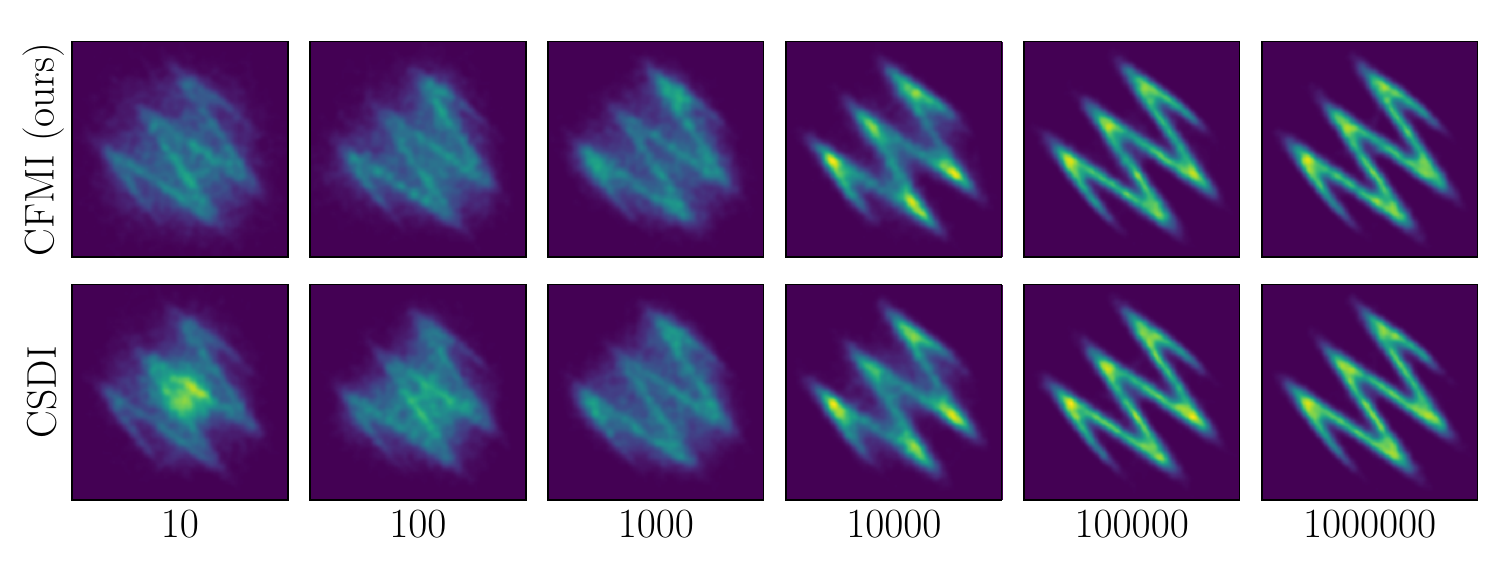}
  \includegraphics[width=0.6\linewidth]{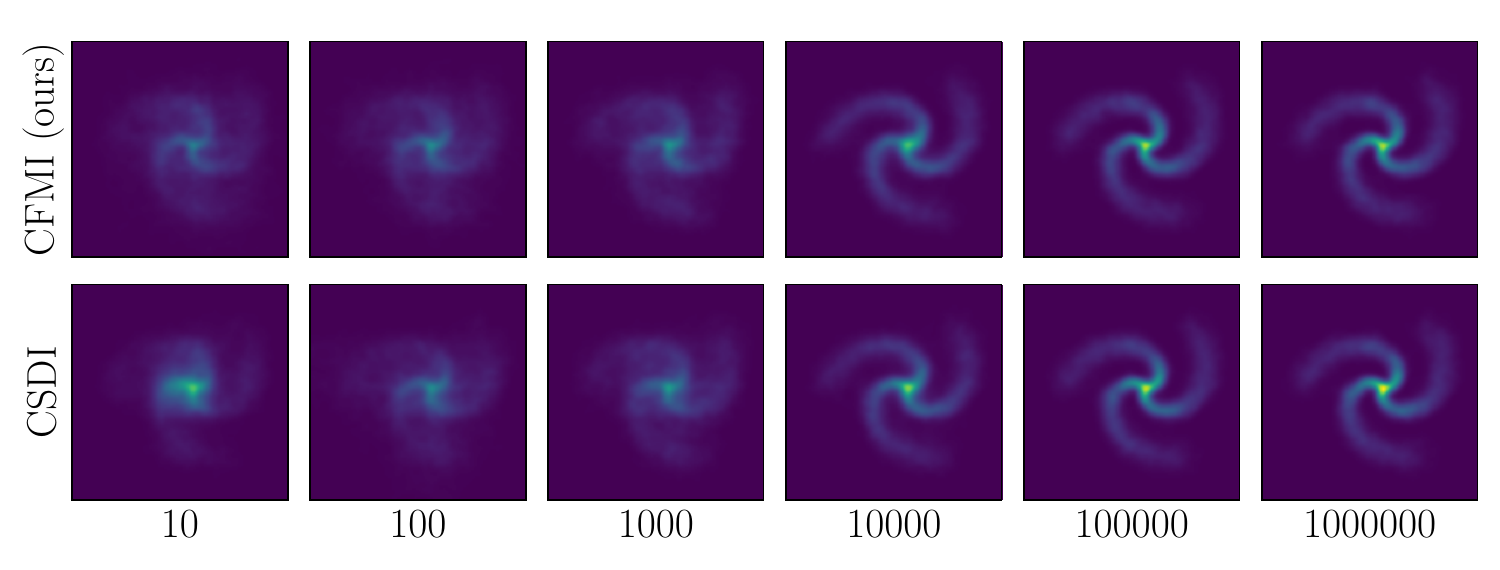}
  \caption{\emph{Comparison of CFMI and CSDI across varying training budgets (10 to 1M steps) on five synthetic data sets.} Contour plots represent kernel density estimates of imputed data sets. 
  Both methods approximate the ground truth distributions well when trained for 1M steps. 
  However, CFMI learns faster on the more complex data sets (panels 1 and 4), particularly evident at smaller training budgets of 10--100 steps.}
  \label{apx:fig:toy_cfmi_vs_csdi_with_diff_budgets}
\end{figure}

\begin{figure}[H]
  \centering
  \includegraphics[width=0.6\linewidth]{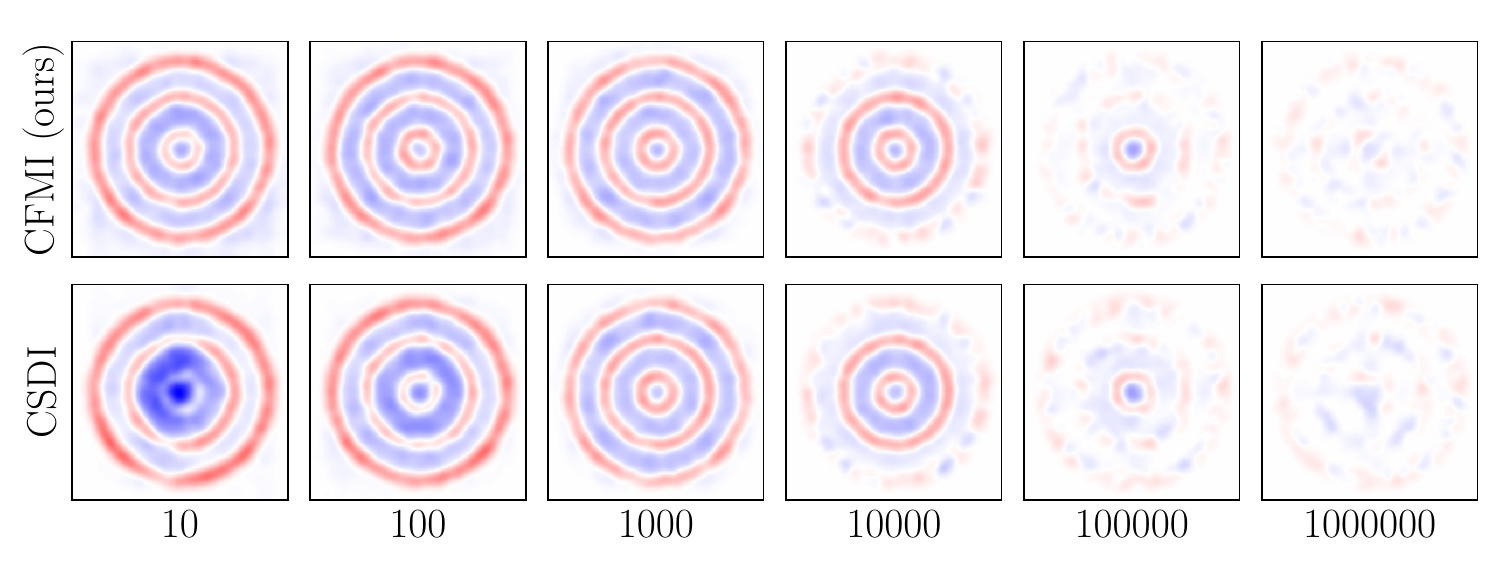}
  \includegraphics[width=0.6\linewidth]{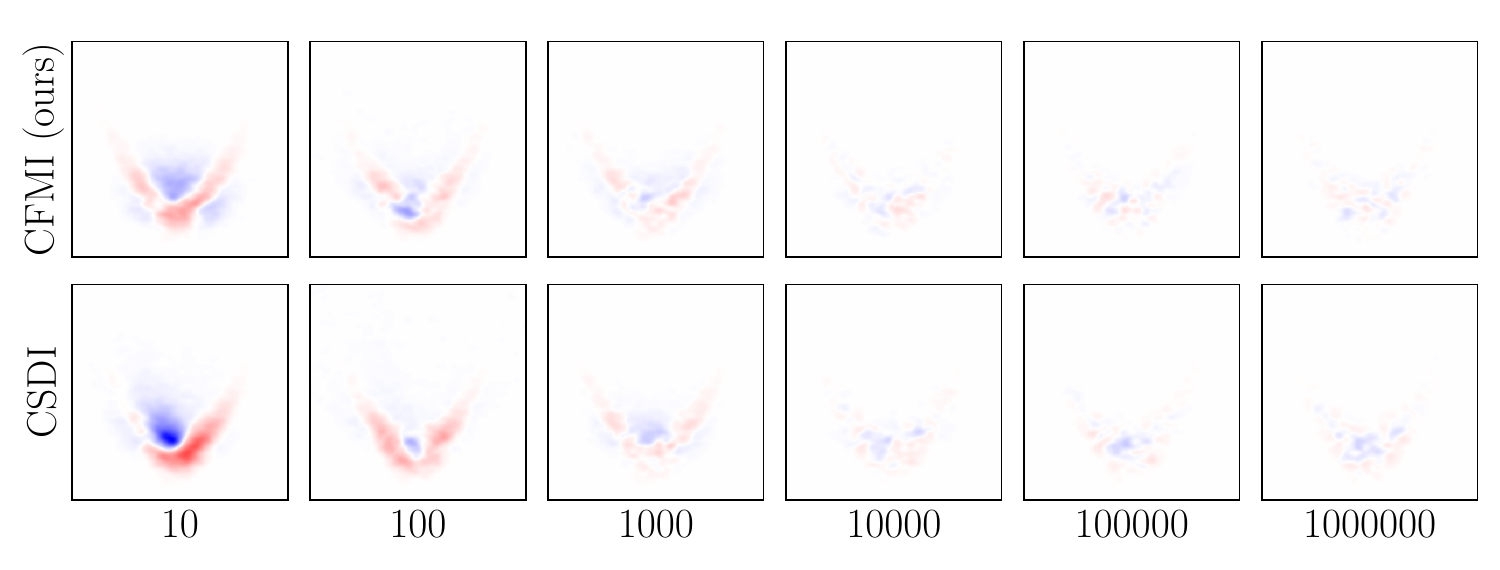}
  \includegraphics[width=0.6\linewidth]{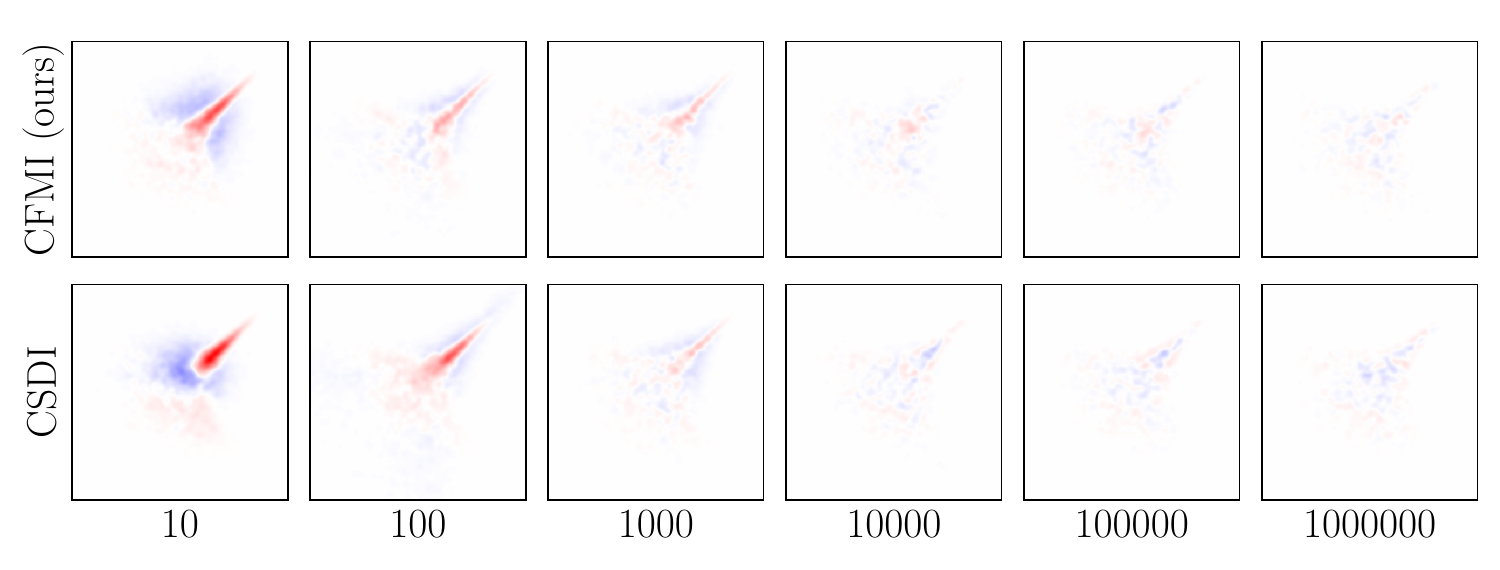}
  \includegraphics[width=0.6\linewidth]{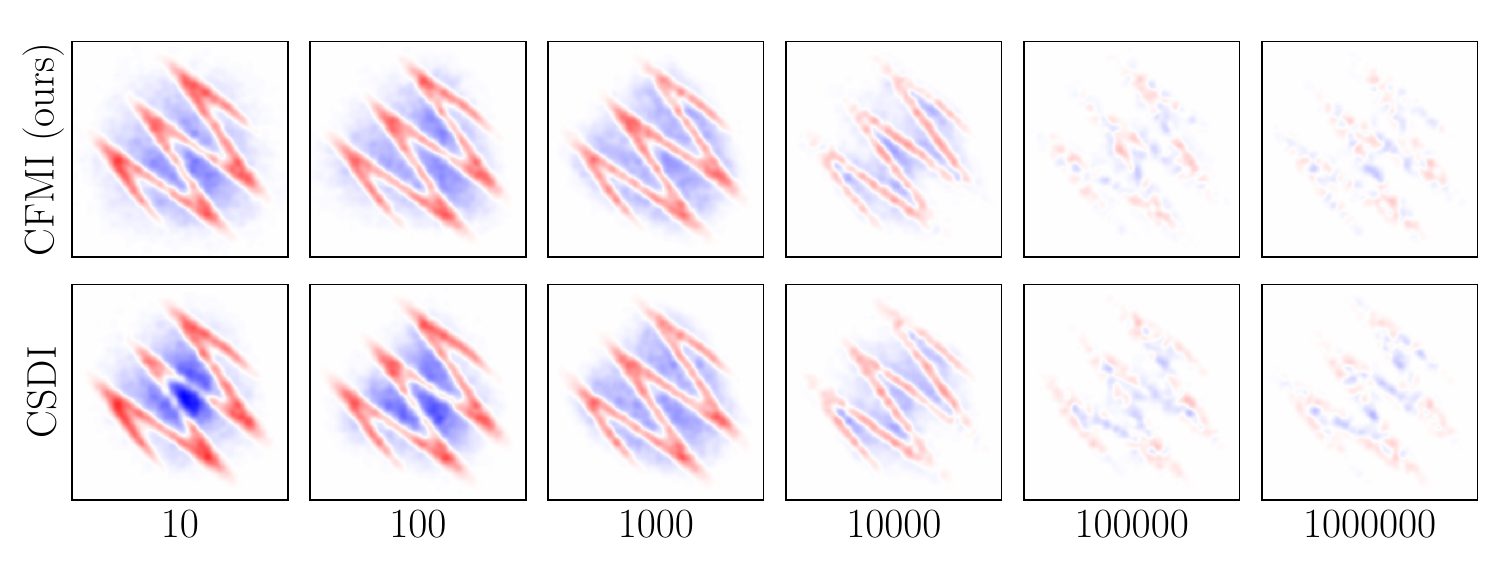}
  \includegraphics[width=0.6\linewidth]{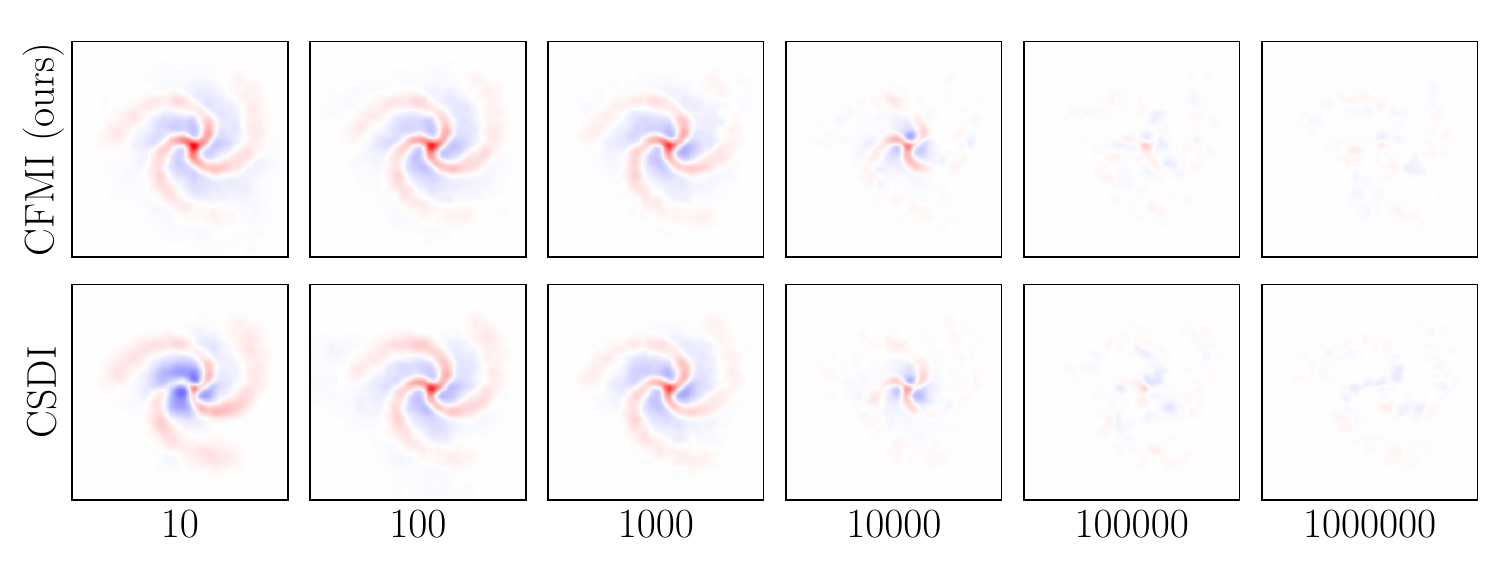}
  \caption{\emph{Comparison of CFMI and CSDI across varying training budgets (10 to 1M steps) on five synthetic data sets.} 
  Contour plots represent the difference between kernel density estimates of imputed data sets and the ground truth: blue and red mean over- and under-estimating, respectively.
  Both methods approximate the ground truth distributions well when trained for 1M steps. 
  However, CFMI generally learns faster, particularly evident at smaller training budgets of 10--100 steps.}
  \label{apx:fig:toy_cfmi_vs_csdi_with_diff_budgets_diff}
\end{figure}

\newpage
\subsection{Tabular UCI data sets}
\label{apx:uci-results}

\subsubsection{Raw results tables}
\label{apx:uci-raw-tables}

\begin{table}[H]
\centering
\caption{Wasserstein-2 results: mean and standard error. MCAR 25\% missingness. }
\resizebox{1.\linewidth}{!}{\begin{tabular}{@{}lcccccccccccc@{}}\toprule
                 & Mean             & OT imputer       & SoftImpute       & MICE             & MissForest       & ForestDiffusion  & GAIN             & HyperImpute      & CSDI             & CFMI (ours)      & CSDI (short)     & CFMI (ours, short)\\\midrule 
airfoil          & $0.315 (0.003)$  & $0.197 (0.002)$  & $0.281 (0.012)$  & $0.233 (0.001)$  & $0.060 (0.003)$  & $0.081 (0.001)$  & $0.294 (0.024)$  & $\underline{\bm{0.051 (0.002)}}$& $\bm{0.052 (0.000)}$& $\bm{0.052 (0.001)}$& $0.112 (0.012)$  & $0.114 (0.010)$  \\ 
banknote         & $0.199 (0.010)$  & $0.030 (0.001)$  & $0.099 (0.003)$  & $0.074 (0.001)$  & $\underline{\bm{0.018 (0.001)}}$& $0.037 (0.001)$  & $0.141 (0.021)$  & $0.024 (0.000)$  & $0.022 (0.000)$  & $0.021 (0.000)$  & $0.035 (0.001)$  & $0.038 (0.001)$  \\ 
blood            & $0.278 (0.004)$  & $0.067 (0.004)$  & $0.093 (0.005)$  & $0.083 (0.005)$  & $\bm{0.044 (0.005)}$& $0.086 (0.006)$  & $0.163 (0.014)$  & $0.057 (0.002)$  & $0.046 (0.002)$  & $\underline{\bm{0.039 (0.005)}}$& $0.052 (0.003)$  & $0.058 (0.003)$  \\ 
cancer           & $3.649 (0.132)$  & $1.001 (0.067)$  & $\underline{\bm{0.619 (0.042)}}$& $0.896 (0.024)$  & $0.773 (0.037)$  & $1.470 (0.043)$  & $1.689 (0.087)$  & $1.105 (0.064)$  & $0.845 (0.047)$  & $0.877 (0.052)$  & $1.573 (0.237)$  & $1.324 (0.039)$  \\ 
california       & $0.766 (0.108)$  & $0.471 (0.055)$  & $0.395 (0.050)$  & $0.463 (0.013)$  & $\underline{\bm{0.238 (0.035)}}$& $1.651 (0.289)$  & $0.733 (0.104)$  & $0.304 (0.044)$  & $0.901 (0.356)$  & $\bm{0.238 (0.027)}$& $3.849 (0.690)$  & $0.292 (0.027)$  \\ 
climate          & $\underline{\bm{2.278 (0.021)}}$& $2.891 (0.025)$  & $2.319 (0.022)$  & $4.181 (0.011)$  & $2.495 (0.025)$  & $3.440 (0.011)$  & $2.604 (0.042)$  & $2.358 (0.024)$  & $2.582 (0.031)$  & $2.871 (0.039)$  & $3.688 (0.020)$  & $4.189 (0.032)$  \\
\bottomrule
\end{tabular}}
\end{table}

\begin{table}[H]
\centering
\caption{Wasserstein-2 results: mean and standard error. MCAR 50\% missingness. }
\resizebox{1.\linewidth}{!}{\begin{tabular}{@{}lcccccccccccc@{}}\toprule
                 & Mean             & OT imputer       & SoftImpute       & MICE             & MissForest       & ForestDiffusion  & GAIN             & HyperImpute      & CSDI             & CFMI (ours)      & CSDI (short)     & CFMI (ours, short)\\\midrule 
airfoil          & $0.605 (0.004)$  & $0.397 (0.004)$  & $0.573 (0.004)$  & $0.491 (0.002)$  & $0.138 (0.005)$  & $0.146 (0.007)$  & $0.535 (0.035)$  & $0.118 (0.004)$  & $\underline{\bm{0.107 (0.002)}}$& $0.112 (0.002)$  & $0.192 (0.020)$  & $0.228 (0.024)$  \\ 
banknote         & $0.384 (0.001)$  & $0.065 (0.002)$  & $0.201 (0.005)$  & $0.137 (0.004)$  & $\underline{\bm{0.045 (0.002)}}$& $0.076 (0.002)$  & $0.259 (0.006)$  & $0.057 (0.007)$  & $0.049 (0.001)$  & $0.049 (0.001)$  & $0.085 (0.003)$  & $0.077 (0.002)$  \\ 
blood            & $0.481 (0.001)$  & $0.122 (0.005)$  & $0.168 (0.006)$  & $0.135 (0.012)$  & $0.114 (0.014)$  & $0.156 (0.003)$  & $0.303 (0.019)$  & $0.133 (0.011)$  & $\underline{\bm{0.075 (0.008)}}$& $0.088 (0.009)$  & $0.211 (0.064)$  & $0.119 (0.010)$  \\ 
cancer           & $7.149 (0.176)$  & $2.601 (0.114)$  & $\underline{\bm{1.906 (0.096)}}$& $2.670 (0.035)$  & $2.404 (0.097)$  & $4.025 (0.126)$  & $5.711 (0.224)$  & $2.564 (0.086)$  & $2.740 (0.110)$  & $2.605 (0.146)$  & $9.708 (1.656)$  & $4.031 (0.046)$  \\ 
california       & $1.125 (0.056)$  & $1.008 (0.099)$  & $0.795 (0.052)$  & $0.923 (0.052)$  & $\bm{0.618 (0.110)}$& $1.805 (0.228)$  & $1.288 (0.061)$  & $0.648 (0.111)$  & $1.130 (0.210)$  & $\underline{\bm{0.535 (0.095)}}$& $5.977 (1.554)$  & $0.662 (0.094)$  \\ 
climate          & $\underline{\bm{4.476 (0.007)}}$& $5.386 (0.004)$  & $4.552 (0.018)$  & $6.233 (0.018)$  & $5.281 (0.024)$  & $5.156 (0.052)$  & $4.841 (0.047)$  & $4.843 (0.034)$  & $4.922 (0.009)$  & $4.996 (0.013)$  & $5.494 (0.034)$  & $6.190 (0.066)$  \\
\bottomrule
\end{tabular}}
\end{table}

\begin{table}[H]
\centering
\caption{Wasserstein-2 results: mean and standard error. MCAR 75\% missingness. }
\resizebox{1.\linewidth}{!}{\begin{tabular}{@{}lcccccccccccc@{}}\toprule
                 & Mean             & OT imputer       & SoftImpute       & MICE             & MissForest       & ForestDiffusion  & GAIN             & HyperImpute      & CSDI             & CFMI (ours)      & CSDI (short)     & CFMI (ours, short)\\\midrule 
airfoil          & $0.846 (0.012)$  & $0.687 (0.017)$  & $0.795 (0.032)$  & $1.051 (0.003)$  & $0.310 (0.029)$  & $0.382 (0.056)$  & $0.775 (0.049)$  & $0.497 (0.027)$  & $\underline{\bm{0.248 (0.017)}}$& $\bm{0.255 (0.030)}$& $0.384 (0.022)$  & $0.389 (0.056)$  \\ 
banknote         & $0.510 (0.027)$  & $0.137 (0.011)$  & $0.293 (0.021)$  & $0.204 (0.004)$  & $0.145 (0.013)$  & $0.197 (0.026)$  & $0.521 (0.099)$  & $0.177 (0.008)$  & $\underline{\bm{0.126 (0.009)}}$& $\bm{0.129 (0.005)}$& $0.418 (0.241)$  & $0.160 (0.013)$  \\ 
blood            & $0.556 (0.059)$  & $\bm{0.224 (0.017)}$& $0.290 (0.023)$  & $\bm{0.225 (0.011)}$& $0.289 (0.022)$  & $0.402 (0.032)$  & $0.438 (0.069)$  & $\bm{0.228 (0.013)}$& $0.242 (0.039)$  & $0.244 (0.044)$  & $7.431 (5.977)$  & $\underline{\bm{0.223 (0.015)}}$\\ 
cancer           & $10.342 (0.096)$ & $4.855 (0.129)$  & $\underline{\bm{4.152 (0.073)}}$& $5.283 (0.023)$  & $4.816 (0.033)$  & $7.710 (0.064)$  & $9.930 (0.674)$  & $4.438 (0.041)$  & $5.917 (0.191)$  & $5.391 (0.163)$  & $1133.831 (120.235)$& $7.792 (0.109)$  \\ 
california       & $2.216 (0.313)$  & $1.884 (0.300)$  & $1.708 (0.288)$  & $1.345 (0.269)$  & $\bm{1.096 (0.277)}$& $1.612 (0.353)$  & $8.120 (2.763)$  & $\bm{1.138 (0.235)}$& $2.156 (0.436)$  & $\underline{\bm{0.918 (0.249)}}$& $159.032 (43.211)$& $1.344 (0.310)$  \\ 
climate          & $6.136 (0.023)$  & $6.820 (0.043)$  & $6.124 (0.010)$  & $6.821 (0.034)$  & $7.210 (0.074)$  & $5.961 (0.025)$  & $7.830 (0.101)$  & $6.898 (0.027)$  & $\bm{5.935 (0.052)}$& $\underline{\bm{5.903 (0.042)}}$& $55.578 (7.254)$ & $6.866 (0.071)$  \\
\bottomrule
\end{tabular}}
\end{table}

\begin{table}[H]
\centering
\caption{Wasserstein-2 results: mean and standard error. MAR 25\% missingness. }
\resizebox{1.\linewidth}{!}{\begin{tabular}{@{}lcccccccccccc@{}}\toprule
                 & Mean             & OT imputer       & SoftImpute       & MICE             & MissForest       & ForestDiffusion  & GAIN             & HyperImpute      & CSDI             & CFMI (ours)      & CSDI (short)     & CFMI (ours, short)\\\midrule 
airfoil          & $0.391 (0.020)$  & $0.258 (0.012)$  & $0.365 (0.012)$  & $0.303 (0.002)$  & $0.080 (0.005)$  & $0.093 (0.012)$  & $0.316 (0.008)$  & $0.070 (0.001)$  & $\underline{\bm{0.059 (0.003)}}$& $0.062 (0.006)$  & $0.117 (0.013)$  & $0.127 (0.016)$  \\ 
banknote         & $0.242 (0.003)$  & $0.044 (0.006)$  & $0.129 (0.005)$  & $0.093 (0.008)$  & $\underline{\bm{0.023 (0.003)}}$& $0.056 (0.010)$  & $0.133 (0.013)$  & $0.063 (0.034)$  & $0.026 (0.003)$  & $0.025 (0.002)$  & $0.041 (0.002)$  & $0.040 (0.001)$  \\ 
blood            & $0.226 (0.018)$  & $0.064 (0.008)$  & $0.094 (0.020)$  & $0.064 (0.010)$  & $0.040 (0.006)$  & $0.085 (0.009)$  & $0.149 (0.018)$  & $0.071 (0.008)$  & $\underline{\bm{0.033 (0.006)}}$& $\bm{0.035 (0.006)}$& $\bm{0.037 (0.005)}$& $0.050 (0.003)$  \\ 
cancer           & $5.740 (0.022)$  & $1.874 (0.047)$  & $\underline{\bm{1.145 (0.004)}}$& $1.360 (0.073)$  & $1.526 (0.041)$  & $3.257 (0.044)$  & $5.293 (0.428)$  & $1.782 (0.041)$  & $1.766 (0.049)$  & $1.740 (0.001)$  & $3.229 (0.967)$  & $1.980 (0.067)$  \\ 
california       & $1.023 (0.198)$  & $0.904 (0.170)$  & $0.810 (0.189)$  & $0.911 (0.116)$  & $0.658 (0.118)$  & $1.712 (0.408)$  & $1.145 (0.056)$  & $0.745 (0.135)$  & $1.290 (0.443)$  & $\underline{\bm{0.527 (0.076)}}$& $2.886 (1.019)$  & $0.687 (0.115)$  \\ 
climate          & $\underline{\bm{2.347 (0.037)}}$& $2.840 (0.028)$  & $2.384 (0.041)$  & $4.071 (0.001)$  & $2.698 (0.023)$  & $3.427 (0.066)$  & $2.967 (0.164)$  & $2.474 (0.044)$  & $2.707 (0.056)$  & $2.952 (0.037)$  & $3.591 (0.019)$  & $3.748 (0.013)$  \\
\bottomrule
\end{tabular}}
\end{table}

\begin{table}[H]
\centering
\caption{Average RMSE results: mean and standard error. MCAR 25\% missingness. }
\resizebox{1.\linewidth}{!}{\begin{tabular}{@{}lcccccccccccc@{}}\toprule
                 & Mean             & OT imputer       & SoftImpute       & MICE             & MissForest       & ForestDiffusion  & GAIN             & HyperImpute      & CSDI             & CFMI (ours)      & CSDI (short)     & CFMI (ours, short)\\\midrule 
airfoil          & $1.009 (0.008)$  & $0.832 (0.003)$  & $0.954 (0.022)$  & $1.212 (0.003)$  & $0.535 (0.010)$  & $0.980 (0.003)$  & $0.977 (0.020)$  & $\underline{\bm{0.513 (0.021)}}$& $0.765 (0.006)$  & $0.719 (0.014)$  & $1.074 (0.011)$  & $1.049 (0.004)$  \\ 
banknote         & $0.979 (0.022)$  & $0.470 (0.013)$  & $0.711 (0.009)$  & $0.765 (0.007)$  & $\underline{\bm{0.412 (0.010)}}$& $0.657 (0.009)$  & $0.831 (0.047)$  & $0.448 (0.014)$  & $0.498 (0.001)$  & $0.489 (0.004)$  & $0.595 (0.012)$  & $0.606 (0.006)$  \\ 
blood            & $1.035 (0.007)$  & $0.689 (0.014)$  & $0.712 (0.013)$  & $0.905 (0.015)$  & $\underline{\bm{0.599 (0.007)}}$& $0.910 (0.004)$  & $0.960 (0.019)$  & $0.633 (0.018)$  & $0.731 (0.016)$  & $0.701 (0.013)$  & $0.803 (0.024)$  & $0.804 (0.015)$  \\ 
cancer           & $0.985 (0.009)$  & $0.514 (0.014)$  & $\underline{\bm{0.399 (0.010)}}$& $0.484 (0.002)$  & $0.449 (0.007)$  & $0.623 (0.005)$  & $0.657 (0.013)$  & $0.537 (0.011)$  & $0.470 (0.009)$  & $0.481 (0.010)$  & $0.644 (0.048)$  & $0.592 (0.005)$  \\ 
california       & $1.075 (0.066)$  & $0.933 (0.073)$  & $0.861 (0.040)$  & $1.060 (0.018)$  & $\underline{\bm{0.786 (0.049)}}$& $1.645 (0.086)$  & $1.103 (0.092)$  & $\bm{0.829 (0.077)}$& $1.300 (0.150)$  & $0.922 (0.037)$  & $2.889 (0.310)$  & $0.998 (0.033)$  \\ 
climate          & $\underline{\bm{1.005 (0.005)}}$& $1.133 (0.005)$  & $1.015 (0.005)$  & $1.413 (0.000)$  & $1.051 (0.006)$  & $1.251 (0.007)$  & $1.076 (0.008)$  & $1.022 (0.005)$  & $1.067 (0.006)$  & $1.131 (0.011)$  & $1.306 (0.005)$  & $1.412 (0.007)$  \\
\bottomrule
\end{tabular}}
\end{table}

\begin{table}[H]
\centering
\caption{Average RMSE results: mean and standard error. MCAR 50\% missingness. }
\resizebox{1.\linewidth}{!}{\begin{tabular}{@{}lcccccccccccc@{}}\toprule
                 & Mean             & OT imputer       & SoftImpute       & MICE             & MissForest       & ForestDiffusion  & GAIN             & HyperImpute      & CSDI             & CFMI (ours)      & CSDI (short)     & CFMI (ours, short)\\\midrule 
airfoil          & $1.000 (0.003)$  & $0.881 (0.003)$  & $0.979 (0.003)$  & $1.363 (0.001)$  & $0.749 (0.012)$  & $1.058 (0.003)$  & $0.995 (0.018)$  & $\underline{\bm{0.726 (0.008)}}$& $0.967 (0.007)$  & $0.932 (0.009)$  & $1.155 (0.007)$  & $1.223 (0.009)$  \\ 
banknote         & $0.996 (0.009)$  & $\bm{0.608 (0.018)}$& $0.781 (0.013)$  & $0.895 (0.009)$  & $\underline{\bm{0.605 (0.012)}}$& $0.800 (0.013)$  & $0.875 (0.008)$  & $0.638 (0.004)$  & $0.699 (0.018)$  & $0.692 (0.020)$  & $0.813 (0.026)$  & $0.812 (0.019)$  \\ 
blood            & $1.023 (0.021)$  & $0.746 (0.014)$  & $0.740 (0.007)$  & $0.963 (0.004)$  & $0.767 (0.040)$  & $0.965 (0.013)$  & $0.970 (0.047)$  & $\underline{\bm{0.684 (0.011)}}$& $0.830 (0.016)$  & $0.858 (0.020)$  & $1.187 (0.135)$  & $0.983 (0.024)$  \\ 
cancer           & $1.005 (0.010)$  & $0.603 (0.016)$  & $\underline{\bm{0.506 (0.017)}}$& $0.620 (0.002)$  & $0.578 (0.013)$  & $0.762 (0.011)$  & $0.899 (0.014)$  & $0.594 (0.014)$  & $0.620 (0.017)$  & $0.602 (0.021)$  & $1.172 (0.065)$  & $0.775 (0.005)$  \\ 
california       & $1.086 (0.039)$  & $1.004 (0.042)$  & $\underline{\bm{0.951 (0.046)}}$& $1.170 (0.030)$  & $\bm{0.966 (0.043)}$& $1.444 (0.022)$  & $1.119 (0.027)$  & $\bm{0.980 (0.030)}$& $1.793 (0.285)$  & $1.092 (0.046)$  & $3.529 (0.333)$  & $1.149 (0.030)$  \\ 
climate          & $\underline{\bm{1.002 (0.000)}}$& $1.171 (0.005)$  & $1.011 (0.001)$  & $1.411 (0.004)$  & $1.165 (0.002)$  & $1.158 (0.007)$  & $1.057 (0.008)$  & $1.061 (0.003)$  & $1.117 (0.003)$  & $1.126 (0.006)$  & $1.262 (0.001)$  & $1.399 (0.015)$  \\
\bottomrule
\end{tabular}}
\end{table}

\begin{table}[H]
\centering
\caption{Average RMSE results: mean and standard error. MCAR 75\% missingness. }
\resizebox{1.\linewidth}{!}{\begin{tabular}{@{}lcccccccccccc@{}}\toprule
                 & Mean             & OT imputer       & SoftImpute       & MICE             & MissForest       & ForestDiffusion  & GAIN             & HyperImpute      & CSDI             & CFMI (ours)      & CSDI (short)     & CFMI (ours, short)\\\midrule 
airfoil          & $1.004 (0.005)$  & $\underline{\bm{0.957 (0.006)}}$& $0.985 (0.012)$  & $1.610 (0.006)$  & $1.000 (0.005)$  & $1.127 (0.011)$  & $1.049 (0.032)$  & $0.968 (0.005)$  & $1.159 (0.004)$  & $1.100 (0.011)$  & $1.264 (0.033)$  & $1.336 (0.027)$  \\ 
banknote         & $0.991 (0.006)$  & $\underline{\bm{0.733 (0.022)}}$& $0.841 (0.017)$  & $1.008 (0.010)$  & $0.812 (0.020)$  & $0.901 (0.023)$  & $1.090 (0.087)$  & $0.800 (0.018)$  & $0.832 (0.022)$  & $0.837 (0.013)$  & $1.005 (0.052)$  & $0.940 (0.012)$  \\ 
blood            & $1.015 (0.027)$  & $0.889 (0.030)$  & $\underline{\bm{0.836 (0.013)}}$& $1.079 (0.020)$  & $1.063 (0.057)$  & $1.236 (0.070)$  & $1.068 (0.031)$  & $\bm{0.843 (0.015)}$& $1.073 (0.078)$  & $1.000 (0.020)$  & $2.989 (0.715)$  & $1.092 (0.052)$  \\ 
cancer           & $1.006 (0.004)$  & $0.734 (0.008)$  & $\underline{\bm{0.647 (0.006)}}$& $0.833 (0.000)$  & $0.759 (0.003)$  & $0.925 (0.006)$  & $1.101 (0.032)$  & $0.696 (0.004)$  & $0.825 (0.013)$  & $0.764 (0.007)$  & $10.231 (0.562)$ & $1.024 (0.005)$  \\ 
california       & $1.063 (0.023)$  & $\bm{1.025 (0.024)}$& $\underline{\bm{1.004 (0.026)}}$& $1.237 (0.022)$  & $1.103 (0.023)$  & $1.211 (0.039)$  & $1.737 (0.211)$  & $1.131 (0.015)$  & $1.485 (0.115)$  & $1.218 (0.028)$  & $8.813 (1.050)$  & $1.322 (0.039)$  \\ 
climate          & $\underline{\bm{1.003 (0.001)}}$& $1.190 (0.001)$  & $1.016 (0.001)$  & $1.416 (0.001)$  & $1.319 (0.008)$  & $1.135 (0.003)$  & $1.211 (0.013)$  & $1.174 (0.002)$  & $1.146 (0.009)$  & $1.106 (0.002)$  & $3.144 (0.182)$  & $1.385 (0.008)$  \\
\bottomrule
\end{tabular}}
\end{table}

\begin{table}[H]
\centering
\caption{Average RMSE results: mean and standard error. MAR 25\% missingness. }
\resizebox{1.\linewidth}{!}{\begin{tabular}{@{}lcccccccccccc@{}}\toprule
                 & Mean             & OT imputer       & SoftImpute       & MICE             & MissForest       & ForestDiffusion  & GAIN             & HyperImpute      & CSDI             & CFMI (ours)      & CSDI (short)     & CFMI (ours, short)\\\midrule 
airfoil          & $0.989 (0.008)$  & $0.840 (0.007)$  & $0.961 (0.013)$  & $1.263 (0.013)$  & $\bm{0.582 (0.012)}$& $0.985 (0.031)$  & $0.930 (0.022)$  & $\underline{\bm{0.567 (0.017)}}$& $0.804 (0.007)$  & $0.733 (0.007)$  & $1.055 (0.003)$  & $1.034 (0.015)$  \\ 
banknote         & $1.014 (0.030)$  & $0.503 (0.038)$  & $0.750 (0.025)$  & $0.829 (0.012)$  & $\underline{\bm{0.443 (0.046)}}$& $0.712 (0.029)$  & $0.765 (0.015)$  & $0.622 (0.100)$  & $0.517 (0.040)$  & $0.515 (0.038)$  & $0.637 (0.008)$  & $0.612 (0.033)$  \\ 
blood            & $1.081 (0.035)$  & $0.735 (0.048)$  & $\bm{0.715 (0.059)}$& $0.924 (0.068)$  & $\underline{\bm{0.665 (0.054)}}$& $1.038 (0.103)$  & $0.994 (0.076)$  & $\bm{0.703 (0.044)}$& $0.787 (0.035)$  & $0.786 (0.043)$  & $1.011 (0.120)$  & $0.883 (0.029)$  \\ 
cancer           & $1.016 (0.004)$  & $0.560 (0.009)$  & $\underline{\bm{0.441 (0.006)}}$& $0.490 (0.010)$  & $0.502 (0.005)$  & $0.737 (0.006)$  & $0.932 (0.039)$  & $0.560 (0.008)$  & $0.529 (0.008)$  & $0.529 (0.004)$  & $0.681 (0.072)$  & $0.604 (0.008)$  \\ 
california       & $1.033 (0.037)$  & $0.915 (0.046)$  & $\bm{0.885 (0.038)}$& $1.106 (0.018)$  & $\underline{\bm{0.833 (0.053)}}$& $1.485 (0.179)$  & $1.146 (0.016)$  & $\bm{0.860 (0.053)}$& $1.180 (0.082)$  & $0.942 (0.047)$  & $3.228 (1.162)$  & $1.016 (0.054)$  \\ 
climate          & $\underline{\bm{1.003 (0.010)}}$& $1.108 (0.004)$  & $1.013 (0.011)$  & $1.415 (0.003)$  & $1.075 (0.007)$  & $1.257 (0.010)$  & $1.144 (0.037)$  & $1.030 (0.010)$  & $1.080 (0.013)$  & $1.137 (0.008)$  & $1.306 (0.005)$  & $1.339 (0.009)$  \\
\bottomrule
\end{tabular}}
\end{table}

\begin{table}[H]
\centering
\caption{CRPS reults: mean and standard error. MCAR 25\% missingness. }
\resizebox{1.\linewidth}{!}{\begin{tabular}{@{}lcccccccccccc@{}}\toprule
                 & Mean             & OT imputer       & SoftImpute       & MICE             & MissForest       & ForestDiffusion  & GAIN             & HyperImpute      & CSDI             & CFMI (ours)      & CSDI (short)     & CFMI (ours, short)\\\midrule 
airfoil          & $0.827 (0.006)$  & $0.604 (0.005)$  & $0.727 (0.021)$  & $0.546 (0.007)$  & $0.279 (0.006)$  & $0.400 (0.004)$  & $0.566 (0.030)$  & $\underline{\bm{0.248 (0.010)}}$& $0.283 (0.006)$  & $0.277 (0.008)$  & $0.462 (0.007)$  & $0.446 (0.014)$  \\ 
banknote         & $0.796 (0.013)$  & $0.301 (0.005)$  & $0.586 (0.007)$  & $0.366 (0.005)$  & $0.265 (0.006)$  & $0.292 (0.005)$  & $0.512 (0.035)$  & $0.269 (0.003)$  & $0.236 (0.002)$  & $\underline{\bm{0.227 (0.004)}}$& $0.284 (0.005)$  & $0.277 (0.004)$  \\ 
blood            & $0.756 (0.008)$  & $0.353 (0.005)$  & $0.429 (0.009)$  & $0.344 (0.004)$  & $0.326 (0.010)$  & $0.364 (0.009)$  & $0.425 (0.002)$  & $0.302 (0.006)$  & $\underline{\bm{0.258 (0.006)}}$& $0.264 (0.012)$  & $0.290 (0.006)$  & $0.293 (0.008)$  \\ 
cancer           & $0.748 (0.008)$  & $0.308 (0.006)$  & $0.241 (0.004)$  & $\underline{\bm{0.187 (0.003)}}$& $0.248 (0.003)$  & $0.280 (0.004)$  & $0.348 (0.005)$  & $0.329 (0.004)$  & $0.254 (0.004)$  & $0.263 (0.004)$  & $0.252 (0.008)$  & $0.243 (0.002)$  \\ 
california       & $0.603 (0.001)$  & $0.343 (0.001)$  & $0.454 (0.001)$  & $0.385 (0.006)$  & $0.294 (0.001)$  & $0.479 (0.015)$  & $0.514 (0.018)$  & $0.289 (0.003)$  & $0.282 (0.001)$  & $\underline{\bm{0.271 (0.002)}}$& $0.336 (0.003)$  & $0.310 (0.002)$  \\ 
climate          & $0.873 (0.005)$  & $0.887 (0.005)$  & $0.871 (0.005)$  & $\underline{\bm{0.702 (0.007)}}$& $0.846 (0.007)$  & $0.741 (0.001)$  & $0.807 (0.009)$  & $0.854 (0.007)$  & $0.841 (0.007)$  & $0.878 (0.008)$  & $\bm{0.707 (0.002)}$& $\bm{0.707 (0.003)}$\\
\bottomrule
\end{tabular}}
\end{table}

\begin{table}[H]
\centering
\caption{CRPS results: mean and standard error. MCAR 50\% missingness. }
\resizebox{1.\linewidth}{!}{\begin{tabular}{@{}lcccccccccccc@{}}\toprule
                 & Mean             & OT imputer       & SoftImpute       & MICE             & MissForest       & ForestDiffusion  & GAIN             & HyperImpute      & CSDI             & CFMI (ours)      & CSDI (short)     & CFMI (ours, short)\\\midrule 
airfoil          & $0.822 (0.003)$  & $0.661 (0.002)$  & $0.771 (0.004)$  & $0.600 (0.003)$  & $0.418 (0.002)$  & $0.463 (0.002)$  & $0.621 (0.038)$  & $\underline{\bm{0.373 (0.002)}}$& $0.416 (0.003)$  & $0.409 (0.000)$  & $0.527 (0.004)$  & $0.527 (0.003)$  \\ 
banknote         & $0.797 (0.005)$  & $0.389 (0.013)$  & $0.637 (0.008)$  & $0.434 (0.009)$  & $0.377 (0.007)$  & $0.375 (0.007)$  & $0.545 (0.018)$  & $0.359 (0.003)$  & $\underline{\bm{0.331 (0.007)}}$& $\bm{0.331 (0.008)}$& $0.382 (0.014)$  & $0.366 (0.008)$  \\ 
blood            & $0.749 (0.008)$  & $0.402 (0.003)$  & $0.482 (0.005)$  & $0.385 (0.004)$  & $0.394 (0.015)$  & $0.413 (0.004)$  & $0.478 (0.008)$  & $0.437 (0.001)$  & $\underline{\bm{0.322 (0.003)}}$& $0.346 (0.003)$  & $0.359 (0.012)$  & $0.369 (0.006)$  \\ 
cancer           & $0.752 (0.003)$  & $0.359 (0.005)$  & $0.309 (0.002)$  & $\underline{\bm{0.248 (0.003)}}$& $0.311 (0.003)$  & $0.367 (0.006)$  & $0.412 (0.002)$  & $0.356 (0.003)$  & $0.351 (0.003)$  & $0.339 (0.004)$  & $0.425 (0.009)$  & $0.330 (0.002)$  \\ 
california       & $0.603 (0.001)$  & $0.421 (0.001)$  & $0.505 (0.001)$  & $0.430 (0.007)$  & $0.377 (0.002)$  & $0.459 (0.016)$  & $0.463 (0.015)$  & $0.357 (0.004)$  & $0.349 (0.000)$  & $\underline{\bm{0.334 (0.002)}}$& $0.410 (0.009)$  & $0.366 (0.002)$  \\ 
climate          & $0.869 (0.000)$  & $0.831 (0.003)$  & $0.868 (0.005)$  & $\underline{\bm{0.697 (0.004)}}$& $0.803 (0.012)$  & $0.750 (0.004)$  & $0.783 (0.004)$  & $0.840 (0.004)$  & $0.805 (0.001)$  & $0.814 (0.003)$  & $0.707 (0.002)$  & $0.727 (0.002)$  \\
\bottomrule
\end{tabular}}
\end{table}

\begin{table}[H]
\centering
\caption{CRPS results: mean and standard error. MCAR 75\% missingness. }
\resizebox{1.\linewidth}{!}{\begin{tabular}{@{}lcccccccccccc@{}}\toprule
                 & Mean             & OT imputer       & SoftImpute       & MICE             & MissForest       & ForestDiffusion  & GAIN             & HyperImpute      & CSDI             & CFMI (ours)      & CSDI (short)     & CFMI (ours, short)\\\midrule 
airfoil          & $0.822 (0.004)$  & $0.750 (0.007)$  & $0.778 (0.019)$  & $0.687 (0.002)$  & $0.543 (0.008)$  & $0.564 (0.009)$  & $0.622 (0.006)$  & $0.635 (0.008)$  & $0.531 (0.003)$  & $\underline{\bm{0.518 (0.004)}}$& $0.574 (0.004)$  & $0.586 (0.008)$  \\ 
banknote         & $0.796 (0.002)$  & $0.471 (0.013)$  & $0.673 (0.013)$  & $0.478 (0.004)$  & $0.476 (0.012)$  & $0.472 (0.022)$  & $0.577 (0.018)$  & $0.548 (0.017)$  & $\bm{0.429 (0.017)}$& $\underline{\bm{0.428 (0.008)}}$& $0.462 (0.017)$  & $0.440 (0.013)$  \\ 
blood            & $0.749 (0.006)$  & $0.496 (0.005)$  & $0.567 (0.015)$  & $\underline{\bm{0.445 (0.005)}}$& $0.519 (0.024)$  & $0.592 (0.041)$  & $0.499 (0.010)$  & $0.481 (0.015)$  & $0.475 (0.004)$  & $0.473 (0.010)$  & $0.537 (0.039)$  & $0.461 (0.011)$  \\ 
cancer           & $0.749 (0.001)$  & $0.423 (0.003)$  & $0.417 (0.001)$  & $\underline{\bm{0.345 (0.002)}}$& $0.382 (0.001)$  & $0.474 (0.003)$  & $0.448 (0.005)$  & $0.413 (0.003)$  & $0.454 (0.007)$  & $0.424 (0.003)$  & $2.270 (0.135)$  & $0.438 (0.003)$  \\ 
california       & $0.603 (0.001)$  & $0.497 (0.001)$  & $0.546 (0.002)$  & $0.455 (0.001)$  & $0.428 (0.004)$  & $0.434 (0.006)$  & $0.550 (0.039)$  & $0.420 (0.002)$  & $0.402 (0.003)$  & $\underline{\bm{0.398 (0.002)}}$& $0.655 (0.019)$  & $0.428 (0.003)$  \\ 
climate          & $0.868 (0.001)$  & $0.806 (0.004)$  & $0.850 (0.004)$  & $\underline{\bm{0.699 (0.001)}}$& $0.790 (0.005)$  & $0.760 (0.003)$  & $0.709 (0.008)$  & $0.823 (0.002)$  & $0.757 (0.002)$  & $0.771 (0.002)$  & $0.912 (0.025)$  & $0.721 (0.005)$  \\
\bottomrule
\end{tabular}}
\end{table}

\begin{table}[H]
\centering
\caption{CRPS results: mean and standard error. MAR 25\% missingness. }
\resizebox{1.\linewidth}{!}{\begin{tabular}{@{}lcccccccccccc@{}}\toprule
                 & Mean             & OT imputer       & SoftImpute       & MICE             & MissForest       & ForestDiffusion  & GAIN             & HyperImpute      & CSDI             & CFMI (ours)      & CSDI (short)     & CFMI (ours, short)\\\midrule 
airfoil          & $0.840 (0.020)$  & $0.650 (0.022)$  & $0.774 (0.021)$  & $0.559 (0.007)$  & $0.323 (0.021)$  & $0.401 (0.023)$  & $0.538 (0.016)$  & $\underline{\bm{0.276 (0.013)}}$& $0.306 (0.009)$  & $0.293 (0.012)$  & $0.463 (0.013)$  & $0.444 (0.017)$  \\ 
banknote         & $0.852 (0.041)$  & $0.336 (0.024)$  & $0.617 (0.024)$  & $0.413 (0.003)$  & $0.294 (0.029)$  & $0.347 (0.029)$  & $0.467 (0.029)$  & $0.405 (0.083)$  & $\underline{\bm{0.251 (0.016)}}$& $\bm{0.253 (0.013)}$& $0.308 (0.007)$  & $0.290 (0.014)$  \\ 
blood            & $0.844 (0.067)$  & $0.448 (0.041)$  & $0.534 (0.067)$  & $0.380 (0.029)$  & $0.390 (0.030)$  & $0.517 (0.077)$  & $0.473 (0.030)$  & $0.469 (0.040)$  & $\underline{\bm{0.313 (0.024)}}$& $\bm{0.336 (0.025)}$& $\bm{0.323 (0.015)}$& $0.355 (0.013)$  \\ 
cancer           & $0.897 (0.002)$  & $0.413 (0.004)$  & $0.314 (0.003)$  & $\underline{\bm{0.231 (0.004)}}$& $0.341 (0.005)$  & $0.462 (0.003)$  & $0.507 (0.029)$  & $0.402 (0.005)$  & $0.355 (0.005)$  & $0.354 (0.002)$  & $0.317 (0.016)$  & $0.298 (0.003)$  \\ 
california       & $0.596 (0.028)$  & $0.386 (0.037)$  & $0.462 (0.036)$  & $0.403 (0.026)$  & $0.329 (0.031)$  & $0.453 (0.032)$  & $0.482 (0.038)$  & $0.315 (0.026)$  & $0.312 (0.041)$  & $\underline{\bm{0.284 (0.026)}}$& $0.370 (0.061)$  & $0.323 (0.029)$  \\ 
climate          & $0.867 (0.006)$  & $0.877 (0.006)$  & $0.866 (0.009)$  & $\underline{\bm{0.695 (0.004)}}$& $0.843 (0.004)$  & $0.746 (0.013)$  & $0.797 (0.011)$  & $0.836 (0.010)$  & $0.824 (0.009)$  & $0.820 (0.003)$  & $0.703 (0.005)$  & $0.717 (0.006)$  \\
\bottomrule
\end{tabular}}
\end{table}

\begin{table}[H]
\centering
\caption{Energy MMD results: mean and standard error. MCAR 25\% missingness. }
\resizebox{1.\linewidth}{!}{\begin{tabular}{@{}lcccccccccccc@{}}\toprule
                 & Mean             & OT imputer       & SoftImpute       & MICE             & MissForest       & ForestDiffusion  & GAIN             & HyperImpute      & CSDI             & CFMI (ours)      & CSDI (short)     & CFMI (ours, short)\\\midrule 
airfoil          & $0.016 (0.000)$  & $0.010 (0.000)$  & $0.014 (0.001)$  & $0.002 (0.000)$  & $0.001 (0.000)$  & $0.001 (0.000)$  & $0.019 (0.002)$  & $0.001 (0.000)$  & $\underline{\bm{0.000 (0.000)}}$& $0.000 (0.000)$  & $0.001 (0.000)$  & $0.001 (0.000)$  \\ 
banknote         & $0.015 (0.000)$  & $0.001 (0.000)$  & $0.009 (0.000)$  & $0.001 (0.000)$  & $0.000 (0.000)$  & $0.001 (0.000)$  & $0.013 (0.004)$  & $0.001 (0.000)$  & $0.000 (0.000)$  & $\underline{\bm{0.000 (0.000)}}$& $0.001 (0.000)$  & $0.001 (0.000)$  \\ 
blood            & $0.017 (0.000)$  & $0.003 (0.000)$  & $0.007 (0.000)$  & $0.002 (0.000)$  & $0.001 (0.000)$  & $0.002 (0.000)$  & $0.017 (0.003)$  & $0.002 (0.000)$  & $\underline{\bm{0.001 (0.000)}}$& $0.001 (0.000)$  & $0.001 (0.000)$  & $0.001 (0.000)$  \\ 
cancer           & $0.033 (0.000)$  & $0.005 (0.000)$  & $0.004 (0.000)$  & $\underline{\bm{0.002 (0.000)}}$& $0.003 (0.000)$  & $0.005 (0.000)$  & $0.009 (0.000)$  & $0.007 (0.000)$  & $0.004 (0.000)$  & $0.004 (0.000)$  & $0.004 (0.000)$  & $0.003 (0.000)$  \\ 
california       & $0.014 (0.000)$  & $0.005 (0.000)$  & $0.008 (0.000)$  & $0.002 (0.000)$  & $0.001 (0.000)$  & $0.017 (0.003)$  & $0.022 (0.006)$  & $0.001 (0.000)$  & $0.000 (0.000)$  & $\underline{\bm{0.000 (0.000)}}$& $0.000 (0.000)$  & $0.000 (0.000)$  \\ 
climate          & $0.021 (0.000)$  & $0.014 (0.000)$  & $0.020 (0.000)$  & $\underline{\bm{0.005 (0.000)}}$& $0.018 (0.000)$  & $0.011 (0.001)$  & $0.018 (0.000)$  & $0.020 (0.000)$  & $0.016 (0.000)$  & $0.012 (0.000)$  & $0.007 (0.000)$  & $0.006 (0.000)$  \\
\bottomrule
\end{tabular}}
\end{table}

\begin{table}[H]
\centering
\caption{Energy MMD results: mean and standard error. MCAR 50\% missingness. }
\resizebox{1.\linewidth}{!}{\begin{tabular}{@{}lcccccccccccc@{}}\toprule
                 & Mean             & OT imputer       & SoftImpute       & MICE             & MissForest       & ForestDiffusion  & GAIN             & HyperImpute      & CSDI             & CFMI (ours)      & CSDI (short)     & CFMI (ours, short)\\\midrule 
airfoil          & $0.064 (0.001)$  & $0.046 (0.001)$  & $0.061 (0.001)$  & $0.010 (0.000)$  & $0.006 (0.000)$  & $0.005 (0.001)$  & $0.062 (0.009)$  & $0.006 (0.000)$  & $0.002 (0.000)$  & $\underline{\bm{0.002 (0.000)}}$& $0.004 (0.001)$  & $0.004 (0.000)$  \\ 
banknote         & $0.056 (0.001)$  & $0.008 (0.000)$  & $0.038 (0.001)$  & $0.005 (0.000)$  & $0.002 (0.000)$  & $0.003 (0.000)$  & $0.044 (0.003)$  & $0.004 (0.001)$  & $0.001 (0.000)$  & $\underline{\bm{0.001 (0.000)}}$& $0.003 (0.001)$  & $0.002 (0.000)$  \\ 
blood            & $0.056 (0.001)$  & $0.010 (0.000)$  & $0.030 (0.001)$  & $0.006 (0.000)$  & $0.007 (0.001)$  & $0.006 (0.001)$  & $0.066 (0.008)$  & $0.016 (0.001)$  & $\underline{\bm{0.002 (0.000)}}$& $0.003 (0.001)$  & $0.003 (0.000)$  & $0.004 (0.001)$  \\ 
cancer           & $0.133 (0.002)$  & $0.018 (0.001)$  & $0.022 (0.002)$  & $\underline{\bm{0.006 (0.000)}}$& $0.009 (0.000)$  & $0.025 (0.002)$  & $0.144 (0.025)$  & $0.024 (0.001)$  & $0.018 (0.001)$  & $0.018 (0.001)$  & $0.043 (0.005)$  & $0.015 (0.001)$  \\ 
california       & $0.066 (0.001)$  & $0.037 (0.001)$  & $0.045 (0.000)$  & $0.007 (0.001)$  & $0.004 (0.000)$  & $0.025 (0.006)$  & $0.095 (0.005)$  & $0.005 (0.000)$  & $0.000 (0.000)$  & $\underline{\bm{0.000 (0.000)}}$& $0.001 (0.000)$  & $0.001 (0.000)$  \\ 
climate          & $0.090 (0.000)$  & $0.042 (0.000)$  & $0.087 (0.000)$  & $\underline{\bm{0.009 (0.000)}}$& $0.043 (0.002)$  & $0.053 (0.001)$  & $0.096 (0.005)$  & $0.071 (0.000)$  & $0.049 (0.000)$  & $0.045 (0.000)$  & $0.020 (0.001)$  & $0.012 (0.001)$  \\
\bottomrule
\end{tabular}}
\end{table}

\begin{table}[H]
\centering
\caption{Energy MMD results: mean and standard error. MCAR 75\% missingness. }
\resizebox{1.\linewidth}{!}{\begin{tabular}{@{}lcccccccccccc@{}}\toprule
                 & Mean             & OT imputer       & SoftImpute       & MICE             & MissForest       & ForestDiffusion  & GAIN             & HyperImpute      & CSDI             & CFMI (ours)      & CSDI (short)     & CFMI (ours, short)\\\midrule 
airfoil          & $0.141 (0.001)$  & $0.119 (0.004)$  & $0.137 (0.004)$  & $0.040 (0.001)$  & $0.020 (0.003)$  & $0.044 (0.017)$  & $0.131 (0.008)$  & $0.050 (0.001)$  & $\bm{0.010 (0.001)}$& $\underline{\bm{0.009 (0.002)}}$& $0.015 (0.000)$  & $0.013 (0.003)$  \\ 
banknote         & $0.109 (0.001)$  & $0.023 (0.001)$  & $0.084 (0.004)$  & $0.008 (0.000)$  & $0.009 (0.001)$  & $0.021 (0.008)$  & $0.132 (0.027)$  & $0.024 (0.002)$  & $0.006 (0.000)$  & $\underline{\bm{0.005 (0.000)}}$& $0.008 (0.001)$  & $0.006 (0.000)$  \\ 
blood            & $0.108 (0.003)$  & $0.024 (0.003)$  & $0.072 (0.004)$  & $\underline{\bm{0.012 (0.001)}}$& $0.026 (0.002)$  & $0.077 (0.032)$  & $0.122 (0.039)$  & $0.035 (0.002)$  & $0.023 (0.001)$  & $0.017 (0.005)$  & $0.015 (0.002)$  & $\bm{0.012 (0.003)}$\\ 
cancer           & $0.355 (0.004)$  & $0.056 (0.002)$  & $0.097 (0.002)$  & $\underline{\bm{0.019 (0.000)}}$& $0.031 (0.002)$  & $0.083 (0.004)$  & $0.240 (0.049)$  & $0.057 (0.003)$  & $0.076 (0.003)$  & $0.083 (0.002)$  & $3.372 (0.144)$  & $0.060 (0.007)$  \\ 
california       & $0.171 (0.000)$  & $0.128 (0.000)$  & $0.142 (0.002)$  & $0.012 (0.000)$  & $0.017 (0.001)$  & $0.019 (0.001)$  & $0.874 (0.268)$  & $0.009 (0.001)$  & $\underline{\bm{0.002 (0.000)}}$& $\bm{0.002 (0.000)}$& $0.043 (0.002)$  & $0.006 (0.000)$  \\ 
climate          & $0.283 (0.001)$  & $0.113 (0.005)$  & $0.257 (0.001)$  & $\underline{\bm{0.017 (0.002)}}$& $0.075 (0.002)$  & $0.171 (0.012)$  & $0.497 (0.011)$  & $0.127 (0.003)$  & $0.119 (0.006)$  & $0.162 (0.003)$  & $0.187 (0.032)$  & $0.039 (0.001)$  \\
\bottomrule
\end{tabular}}
\end{table}

\begin{table}[H]
\centering
\caption{Energy MMD results: mean and standard error. MAR 25\% missingness. }
\resizebox{1.\linewidth}{!}{\begin{tabular}{@{}lcccccccccccc@{}}\toprule
                 & Mean             & OT imputer       & SoftImpute       & MICE             & MissForest       & ForestDiffusion  & GAIN             & HyperImpute      & CSDI             & CFMI (ours)      & CSDI (short)     & CFMI (ours, short)\\\midrule 
airfoil          & $0.031 (0.004)$  & $0.021 (0.001)$  & $0.027 (0.002)$  & $0.006 (0.001)$  & $0.002 (0.000)$  & $0.002 (0.001)$  & $0.027 (0.003)$  & $0.002 (0.000)$  & $\underline{\bm{0.001 (0.000)}}$& $0.001 (0.000)$  & $0.002 (0.001)$  & $0.002 (0.000)$  \\ 
banknote         & $0.036 (0.003)$  & $0.004 (0.001)$  & $0.016 (0.001)$  & $0.004 (0.001)$  & $0.001 (0.000)$  & $0.004 (0.002)$  & $0.015 (0.002)$  & $0.007 (0.006)$  & $\bm{0.001 (0.000)}$& $\underline{\bm{0.001 (0.000)}}$& $0.001 (0.000)$  & $0.001 (0.000)$  \\ 
blood            & $0.032 (0.004)$  & $0.007 (0.002)$  & $0.013 (0.003)$  & $0.003 (0.000)$  & $0.003 (0.000)$  & $0.010 (0.004)$  & $0.027 (0.007)$  & $0.007 (0.001)$  & $\underline{\bm{0.001 (0.000)}}$& $\bm{0.001 (0.000)}$& $\bm{0.001 (0.000)}$& $\bm{0.001 (0.000)}$\\ 
cancer           & $0.074 (0.003)$  & $0.010 (0.000)$  & $0.007 (0.000)$  & $\underline{\bm{0.002 (0.000)}}$& $0.005 (0.000)$  & $0.031 (0.001)$  & $0.027 (0.005)$  & $0.010 (0.000)$  & $0.010 (0.001)$  & $0.010 (0.000)$  & $0.005 (0.001)$  & $0.004 (0.000)$  \\ 
california       & $0.023 (0.003)$  & $0.010 (0.002)$  & $0.011 (0.002)$  & $0.003 (0.001)$  & $0.002 (0.000)$  & $0.016 (0.007)$  & $0.027 (0.002)$  & $0.002 (0.001)$  & $0.000 (0.000)$  & $\underline{\bm{0.000 (0.000)}}$& $0.001 (0.000)$  & $0.001 (0.000)$  \\ 
climate          & $0.024 (0.001)$  & $0.018 (0.001)$  & $0.024 (0.001)$  & $\underline{\bm{0.006 (0.000)}}$& $0.019 (0.001)$  & $0.012 (0.001)$  & $0.021 (0.001)$  & $0.023 (0.001)$  & $0.017 (0.001)$  & $0.014 (0.001)$  & $0.008 (0.001)$  & $0.008 (0.000)$  \\
\bottomrule
\end{tabular}}
\end{table}

\begin{table}[H]
\centering
\caption{Gaussian MMD results: mean and standard error ($\times 10^{3}$). MCAR 25\% missingness. }
\resizebox{1.\linewidth}{!}{\begin{tabular}{@{}lcccccccccccc@{}}\toprule
                 & Mean             & OT imputer       & SoftImpute       & MICE             & MissForest       & ForestDiffusion  & GAIN             & HyperImpute      & CSDI             & CFMI (ours)      & CSDI (short)     & CFMI (ours, short)\\\midrule 
airfoil          & $0.088 (0.007)$  & $0.095 (0.009)$  & $0.076 (0.007)$  & $0.013 (0.001)$  & $0.010 (0.002)$  & $0.023 (0.002)$  & $0.397 (0.101)$  & $0.013 (0.004)$  & $0.005 (0.001)$  & $\underline{\bm{0.004 (0.000)}}$& $0.014 (0.003)$  & $0.020 (0.006)$  \\ 
banknote         & $0.237 (0.013)$  & $0.033 (0.002)$  & $0.151 (0.007)$  & $0.010 (0.002)$  & $\bm{0.005 (0.002)}$& $0.012 (0.001)$  & $0.376 (0.179)$  & $0.008 (0.001)$  & $0.007 (0.000)$  & $\underline{\bm{0.005 (0.000)}}$& $0.012 (0.003)$  & $0.010 (0.001)$  \\ 
blood            & $0.612 (0.037)$  & $0.074 (0.002)$  & $0.181 (0.011)$  & $0.050 (0.004)$  & $0.039 (0.003)$  & $0.053 (0.006)$  & $1.243 (0.249)$  & $0.088 (0.011)$  & $\underline{\bm{0.021 (0.002)}}$& $0.028 (0.006)$  & $\bm{0.022 (0.001)}$& $0.031 (0.004)$  \\ 
cancer           & $0.010 (0.000)$  & $0.002 (0.000)$  & $\underline{\bm{0.001 (0.000)}}$& $0.001 (0.000)$  & $0.001 (0.000)$  & $0.007 (0.001)$  & $0.004 (0.000)$  & $0.002 (0.000)$  & $0.002 (0.000)$  & $0.003 (0.001)$  & $0.004 (0.001)$  & $0.002 (0.000)$  \\ 
california       & $0.066 (0.001)$  & $0.064 (0.001)$  & $0.053 (0.001)$  & $0.005 (0.002)$  & $0.008 (0.000)$  & $0.289 (0.135)$  & $0.454 (0.275)$  & $0.006 (0.000)$  & $0.001 (0.000)$  & $\underline{\bm{0.000 (0.000)}}$& $0.003 (0.000)$  & $0.002 (0.000)$  \\ 
climate          & $0.007 (0.000)$  & $0.007 (0.001)$  & $\bm{0.006 (0.000)}$& $0.007 (0.000)$  & $0.007 (0.001)$  & $0.019 (0.003)$  & $0.010 (0.001)$  & $0.007 (0.001)$  & $\bm{0.006 (0.001)}$& $\underline{\bm{0.005 (0.001)}}$& $0.008 (0.001)$  & $0.010 (0.001)$  \\
\bottomrule
\end{tabular}}
\end{table}

\begin{table}[H]
\centering
\caption{Gaussian MMD results: mean and standard error ($\times 10^{3}$). MCAR 50\% missingness. }
\resizebox{1.\linewidth}{!}{\begin{tabular}{@{}lcccccccccccc@{}}\toprule
                 & Mean             & OT imputer       & SoftImpute       & MICE             & MissForest       & ForestDiffusion  & GAIN             & HyperImpute      & CSDI             & CFMI (ours)      & CSDI (short)     & CFMI (ours, short)\\\midrule 
airfoil          & $0.298 (0.014)$  & $0.429 (0.013)$  & $0.281 (0.013)$  & $0.081 (0.006)$  & $0.076 (0.007)$  & $0.093 (0.015)$  & $0.888 (0.213)$  & $0.074 (0.015)$  & $0.033 (0.015)$  & $\underline{\bm{0.022 (0.004)}}$& $0.049 (0.006)$  & $0.090 (0.019)$  \\ 
banknote         & $0.817 (0.023)$  & $0.191 (0.025)$  & $0.581 (0.024)$  & $0.054 (0.012)$  & $0.033 (0.013)$  & $0.079 (0.032)$  & $1.137 (0.243)$  & $0.089 (0.024)$  & $0.033 (0.013)$  & $\underline{\bm{0.022 (0.009)}}$& $0.103 (0.039)$  & $0.054 (0.019)$  \\ 
blood            & $1.926 (0.026)$  & $0.390 (0.066)$  & $0.763 (0.058)$  & $0.154 (0.042)$  & $0.325 (0.135)$  & $0.356 (0.127)$  & $4.707 (0.872)$  & $0.456 (0.049)$  & $\underline{\bm{0.090 (0.028)}}$& $0.142 (0.053)$  & $0.123 (0.024)$  & $0.167 (0.052)$  \\ 
cancer           & $0.035 (0.002)$  & $0.009 (0.002)$  & $\underline{\bm{0.005 (0.001)}}$& $0.006 (0.001)$  & $0.008 (0.001)$  & $0.043 (0.003)$  & $0.385 (0.099)$  & $0.008 (0.001)$  & $0.015 (0.003)$  & $0.016 (0.004)$  & $0.131 (0.008)$  & $0.020 (0.004)$  \\ 
california       & $0.255 (0.004)$  & $0.333 (0.004)$  & $0.234 (0.005)$  & $0.017 (0.003)$  & $0.066 (0.019)$  & $0.299 (0.093)$  & $1.799 (0.197)$  & $0.033 (0.006)$  & $0.006 (0.001)$  & $\underline{\bm{0.003 (0.000)}}$& $0.012 (0.001)$  & $0.014 (0.002)$  \\ 
climate          & $0.017 (0.002)$  & $0.022 (0.002)$  & $\underline{\bm{0.012 (0.001)}}$& $0.017 (0.001)$  & $0.022 (0.006)$  & $0.084 (0.009)$  & $0.110 (0.023)$  & $0.016 (0.001)$  & $0.017 (0.001)$  & $0.018 (0.001)$  & $0.019 (0.004)$  & $0.027 (0.001)$  \\
\bottomrule
\end{tabular}}
\end{table}

\begin{table}[H]
\centering
\caption{Gaussian MMD results: mean and standard error ($\times 10^{3}$). MCAR 75\% missingness. }
\resizebox{1.\linewidth}{!}{\begin{tabular}{@{}lcccccccccccc@{}}\toprule
                 & Mean             & OT imputer       & SoftImpute       & MICE             & MissForest       & ForestDiffusion  & GAIN             & HyperImpute      & CSDI             & CFMI (ours)      & CSDI (short)     & CFMI (ours, short)\\\midrule 
airfoil          & $0.556 (0.024)$  & $1.078 (0.047)$  & $0.531 (0.020)$  & $0.379 (0.021)$  & $0.297 (0.071)$  & $1.168 (0.542)$  & $1.503 (0.176)$  & $\bm{0.238 (0.007)}$& $\bm{0.191 (0.058)}$& $\underline{\bm{0.181 (0.083)}}$& $\bm{0.259 (0.030)}$& $0.380 (0.110)$  \\ 
banknote         & $1.533 (0.029)$  & $0.518 (0.019)$  & $1.184 (0.076)$  & $\underline{\bm{0.095 (0.027)}}$& $0.128 (0.005)$  & $0.931 (0.469)$  & $5.887 (1.980)$  & $0.325 (0.031)$  & $0.177 (0.036)$  & $0.153 (0.035)$  & $0.273 (0.037)$  & $0.176 (0.017)$  \\ 
blood            & $3.543 (0.341)$  & $1.001 (0.040)$  & $2.074 (0.321)$  & $\underline{\bm{0.412 (0.097)}}$& $1.282 (0.308)$  & $6.898 (2.709)$  & $9.169 (4.484)$  & $1.076 (0.054)$  & $1.544 (0.228)$  & $1.130 (0.443)$  & $0.691 (0.219)$  & $0.518 (0.066)$  \\ 
cancer           & $0.079 (0.006)$  & $0.036 (0.006)$  & $\underline{\bm{0.016 (0.001)}}$& $0.023 (0.001)$  & $0.062 (0.010)$  & $0.173 (0.018)$  & $0.720 (0.189)$  & $0.031 (0.001)$  & $0.109 (0.033)$  & $0.077 (0.012)$  & $26.618 (1.470)$ & $0.167 (0.043)$  \\ 
california       & $0.553 (0.013)$  & $0.637 (0.003)$  & $0.554 (0.015)$  & $0.033 (0.003)$  & $0.331 (0.026)$  & $0.528 (0.017)$  & $41.075 (17.530)$& $0.062 (0.012)$  & $\underline{\bm{0.018 (0.005)}}$& $0.035 (0.007)$  & $1.154 (0.035)$  & $0.159 (0.019)$  \\ 
climate          & $0.055 (0.009)$  & $0.096 (0.021)$  & $\underline{\bm{0.023 (0.002)}}$& $0.052 (0.009)$  & $0.136 (0.007)$  & $0.237 (0.027)$  & $1.403 (0.058)$  & $0.065 (0.017)$  & $0.088 (0.014)$  & $0.125 (0.015)$  & $1.365 (0.255)$  & $0.134 (0.007)$  \\
\bottomrule
\end{tabular}}
\end{table}

\begin{table}[H]
\centering
\caption{Gaussian MMD results: mean and standard error ($\times 10^{3}$). MAR 25\% missingness. }
\resizebox{1.\linewidth}{!}{\begin{tabular}{@{}lcccccccccccc@{}}\toprule
                 & Mean             & OT imputer       & SoftImpute       & MICE             & MissForest       & ForestDiffusion  & GAIN             & HyperImpute      & CSDI             & CFMI (ours)      & CSDI (short)     & CFMI (ours, short)\\\midrule 
airfoil          & $0.420 (0.160)$  & $0.346 (0.028)$  & $0.287 (0.079)$  & $0.055 (0.005)$  & $0.026 (0.012)$  & $0.061 (0.020)$  & $0.551 (0.114)$  & $0.016 (0.005)$  & $\underline{\bm{0.006 (0.002)}}$& $\bm{0.008 (0.004)}$& $0.028 (0.013)$  & $0.021 (0.005)$  \\ 
banknote         & $1.192 (0.261)$  & $0.122 (0.040)$  & $0.254 (0.039)$  & $0.139 (0.052)$  & $\bm{0.020 (0.010)}$& $0.213 (0.127)$  & $0.424 (0.046)$  & $0.395 (0.346)$  & $\bm{0.019 (0.005)}$& $\underline{\bm{0.018 (0.003)}}$& $0.050 (0.017)$  & $0.026 (0.007)$  \\ 
blood            & $2.132 (0.533)$  & $0.412 (0.144)$  & $0.546 (0.150)$  & $0.143 (0.072)$  & $0.150 (0.036)$  & $0.993 (0.433)$  & $2.074 (0.735)$  & $0.340 (0.086)$  & $\bm{0.080 (0.036)}$& $\bm{0.077 (0.032)}$& $\underline{\bm{0.076 (0.016)}}$& $\bm{0.079 (0.015)}$\\ 
cancer           & $0.219 (0.017)$  & $0.016 (0.001)$  & $0.007 (0.000)$  & $\underline{\bm{0.002 (0.000)}}$& $0.008 (0.001)$  & $0.129 (0.005)$  & $0.098 (0.021)$  & $0.014 (0.001)$  & $0.028 (0.003)$  & $0.028 (0.002)$  & $0.015 (0.005)$  & $0.008 (0.001)$  \\ 
california       & $0.237 (0.102)$  & $0.158 (0.037)$  & $0.091 (0.027)$  & $0.016 (0.005)$  & $0.029 (0.006)$  & $0.391 (0.243)$  & $0.539 (0.065)$  & $0.043 (0.019)$  & $0.006 (0.002)$  & $\underline{\bm{0.003 (0.000)}}$& $0.012 (0.004)$  & $0.012 (0.003)$  \\ 
climate          & $0.006 (0.001)$  & $0.010 (0.002)$  & $\underline{\bm{0.005 (0.001)}}$& $0.009 (0.001)$  & $0.008 (0.001)$  & $0.024 (0.002)$  & $0.022 (0.003)$  & $0.007 (0.001)$  & $0.007 (0.002)$  & $0.009 (0.002)$  & $0.011 (0.002)$  & $0.015 (0.003)$  \\
\bottomrule
\end{tabular}}
\end{table}

\begin{table}[H]
\centering
\caption{Laplacian MMD results: mean and standard error ($\times 10^{3}$). MCAR 25\% missingness. }
\resizebox{1.\linewidth}{!}{\begin{tabular}{@{}lcccccccccccc@{}}\toprule
                 & Mean             & OT imputer       & SoftImpute       & MICE             & MissForest       & ForestDiffusion  & GAIN             & HyperImpute      & CSDI             & CFMI (ours)      & CSDI (short)     & CFMI (ours, short)\\\midrule 
airfoil          & $1.845 (0.038)$  & $1.146 (0.043)$  & $1.642 (0.073)$  & $0.264 (0.013)$  & $0.137 (0.004)$  & $0.115 (0.012)$  & $1.920 (0.222)$  & $0.118 (0.009)$  & $\underline{\bm{0.046 (0.001)}}$& $0.048 (0.003)$  & $0.129 (0.004)$  & $0.122 (0.002)$  \\ 
banknote         & $2.462 (0.059)$  & $0.217 (0.009)$  & $1.431 (0.025)$  & $0.226 (0.010)$  & $0.067 (0.010)$  & $0.098 (0.004)$  & $1.866 (0.482)$  & $0.092 (0.003)$  & $0.049 (0.002)$  & $\underline{\bm{0.044 (0.002)}}$& $0.079 (0.009)$  & $0.080 (0.004)$  \\ 
blood            & $3.578 (0.077)$  & $0.553 (0.036)$  & $1.643 (0.056)$  & $0.415 (0.020)$  & $0.317 (0.014)$  & $0.366 (0.006)$  & $2.939 (0.432)$  & $0.481 (0.033)$  & $\underline{\bm{0.130 (0.019)}}$& $\bm{0.145 (0.023)}$& $0.191 (0.015)$  & $0.224 (0.013)$  \\ 
cancer           & $0.784 (0.004)$  & $0.118 (0.003)$  & $0.103 (0.002)$  & $\underline{\bm{0.056 (0.001)}}$& $0.068 (0.001)$  & $0.116 (0.005)$  & $0.215 (0.006)$  & $0.163 (0.003)$  & $0.087 (0.002)$  & $0.087 (0.004)$  & $0.085 (0.004)$  & $0.072 (0.001)$  \\ 
california       & $1.562 (0.017)$  & $0.512 (0.007)$  & $0.843 (0.004)$  & $0.226 (0.044)$  & $0.089 (0.002)$  & $1.785 (0.301)$  & $2.050 (0.491)$  & $0.103 (0.001)$  & $0.010 (0.000)$  & $\underline{\bm{0.006 (0.000)}}$& $0.023 (0.001)$  & $0.021 (0.001)$  \\ 
climate          & $0.572 (0.007)$  & $0.383 (0.011)$  & $0.563 (0.006)$  & $\underline{\bm{0.145 (0.000)}}$& $0.504 (0.010)$  & $0.288 (0.014)$  & $0.501 (0.002)$  & $0.557 (0.007)$  & $0.438 (0.006)$  & $0.345 (0.012)$  & $0.196 (0.004)$  & $0.163 (0.004)$  \\
\bottomrule
\end{tabular}}
\end{table}

\begin{table}[H]
\centering
\caption{Laplacian MMD results: mean and standard error ($\times 10^{3}$). MCAR 50\% missingness. }
\resizebox{1.\linewidth}{!}{\begin{tabular}{@{}lcccccccccccc@{}}\toprule
                 & Mean             & OT imputer       & SoftImpute       & MICE             & MissForest       & ForestDiffusion  & GAIN             & HyperImpute      & CSDI             & CFMI (ours)      & CSDI (short)     & CFMI (ours, short)\\\midrule 
airfoil          & $7.561 (0.133)$  & $5.205 (0.108)$  & $7.249 (0.132)$  & $1.184 (0.046)$  & $0.701 (0.048)$  & $0.479 (0.056)$  & $6.858 (0.969)$  & $0.611 (0.028)$  & $0.207 (0.042)$  & $\underline{\bm{0.188 (0.012)}}$& $0.425 (0.064)$  & $0.412 (0.010)$  \\ 
banknote         & $9.117 (0.123)$  & $1.178 (0.059)$  & $6.191 (0.168)$  & $0.750 (0.068)$  & $0.315 (0.011)$  & $0.371 (0.042)$  & $6.783 (0.333)$  & $0.533 (0.080)$  & $0.194 (0.027)$  & $\underline{\bm{0.169 (0.018)}}$& $0.409 (0.069)$  & $0.259 (0.022)$  \\ 
blood            & $11.969 (0.144)$ & $2.139 (0.038)$  & $6.638 (0.216)$  & $1.214 (0.075)$  & $1.327 (0.166)$  & $1.182 (0.131)$  & $11.476 (0.914)$ & $3.407 (0.060)$  & $\underline{\bm{0.411 (0.018)}}$& $0.476 (0.064)$  & $0.669 (0.030)$  & $0.699 (0.045)$  \\ 
cancer           & $3.163 (0.043)$  & $0.425 (0.029)$  & $0.536 (0.045)$  & $\underline{\bm{0.143 (0.004)}}$& $0.212 (0.010)$  & $0.574 (0.048)$  & $3.116 (0.507)$  & $0.570 (0.034)$  & $0.410 (0.028)$  & $0.429 (0.031)$  & $0.917 (0.125)$  & $0.337 (0.015)$  \\ 
california       & $7.197 (0.070)$  & $3.811 (0.044)$  & $4.936 (0.050)$  & $0.787 (0.104)$  & $0.441 (0.020)$  & $2.637 (0.652)$  & $9.165 (0.460)$  & $0.483 (0.024)$  & $0.048 (0.001)$  & $\underline{\bm{0.032 (0.003)}}$& $0.108 (0.005)$  & $0.108 (0.004)$  \\ 
climate          & $2.511 (0.008)$  & $1.155 (0.004)$  & $2.439 (0.007)$  & $\underline{\bm{0.227 (0.008)}}$& $1.183 (0.053)$  & $1.412 (0.031)$  & $2.600 (0.127)$  & $1.986 (0.007)$  & $1.359 (0.006)$  & $1.244 (0.007)$  & $0.548 (0.015)$  & $0.318 (0.023)$  \\
\bottomrule
\end{tabular}}
\end{table}

\begin{table}[H]
\centering
\caption{Laplacian MMD results: mean and standard error ($\times 10^{3}$). MCAR 75\% missingness. }
\resizebox{1.\linewidth}{!}{\begin{tabular}{@{}lcccccccccccc@{}}\toprule
                 & Mean             & OT imputer       & SoftImpute       & MICE             & MissForest       & ForestDiffusion  & GAIN             & HyperImpute      & CSDI             & CFMI (ours)      & CSDI (short)     & CFMI (ours, short)\\\midrule 
airfoil          & $16.771 (0.140)$ & $13.523 (0.413)$ & $16.190 (0.436)$ & $4.579 (0.139)$  & $2.209 (0.238)$  & $4.332 (1.503)$  & $14.622 (0.703)$ & $5.911 (0.172)$  & $\bm{1.004 (0.099)}$& $\underline{\bm{0.955 (0.216)}}$& $1.657 (0.017)$  & $1.287 (0.219)$  \\ 
banknote         & $18.224 (0.323)$ & $3.630 (0.277)$  & $13.916 (0.816)$ & $1.395 (0.028)$  & $1.423 (0.112)$  & $2.946 (1.002)$  & $18.326 (3.290)$ & $3.997 (0.307)$  & $0.910 (0.079)$  & $\underline{\bm{0.778 (0.053)}}$& $1.262 (0.172)$  & $0.879 (0.040)$  \\ 
blood            & $23.310 (0.261)$ & $4.875 (0.563)$  & $15.893 (0.841)$ & $\bm{2.490 (0.238)}$& $4.819 (0.349)$  & $11.409 (5.023)$ & $20.851 (5.092)$ & $7.399 (0.084)$  & $3.760 (0.405)$  & $3.016 (0.703)$  & $\bm{2.646 (0.374)}$& $\underline{\bm{2.305 (0.497)}}$\\ 
cancer           & $8.490 (0.091)$  & $1.311 (0.042)$  & $2.326 (0.054)$  & $\underline{\bm{0.431 (0.009)}}$& $0.691 (0.032)$  & $1.844 (0.088)$  & $5.140 (1.013)$  & $1.357 (0.072)$  & $1.743 (0.050)$  & $1.937 (0.034)$  & $49.929 (2.173)$ & $1.297 (0.143)$  \\ 
california       & $18.940 (0.099)$ & $13.778 (0.098)$ & $15.714 (0.352)$ & $1.378 (0.030)$  & $1.642 (0.115)$  & $1.669 (0.072)$  & $57.738 (12.064)$& $1.019 (0.131)$  & $\underline{\bm{0.160 (0.017)}}$& $\bm{0.165 (0.011)}$& $1.528 (0.009)$  & $0.517 (0.020)$  \\ 
climate          & $7.902 (0.028)$  & $3.072 (0.119)$  & $7.182 (0.014)$  & $\underline{\bm{0.422 (0.043)}}$& $1.971 (0.063)$  & $4.594 (0.306)$  & $12.668 (0.290)$ & $3.491 (0.078)$  & $3.270 (0.167)$  & $4.424 (0.085)$  & $4.075 (0.694)$  & $0.984 (0.028)$  \\
\bottomrule
\end{tabular}}
\end{table}

\begin{table}[H]
\centering
\caption{Laplacian MMD results: mean and standard error ($\times 10^{3}$). MAR 25\% missingness. }
\resizebox{1.\linewidth}{!}{\begin{tabular}{@{}lcccccccccccc@{}}\toprule
                 & Mean             & OT imputer       & SoftImpute       & MICE             & MissForest       & ForestDiffusion  & GAIN             & HyperImpute      & CSDI             & CFMI (ours)      & CSDI (short)     & CFMI (ours, short)\\\midrule 
airfoil          & $3.467 (0.303)$  & $2.265 (0.097)$  & $3.067 (0.123)$  & $0.639 (0.109)$  & $0.266 (0.047)$  & $0.224 (0.064)$  & $2.813 (0.273)$  & $0.188 (0.018)$  & $\underline{\bm{0.064 (0.010)}}$& $0.077 (0.022)$  & $0.206 (0.067)$  & $0.179 (0.039)$  \\ 
banknote         & $5.289 (0.297)$  & $0.604 (0.164)$  & $2.551 (0.219)$  & $0.599 (0.125)$  & $0.134 (0.015)$  & $0.545 (0.282)$  & $2.320 (0.348)$  & $0.905 (0.667)$  & $\bm{0.109 (0.023)}$& $\underline{\bm{0.103 (0.012)}}$& $0.209 (0.035)$  & $0.142 (0.011)$  \\ 
blood            & $5.752 (0.418)$  & $1.203 (0.260)$  & $2.729 (0.497)$  & $0.515 (0.026)$  & $0.684 (0.075)$  & $1.387 (0.462)$  & $4.458 (0.887)$  & $1.466 (0.239)$  & $\underline{\bm{0.224 (0.044)}}$& $\bm{0.248 (0.042)}$& $\bm{0.232 (0.024)}$& $0.272 (0.035)$  \\ 
cancer           & $1.581 (0.051)$  & $0.222 (0.007)$  & $0.156 (0.004)$  & $\underline{\bm{0.056 (0.001)}}$& $0.109 (0.006)$  & $0.635 (0.019)$  & $0.567 (0.091)$  & $0.225 (0.010)$  & $0.220 (0.011)$  & $0.216 (0.007)$  & $0.114 (0.009)$  & $0.095 (0.004)$  \\ 
california       & $2.344 (0.243)$  & $0.994 (0.240)$  & $1.200 (0.207)$  & $0.304 (0.098)$  & $0.161 (0.028)$  & $1.598 (0.703)$  & $2.553 (0.160)$  & $0.224 (0.056)$  & $0.020 (0.003)$  & $\underline{\bm{0.011 (0.001)}}$& $0.047 (0.011)$  & $0.058 (0.004)$  \\ 
climate          & $0.670 (0.020)$  & $0.492 (0.025)$  & $0.660 (0.019)$  & $\underline{\bm{0.148 (0.005)}}$& $0.524 (0.024)$  & $0.328 (0.019)$  & $0.570 (0.035)$  & $0.631 (0.020)$  & $0.479 (0.015)$  & $0.383 (0.017)$  & $0.228 (0.018)$  & $0.221 (0.006)$  \\
\bottomrule
\end{tabular}}
\end{table}

\begin{table}[H]
\centering
\caption{Classifier 1-AUROC results: mean and standard error. MCAR 25\% missingness. }
\resizebox{1.\linewidth}{!}{\begin{tabular}{@{}lcccccccccccc@{}}\toprule
                 & Mean             & OT imputer       & SoftImpute       & MICE             & MissForest       & ForestDiffusion  & GAIN             & HyperImpute      & CSDI             & CFMI (ours)      & CSDI (short)     & CFMI (ours, short)\\\midrule 
airfoil          & ---              & ---              & ---              & ---              & ---              & ---              & ---              & ---              & ---              & ---              & ---              & ---              \\ 
banknote         & $0.054 (0.005)$  & $0.000 (0.000)$  & $0.015 (0.003)$  & $0.001 (0.000)$  & $0.000 (0.000)$  & $0.002 (0.000)$  & $0.011 (0.002)$  & $\underline{\bm{0.000 (0.000)}}$& $0.000 (0.000)$  & $0.001 (0.000)$  & $0.001 (0.000)$  & $0.002 (0.000)$  \\ 
blood            & $0.270 (0.007)$  & $0.213 (0.004)$  & $0.249 (0.005)$  & $0.252 (0.002)$  & $0.226 (0.006)$  & $0.239 (0.005)$  & $\underline{\bm{0.186 (0.015)}}$& $0.227 (0.005)$  & $0.234 (0.005)$  & $0.234 (0.006)$  & $0.250 (0.012)$  & $0.240 (0.010)$  \\ 
cancer           & $0.002 (0.001)$  & $0.002 (0.000)$  & $0.002 (0.000)$  & $0.002 (0.000)$  & $0.002 (0.000)$  & $\underline{\bm{0.000 (0.000)}}$& $0.002 (0.000)$  & $0.002 (0.000)$  & $0.001 (0.000)$  & $0.001 (0.000)$  & $0.002 (0.000)$  & $0.002 (0.000)$  \\ 
california       & ---              & ---              & ---              & ---              & ---              & ---              & ---              & ---              & ---              & ---              & ---              & ---              \\ 
climate          & $0.066 (0.004)$  & $0.064 (0.008)$  & $0.065 (0.005)$  & $0.107 (0.005)$  & $0.025 (0.003)$  & $\underline{\bm{0.019 (0.004)}}$& $0.033 (0.003)$  & $0.040 (0.003)$  & $0.048 (0.005)$  & $0.030 (0.003)$  & $0.067 (0.007)$  & $0.090 (0.018)$  \\
\bottomrule
\end{tabular}}
\end{table}

\begin{table}[H]
\centering
\caption{Classifier 1-AUROC results: mean and standard error. MCAR 50\% missingness. }
\resizebox{1.\linewidth}{!}{\input{figures/uci/umis50_1maucroc_raw_table}}
\end{table}

\begin{table}[H]
\centering
\caption{Classifier 1-AUROC results: mean and standard error. MCAR 75\% missingness. }
\resizebox{1.\linewidth}{!}{\input{figures/uci/umis75_1maucroc_raw_table}}
\end{table}

\begin{table}[H]
\centering
\caption{Classifier 1-AUROC results: mean and standard error. MAR 25\% missingness. }
\resizebox{1.\linewidth}{!}{\input{figures/uci/marlogmis25_1maucroc_raw_table}}
\end{table}

\begin{table}[H]
\centering
\caption{Regression CRPS results: mean and standard error. MCAR 25\% missingness. }
\resizebox{1.\linewidth}{!}{\input{figures/uci/umis25_regcrps_raw_table}}
\end{table}

\begin{table}[H]
\centering
\caption{Regression CRPS results: mean and standard error. MCAR 50\% missingness. }
\resizebox{1.\linewidth}{!}{\input{figures/uci/umis50_regcrps_raw_table}}
\end{table}

\begin{table}[H]
\centering
\caption{Regression CRPS results: mean and standard error. MCAR 75\% missingness. }
\resizebox{1.\linewidth}{!}{\input{figures/uci/umis75_regcrps_raw_table}}
\end{table}

\begin{table}[H]
\centering
\caption{Regression CRPS results: mean and standard error. MAR 25\% missingness. }
\resizebox{1.\linewidth}{!}{\input{figures/uci/marlogmis25_regcrps_raw_table}}
\end{table}

\begin{table}[H]
\centering
\caption{Regression parameter percent bias results: mean and standard error. MCAR 25\% missingness. }
\resizebox{1.\linewidth}{!}{\input{figures/uci/umis25_regpb_raw_table}}
\end{table}

\begin{table}[H]
\centering
\caption{Regression parameter percent bias results: mean and standard error. MCAR 50\% missingness. }
\resizebox{1.\linewidth}{!}{\input{figures/uci/umis50_regpb_raw_table}}
\end{table}

\begin{table}[H]
\centering
\caption{Regression parameter percent bias results: mean and standard error. MCAR 75\% missingness. }
\resizebox{1.\linewidth}{!}{\input{figures/uci/umis75_regpb_raw_table}}
\end{table}

\begin{table}[H]
\centering
\caption{Regression parameter percent bias results: mean and standard error. MAR 25\% missingness. }
\resizebox{1.\linewidth}{!}{\input{figures/uci/marlogmis25_regpb_raw_table}}
\end{table}

\begin{table}[H]
\centering
\caption{Regression parameter 1-CR results: mean and standard error. MCAR 25\% missingness. }
\resizebox{1.\linewidth}{!}{\input{figures/uci/umis25_regparamcr_raw_table}}
\end{table}

\begin{table}[H]
\centering
\caption{Regression parameter 1-CR results: mean and standard error. MCAR 50\% missingness. }
\resizebox{1.\linewidth}{!}{\input{figures/uci/umis50_regparamcr_raw_table}}
\end{table}

\begin{table}[H]
\centering
\caption{Regression parameter 1-CR results: mean and standard error. MCAR 75\% missingness. }
\resizebox{1.\linewidth}{!}{\input{figures/uci/umis75_regparamcr_raw_table}}
\end{table}

\begin{table}[H]
\centering
\caption{Regression parameter 1-CR results: mean and standard error. MAR 25\% missingness. }
\resizebox{1.\linewidth}{!}{\input{figures/uci/marlogmis25_regparamcr_raw_table}}
\end{table}

\begin{table}[H]
\centering
\caption{Regression parameter AW results: mean and standard error. MCAR 25\% missingness. }
\resizebox{1.\linewidth}{!}{\input{figures/uci/umis25_regparamaw_raw_table}}
\end{table}

\begin{table}[H]
\centering
\caption{Regression parameter AW results: mean and standard error. MCAR 50\% missingness. }
\resizebox{1.\linewidth}{!}{\input{figures/uci/umis50_regparamaw_raw_table}}
\end{table}

\begin{table}[H]
\centering
\caption{Regression parameter AW results: mean and standard error. MCAR 75\% missingness. }
\resizebox{1.\linewidth}{!}{\input{figures/uci/umis75_regparamaw_raw_table}}
\end{table}

\begin{table}[H]
\centering
\caption{Regression parameter AW results: mean and standard error. MAR 25\% missingness. }
\resizebox{1.\linewidth}{!}{\input{figures/uci/marlogmis25_regparamaw_raw_table}}
\end{table}

\subsubsection{Raw results box plots}
\label{apx:uci-raw-boxplots}

\begin{figure}[H]
  \centering
  \includegraphics[width=0.9\linewidth]{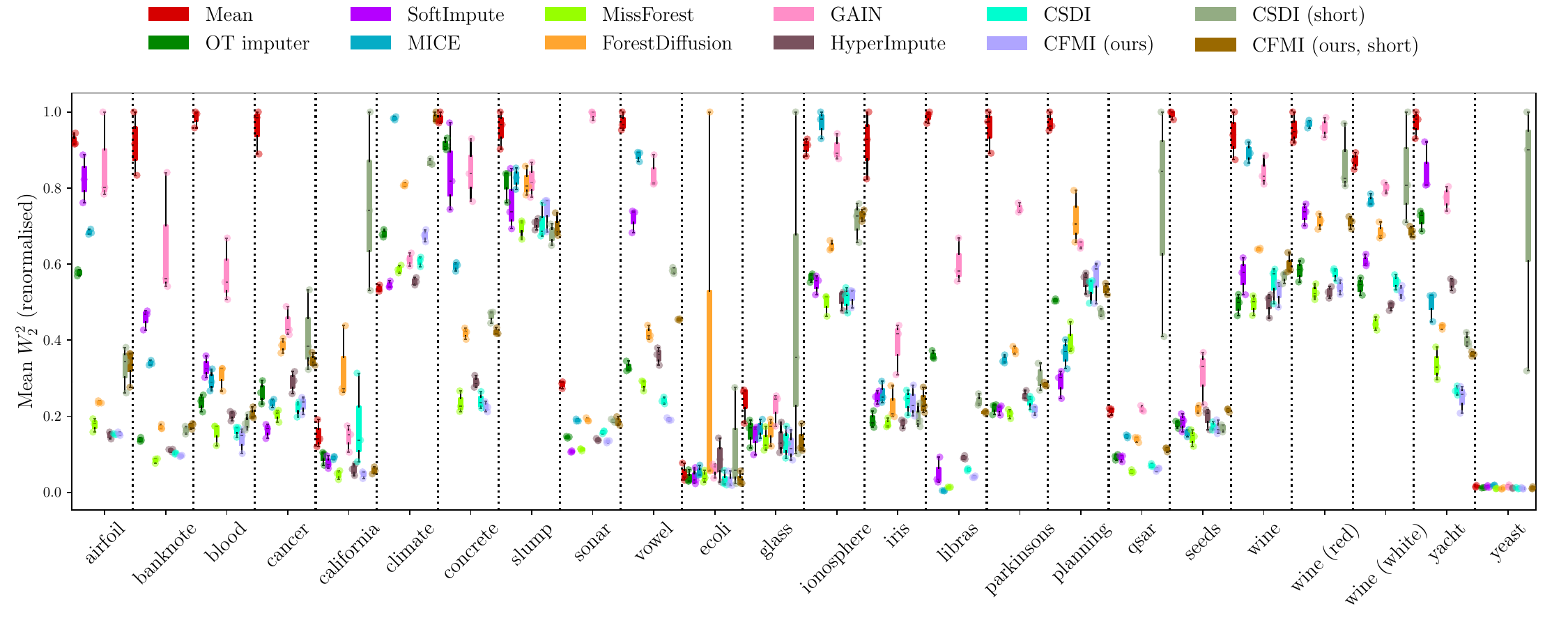}
  \caption{Wasserstein-2 results: box-plot of 3 runs. MCAR 25\% missingness.}
\end{figure}

\begin{figure}[H]
  \centering
  \includegraphics[width=0.9\linewidth]{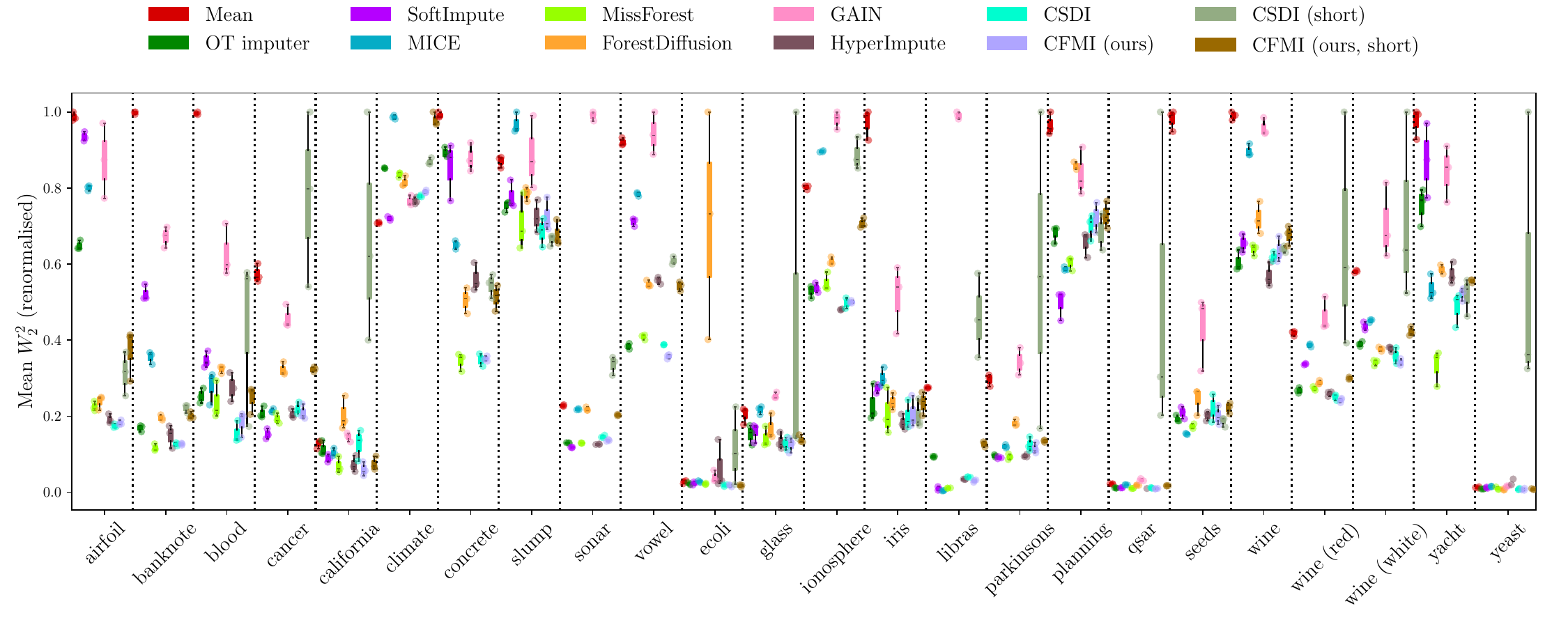}
  \caption{Wasserstein-2 results: box-plot of 3 runs. MCAR 50\% missingness.}
\end{figure}

\begin{figure}[H]
  \centering
  \includegraphics[width=0.9\linewidth]{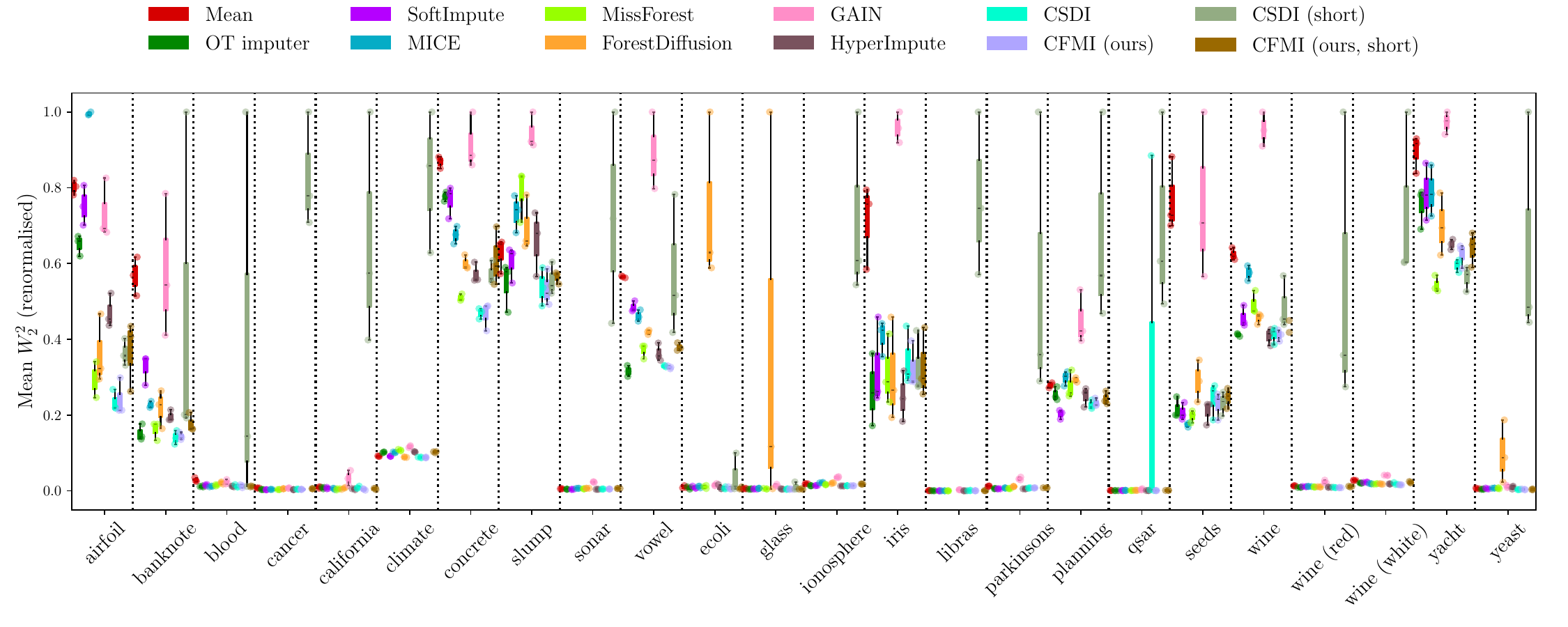}
  \caption{Wasserstein-2 results: box-plot of 3 runs. MCAR 75\% missingness.}
\end{figure}

\begin{figure}[H]
  \centering
  \includegraphics[width=0.9\linewidth]{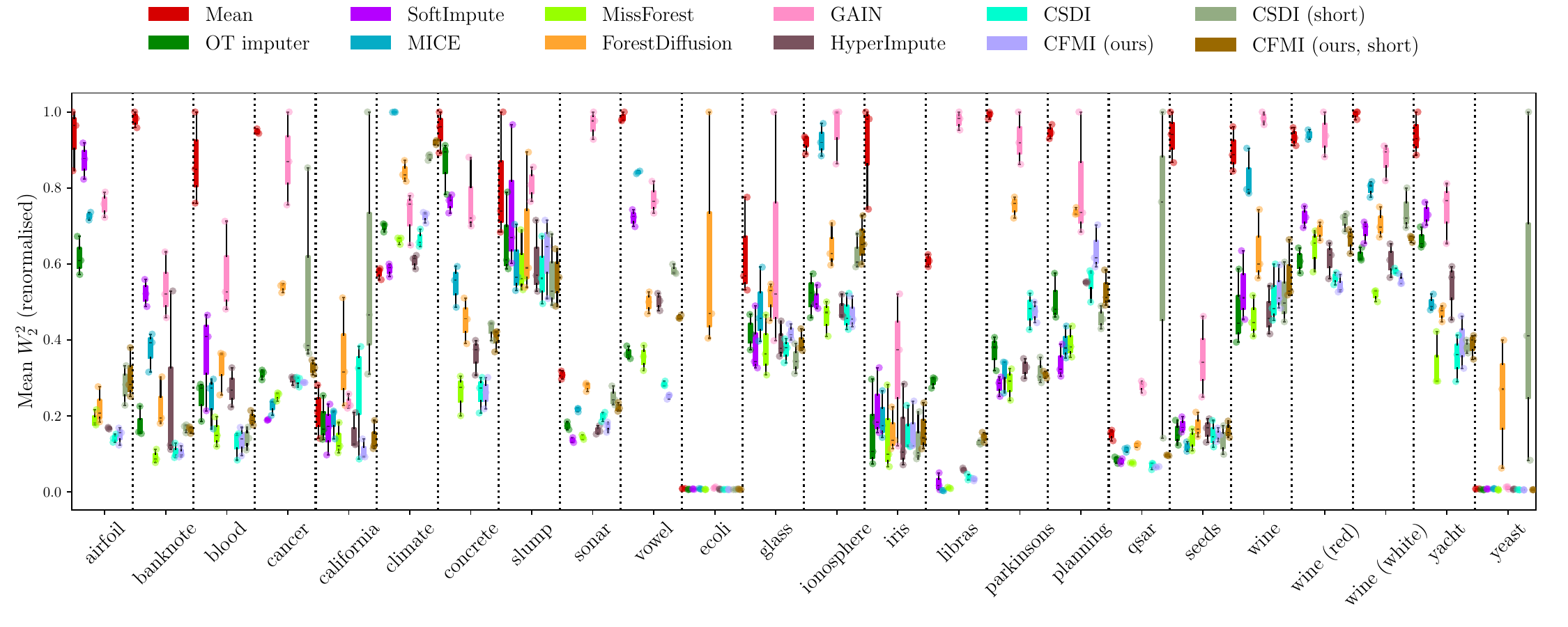}
  \caption{Wasserstein-2 results: box-plot of 3 runs. MAR 25\% missingness.}
\end{figure}

\begin{figure}[H]
  \centering
  \includegraphics[width=0.9\linewidth]{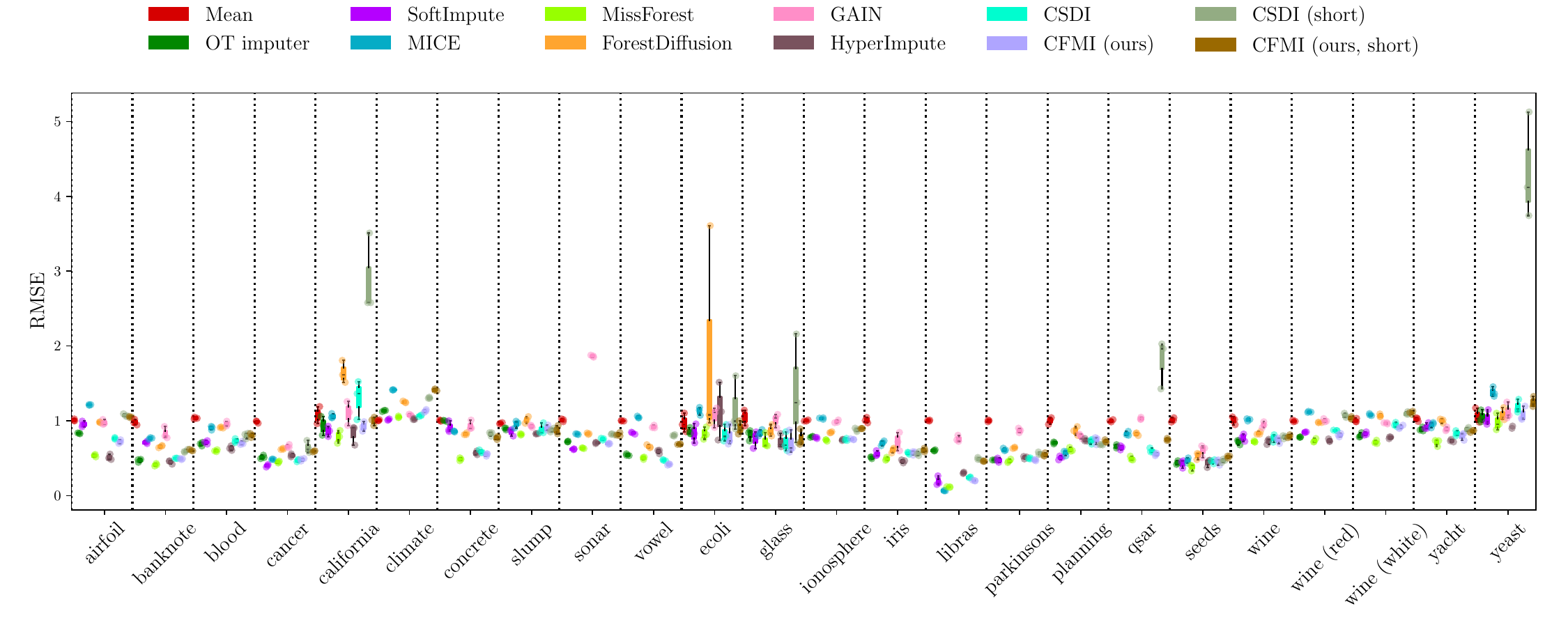}
  \caption{Average RMSE results: box-plot of 3 runs. MCAR 25\% missingness.}
\end{figure}

\begin{figure}[H]
  \centering
  \includegraphics[width=0.9\linewidth]{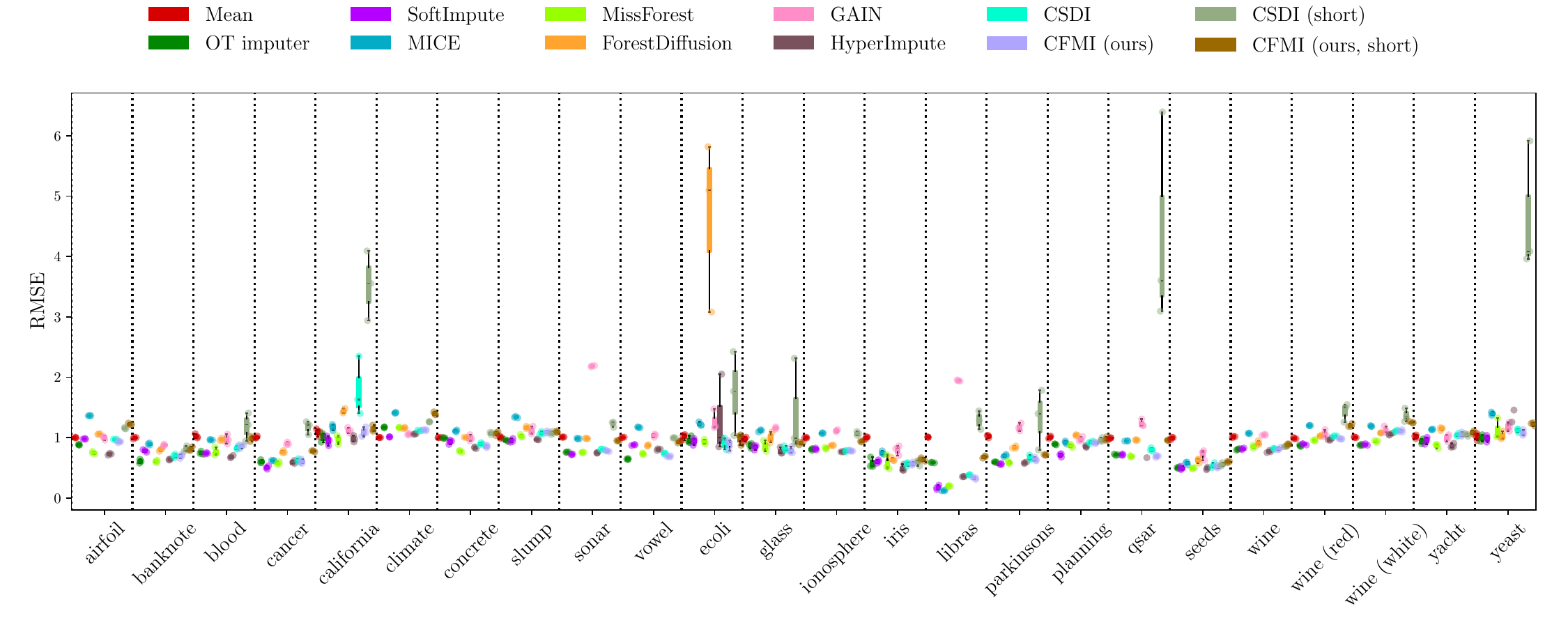}
  \caption{Average RMSE results: box-plot of 3 runs. MCAR 50\% missingness.}
\end{figure}

\begin{figure}[H]
  \centering
  \includegraphics[width=0.9\linewidth]{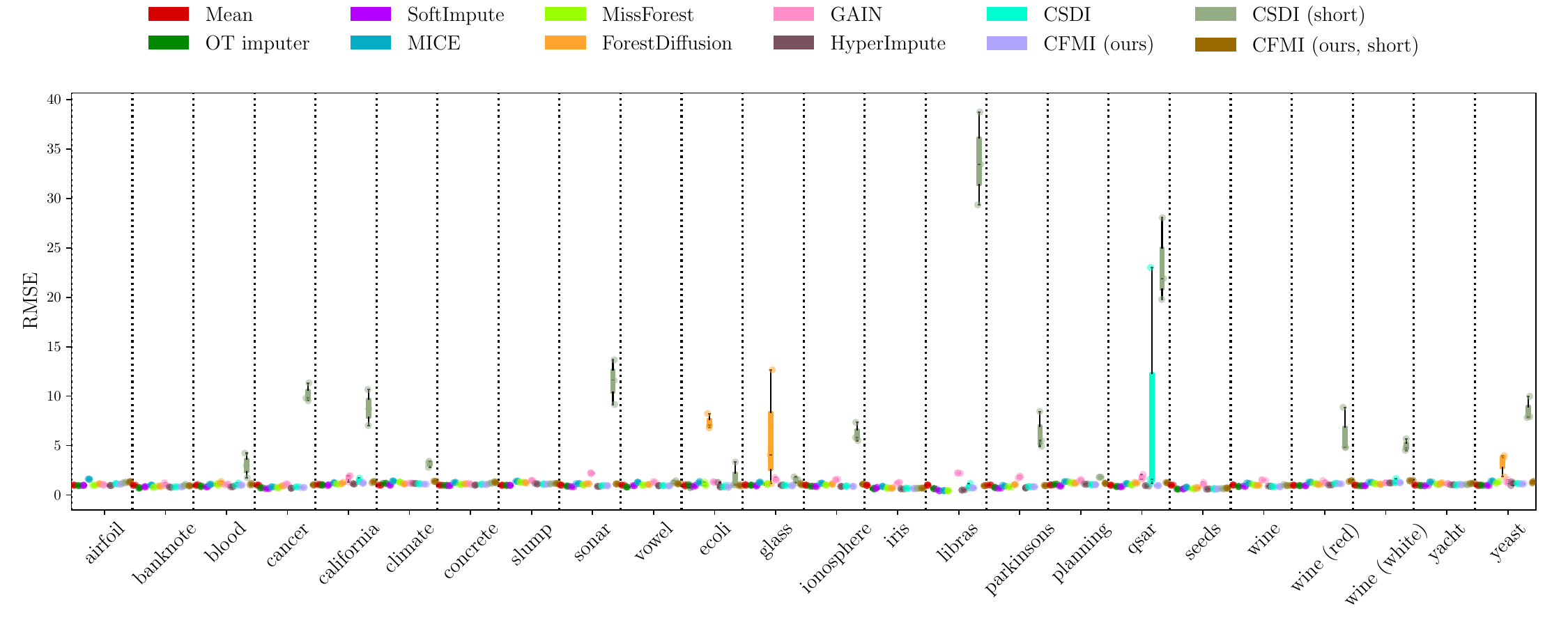}
  \caption{Average RMSE results: box-plot of 3 runs. MCAR 75\% missingness.}
\end{figure}

\begin{figure}[H]
  \centering
  \includegraphics[width=0.9\linewidth]{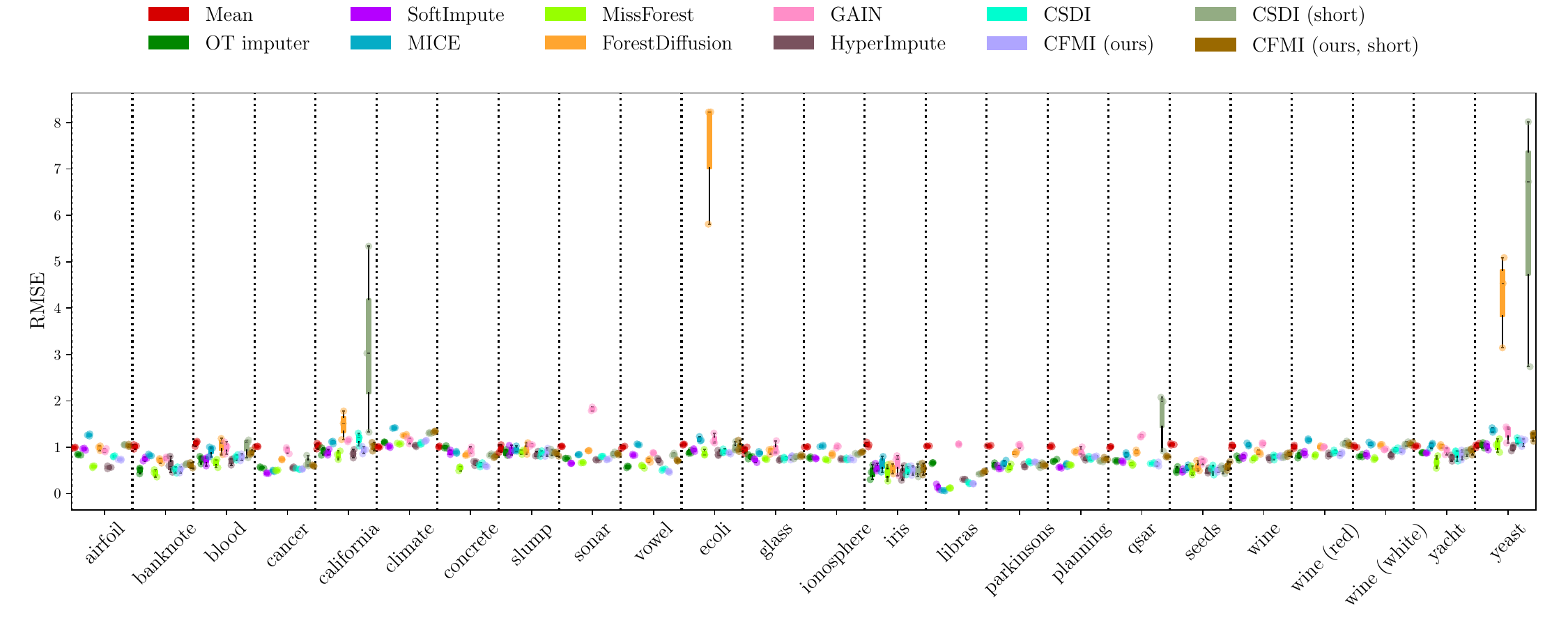}
  \caption{Average RMSE results: box-plot of 3 runs. MAR 25\% missingness.}
\end{figure}

\begin{figure}[H]
  \centering
  \includegraphics[width=0.9\linewidth]{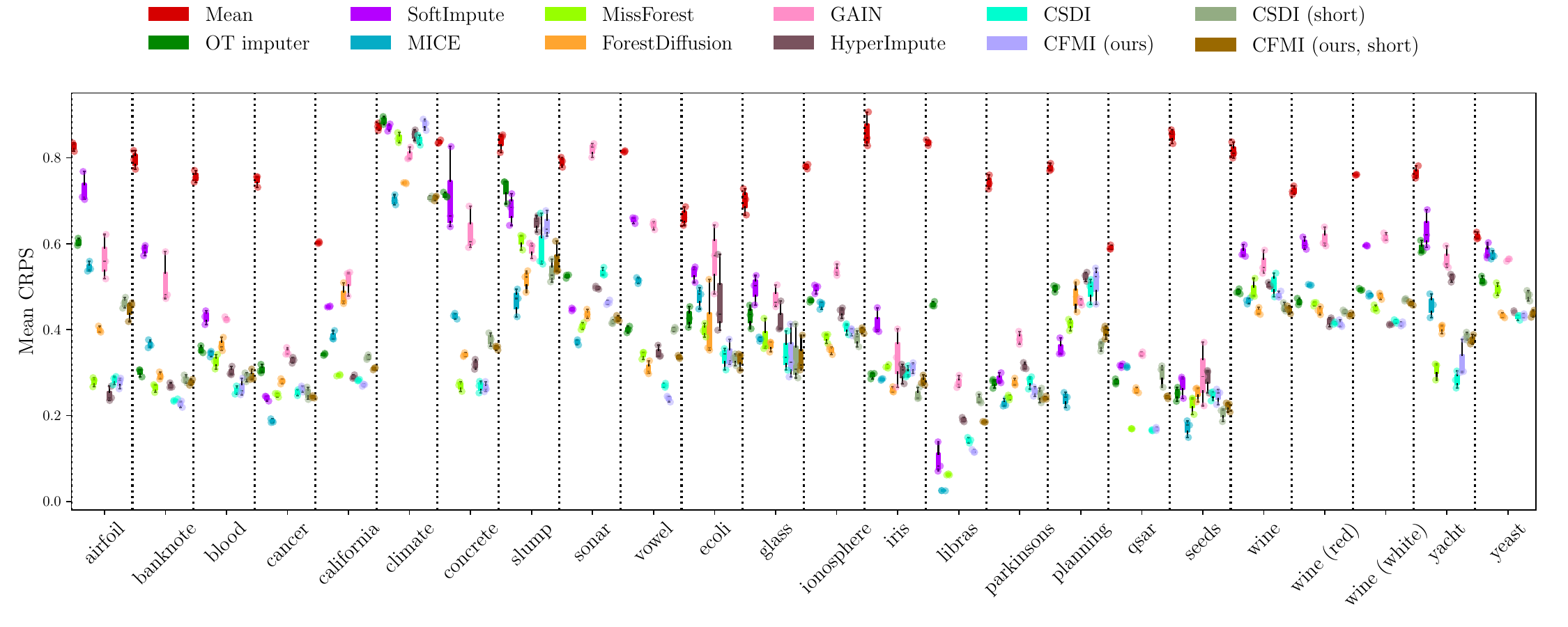}
  \caption{CRPS results: box-plot of 3 runs. MCAR 25\% missingness.}
\end{figure}

\begin{figure}[H]
  \centering
  \includegraphics[width=0.9\linewidth]{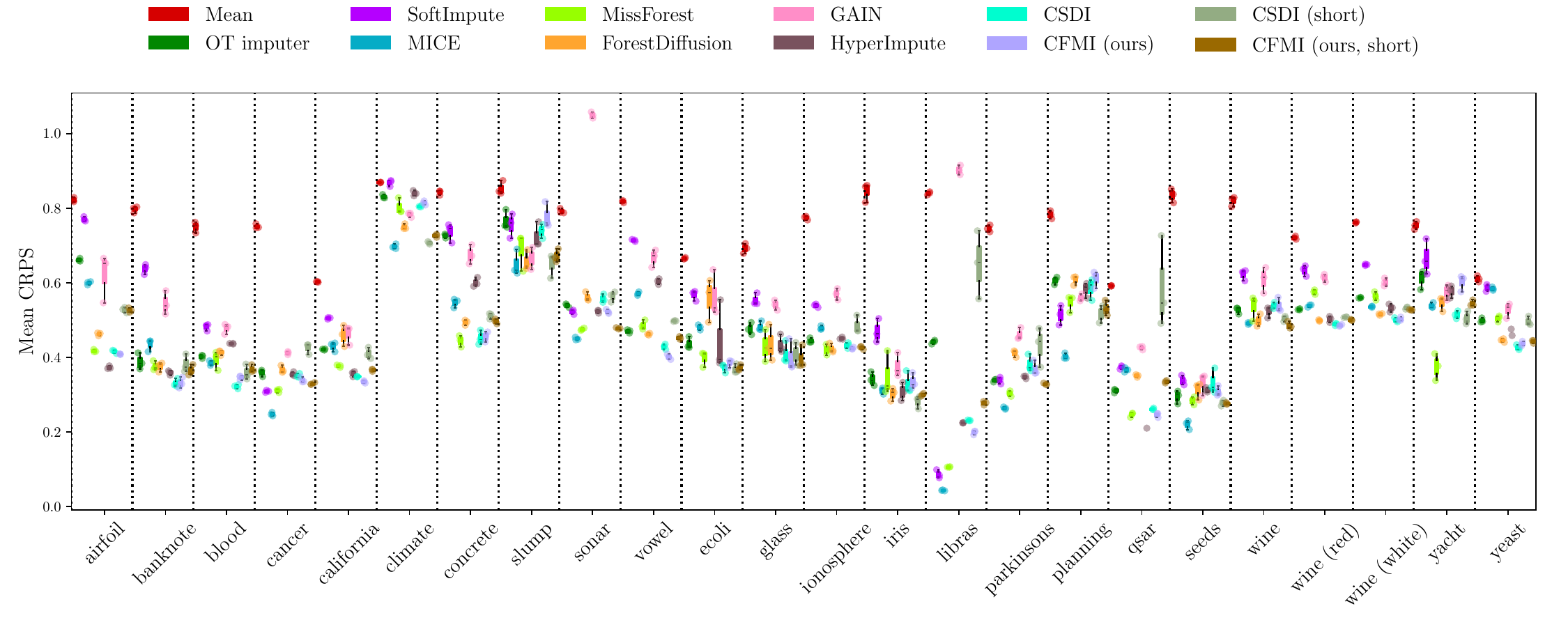}
  \caption{CRPS results: box-plot of 3 runs. MCAR 50\% missingness.}
\end{figure}

\begin{figure}[H]
  \centering
  \includegraphics[width=0.9\linewidth]{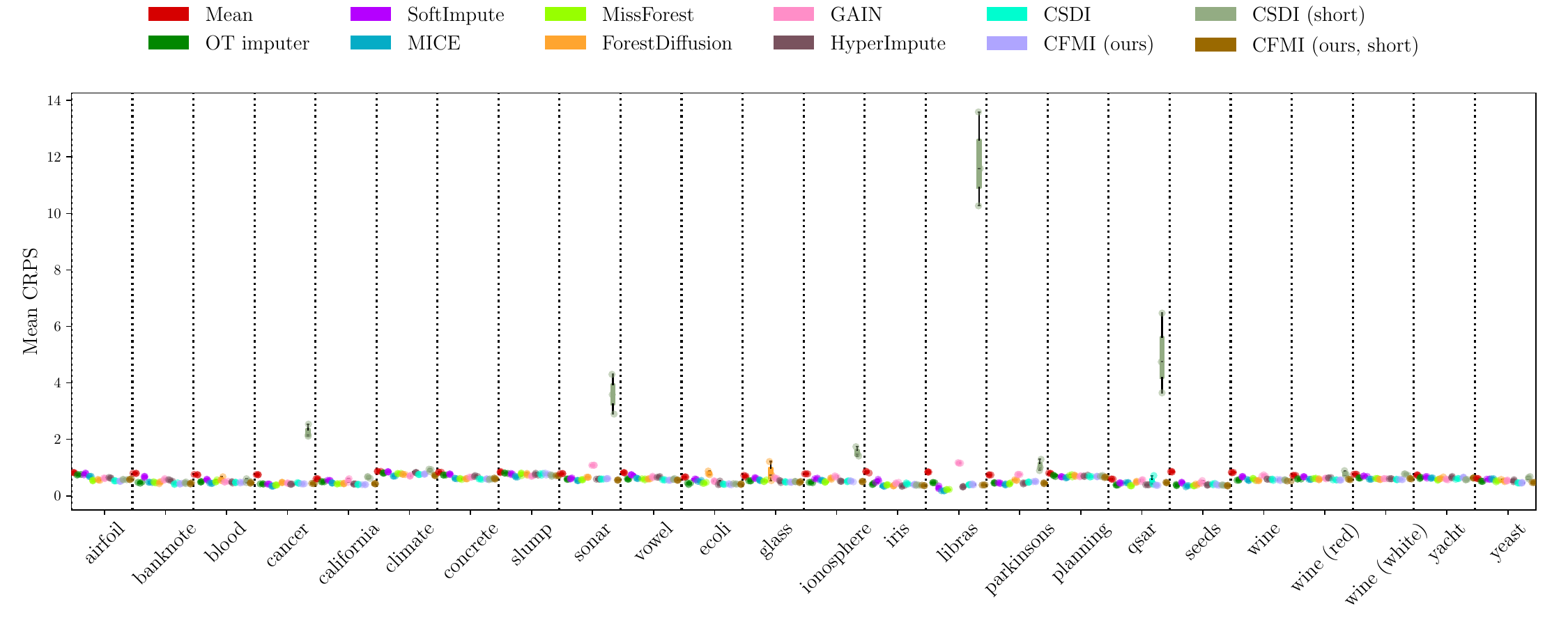}
  \caption{CRPS results: box-plot of 3 runs. MCAR 75\% missingness.}
\end{figure}

\begin{figure}[H]
  \centering
  \includegraphics[width=0.9\linewidth]{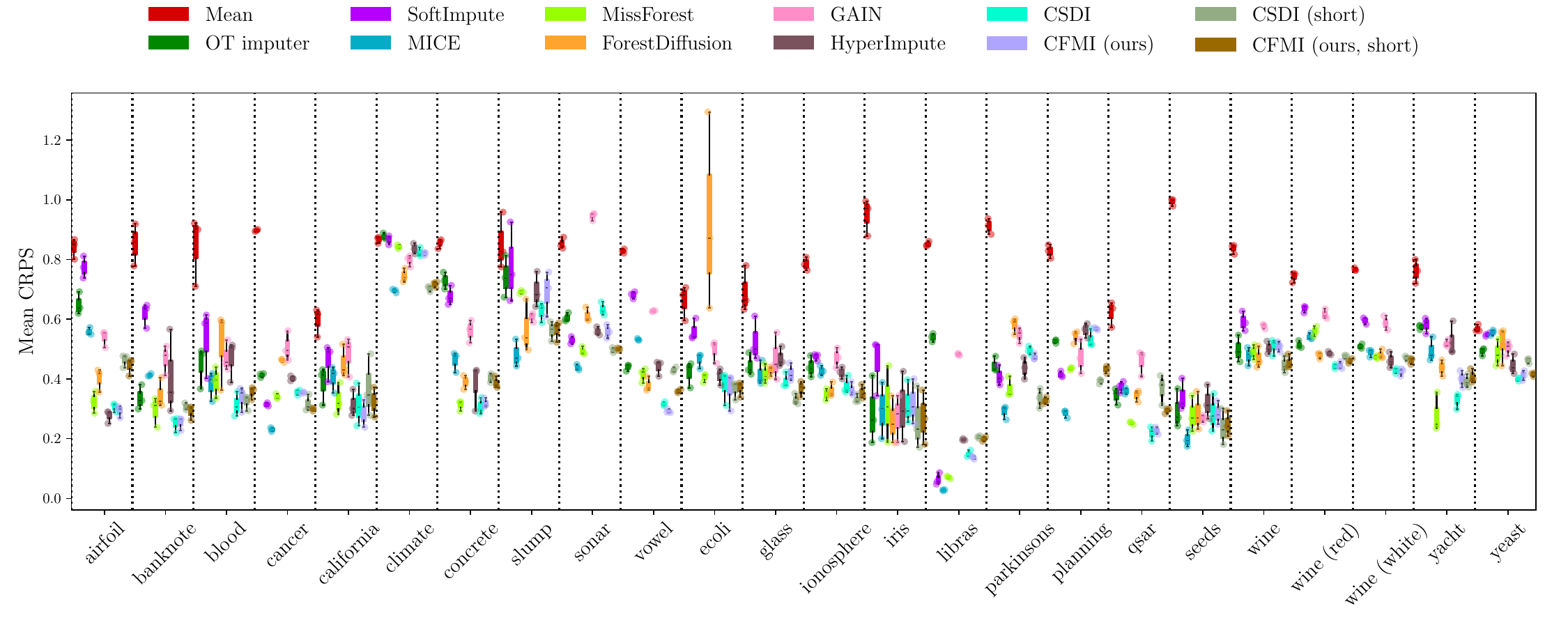}
  \caption{CRPS results: box-plot of 3 runs. MAR 25\% missingness.}
\end{figure}

\begin{figure}[H]
  \centering
  \includegraphics[width=0.9\linewidth]{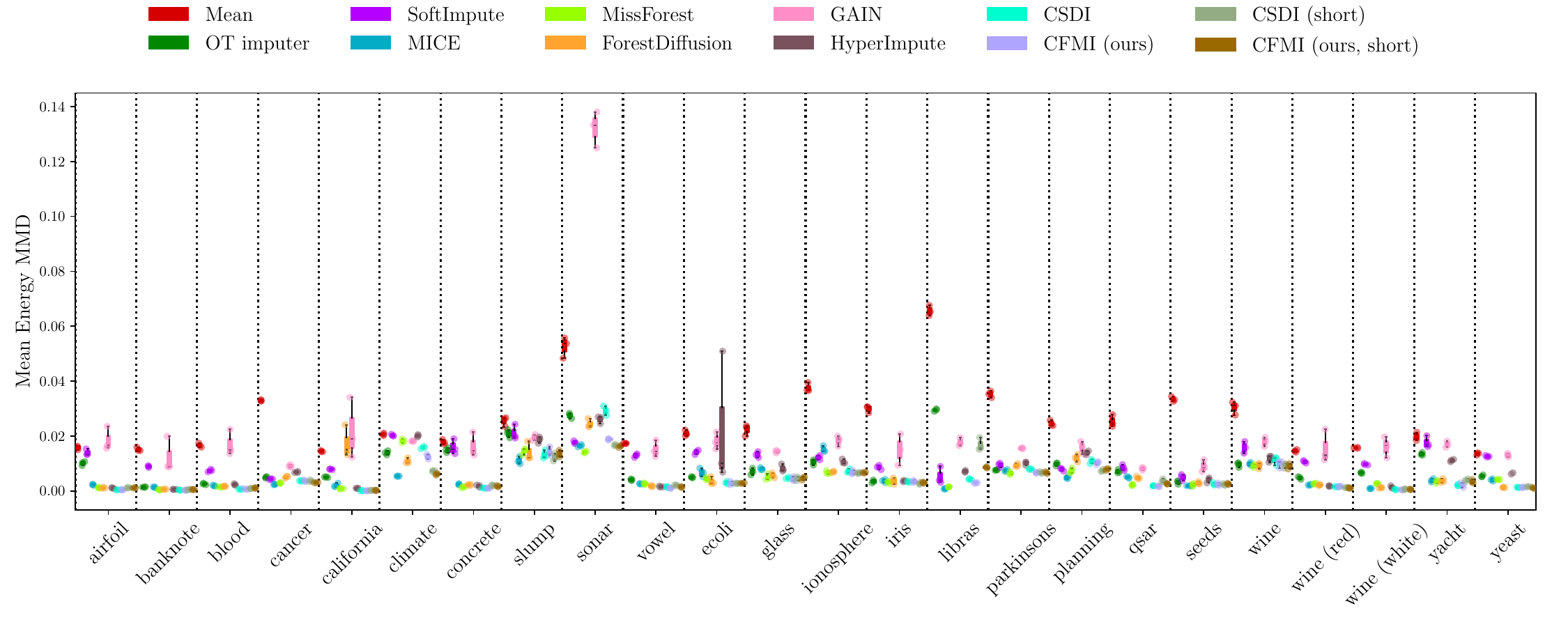}
  \caption{Energy MMD results: box-plot of 3 runs. MCAR 25\% missingness.}
\end{figure}

\begin{figure}[H]
  \centering
  \includegraphics[width=0.9\linewidth]{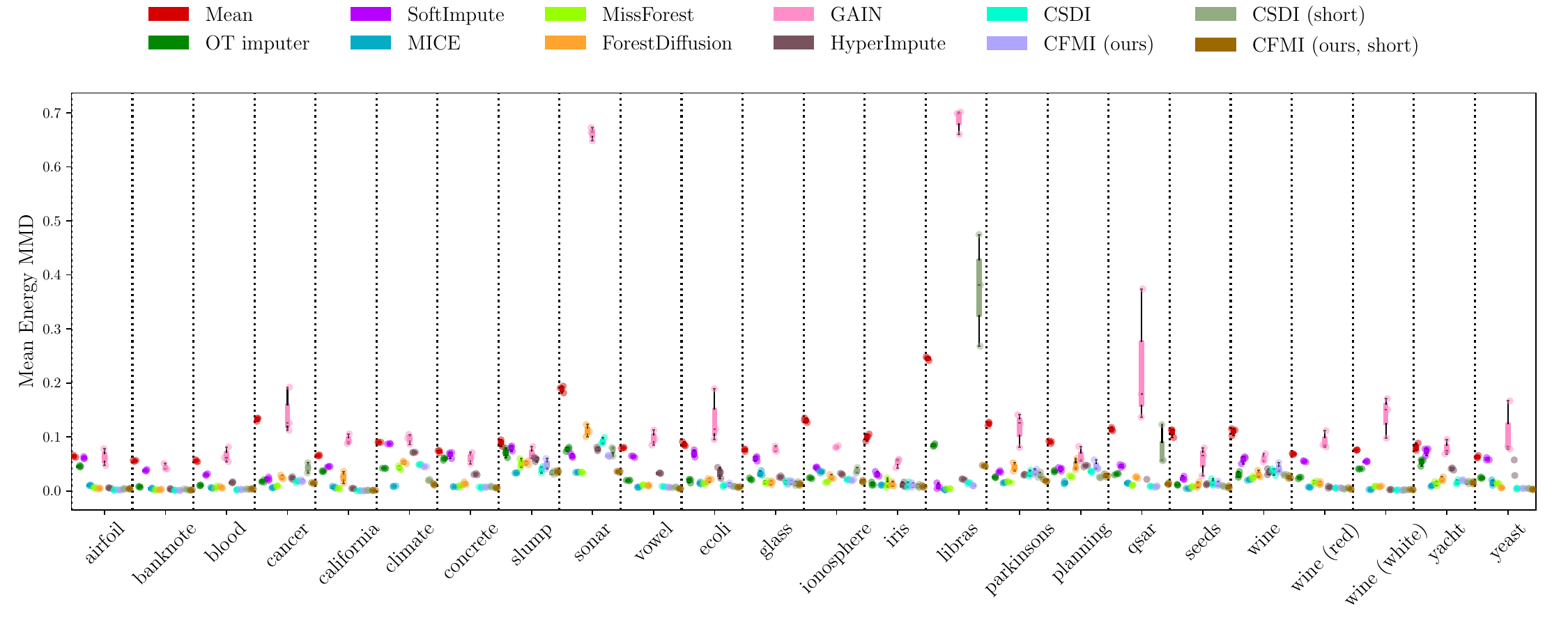}
  \caption{Energy MMD results: box-plot of 3 runs. MCAR 50\% missingness.}
\end{figure}

\begin{figure}[H]
  \centering
  \includegraphics[width=0.9\linewidth]{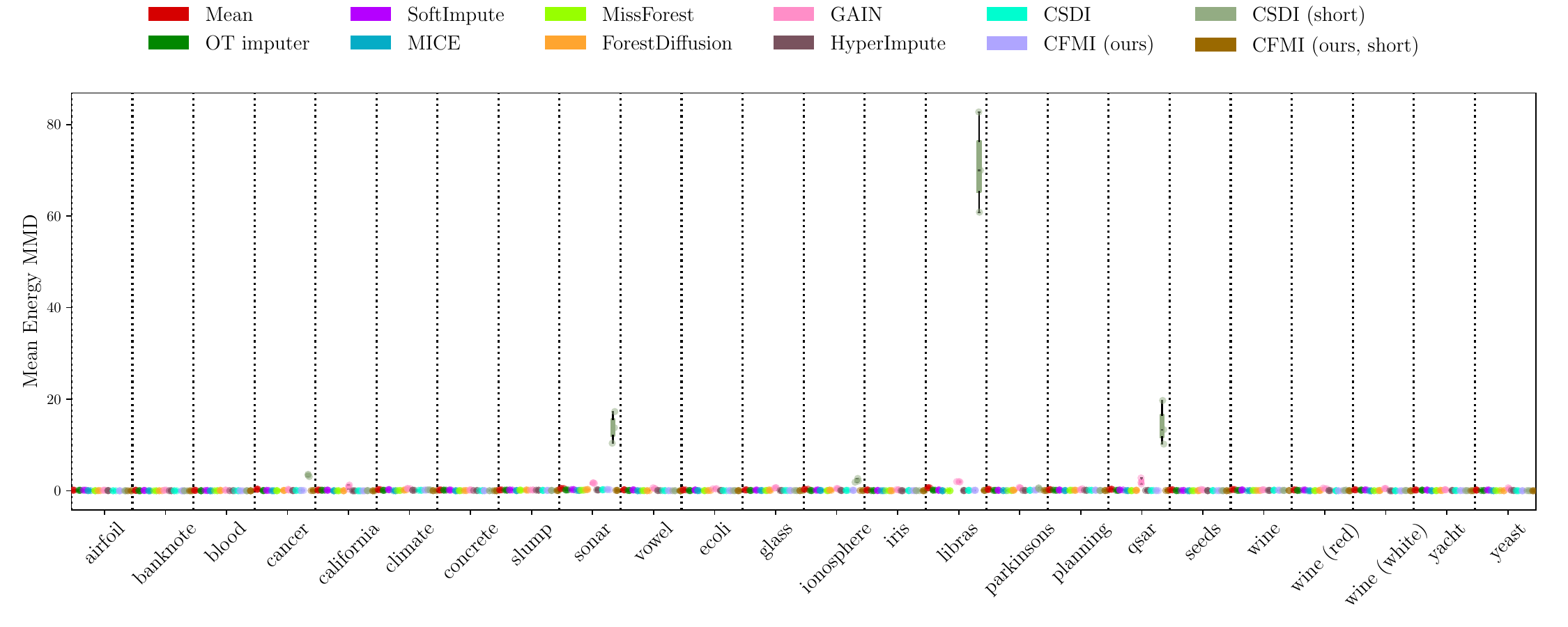}
  \caption{Energy MMD results: box-plot of 3 runs. MCAR 75\% missingness.}
\end{figure}

\begin{figure}[H]
  \centering
  \includegraphics[width=0.9\linewidth]{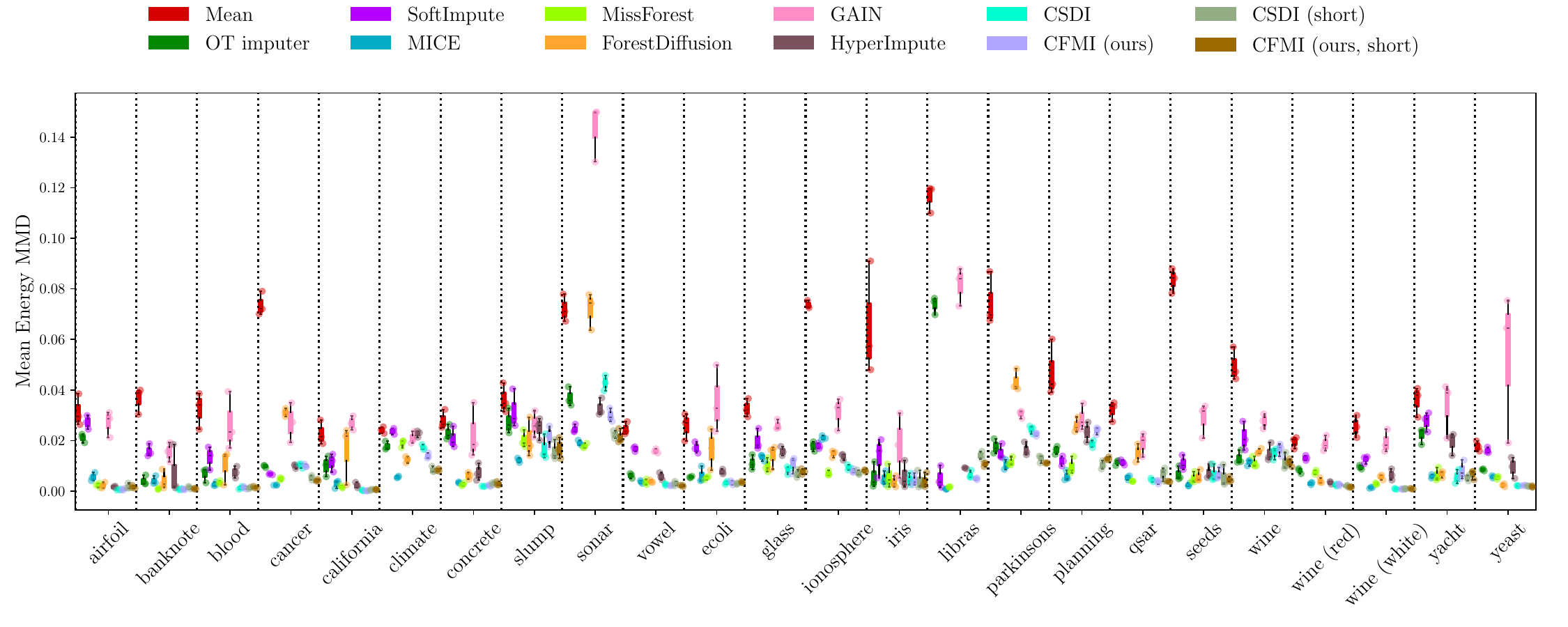}
  \caption{Energy MMD results: box-plot of 3 runs. MAR 25\% missingness.}
\end{figure}

\begin{figure}[H]
  \centering
  \includegraphics[width=0.9\linewidth]{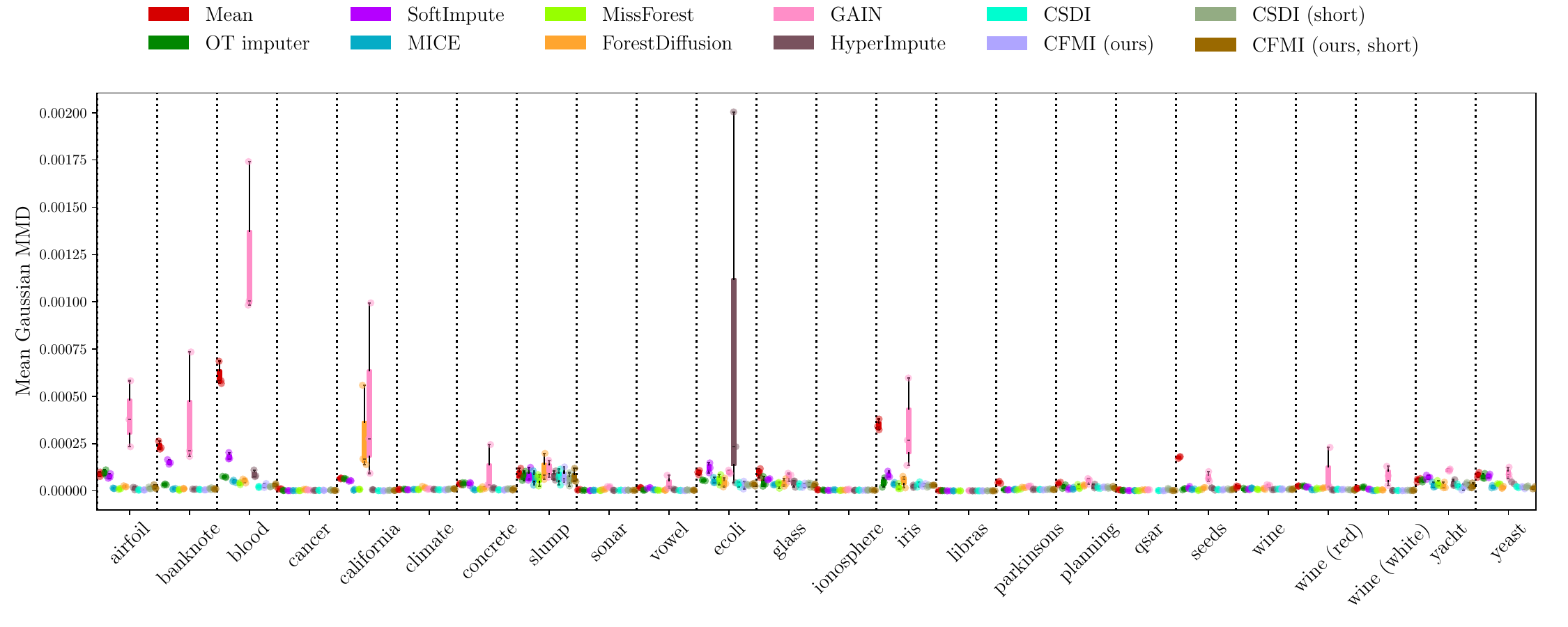}
  \caption{Gaussian MMD results: box-plot of 3 runs. MCAR 25\% missingness.}
\end{figure}

\begin{figure}[H]
  \centering
  \includegraphics[width=0.9\linewidth]{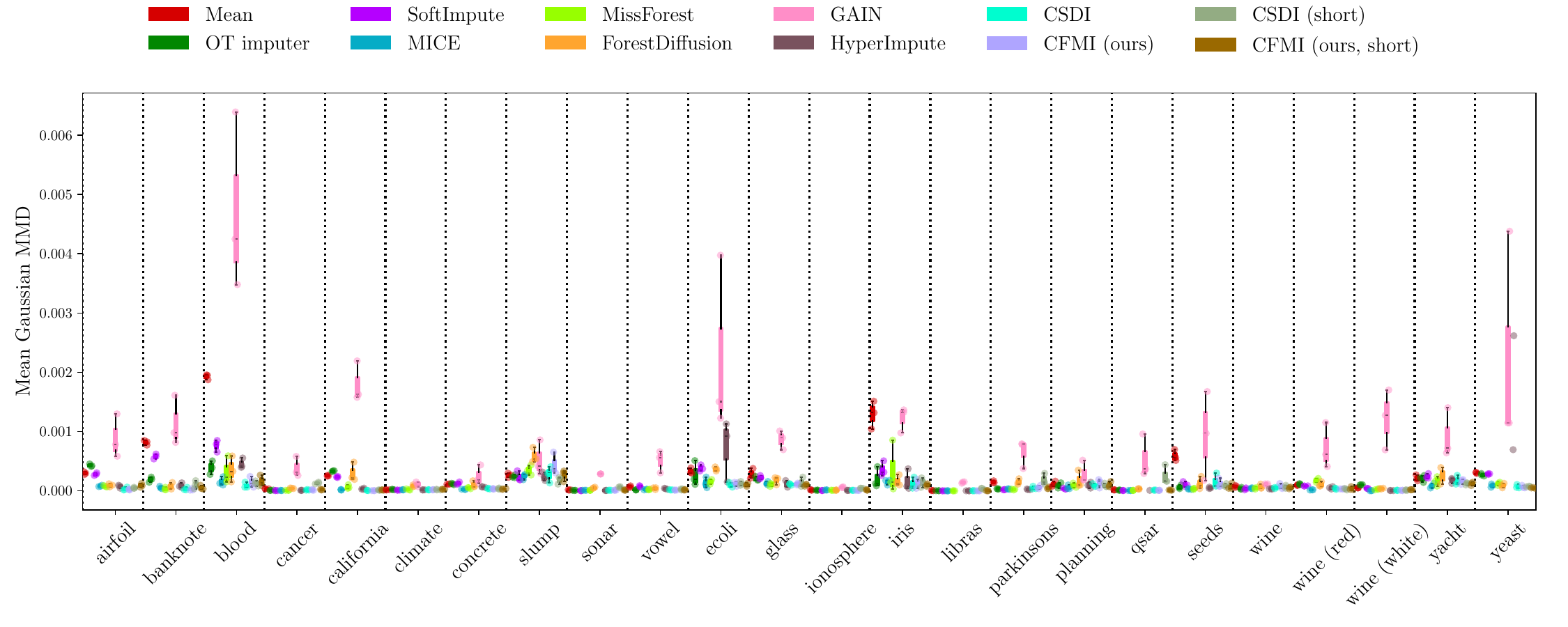}
  \caption{Gaussian MMD results: box-plot of 3 runs. MCAR 50\% missingness.}
\end{figure}

\begin{figure}[H]
  \centering
  \includegraphics[width=0.9\linewidth]{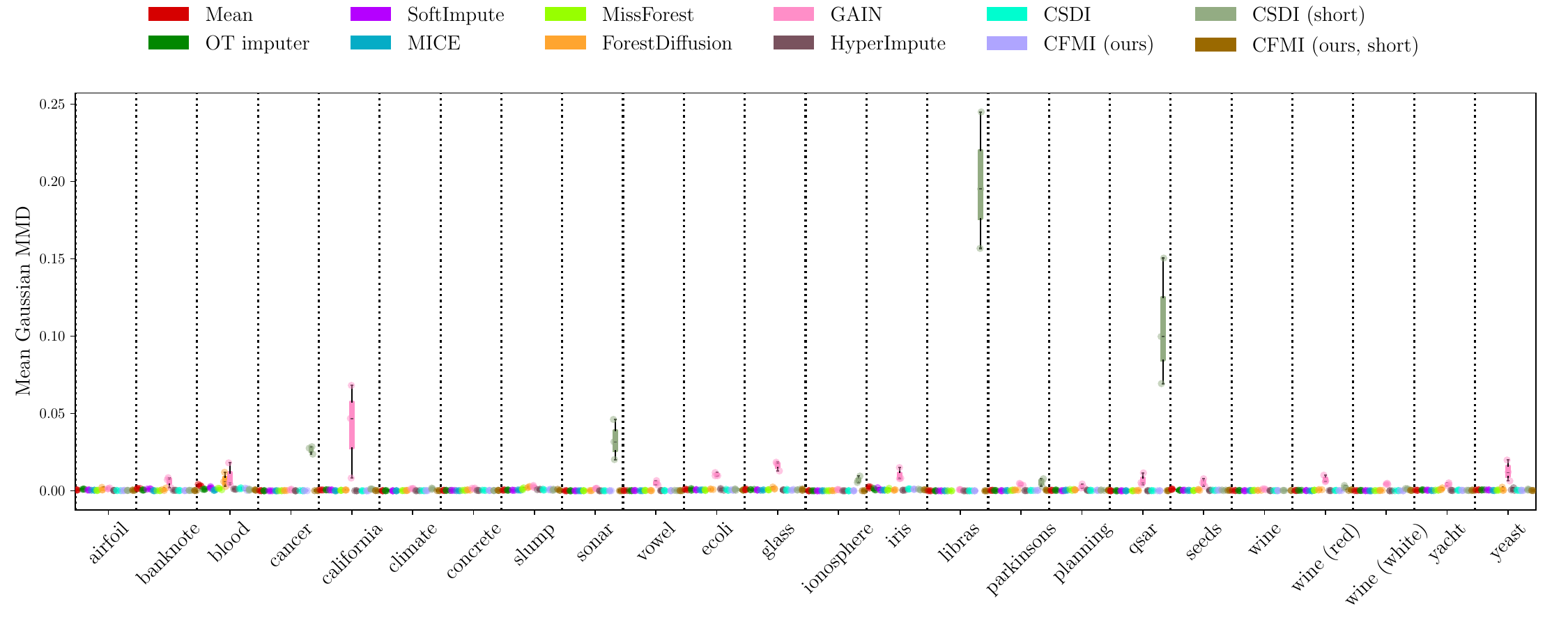}
  \caption{Gaussian MMD results: box-plot of 3 runs. MCAR 75\% missingness.}
\end{figure}

\begin{figure}[H]
  \centering
  \includegraphics[width=0.9\linewidth]{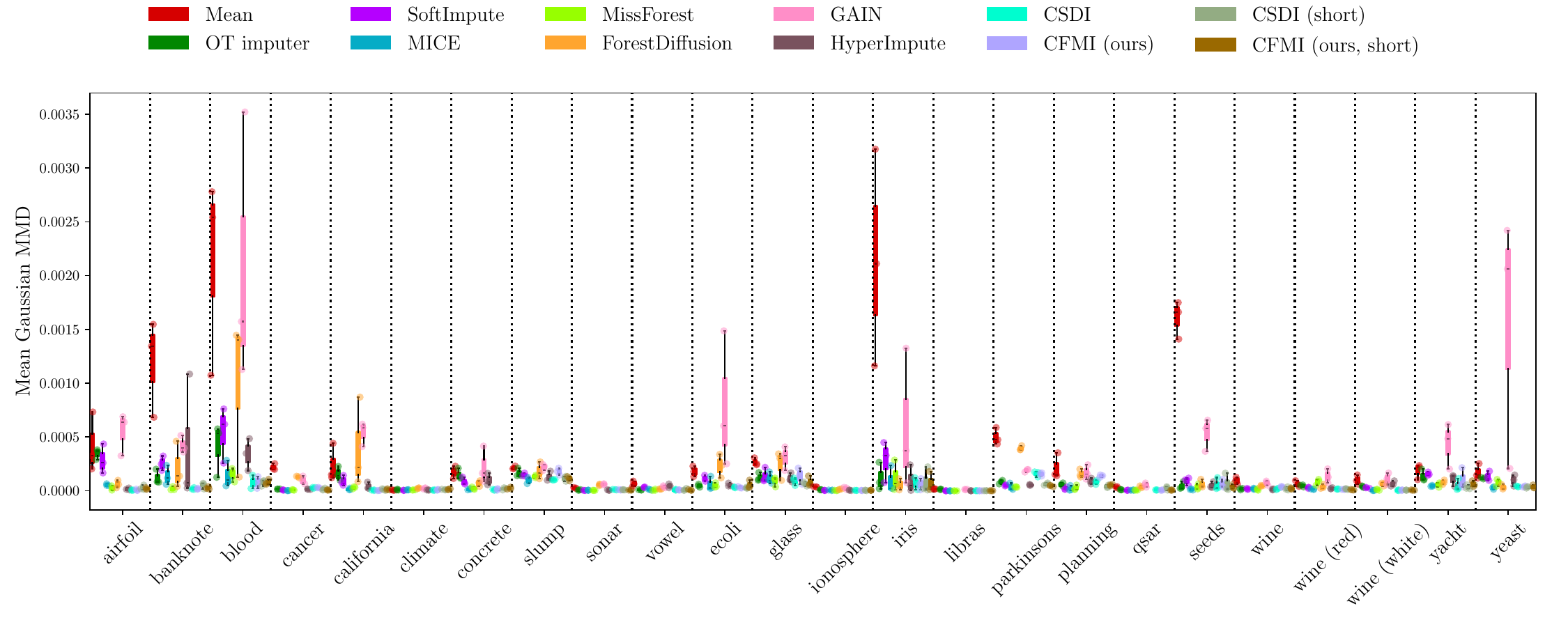}
  \caption{Gaussian MMD results: box-plot of 3 runs. MAR 25\% missingness.}
\end{figure}

\begin{figure}[H]
  \centering
  \includegraphics[width=0.9\linewidth]{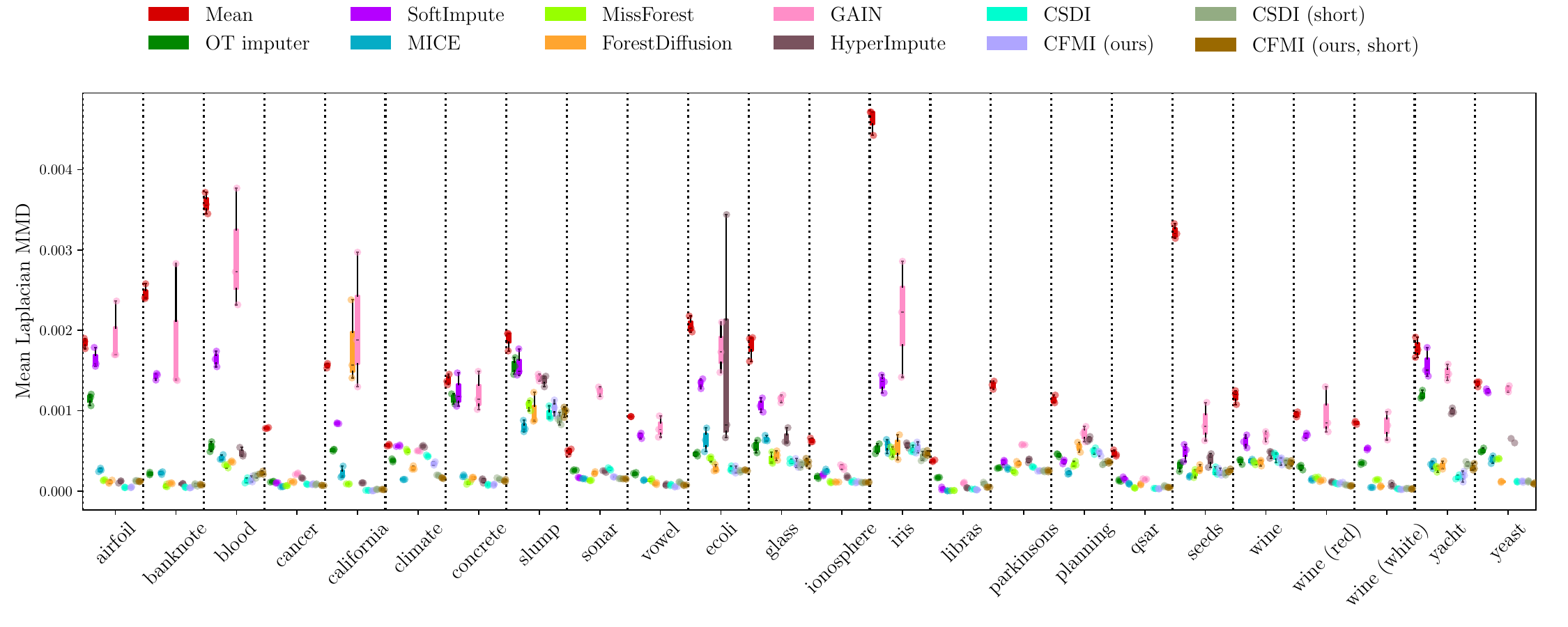}
  \caption{Laplacian MMD results: box-plot of 3 runs. MCAR 25\% missingness.}
\end{figure}

\begin{figure}[H]
  \centering
  \includegraphics[width=0.9\linewidth]{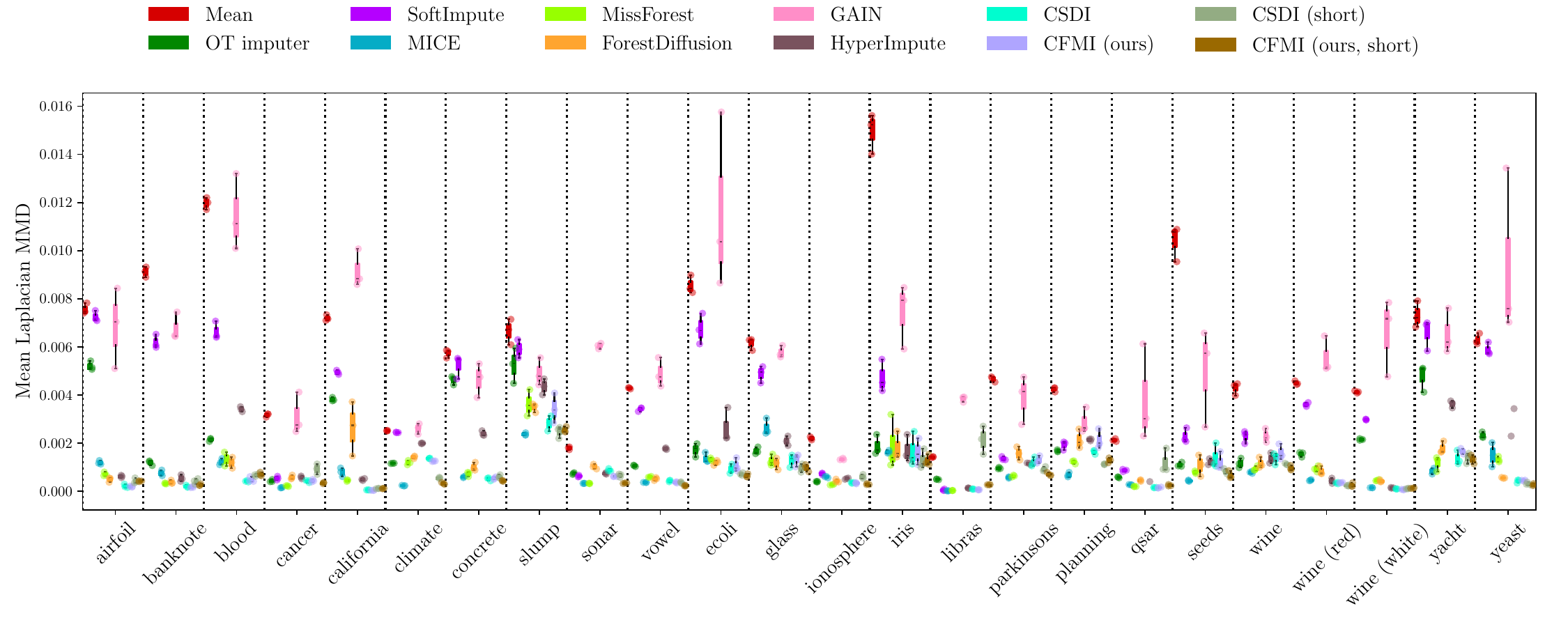}
  \caption{Laplacian MMD results: box-plot of 3 runs. MCAR 50\% missingness.}
\end{figure}

\begin{figure}[H]
  \centering
  \includegraphics[width=0.9\linewidth]{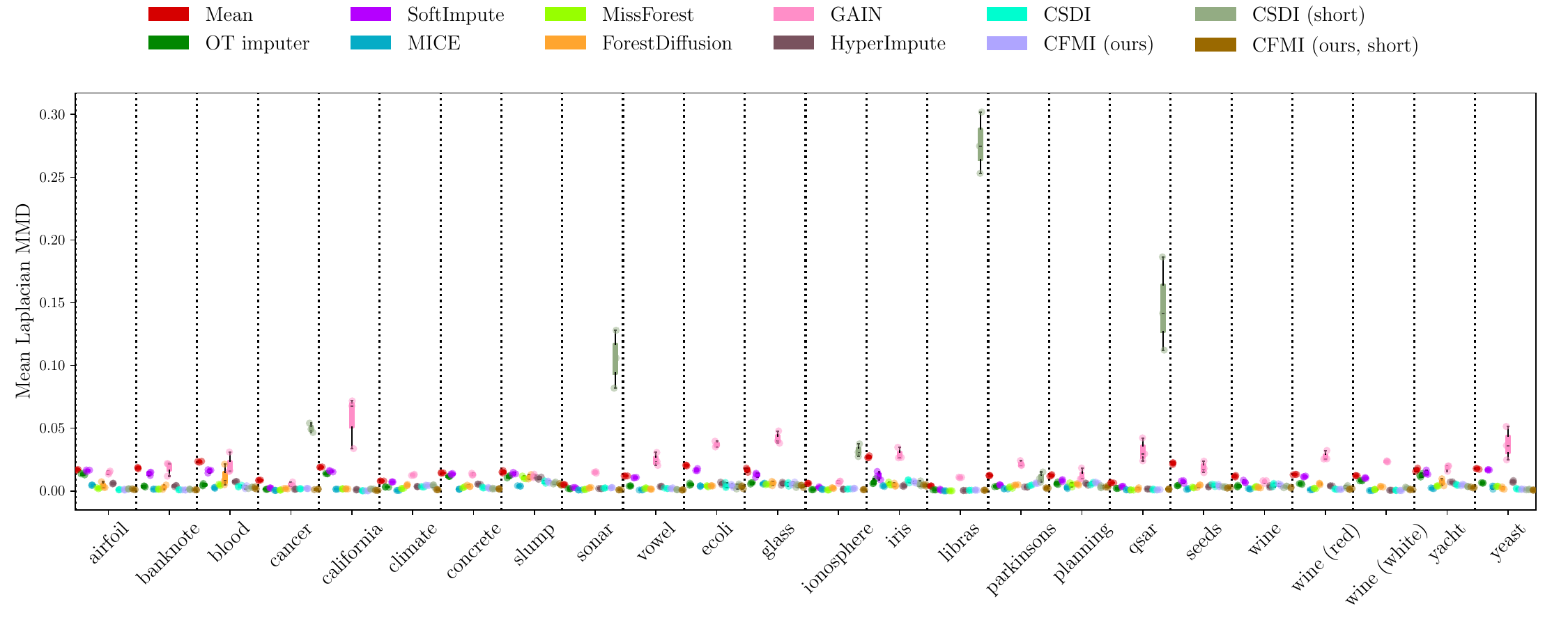}
  \caption{Laplacian MMD results: box-plot of 3 runs. MCAR 75\% missingness.}
\end{figure}

\begin{figure}[H]
  \centering
  \includegraphics[width=0.9\linewidth]{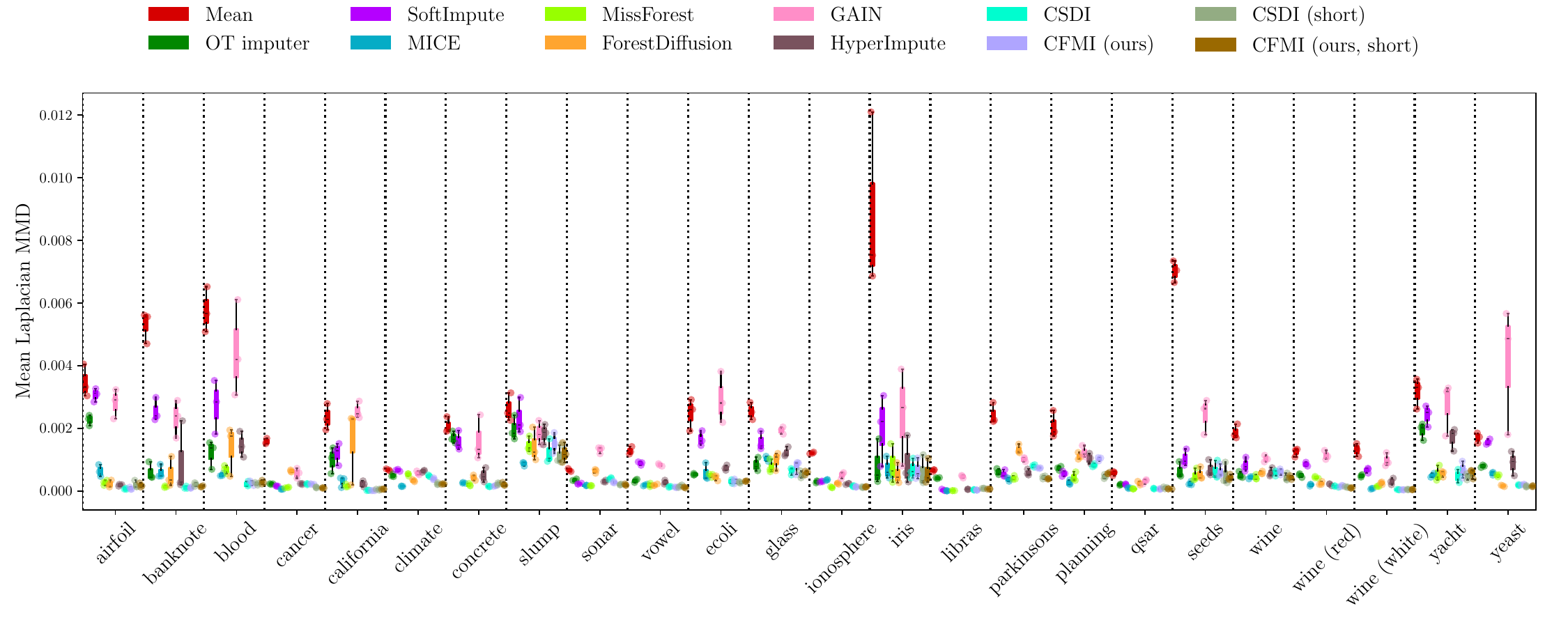}
  \caption{Laplacian MMD results: box-plot of 3 runs. MAR 25\% missingness.}
\end{figure}

\begin{figure}[H]
  \centering
  \includegraphics[width=0.9\linewidth]{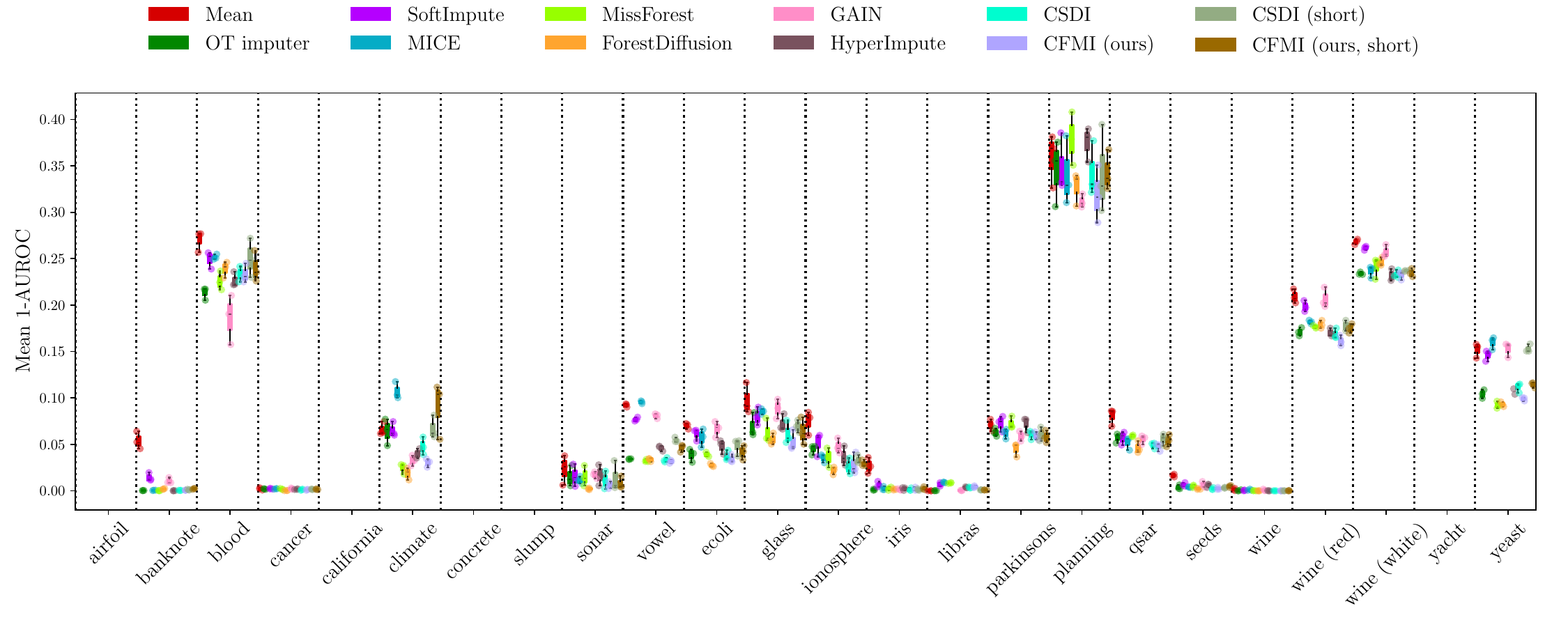}
  \caption{Classifier 1-AUROC results: box-plot of 3 runs. MCAR 25\% missingness.}
\end{figure}

\begin{figure}[H]
  \centering
  \includegraphics[width=0.9\linewidth]{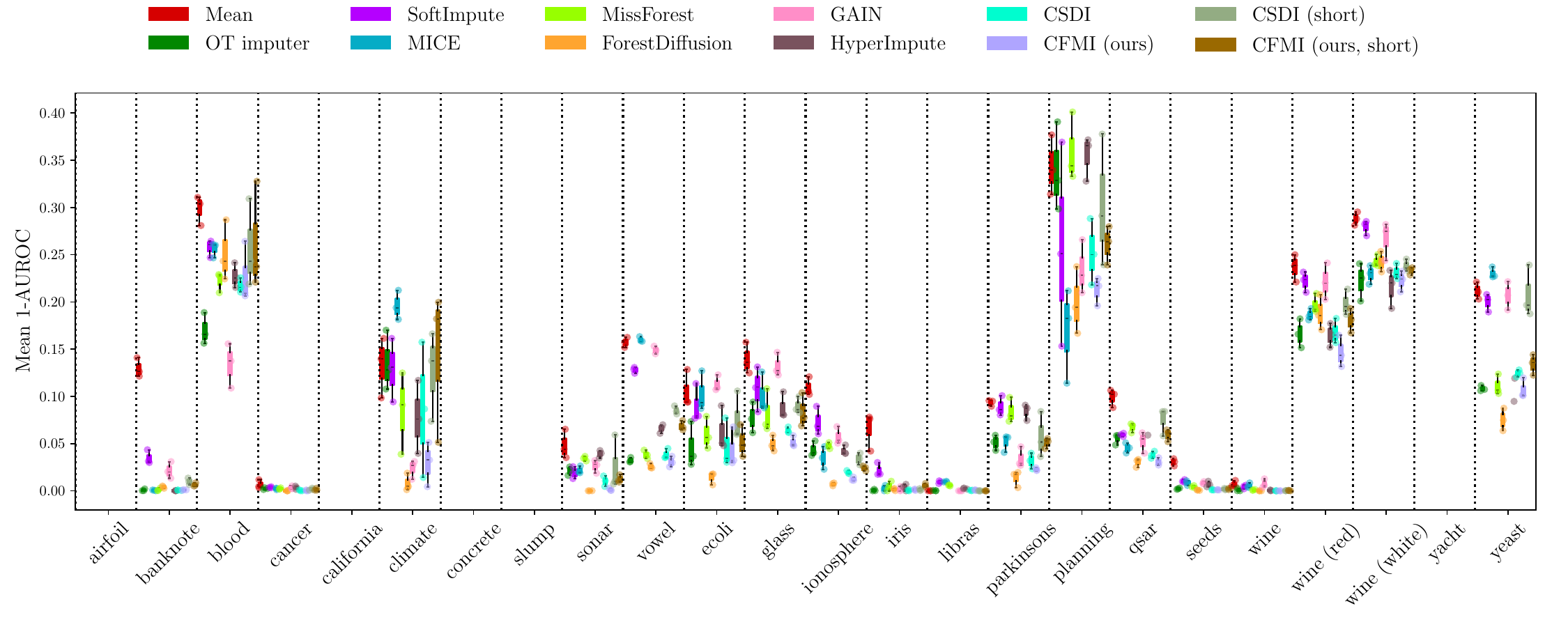}
  \caption{Classifier 1-AUROC results: box-plot of 3 runs. MCAR 50\% missingness.}
\end{figure}

\begin{figure}[H]
  \centering
  \includegraphics[width=0.9\linewidth]{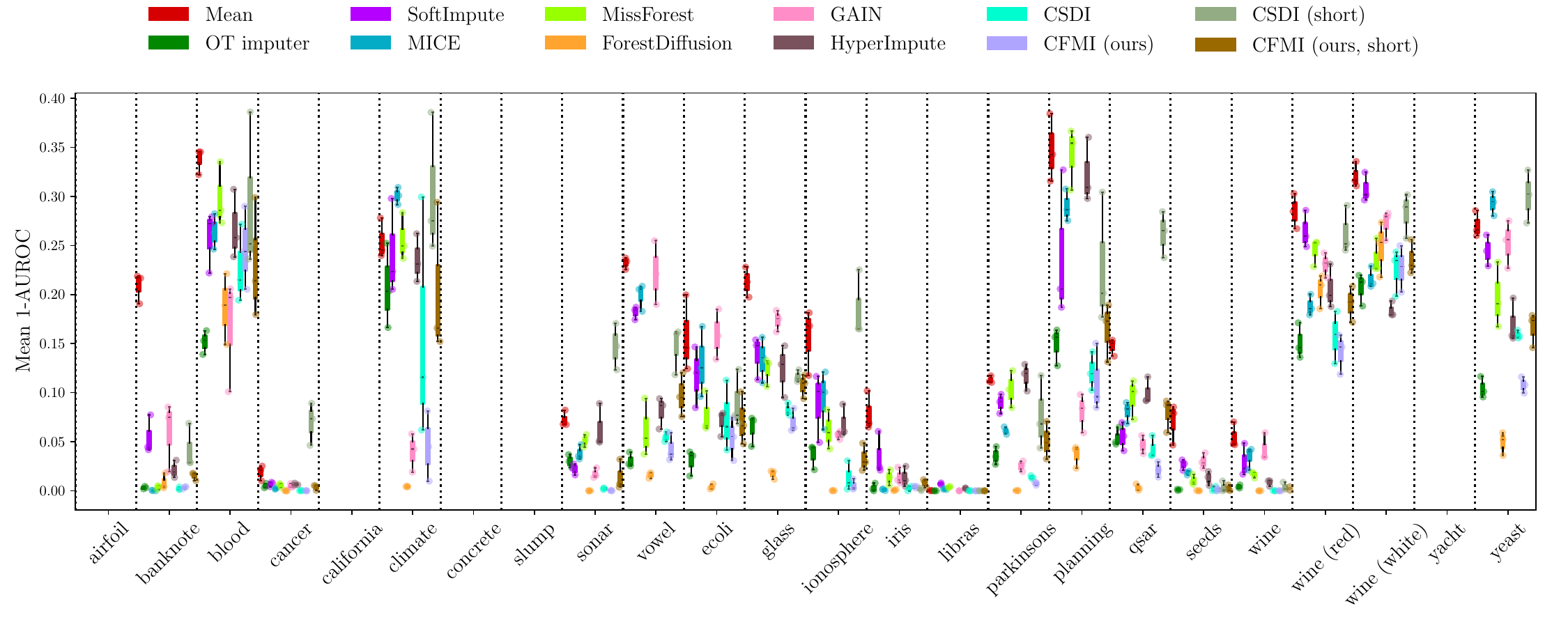}
  \caption{Classifier 1-AUROC results: box-plot of 3 runs. MCAR 75\% missingness.}
\end{figure}

\begin{figure}[H]
  \centering
  \includegraphics[width=0.9\linewidth]{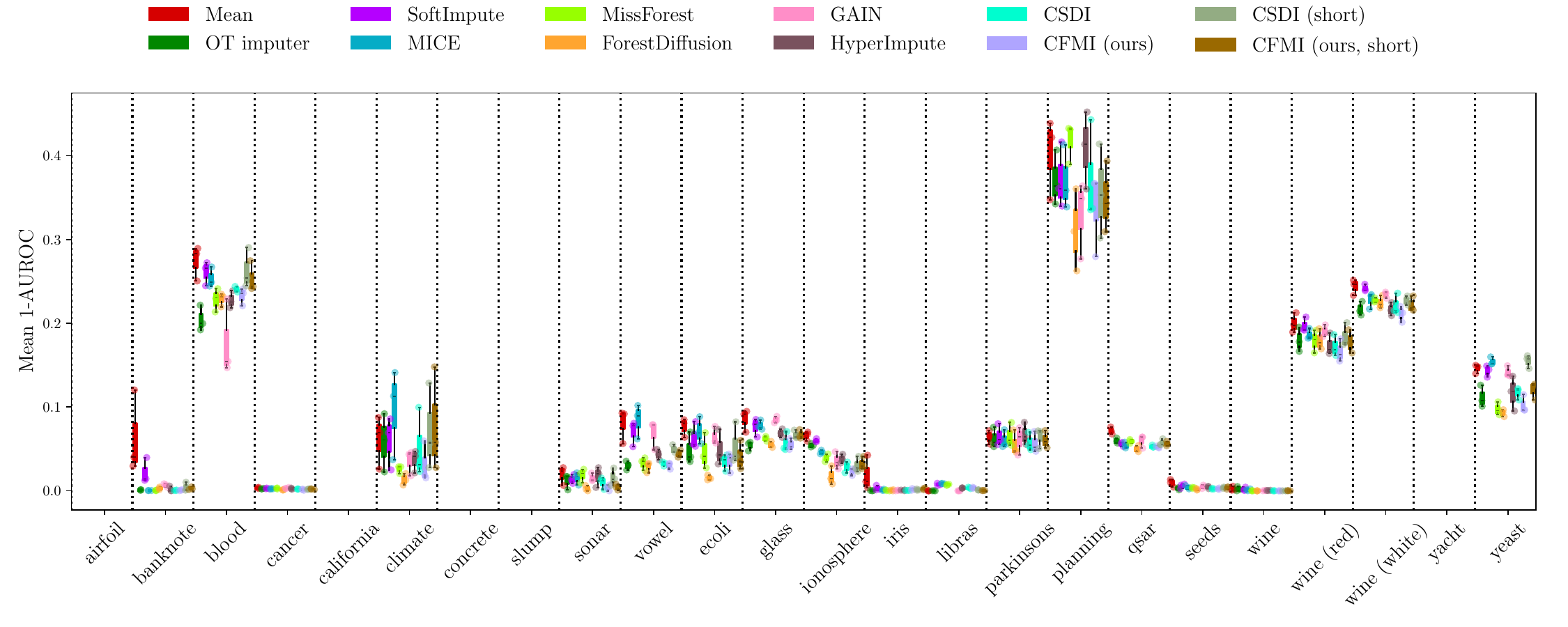}
  \caption{Classifier 1-AUROC results: box-plot of 3 runs. MAR 25\% missingness.}
\end{figure}

\begin{figure}[H]
  \centering
  \includegraphics[width=0.9\linewidth]{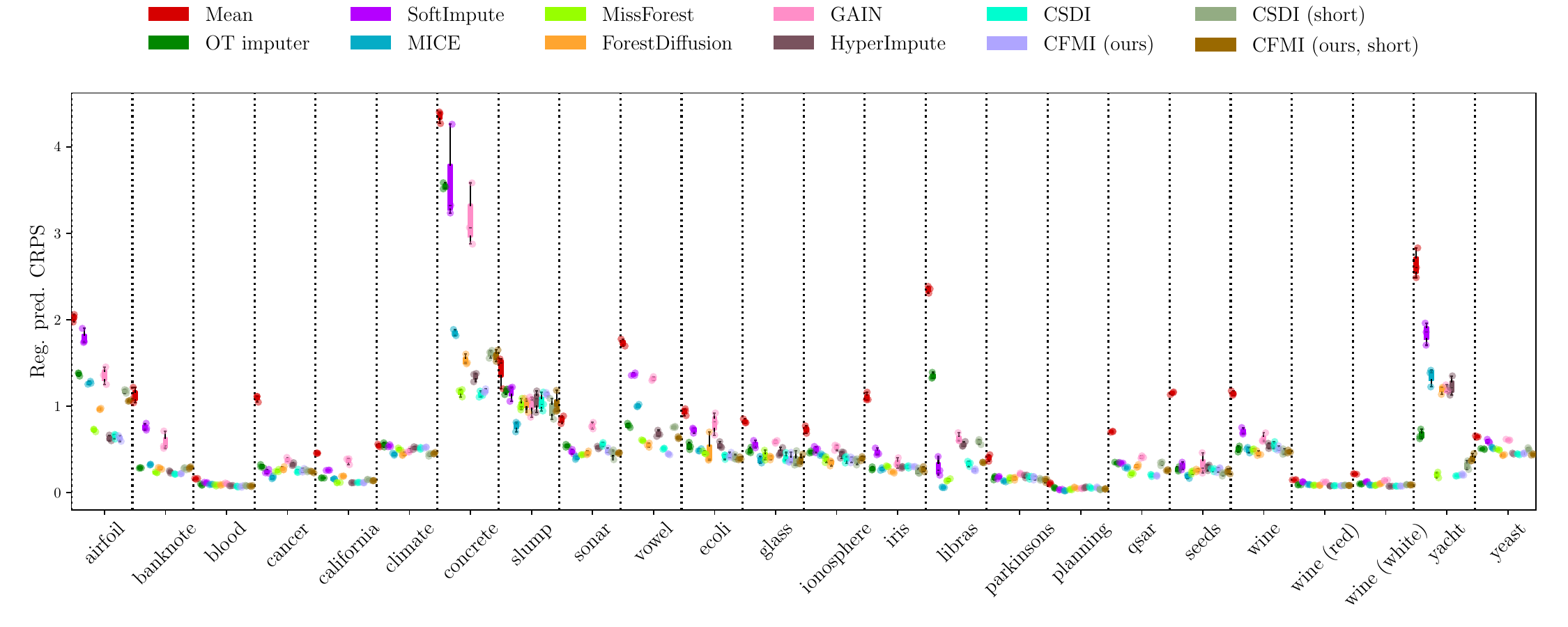}
  \caption{Regression CRPS results: box-plot of 3 runs. MCAR 25\% missingness.}
\end{figure}

\begin{figure}[H]
  \centering
  \includegraphics[width=0.9\linewidth]{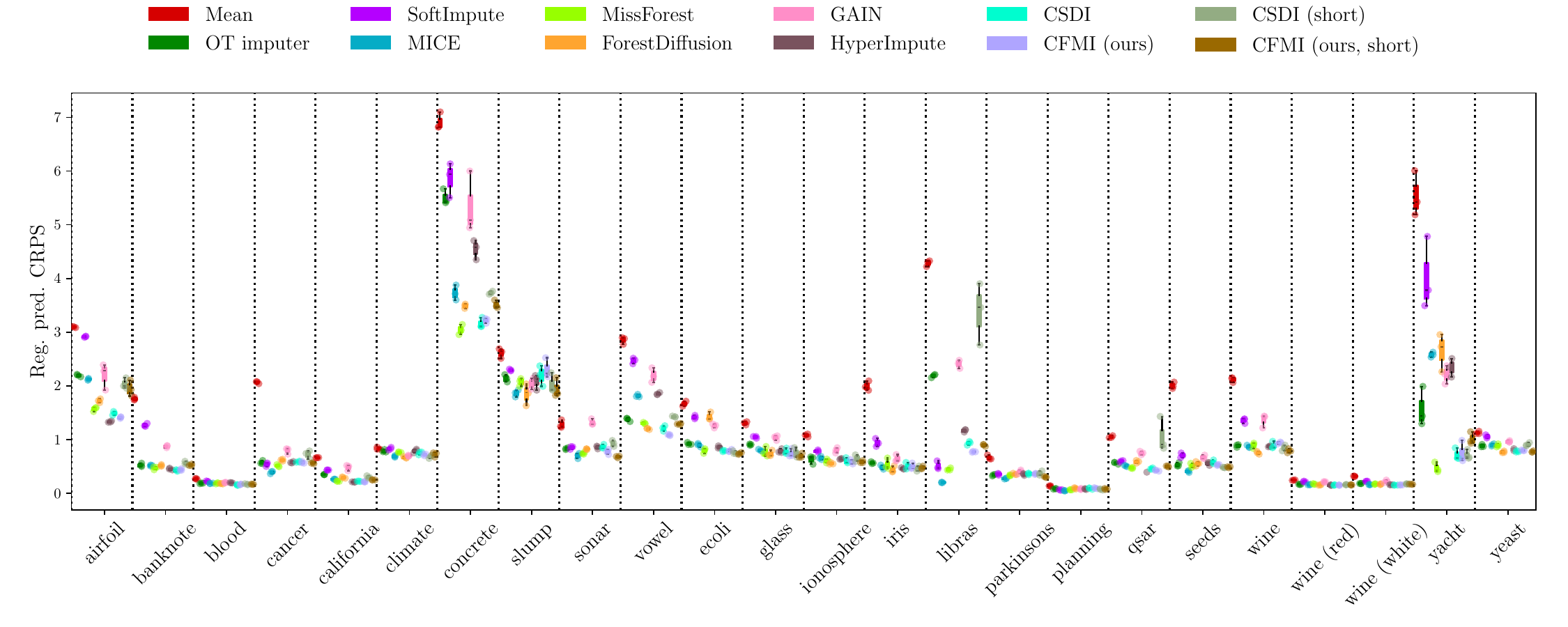}
  \caption{Regression CRPS results: box-plot of 3 runs. MCAR 50\% missingness.}
\end{figure}

\begin{figure}[H]
  \centering
  \includegraphics[width=0.9\linewidth]{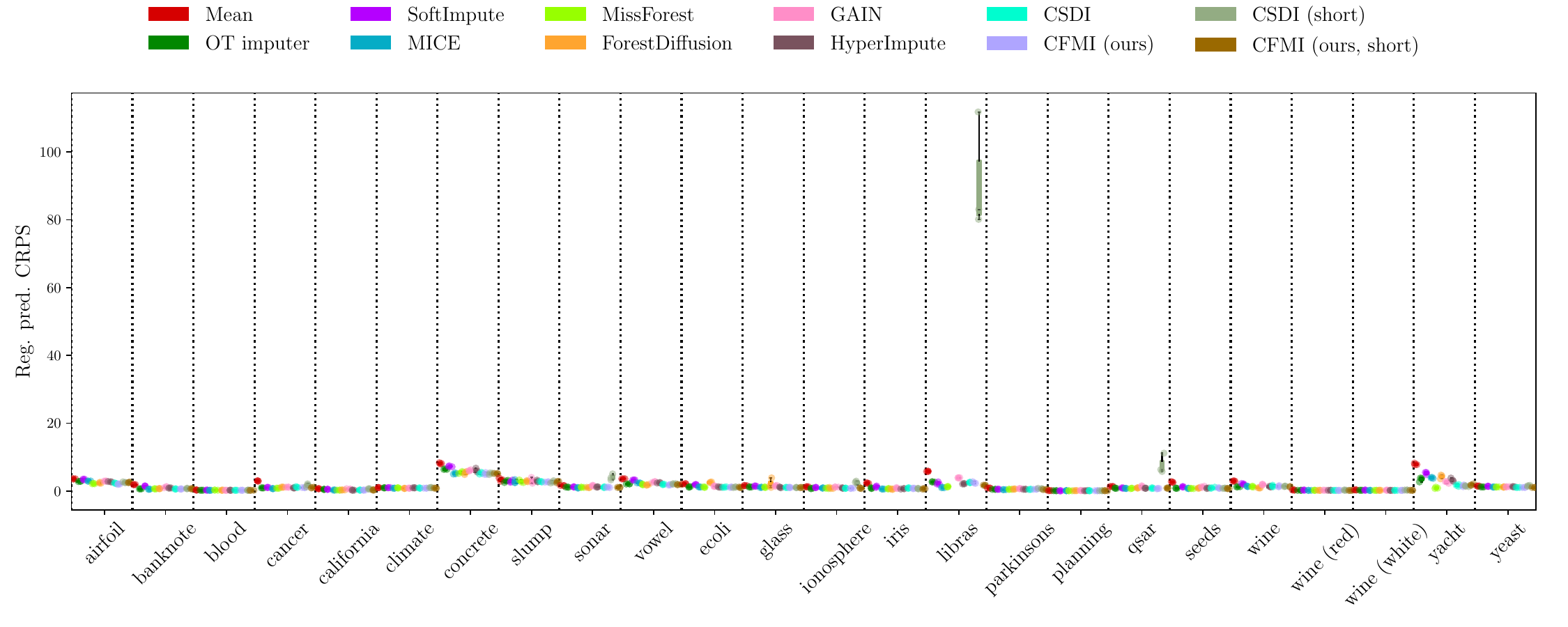}
  \caption{Regression CRPS results: box-plot of 3 runs. MCAR 75\% missingness.}
\end{figure}

\begin{figure}[H]
  \centering
  \includegraphics[width=0.9\linewidth]{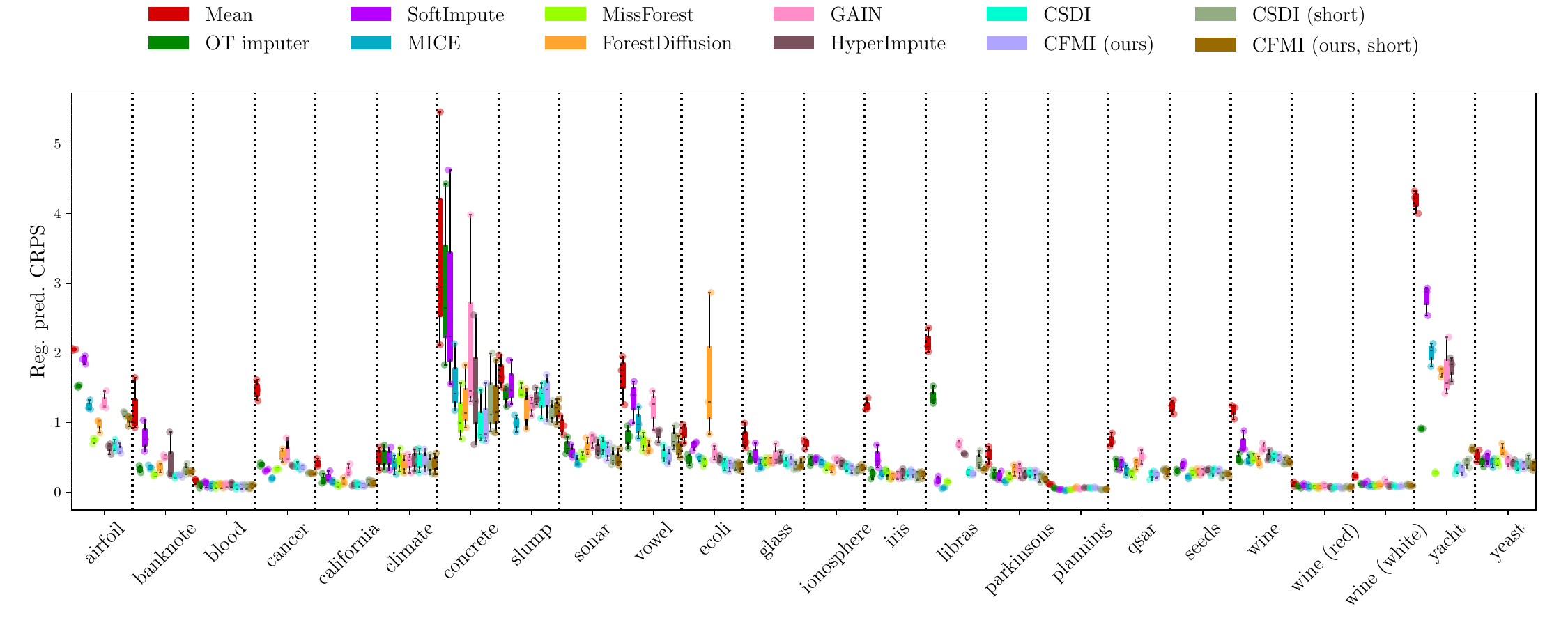}
  \caption{Regression CRPS results: box-plot of 3 runs. MAR 25\% missingness.}
\end{figure}

\begin{figure}[H]
  \centering
  \includegraphics[width=0.9\linewidth]{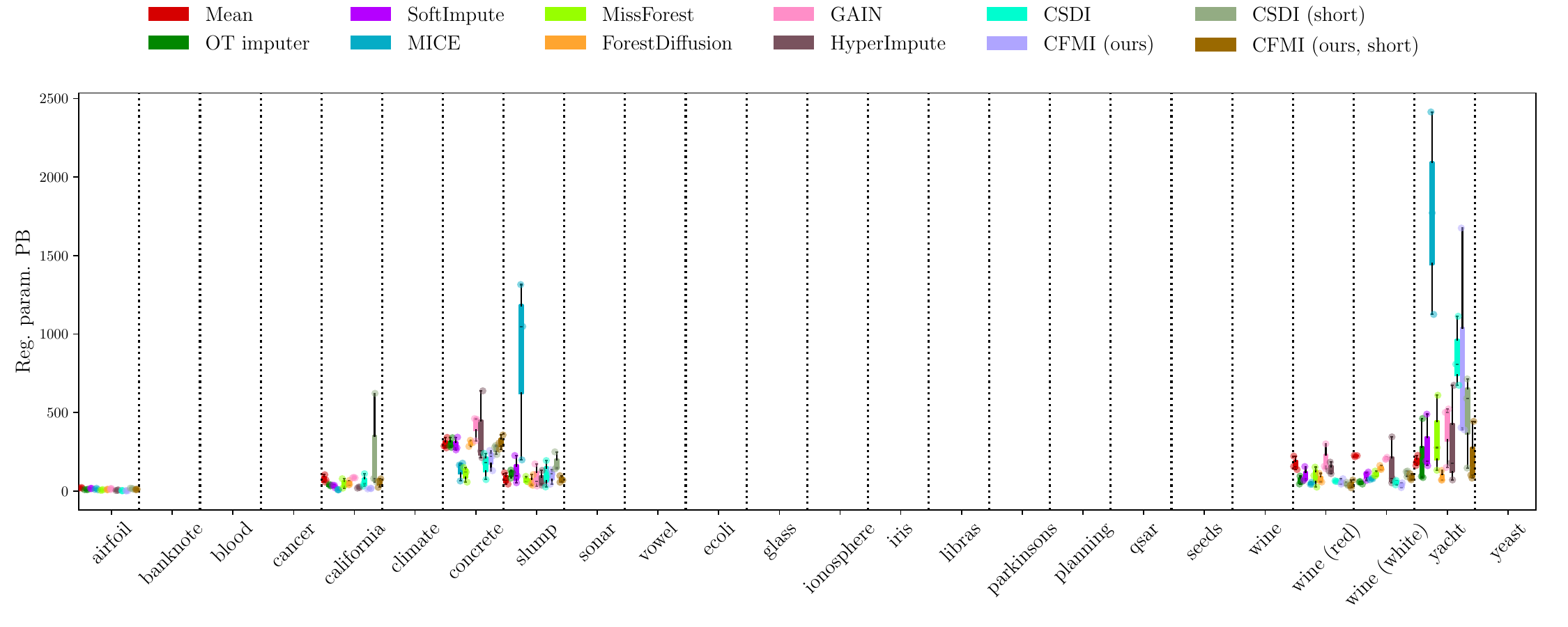}
  \caption{Regression parameter percent bias results: box-plot of 3 runs. MCAR 25\% missingness.}
\end{figure}

\begin{figure}[H]
  \centering
  \includegraphics[width=0.9\linewidth]{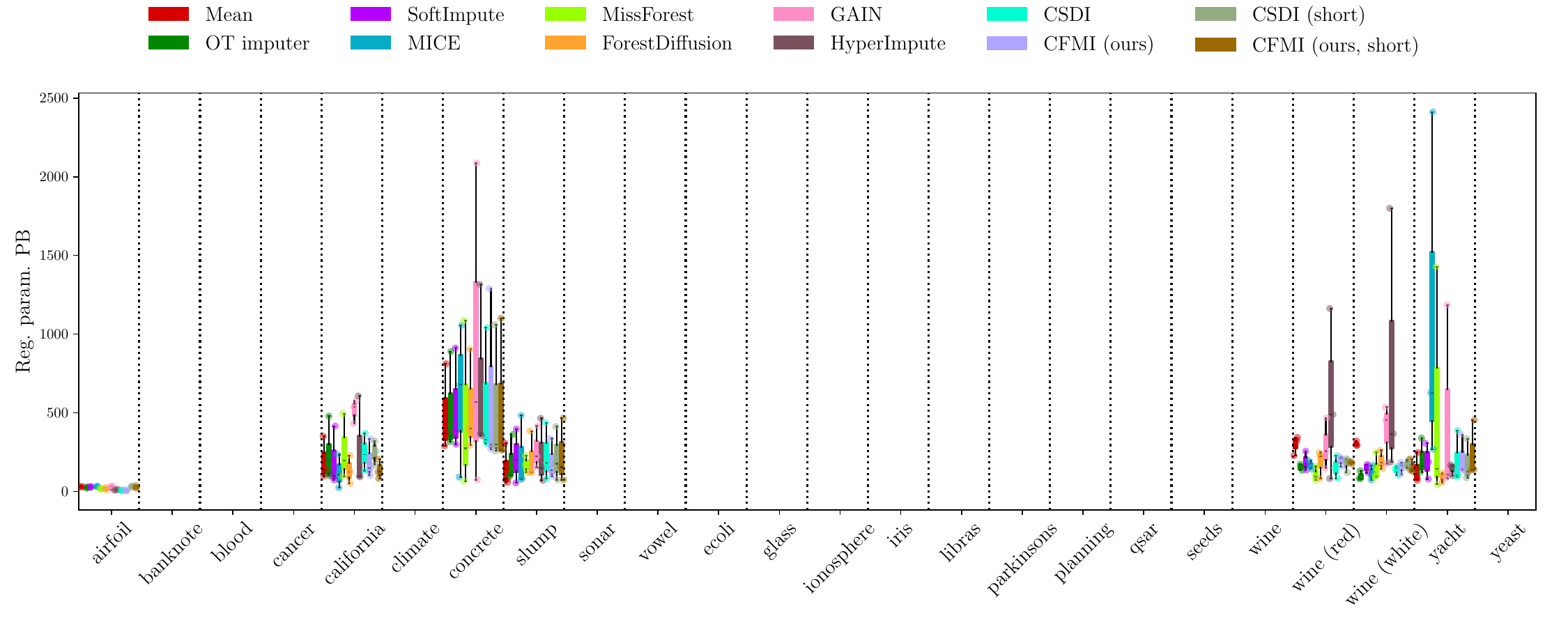}
  \caption{Regression parameter percent bias results: box-plot of 3 runs. MCAR 50\% missingness.}
\end{figure}

\begin{figure}[H]
  \centering
  \includegraphics[width=0.9\linewidth]{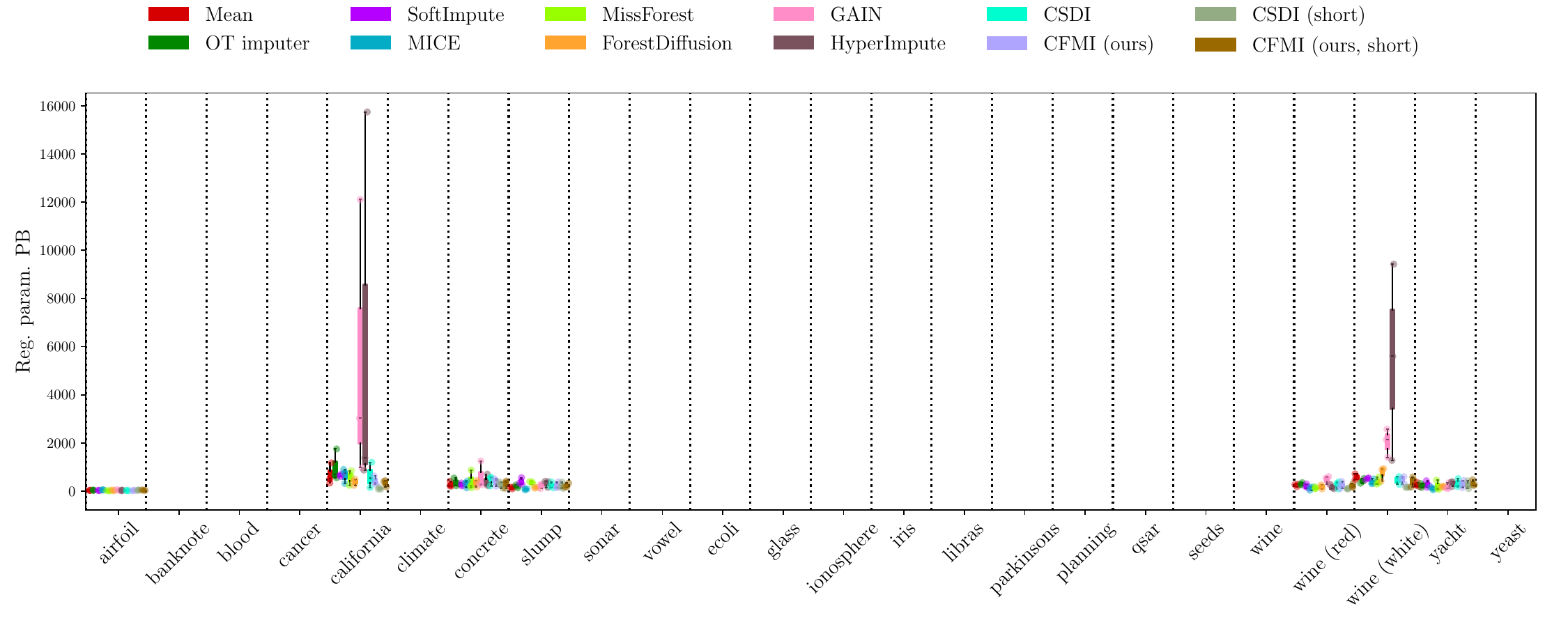}
  \caption{Regression parameter percent bias results: box-plot of 3 runs. MCAR 75\% missingness.}
\end{figure}

\begin{figure}[H]
  \centering
  \includegraphics[width=0.9\linewidth]{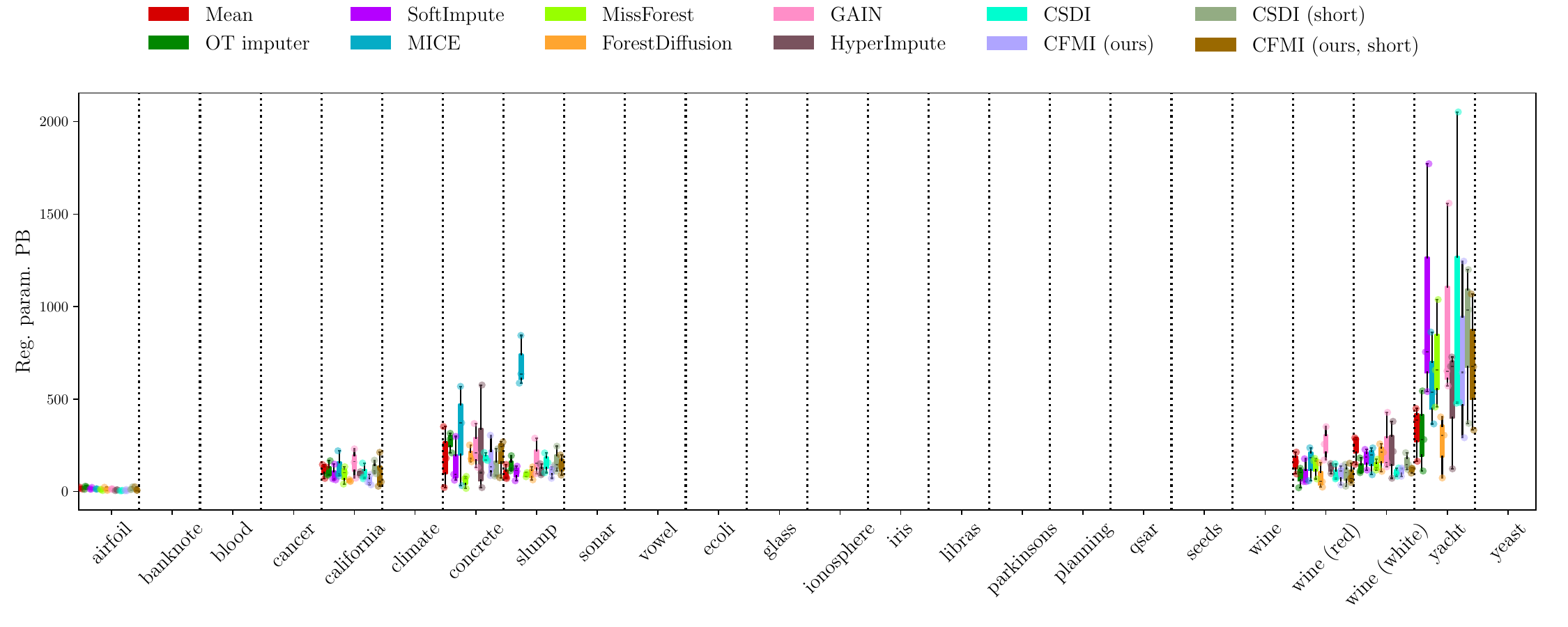}
  \caption{Regression parameter percent bias results: box-plot of 3 runs. MAR 25\% missingness.}
\end{figure}

\begin{figure}[H]
  \centering
  \includegraphics[width=0.9\linewidth]{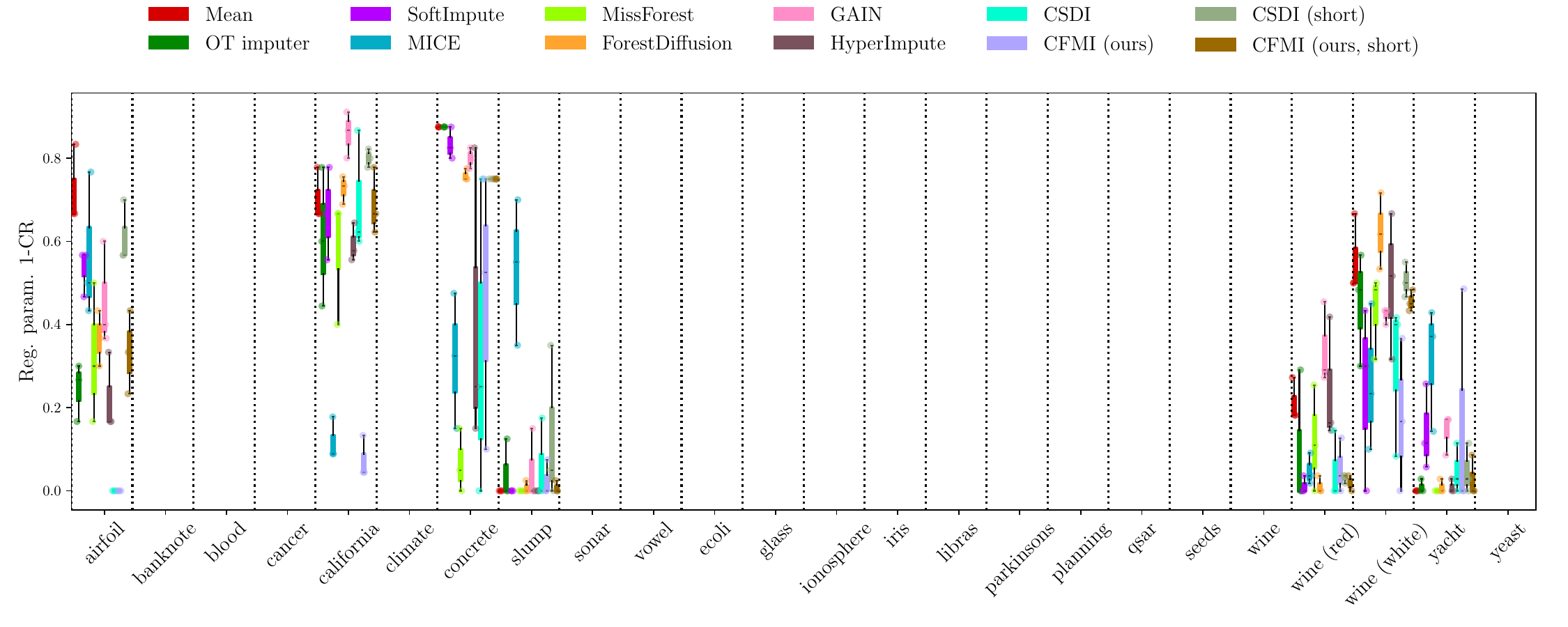}
  \caption{Regression parameter 1-CR results: box-plot of 3 runs. MCAR 25\% missingness.}
\end{figure}

\begin{figure}[H]
  \centering
  \includegraphics[width=0.9\linewidth]{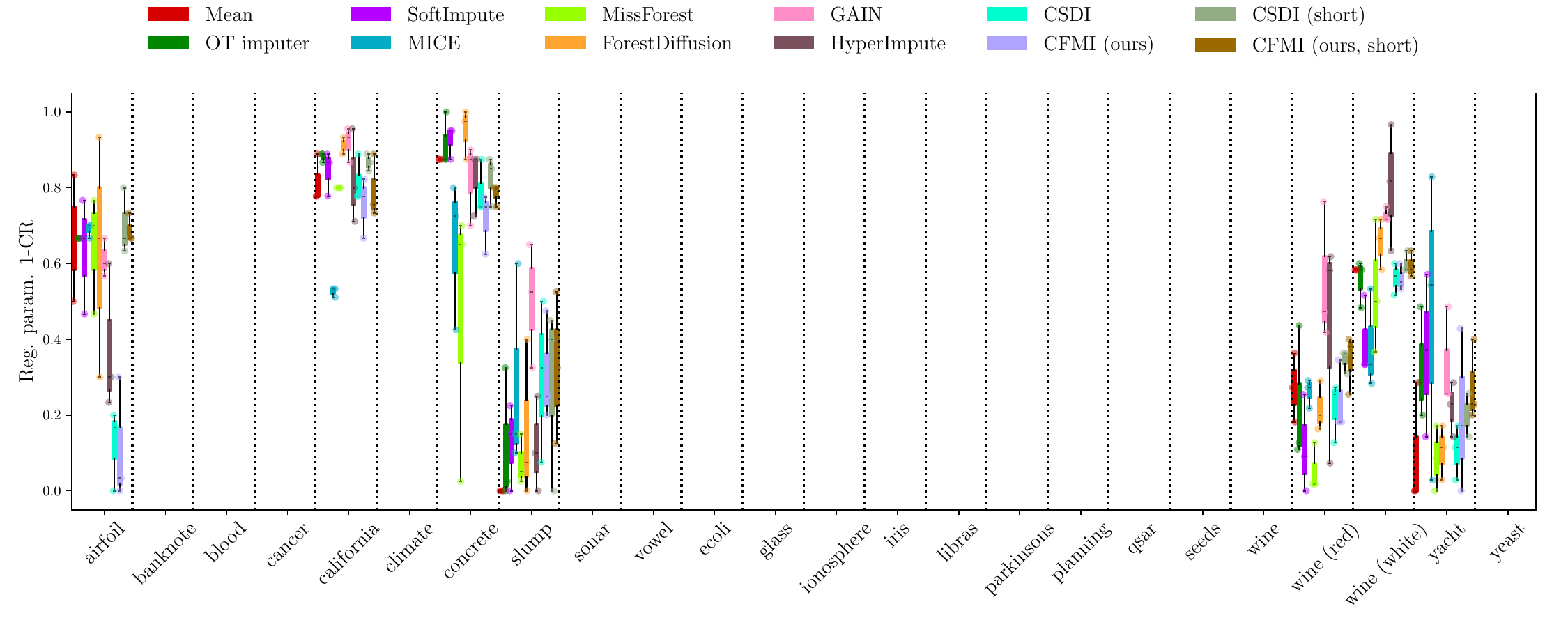}
  \caption{Regression parameter 1-CR results: box-plot of 3 runs. MCAR 50\% missingness.}
\end{figure}

\begin{figure}[H]
  \centering
  \includegraphics[width=0.9\linewidth]{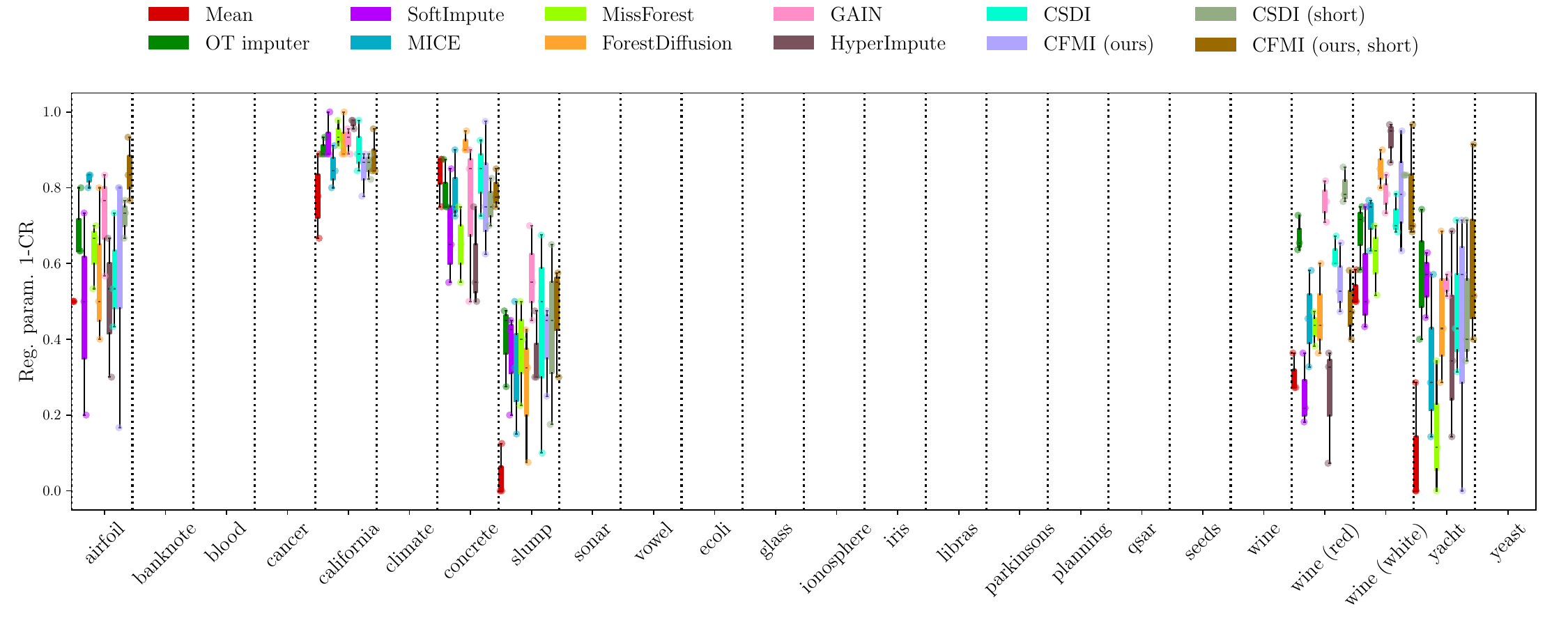}
  \caption{Regression parameter 1-CR results: box-plot of 3 runs. MCAR 75\% missingness.}
\end{figure}

\begin{figure}[H]
  \centering
  \includegraphics[width=0.9\linewidth]{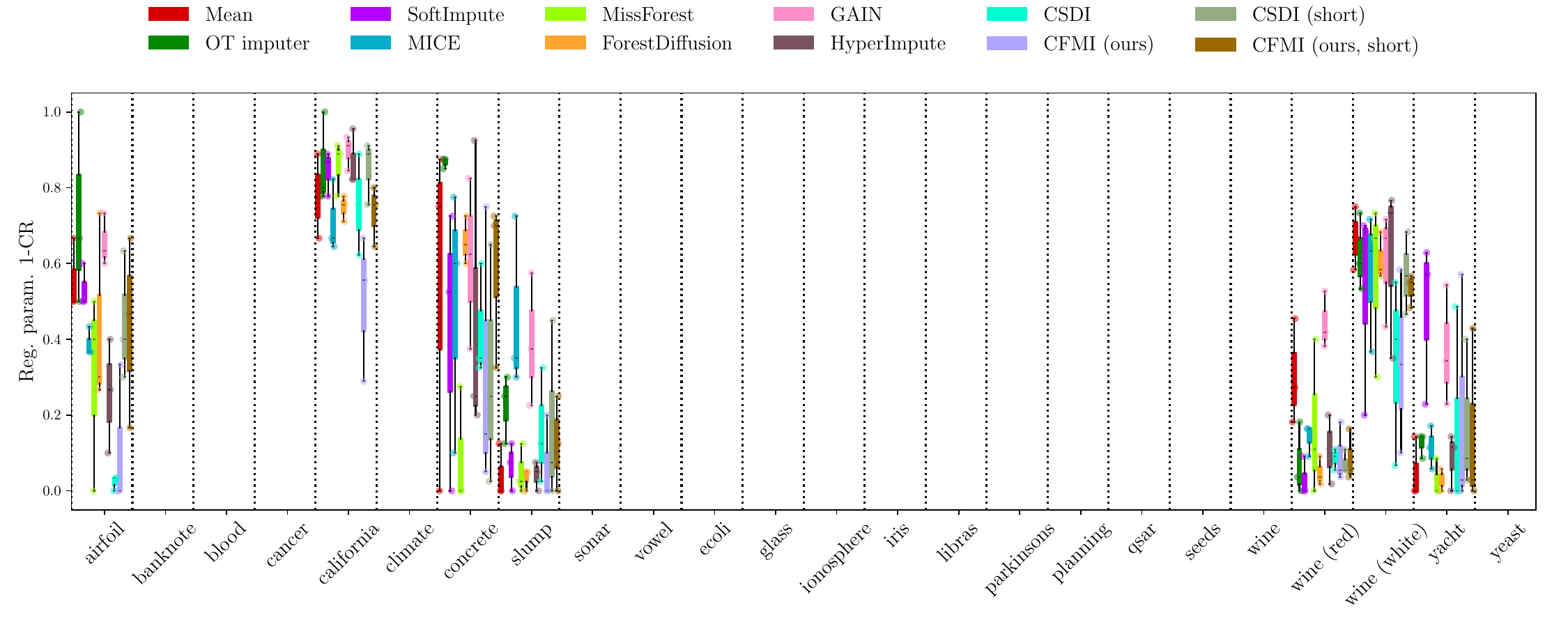}
  \caption{Regression parameter 1-CR results: box-plot of 3 runs. MAR 25\% missingness.}
\end{figure}

\begin{figure}[H]
  \centering
  \includegraphics[width=0.9\linewidth]{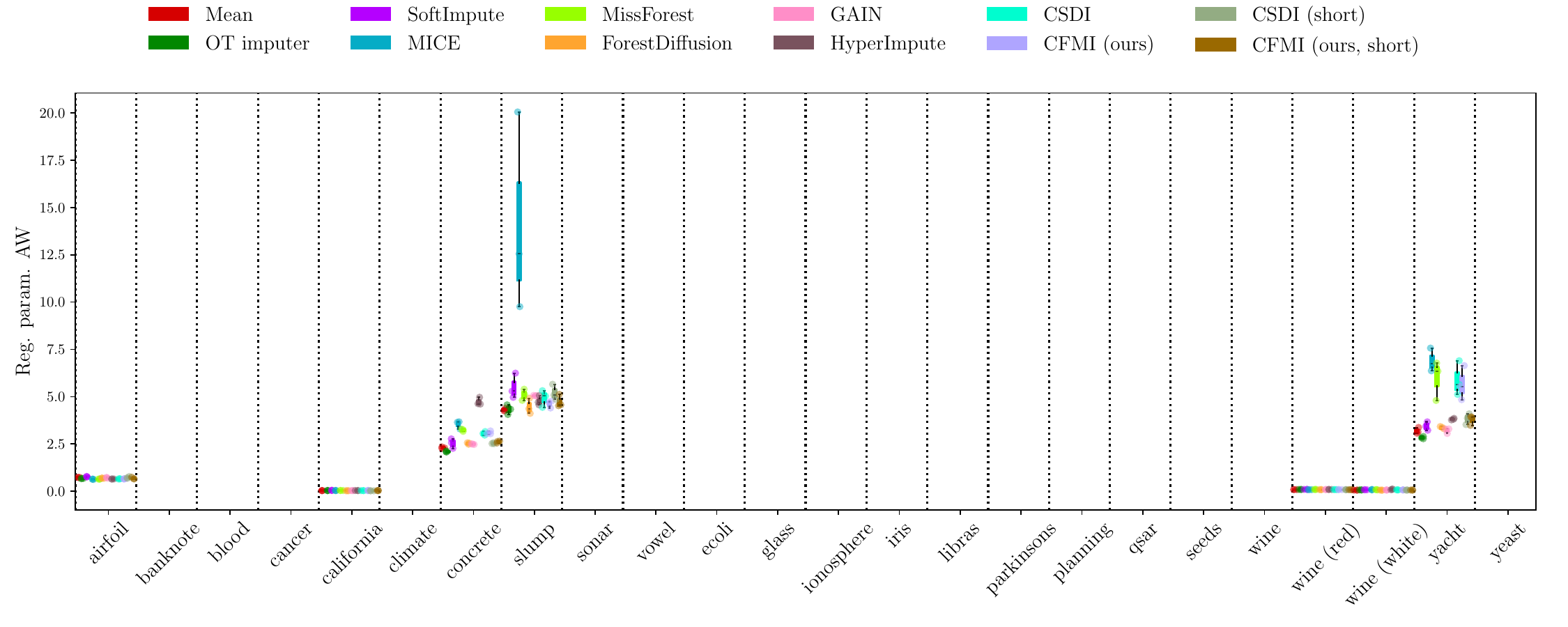}
  \caption{Regression parameter AW results: box-plot of 3 runs. MCAR 25\% missingness.}
\end{figure}

\begin{figure}[H]
  \centering
  \includegraphics[width=0.9\linewidth]{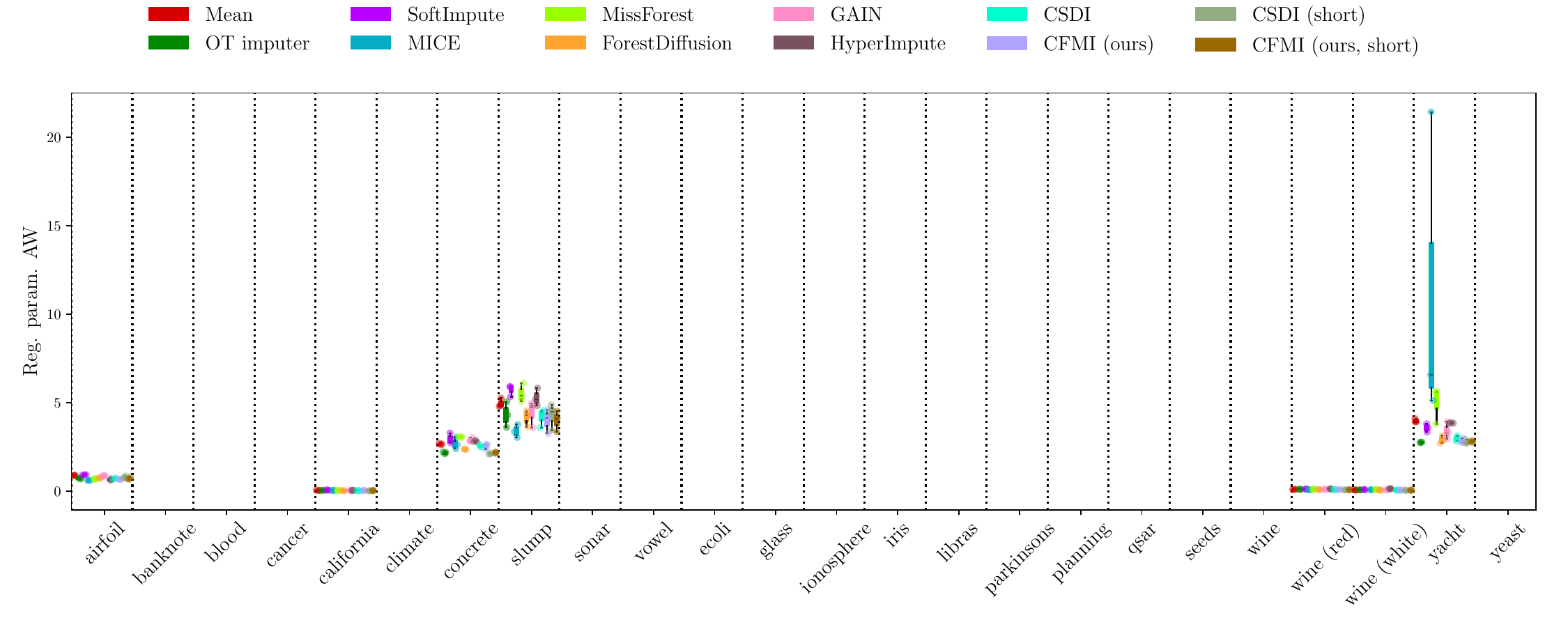}
  \caption{Regression parameter AW results: box-plot of 3 runs. MCAR 50\% missingness.}
\end{figure}

\begin{figure}[H]
  \centering
  \includegraphics[width=0.9\linewidth]{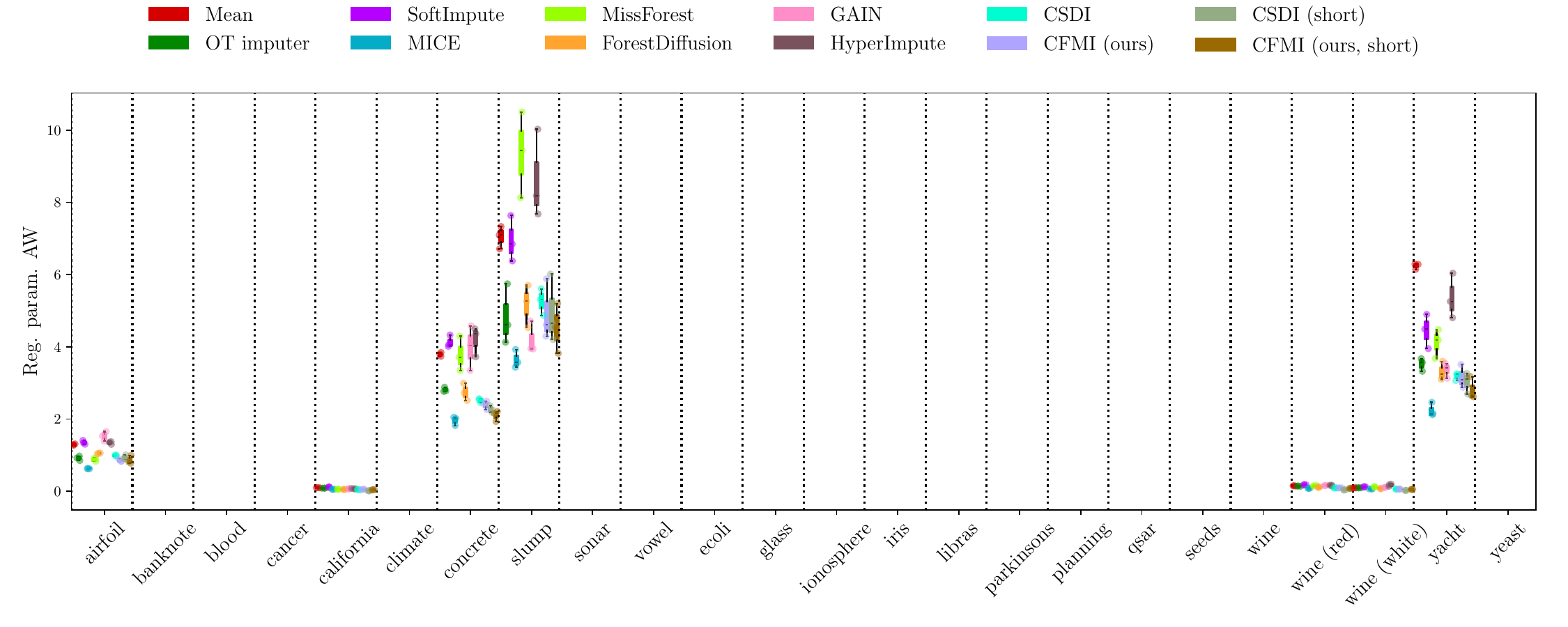}
  \caption{Regression parameter AW results: box-plot of 3 runs. MCAR 75\% missingness.}
\end{figure}

\begin{figure}[H]
  \centering
  \includegraphics[width=0.9\linewidth]{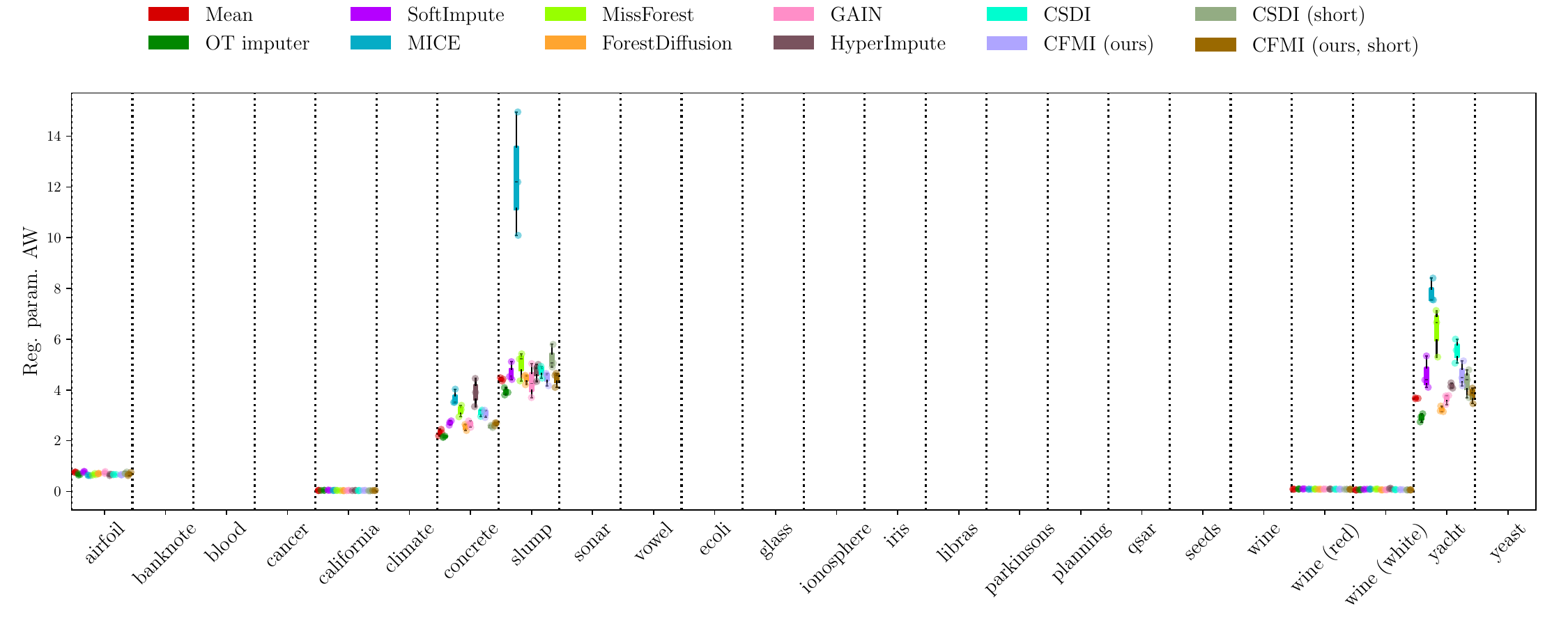}
  \caption{Regression parameter AW results: box-plot of 3 runs. MAR 25\% missingness.}
\end{figure}

\subsubsection{Average relative performance matrices}
\label{apx:uci-relperformance}

\begin{minipage}{.48\textwidth}
\begin{figure}[H]
  \centering
  \includegraphics[width=1.\linewidth]{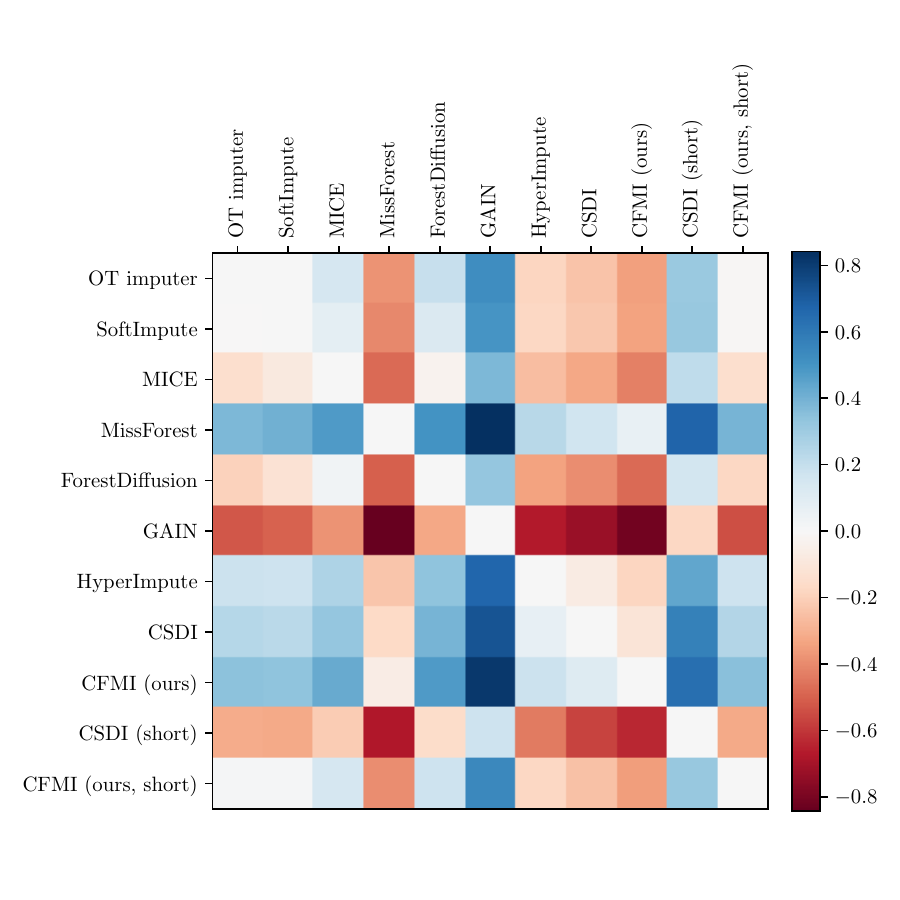}
  \caption{Wasserstein-2 results: average relative distance (blue means row method outperforms the column method). MCAR 25\% missingness.}
\end{figure}
\end{minipage}
\begin{minipage}{.48\textwidth}
\begin{figure}[H]
  \centering
  \includegraphics[width=1.\linewidth]{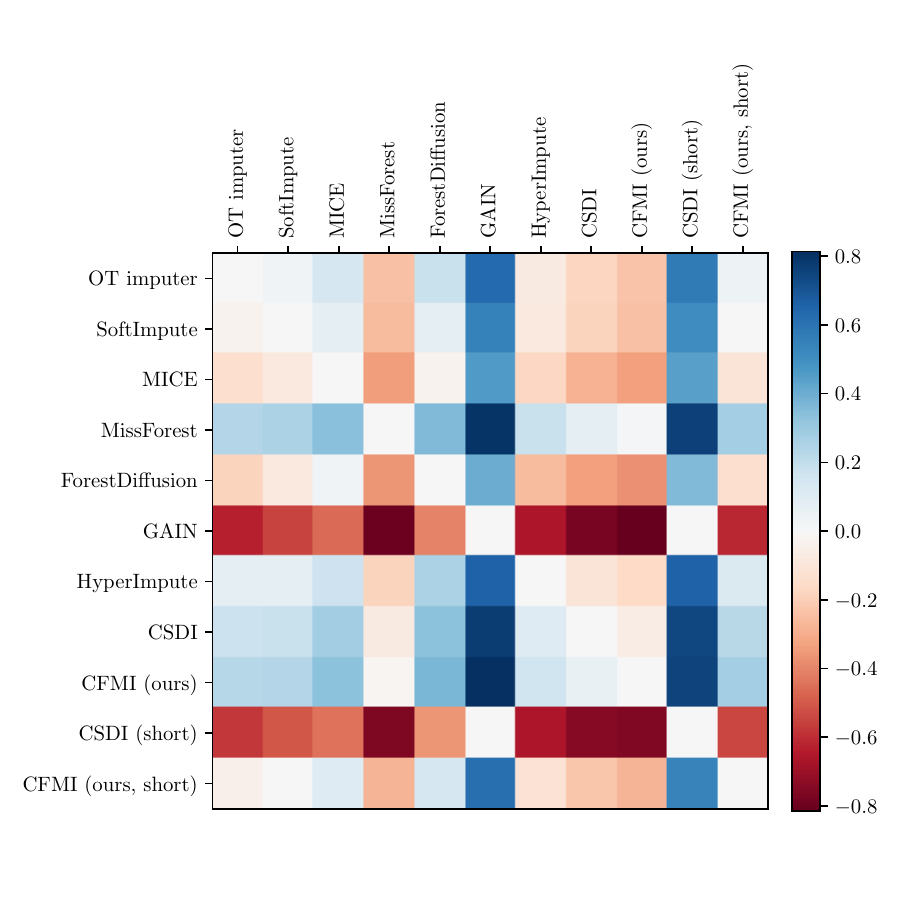}
  \caption{Wasserstein-2 results: average relative distance (blue means row method outperforms the column method). MCAR 50\% missingness.}
\end{figure}
\end{minipage}
\begin{minipage}{.48\textwidth}
\begin{figure}[H]
  \centering
  \includegraphics[width=1.\linewidth]{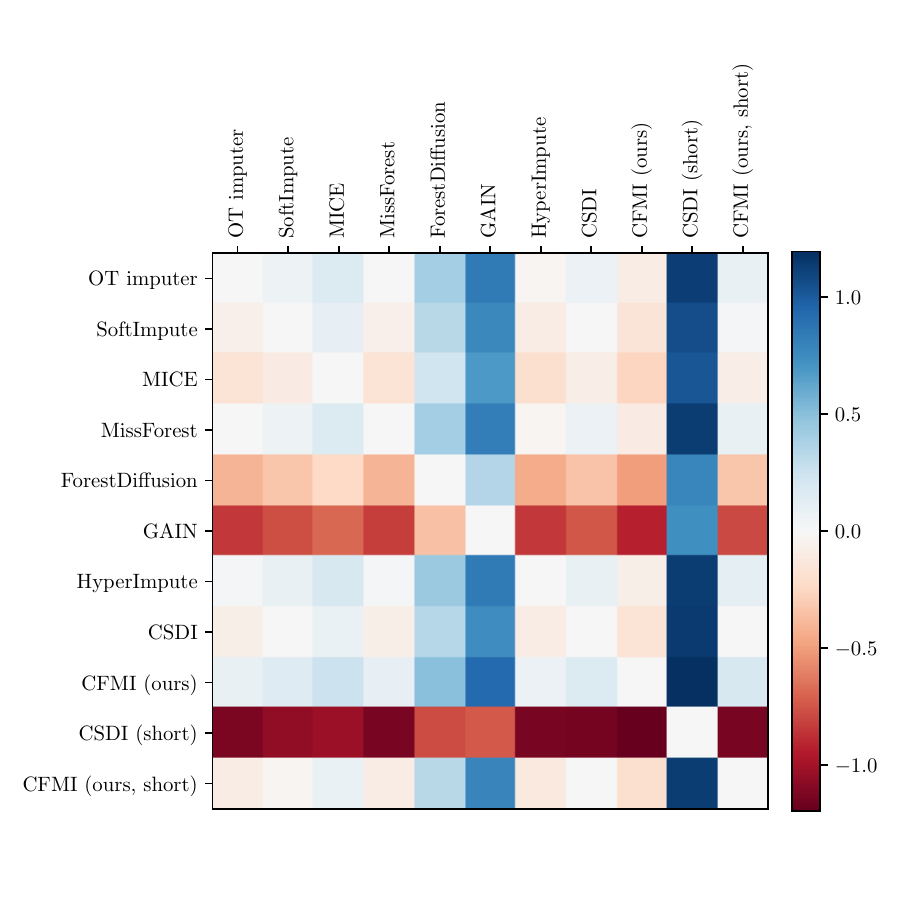}
  \caption{Wasserstein-2 results: average relative distance (blue means row method outperforms the column method). MCAR 75\% missingness.}
\end{figure}
\end{minipage}
\begin{minipage}{.48\textwidth}
\begin{figure}[H]
  \centering
  \includegraphics[width=1.\linewidth]{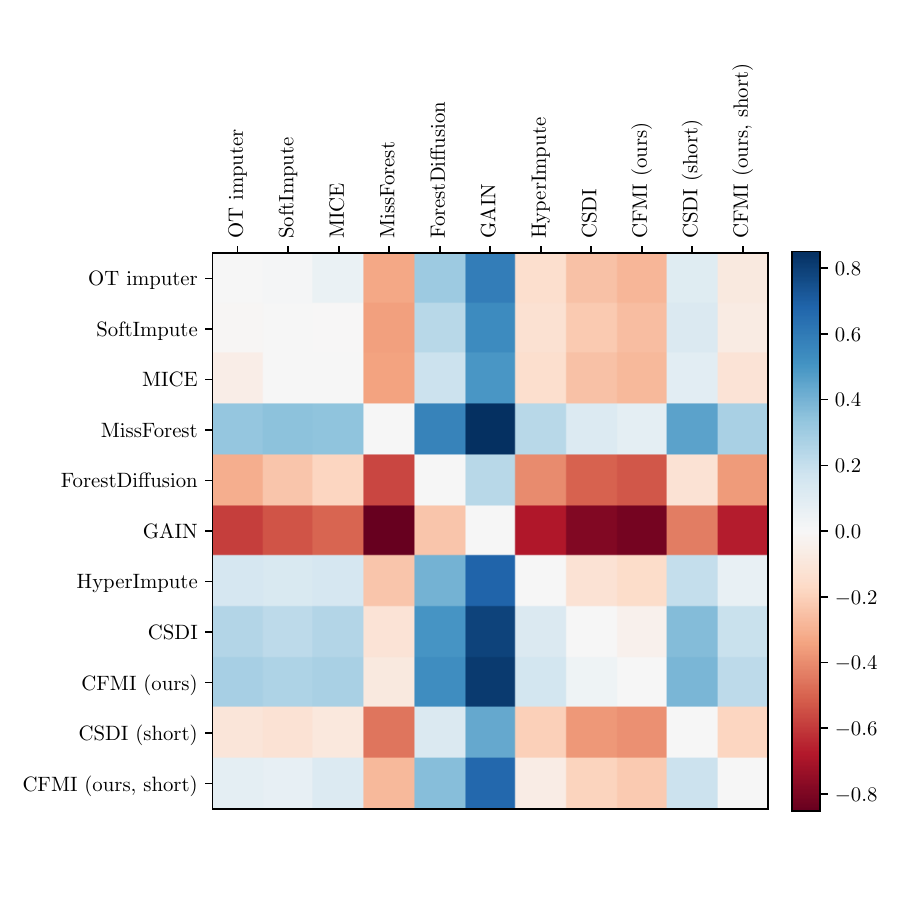}
  \caption{Wasserstein-2 results: average relative distance (blue means row method outperforms the column method). MAR 25\% missingness.}
\end{figure}
\end{minipage}
\begin{minipage}{.48\textwidth}
\begin{figure}[H]
  \centering
  \includegraphics[width=1.\linewidth]{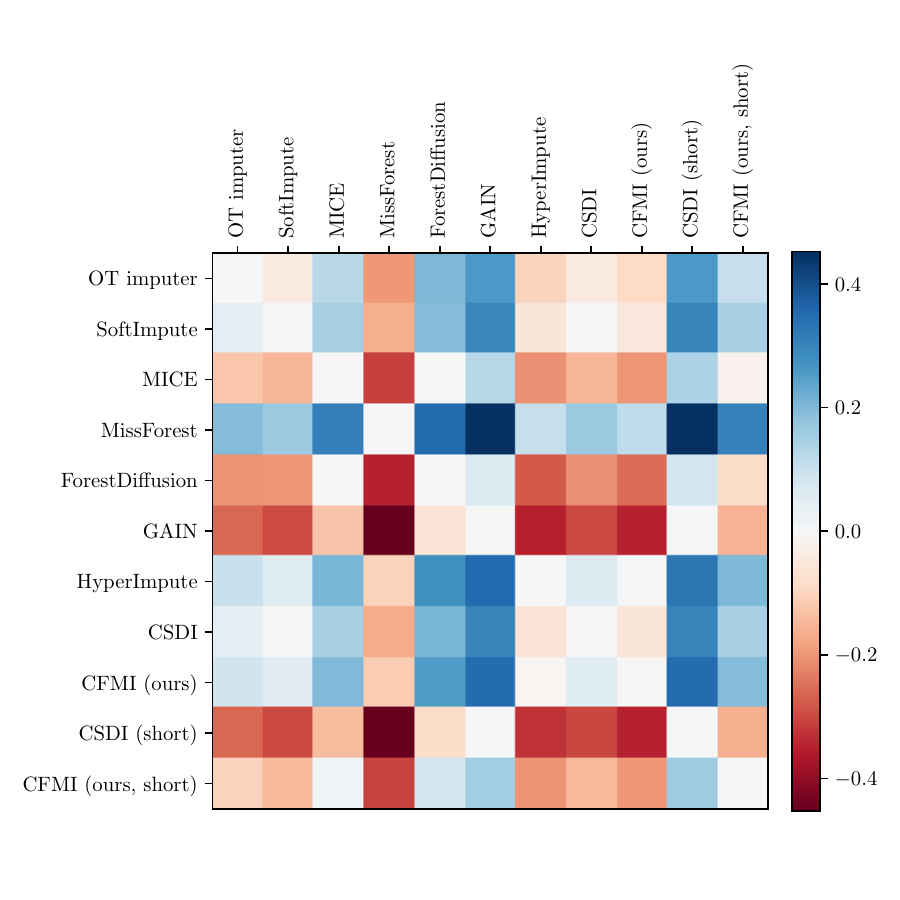}
  \caption{Average RMSE results: average relative distance (blue means row method outperforms the column method). MCAR 25\% missingness.}
\end{figure}
\end{minipage}
\begin{minipage}{.48\textwidth}
\begin{figure}[H]
  \centering
  \includegraphics[width=1.\linewidth]{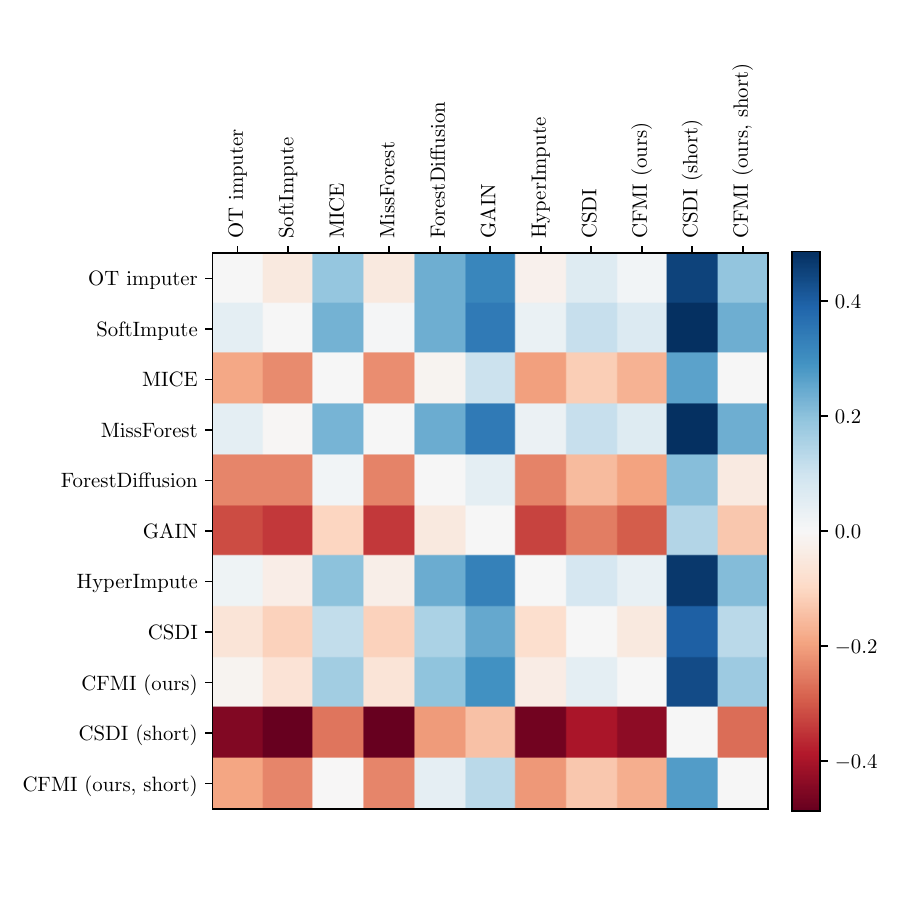}
  \caption{Average RMSE results: average relative distance (blue means row method outperforms the column method). MCAR 50\% missingness.}
\end{figure}
\end{minipage}
\begin{minipage}{.48\textwidth}
\begin{figure}[H]
  \centering
  \includegraphics[width=1.\linewidth]{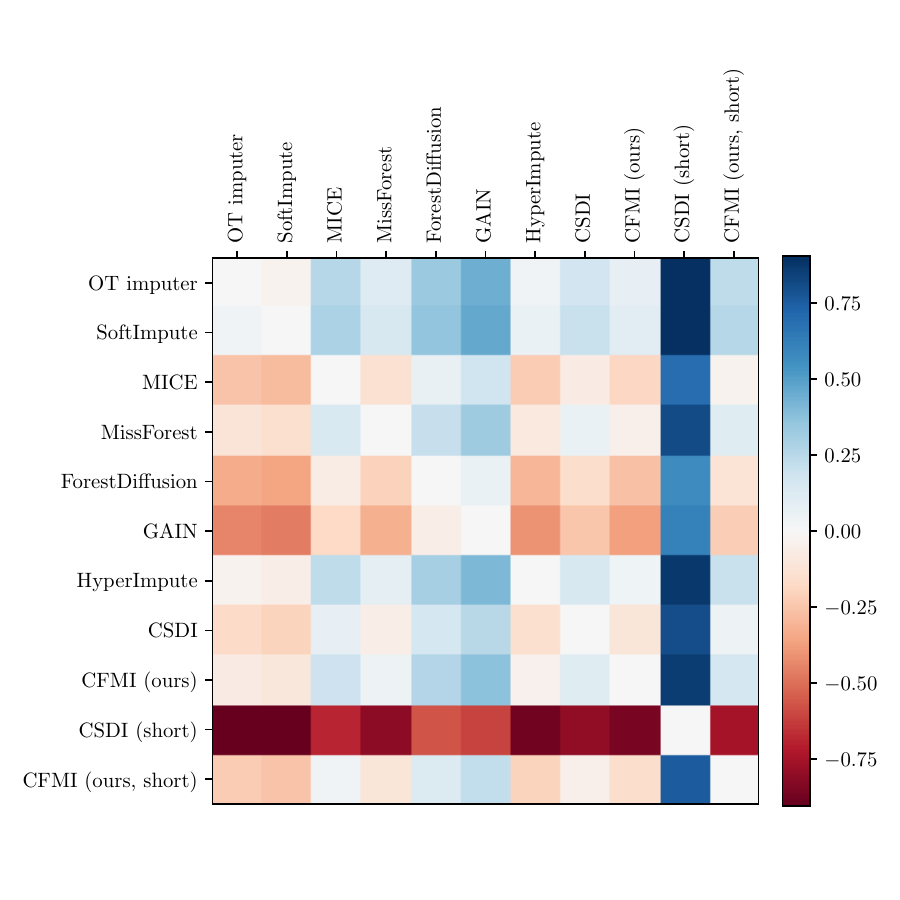}
  \caption{Average RMSE results: average relative distance (blue means row method outperforms the column method). MCAR 75\% missingness.}
\end{figure}
\end{minipage}
\begin{minipage}{.48\textwidth}
\begin{figure}[H]
  \centering
  \includegraphics[width=1.\linewidth]{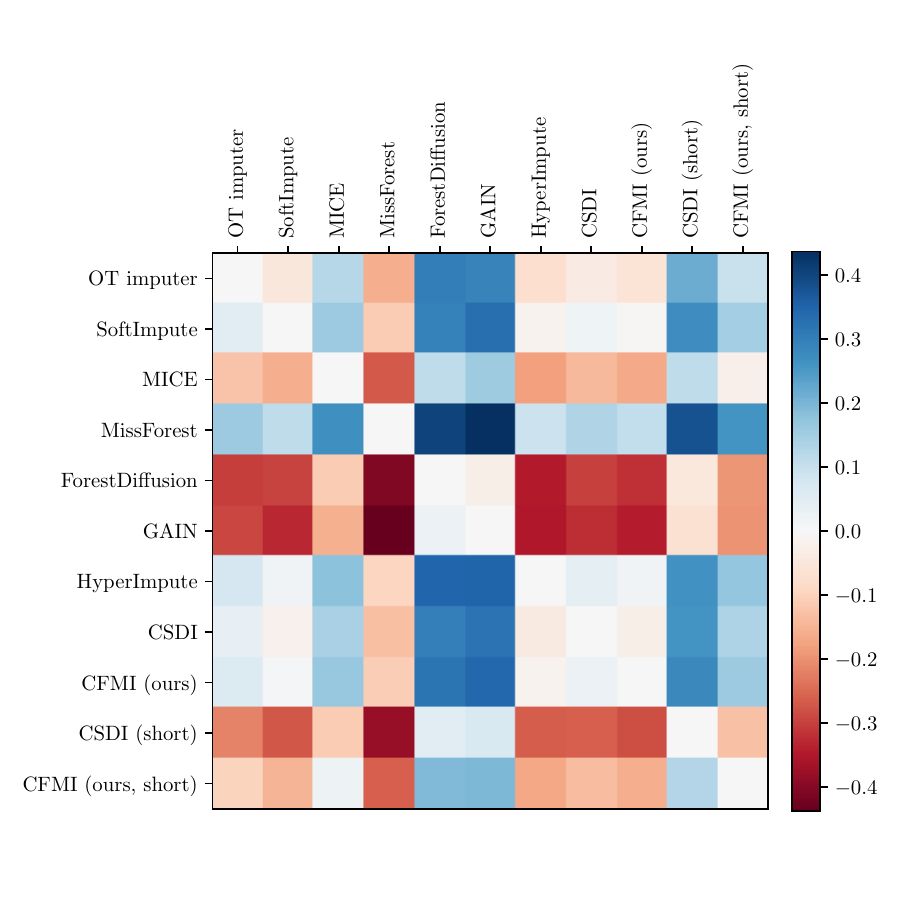}
  \caption{Average RMSE results: average relative distance (blue means row method outperforms the column method). MAR 25\% missingness.}
\end{figure}
\end{minipage}
\begin{minipage}{.48\textwidth}
\begin{figure}[H]
  \centering
  \includegraphics[width=1.\linewidth]{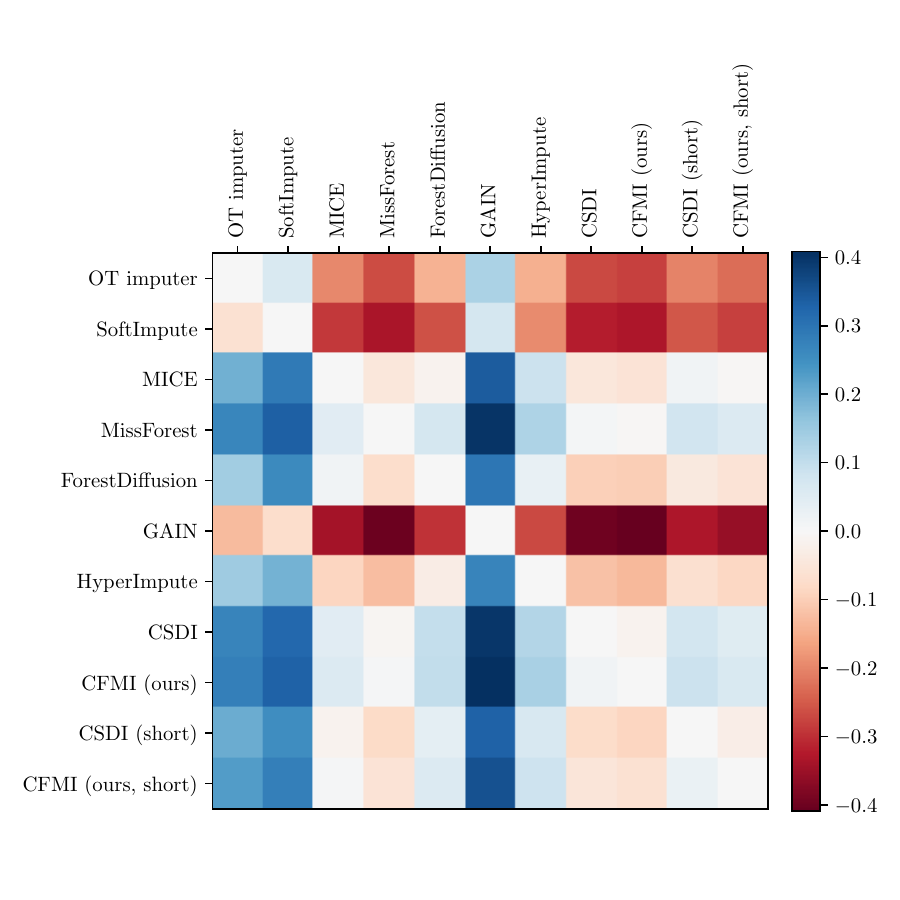}
  \caption{CRPS results: average relative distance (blue means row method outperforms the column method). MCAR 25\% missingness.}
\end{figure}
\end{minipage}
\begin{minipage}{.48\textwidth}
\begin{figure}[H]
  \centering
  \includegraphics[width=1.\linewidth]{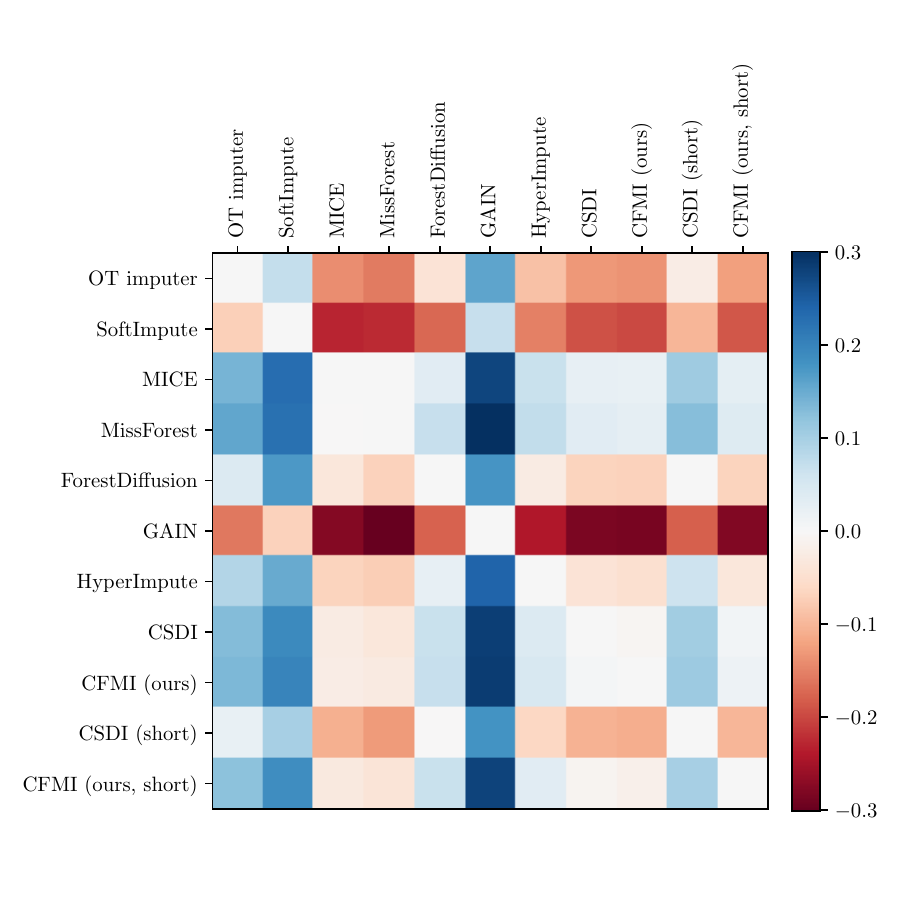}
  \caption{CRPS results: average relative distance (blue means row method outperforms the column method). MCAR 50\% missingness.}
\end{figure}
\end{minipage}
\begin{minipage}{.48\textwidth}
\begin{figure}[H]
  \centering
  \includegraphics[width=1.\linewidth]{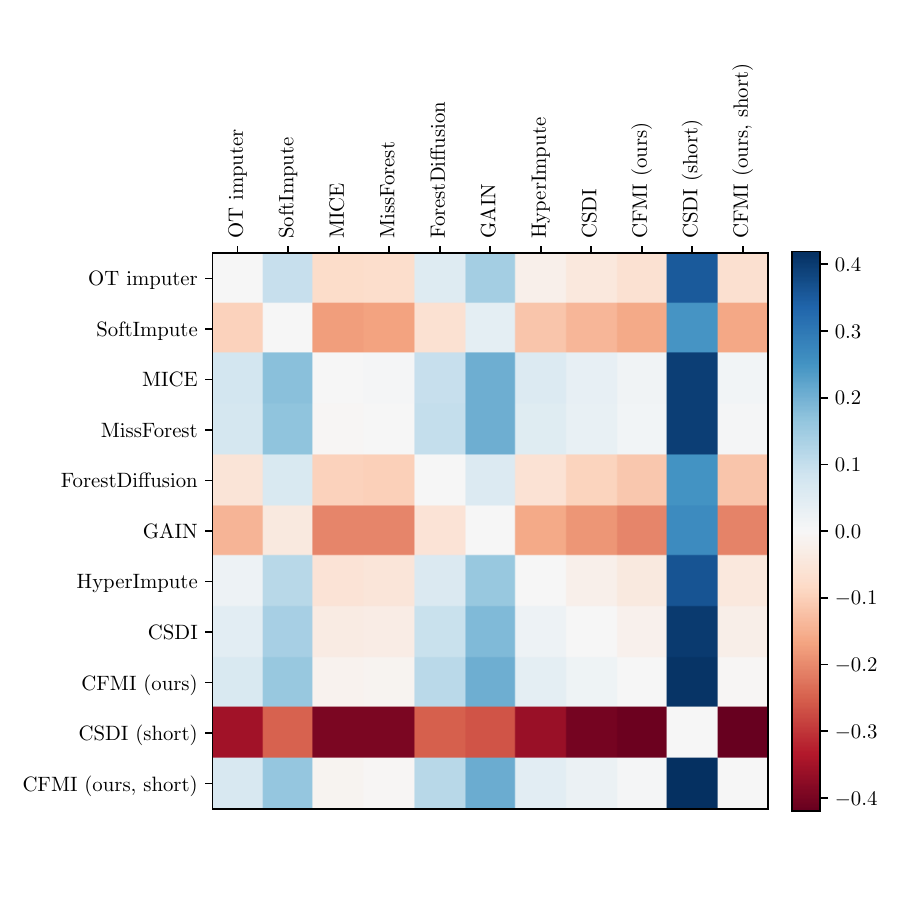}
  \caption{CRPS results: average relative distance (blue means row method outperforms the column method). MCAR 75\% missingness.}
\end{figure}
\end{minipage}
\begin{minipage}{.48\textwidth}
\begin{figure}[H]
  \centering
  \includegraphics[width=1.\linewidth]{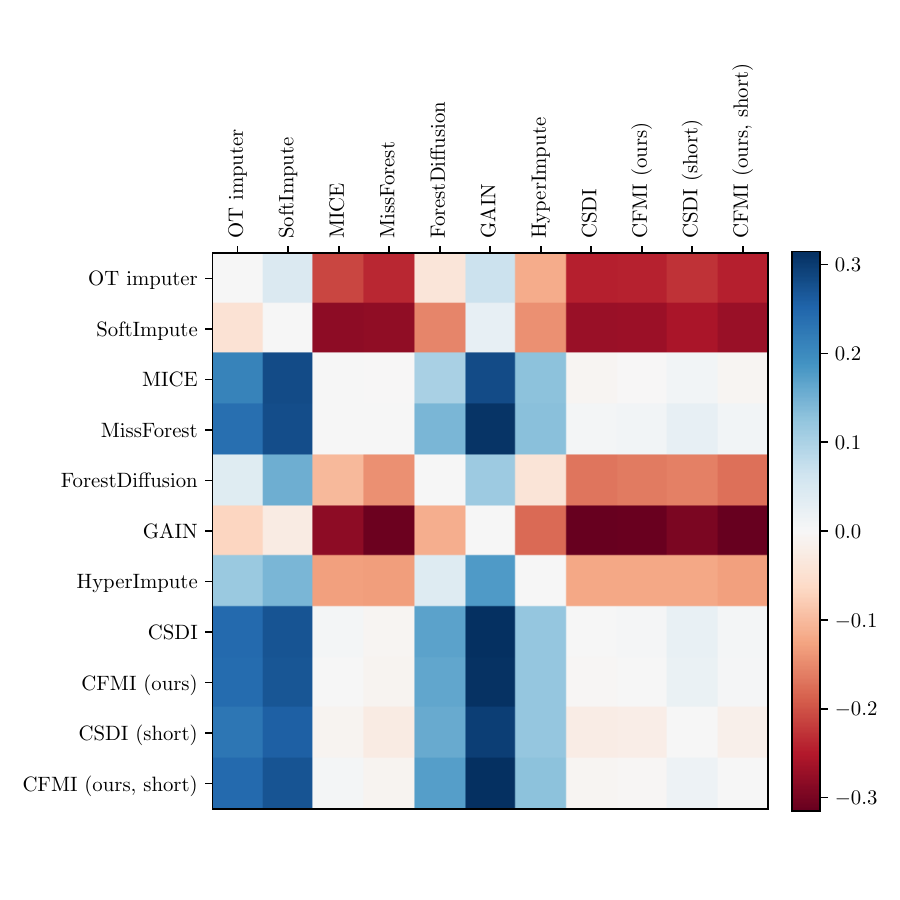}
  \caption{CRPS results: average relative distance (blue means row method outperforms the column method). MAR 25\% missingness.}
\end{figure}
\end{minipage}
\begin{minipage}{.48\textwidth}
\begin{figure}[H]
  \centering
  \includegraphics[width=1.\linewidth]{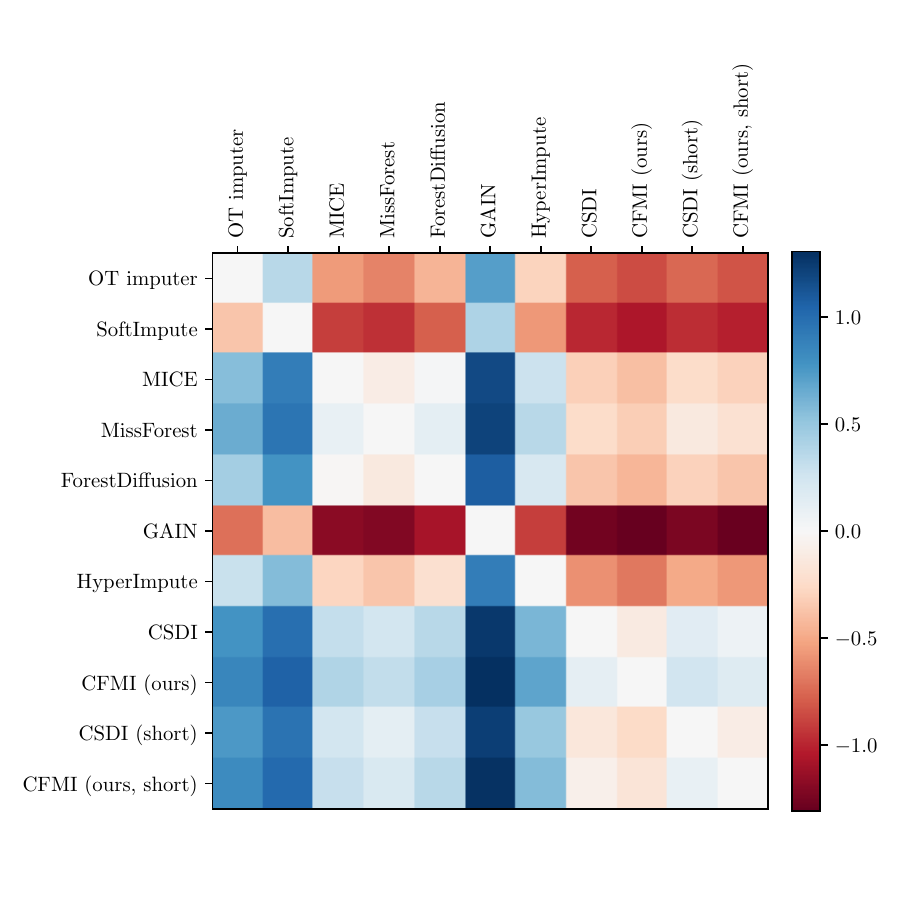}
  \caption{Energy MMD results: average relative distance (blue means row method outperforms the column method). MCAR 25\% missingness.}
\end{figure}
\end{minipage}
\begin{minipage}{.48\textwidth}
\begin{figure}[H]
  \centering
  \includegraphics[width=1.\linewidth]{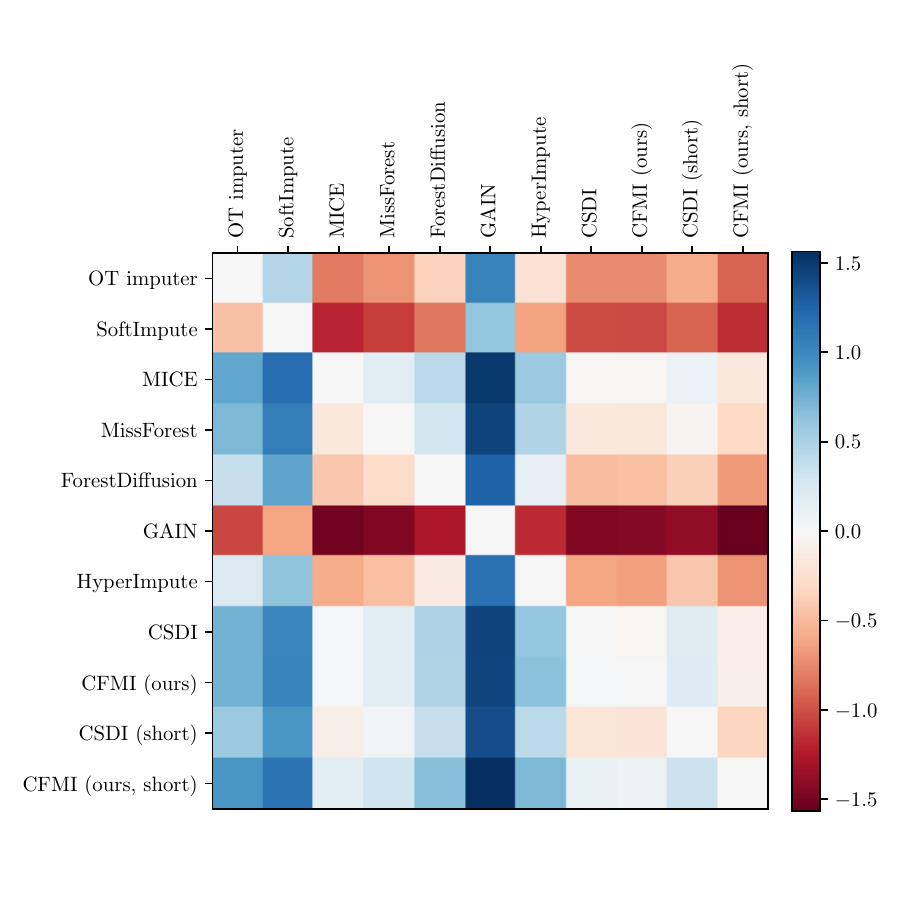}
  \caption{Energy MMD results: average relative distance (blue means row method outperforms the column method). MCAR 50\% missingness.}
\end{figure}
\end{minipage}
\begin{minipage}{.48\textwidth}
\begin{figure}[H]
  \centering
  \includegraphics[width=1.\linewidth]{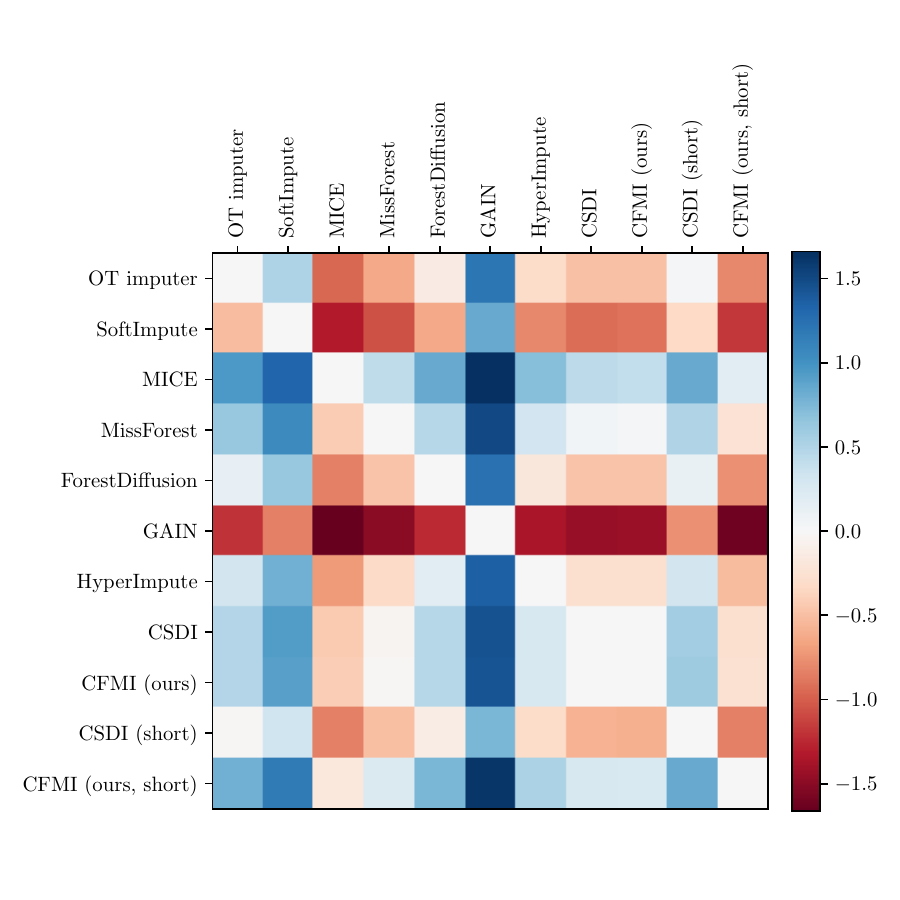}
  \caption{Energy MMD results: average relative distance (blue means row method outperforms the column method). MCAR 75\% missingness.}
\end{figure}
\end{minipage}
\begin{minipage}{.48\textwidth}
\begin{figure}[H]
  \centering
  \includegraphics[width=1.\linewidth]{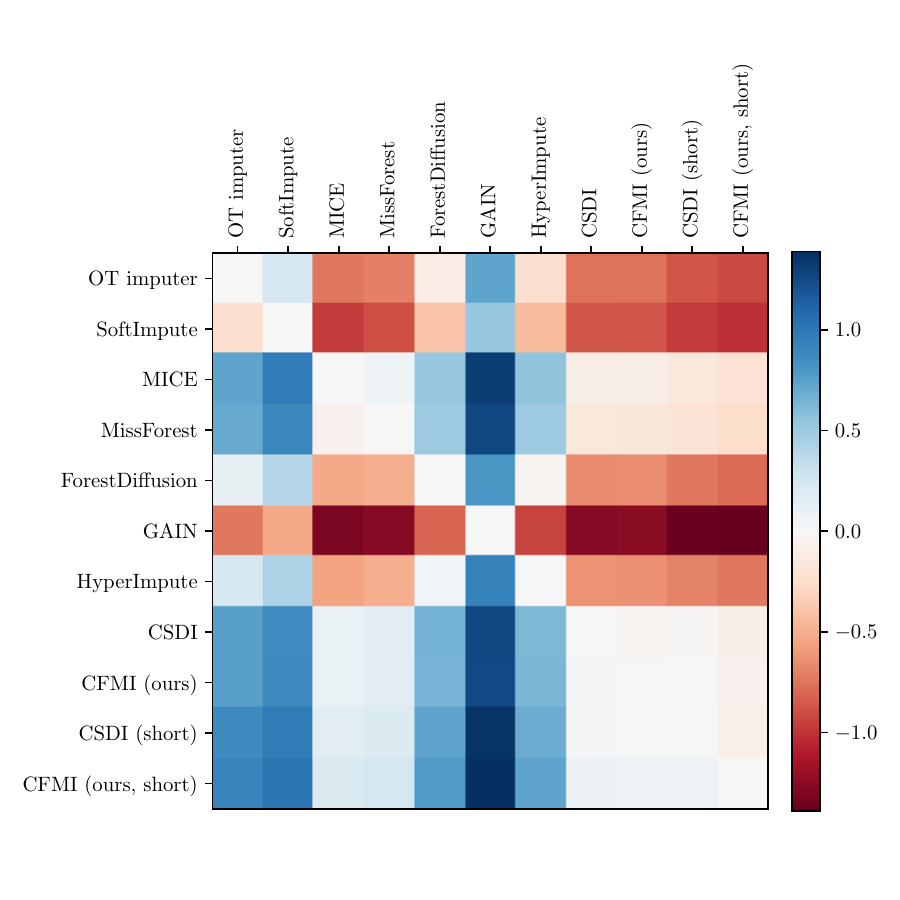}
  \caption{Energy MMD results: average relative distance (blue means row method outperforms the column method). MAR 25\% missingness.}
\end{figure}
\end{minipage}
\begin{minipage}{.48\textwidth}
\begin{figure}[H]
  \centering
  \includegraphics[width=1.\linewidth]{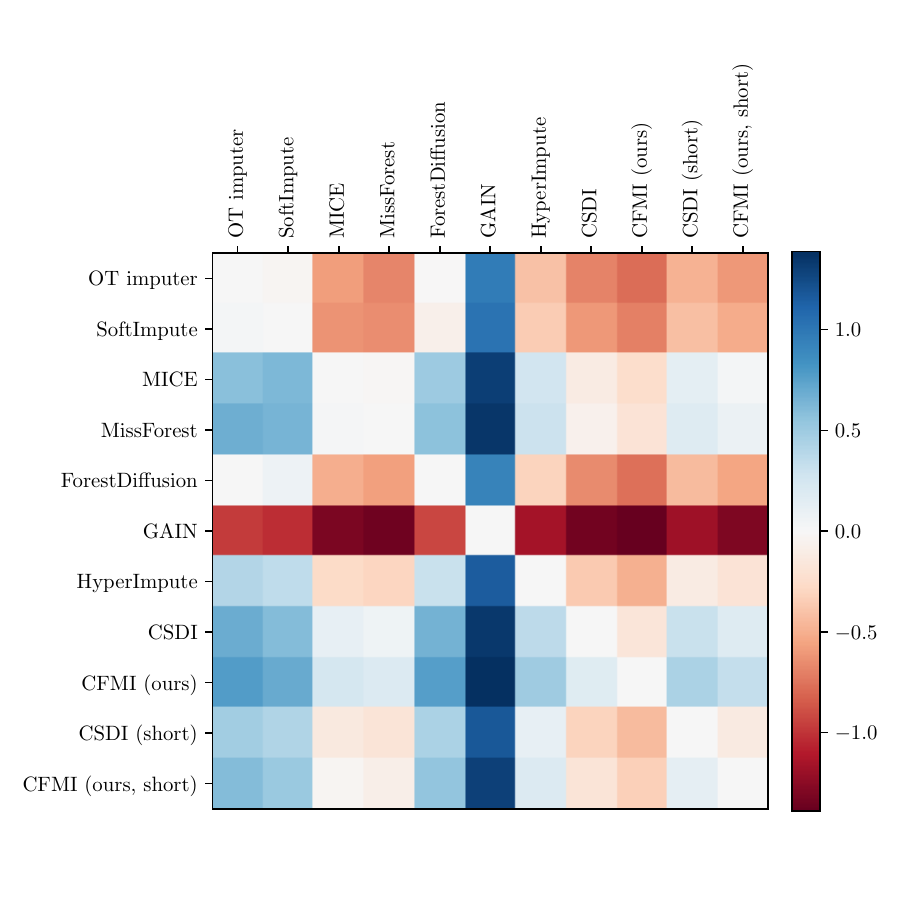}
  \caption{Gaussian MMD results: average relative distance (blue means row method outperforms the column method). MCAR 25\% missingness.}
\end{figure}
\end{minipage}
\begin{minipage}{.48\textwidth}
\begin{figure}[H]
  \centering
  \includegraphics[width=1.\linewidth]{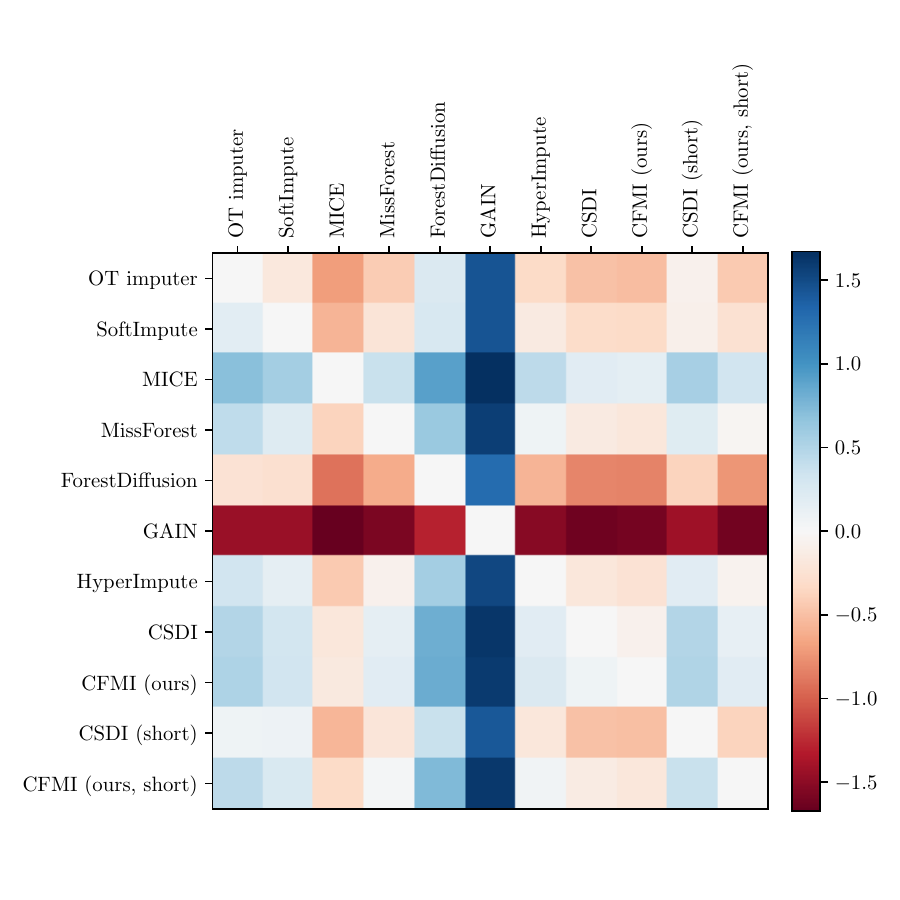}
  \caption{Gaussian MMD results: average relative distance (blue means row method outperforms the column method). MCAR 50\% missingness.}
\end{figure}
\end{minipage}
\begin{minipage}{.48\textwidth}
\begin{figure}[H]
  \centering
  \includegraphics[width=1.\linewidth]{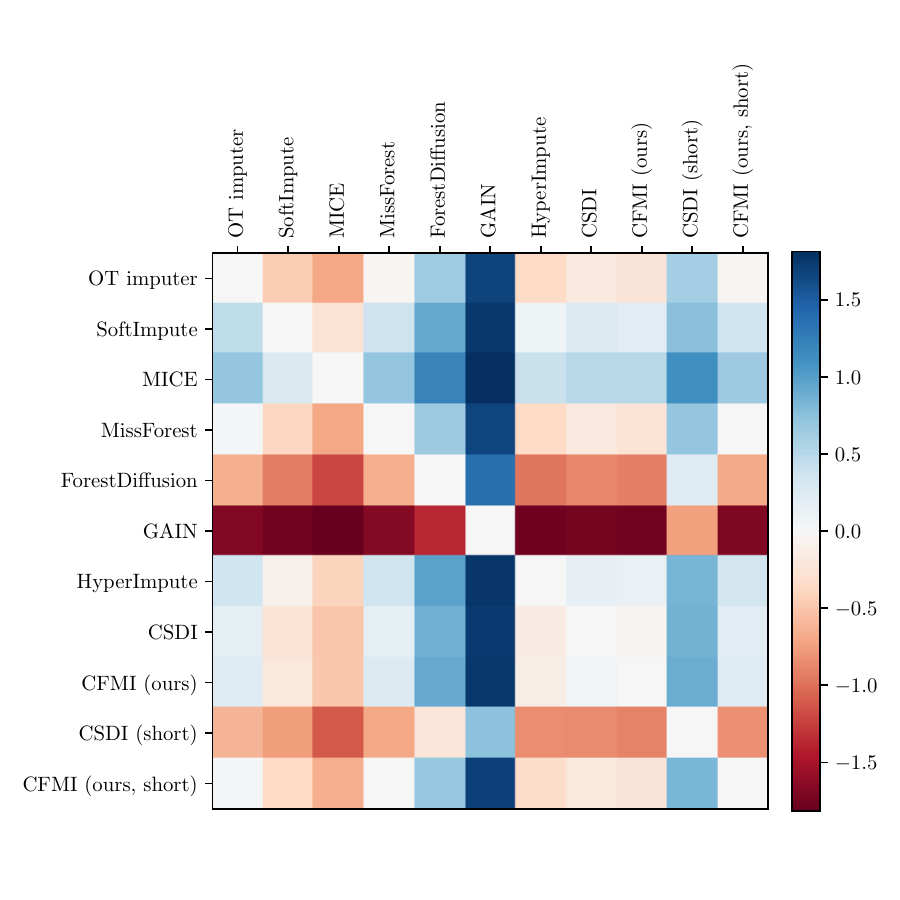}
  \caption{Gaussian MMD results: average relative distance (blue means row method outperforms the column method). MCAR 75\% missingness.}
\end{figure}
\end{minipage}
\begin{minipage}{.48\textwidth}
\begin{figure}[H]
  \centering
  \includegraphics[width=1.\linewidth]{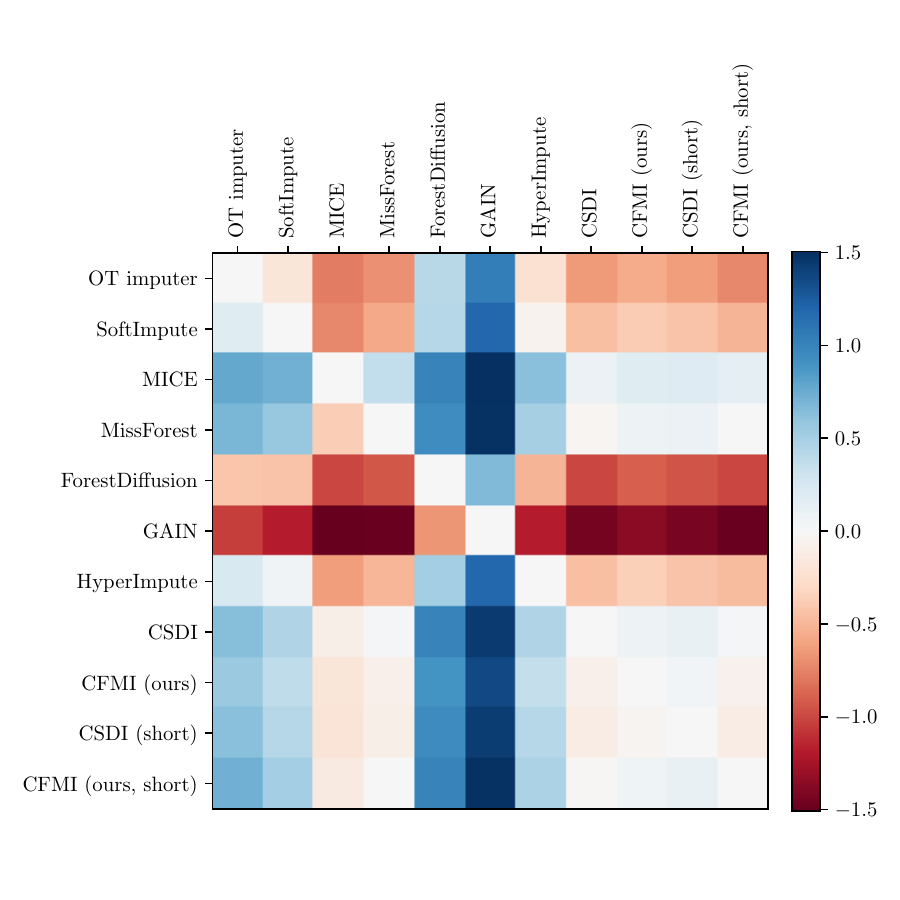}
  \caption{Gaussian MMD results: average relative distance (blue means row method outperforms the column method). MAR 25\% missingness.}
\end{figure}
\end{minipage}
\begin{minipage}{.48\textwidth}
\begin{figure}[H]
  \centering
  \includegraphics[width=1.\linewidth]{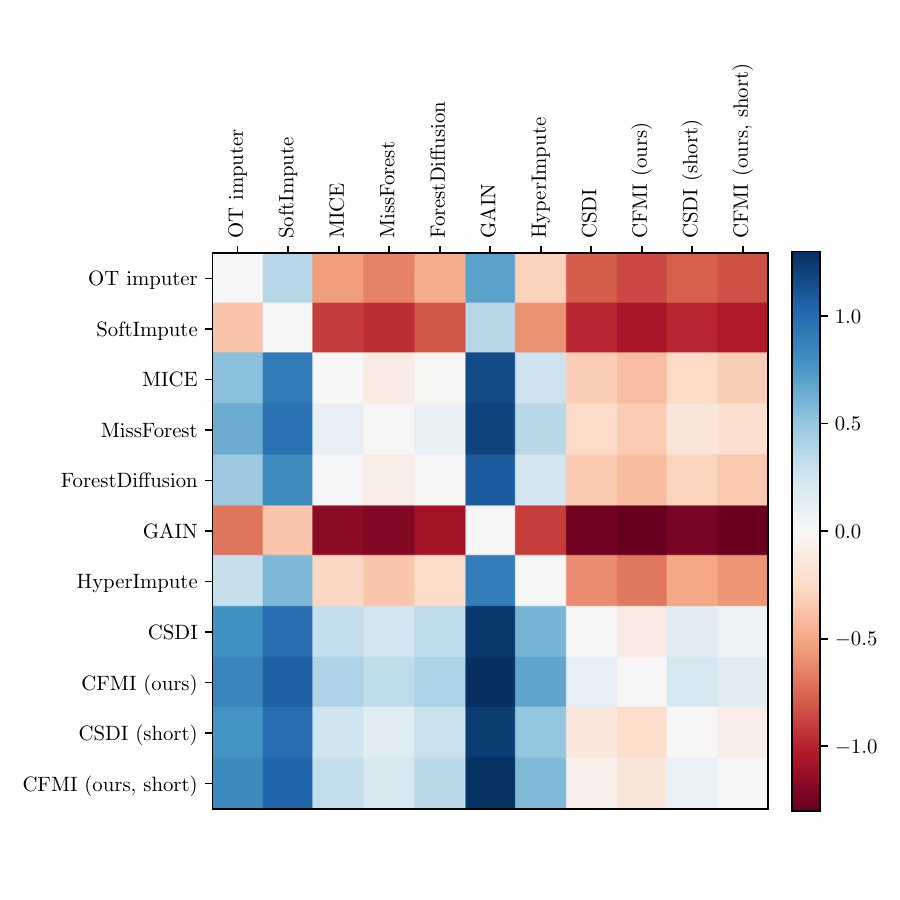}
  \caption{Laplacian MMD results: average relative distance (blue means row method outperforms the column method). MCAR 25\% missingness.}
\end{figure}
\end{minipage}
\begin{minipage}{.48\textwidth}
\begin{figure}[H]
  \centering
  \includegraphics[width=1.\linewidth]{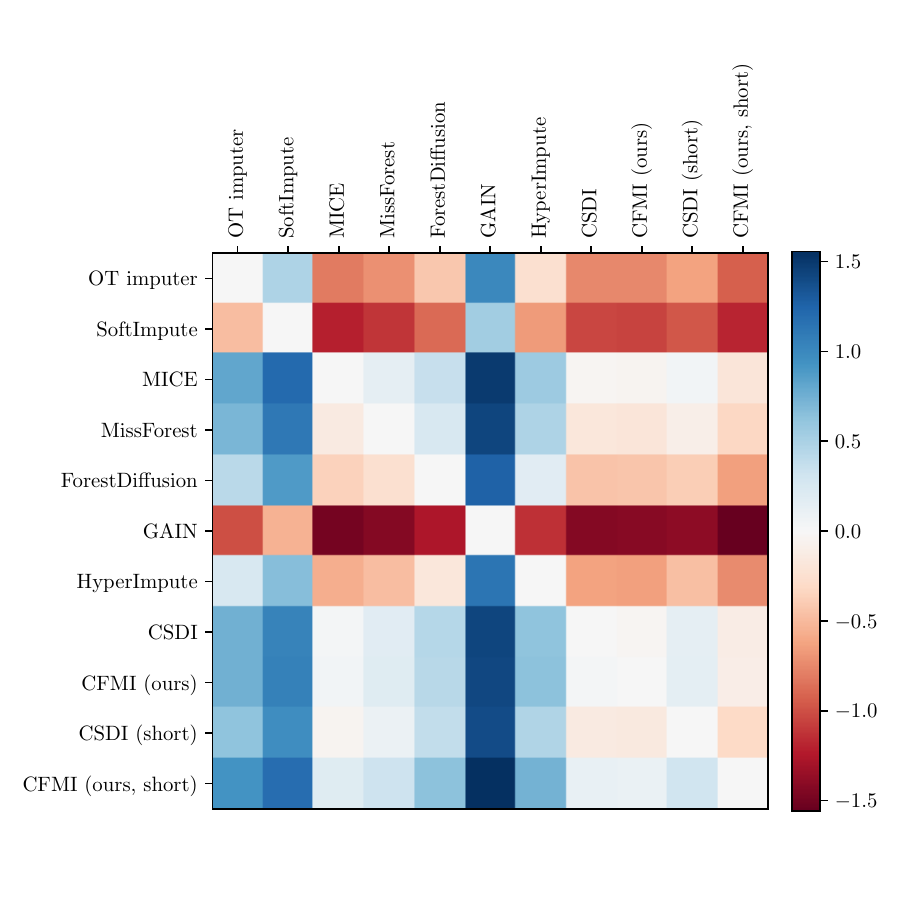}
  \caption{Laplacian MMD results: average relative distance (blue means row method outperforms the column method). MCAR 50\% missingness.}
\end{figure}
\end{minipage}
\begin{minipage}{.48\textwidth}
\begin{figure}[H]
  \centering
  \includegraphics[width=1.\linewidth]{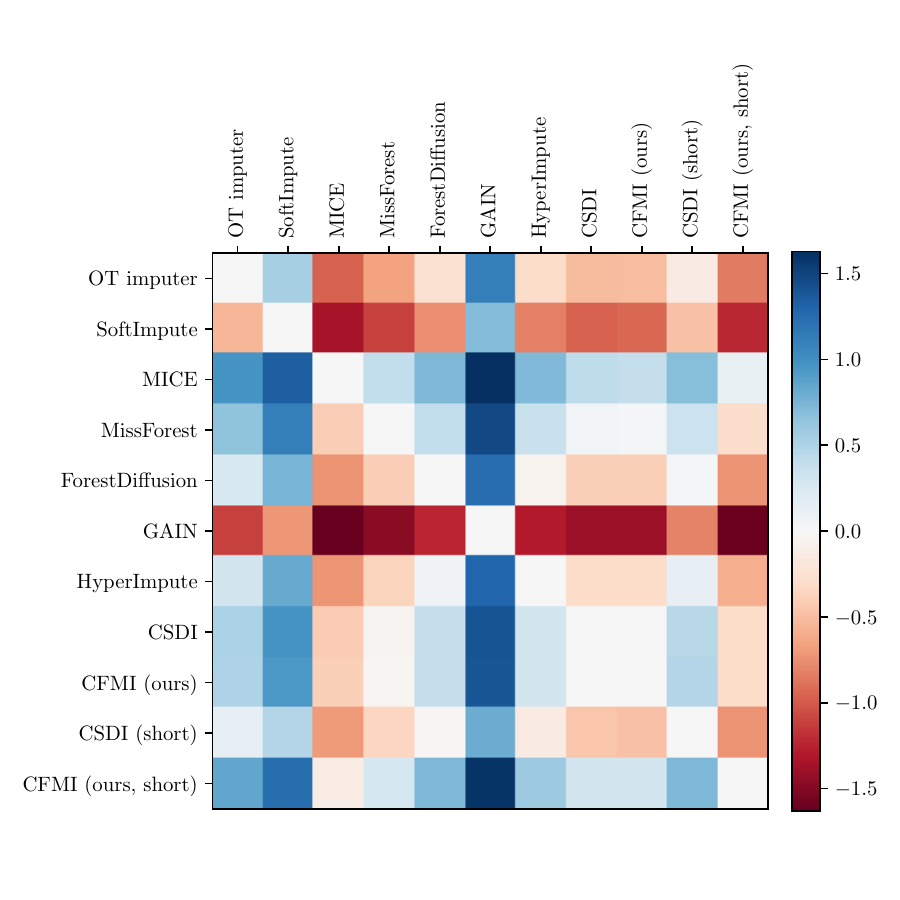}
  \caption{Laplacian MMD results: average relative distance (blue means row method outperforms the column method). MCAR 75\% missingness.}
\end{figure}
\end{minipage}
\begin{minipage}{.48\textwidth}
\begin{figure}[H]
  \centering
  \includegraphics[width=1.\linewidth]{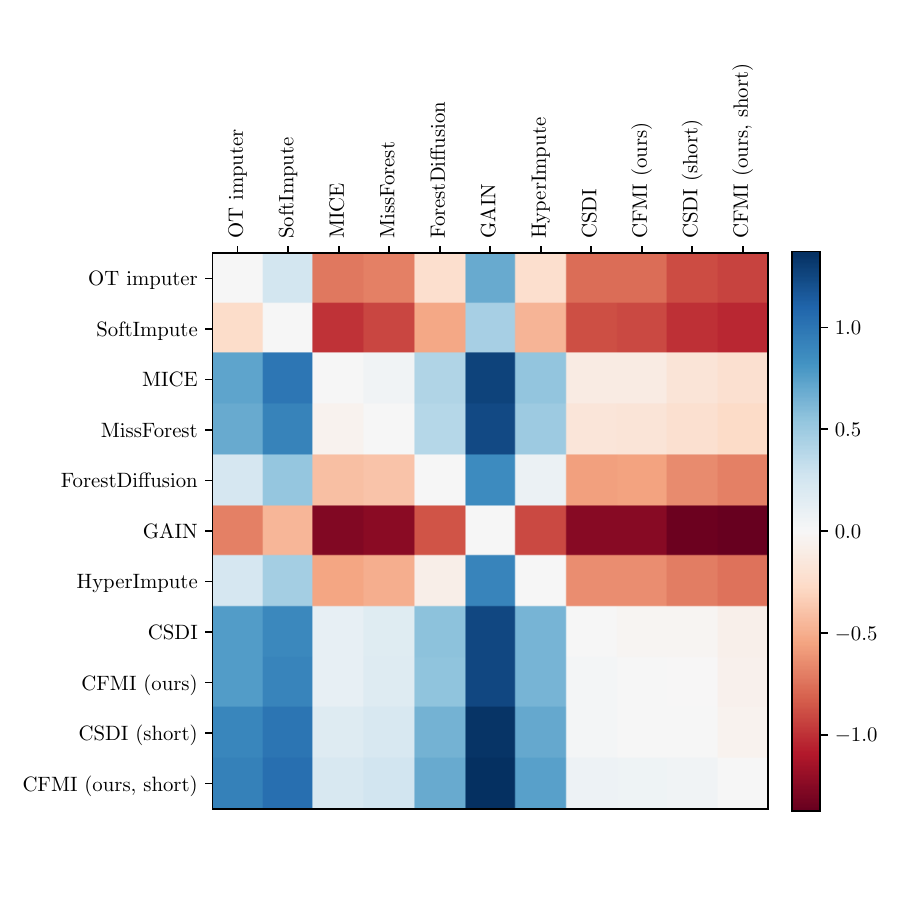}
  \caption{Laplacian MMD results: average relative distance (blue means row method outperforms the column method). MAR 25\% missingness.}
\end{figure}
\end{minipage}
\begin{minipage}{.48\textwidth}
\begin{figure}[H]
  \centering
  \includegraphics[width=1.\linewidth]{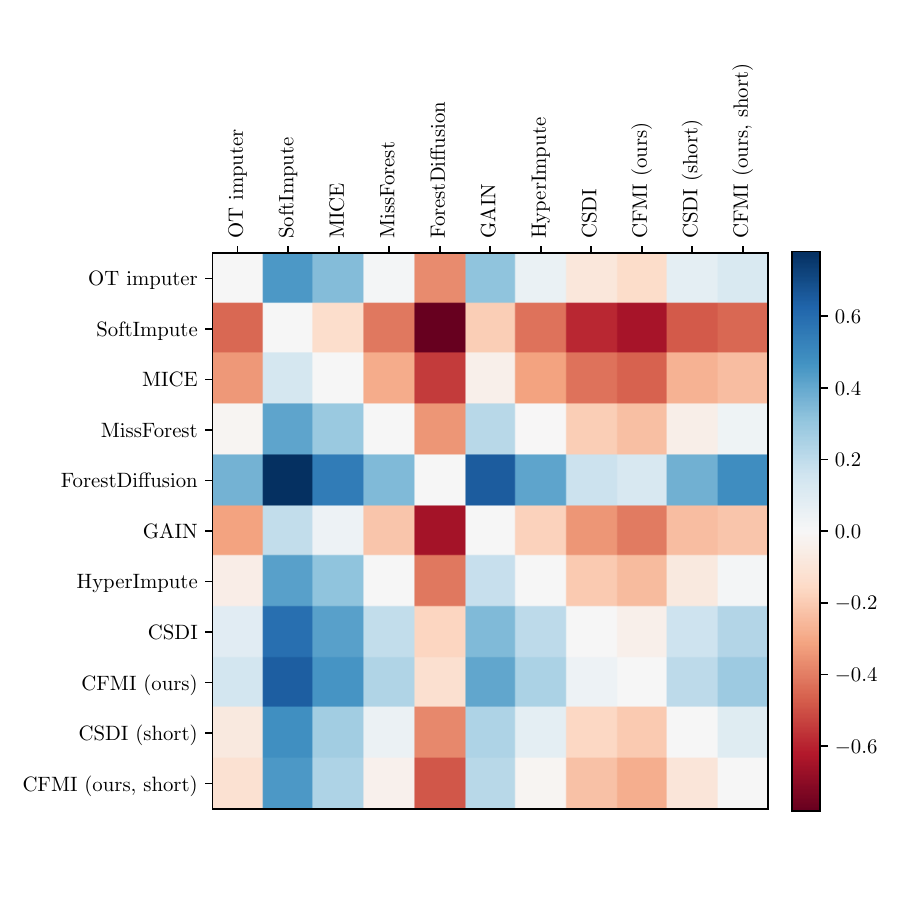}
  \caption{Classifier 1-AUROC results: average relative distance (blue means row method outperforms the column method). MCAR 25\% missingness.}
\end{figure}
\end{minipage}
\begin{minipage}{.48\textwidth}
\begin{figure}[H]
  \centering
  \includegraphics[width=1.\linewidth]{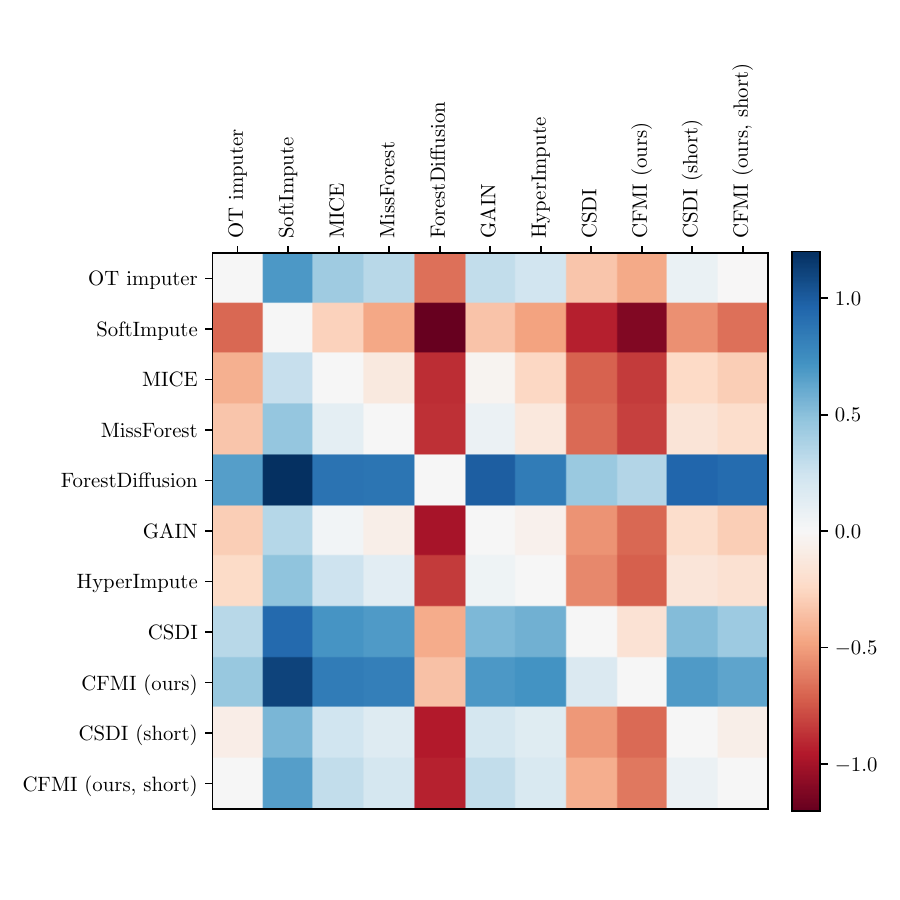}
  \caption{Classifier 1-AUROC results: average relative distance (blue means row method outperforms the column method). MCAR 50\% missingness.}
\end{figure}
\end{minipage}
\begin{minipage}{.48\textwidth}
\begin{figure}[H]
  \centering
  \includegraphics[width=1.\linewidth]{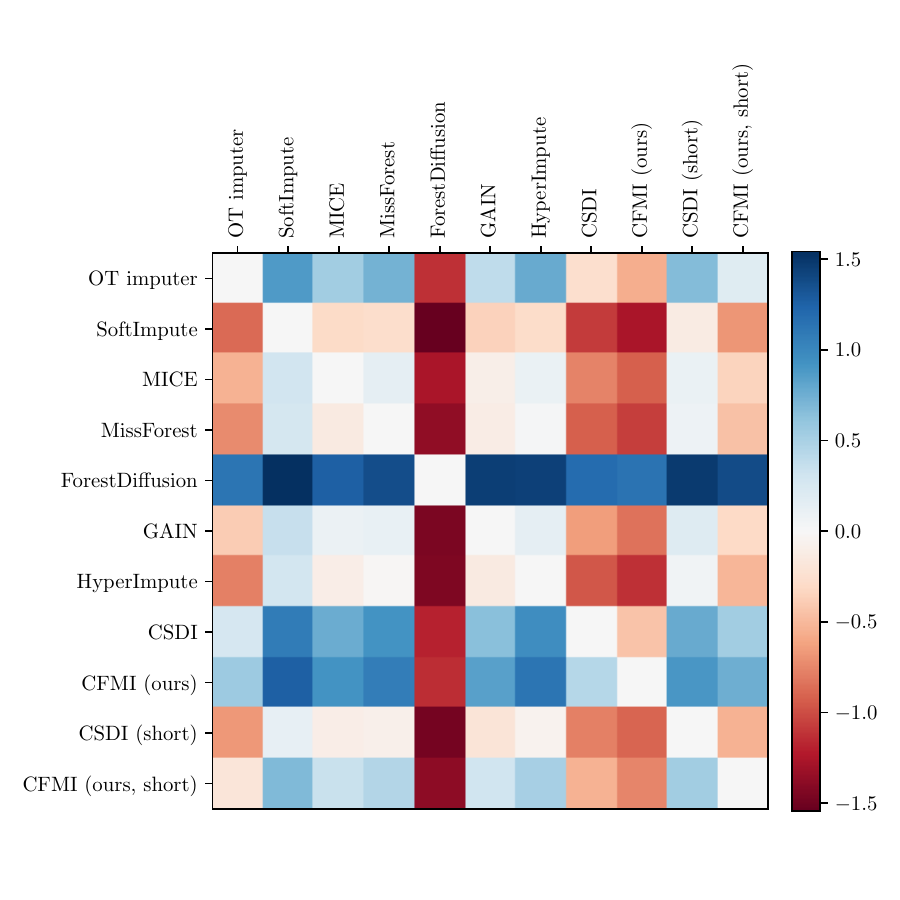}
  \caption{Classifier 1-AUROC results: average relative distance (blue means row method outperforms the column method). MCAR 75\% missingness.}
\end{figure}
\end{minipage}
\begin{minipage}{.48\textwidth}
\begin{figure}[H]
  \centering
  \includegraphics[width=1.\linewidth]{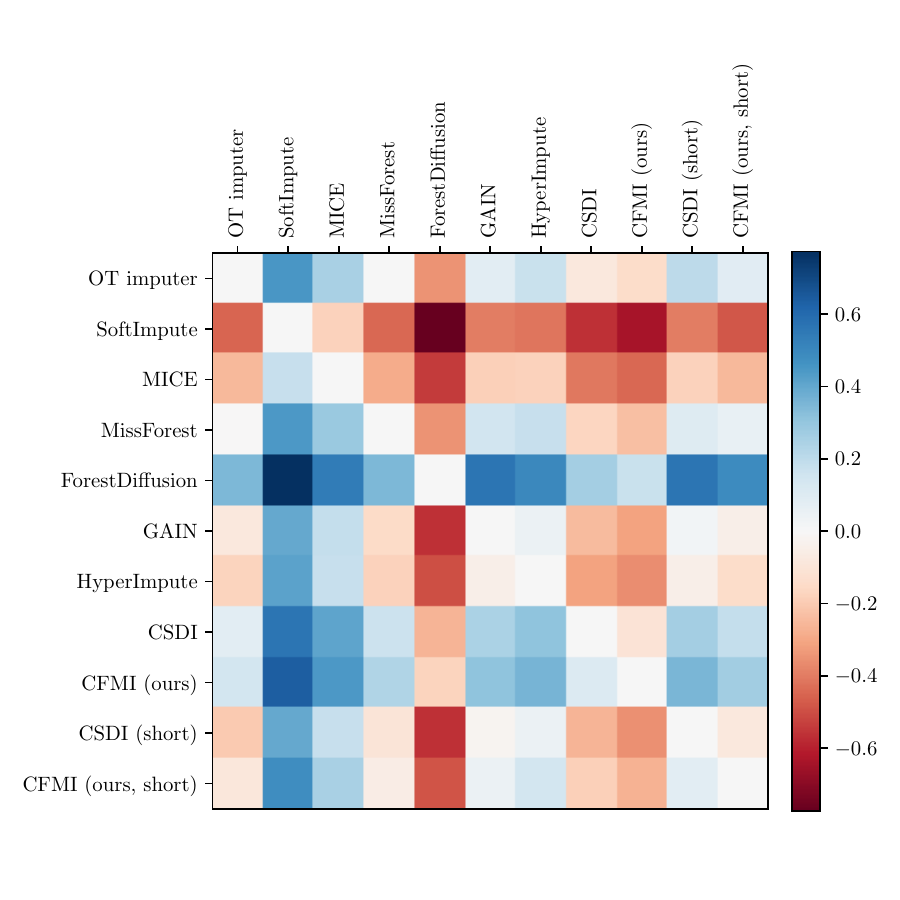}
  \caption{Classifier 1-AUROC results: average relative distance (blue means row method outperforms the column method). MAR 25\% missingness.}
\end{figure}
\end{minipage}
\begin{minipage}{.48\textwidth}
\begin{figure}[H]
  \centering
  \includegraphics[width=1.\linewidth]{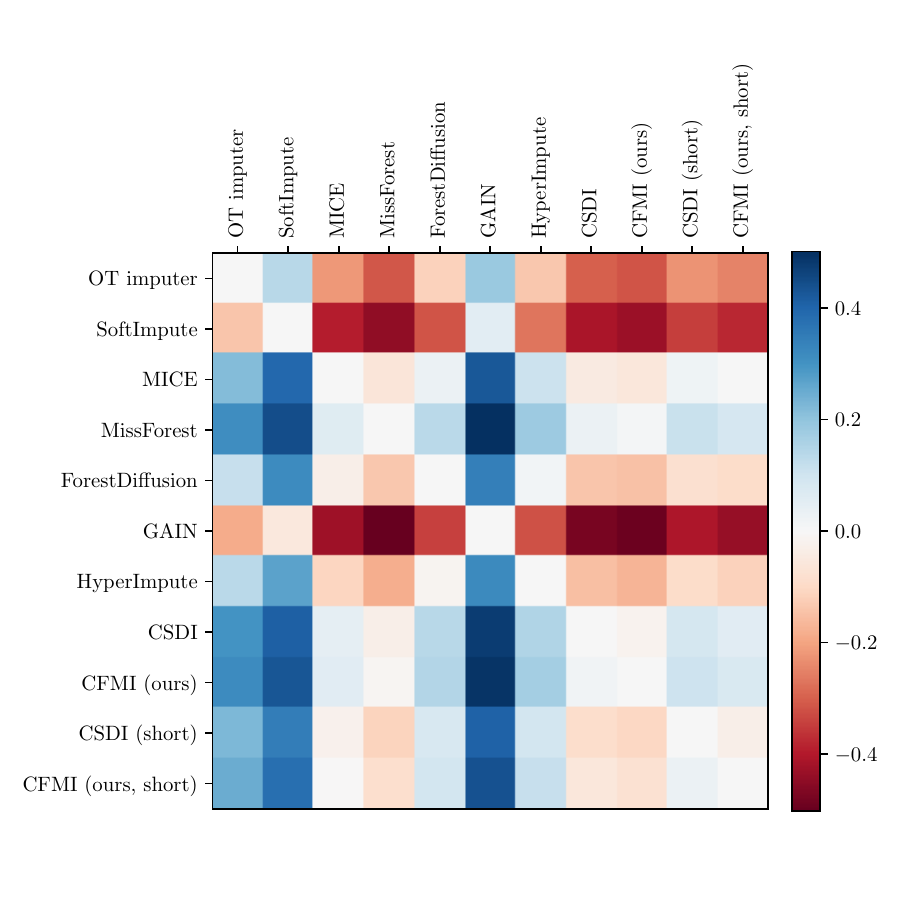}
  \caption{Regression CRPS results: average relative distance (blue means row method outperforms the column method). MCAR 25\% missingness.}
\end{figure}
\end{minipage}
\begin{minipage}{.48\textwidth}
\begin{figure}[H]
  \centering
  \includegraphics[width=1.\linewidth]{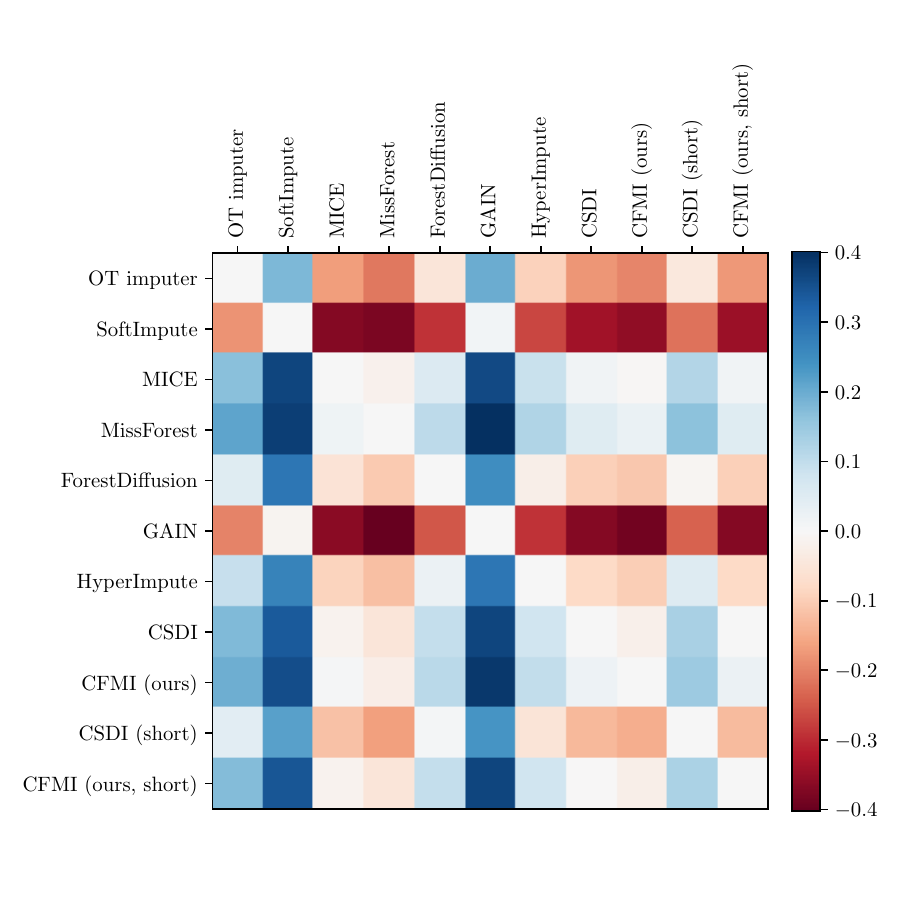}
  \caption{Regression CRPS results: average relative distance (blue means row method outperforms the column method). MCAR 50\% missingness.}
\end{figure}
\end{minipage}
\begin{minipage}{.48\textwidth}
\begin{figure}[H]
  \centering
  \includegraphics[width=1.\linewidth]{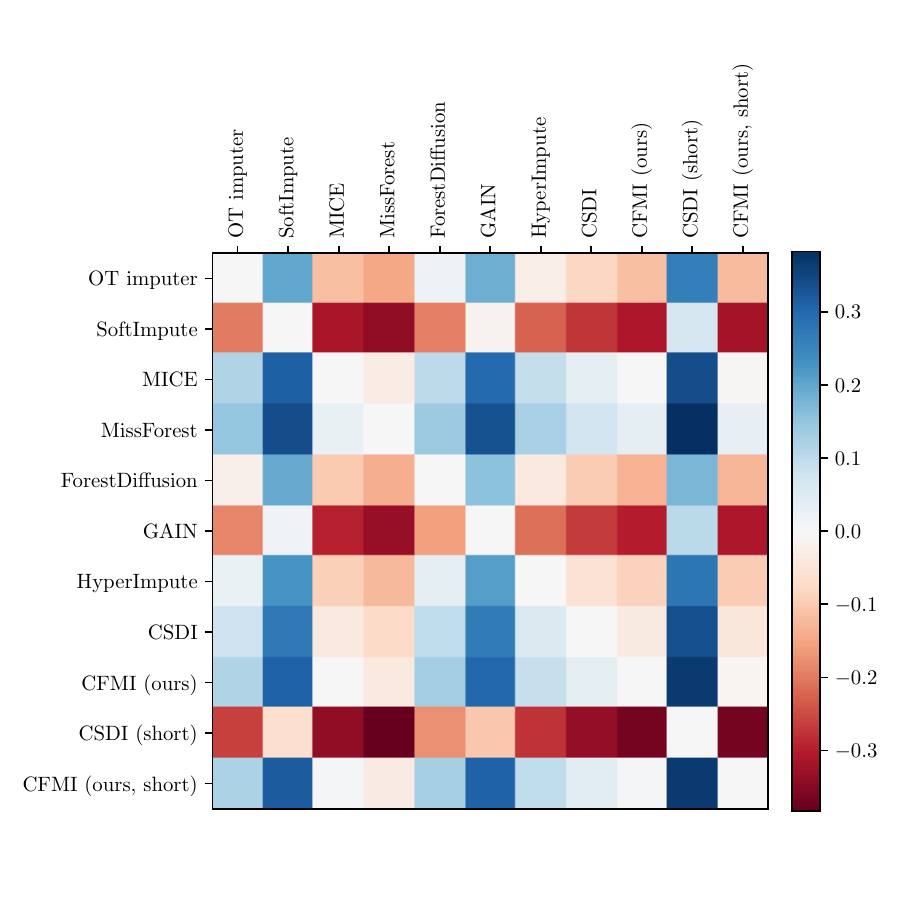}
  \caption{Regression CRPS results: average relative distance (blue means row method outperforms the column method). MCAR 75\% missingness.}
\end{figure}
\end{minipage}
\begin{minipage}{.48\textwidth}
\begin{figure}[H]
  \centering
  \includegraphics[width=1.\linewidth]{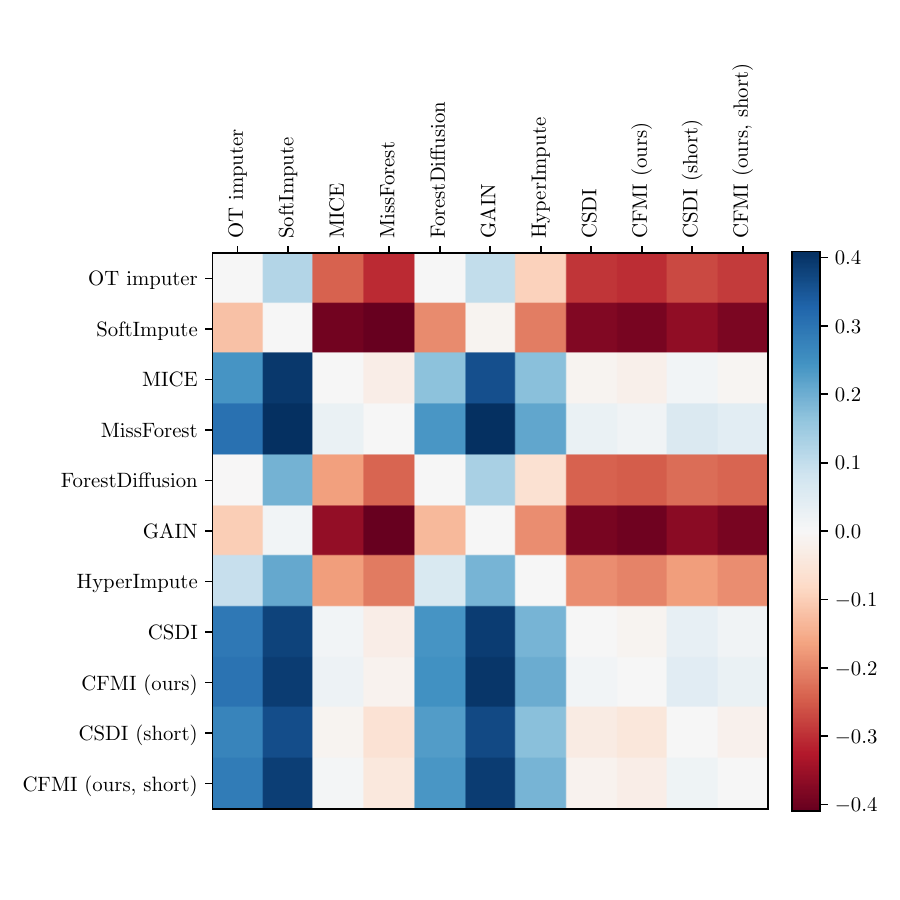}
  \caption{Regression CRPS results: average relative distance (blue means row method outperforms the column method). MAR 25\% missingness.}
\end{figure}
\end{minipage}
\begin{minipage}{.48\textwidth}
\begin{figure}[H]
  \centering
  \includegraphics[width=1.\linewidth]{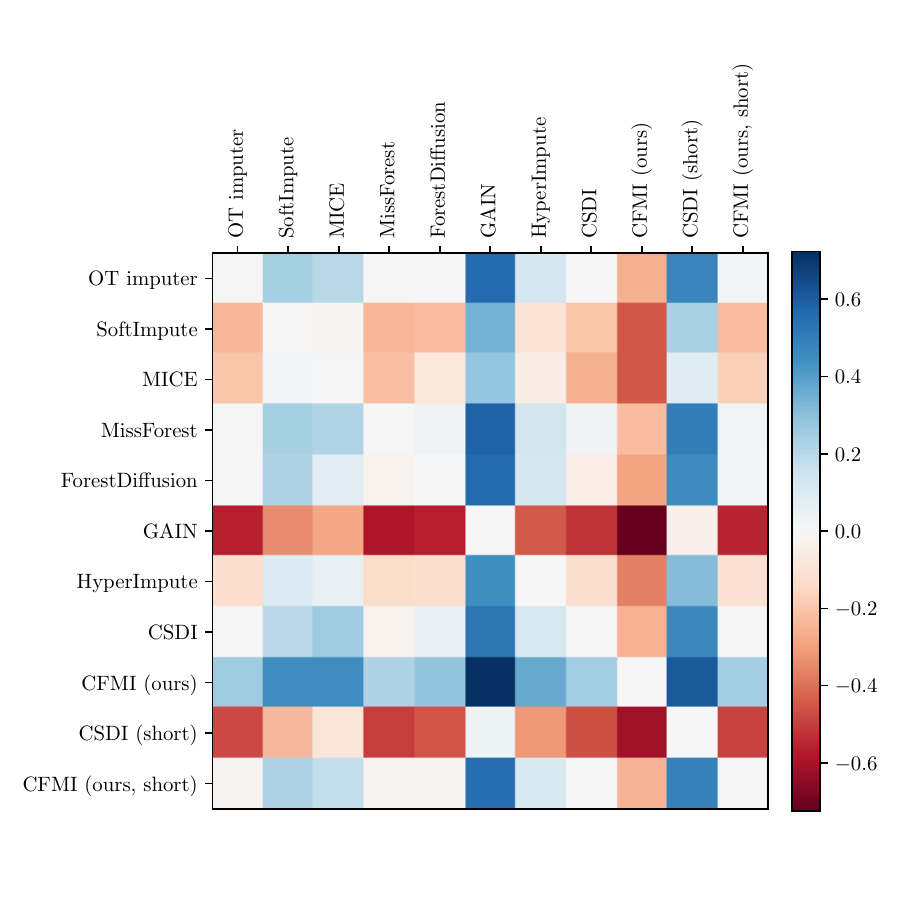}
  \caption{Regression parameter percent bias results: average relative distance (blue means row method outperforms the column method). MCAR 25\% missingness.}
\end{figure}
\end{minipage}
\begin{minipage}{.48\textwidth}
\begin{figure}[H]
  \centering
  \includegraphics[width=1.\linewidth]{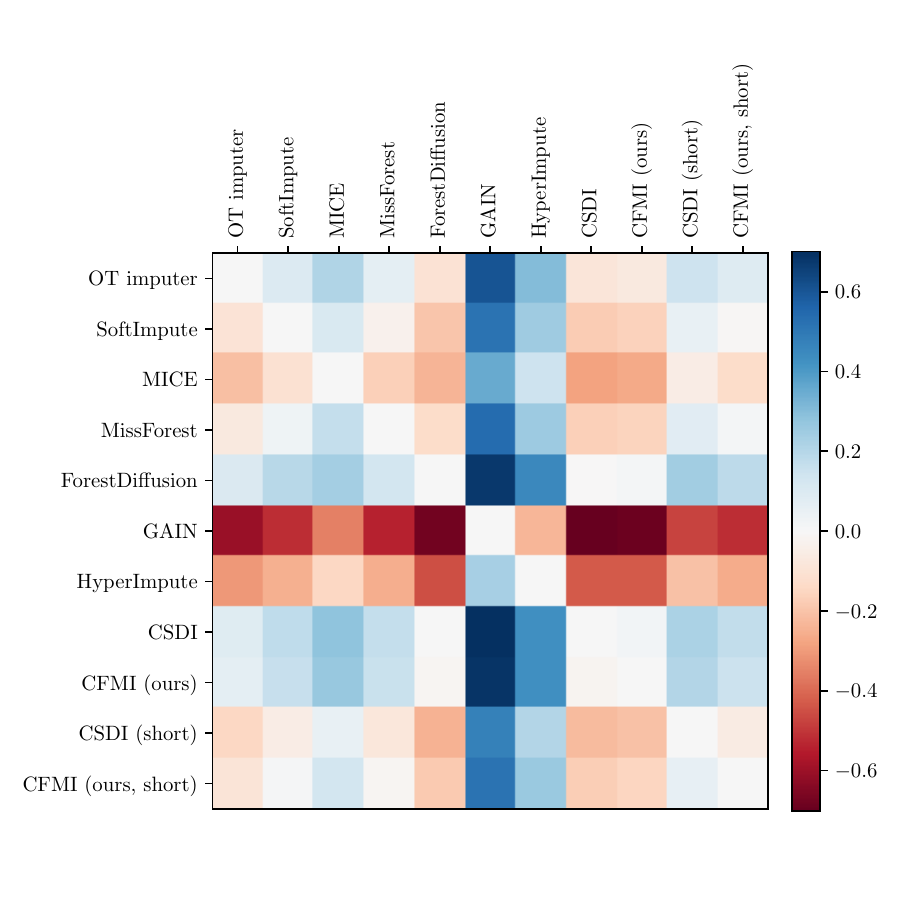}
  \caption{Regression parameter percent bias results: average relative distance (blue means row method outperforms the column method). MCAR 50\% missingness.}
\end{figure}
\end{minipage}
\begin{minipage}{.48\textwidth}
\begin{figure}[H]
  \centering
  \includegraphics[width=1.\linewidth]{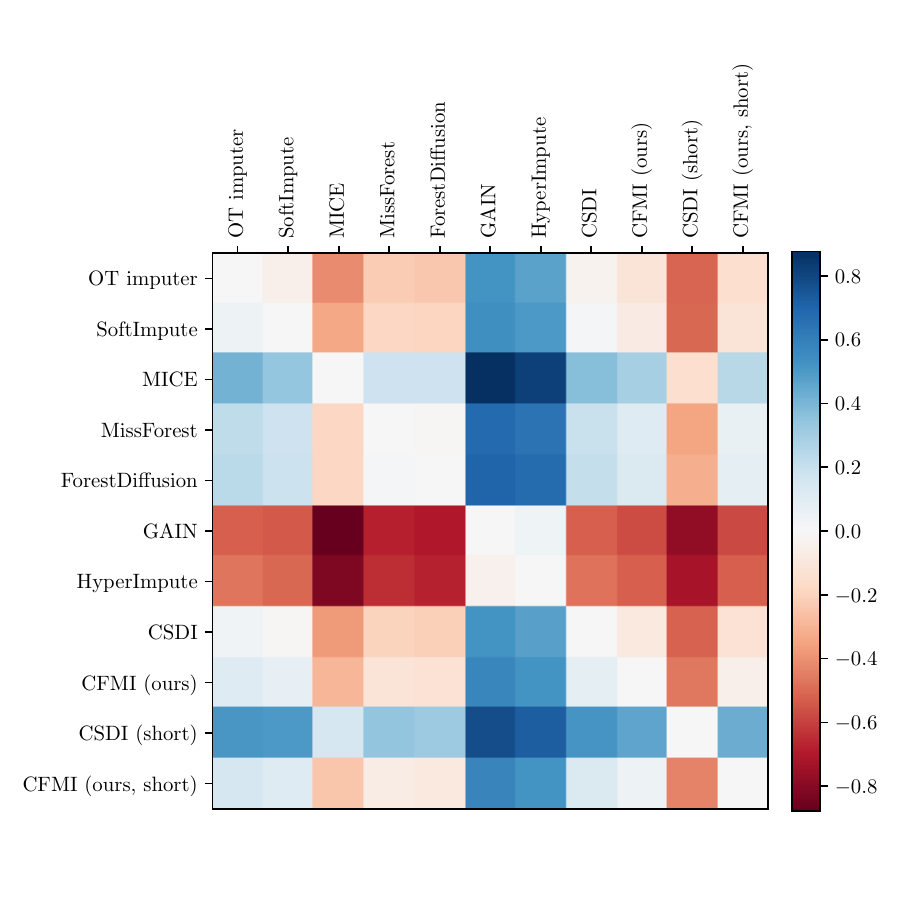}
  \caption{Regression parameter percent bias results: average relative distance (blue means row method outperforms the column method). MCAR 75\% missingness.}
\end{figure}
\end{minipage}
\begin{minipage}{.48\textwidth}
\begin{figure}[H]
  \centering
  \includegraphics[width=1.\linewidth]{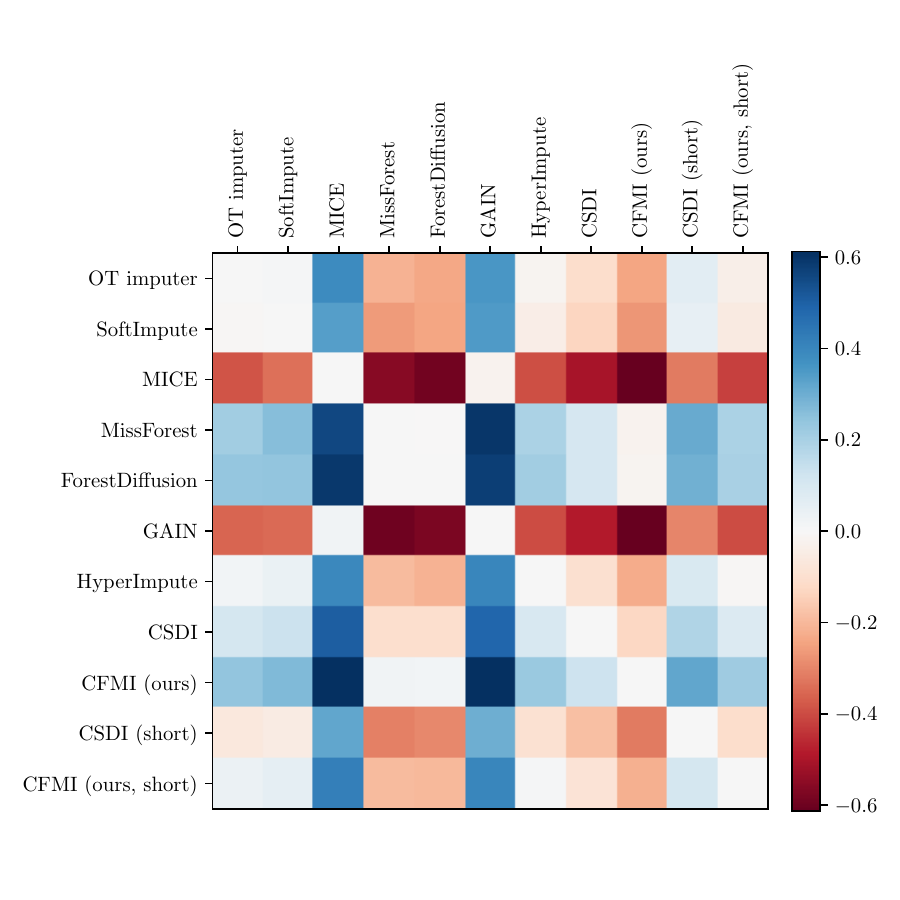}
  \caption{Regression parameter percent bias results: average relative distance (blue means row method outperforms the column method). MAR 25\% missingness.}
\end{figure}
\end{minipage}
\begin{minipage}{.48\textwidth}
\begin{figure}[H]
  \centering
  \includegraphics[width=1.\linewidth]{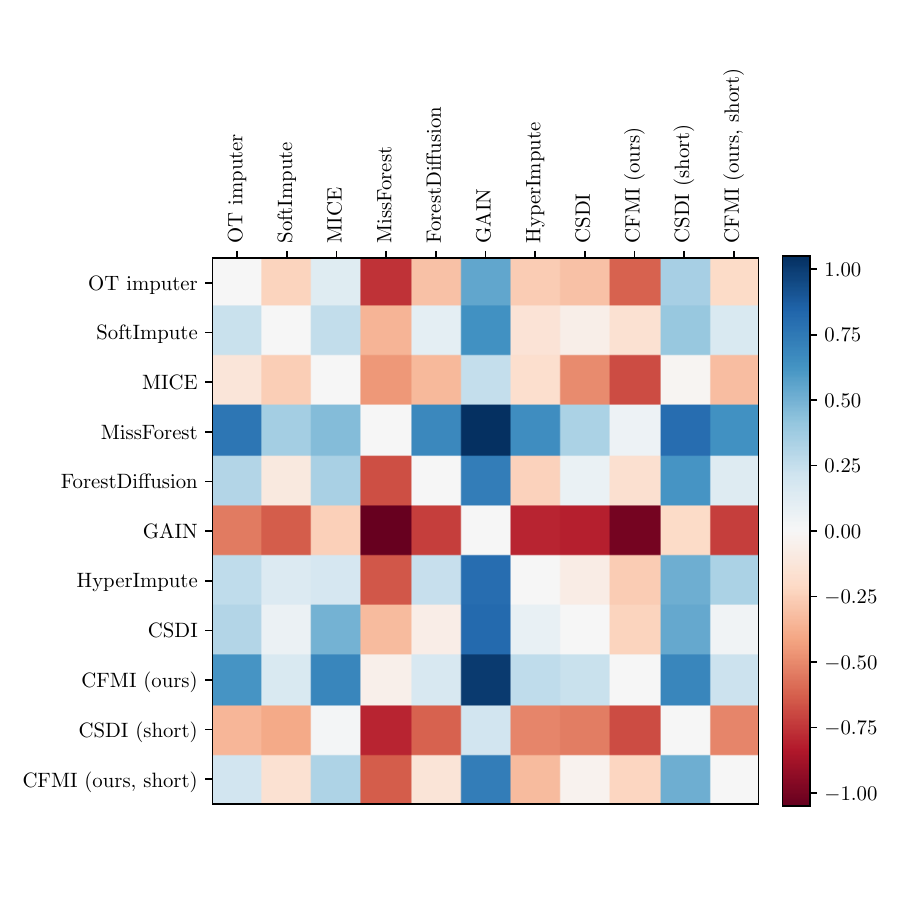}
  \caption{Regression parameter 1-CR results: average relative distance (blue means row method outperforms the column method). MCAR 25\% missingness.}
\end{figure}
\end{minipage}
\begin{minipage}{.48\textwidth}
\begin{figure}[H]
  \centering
  \includegraphics[width=1.\linewidth]{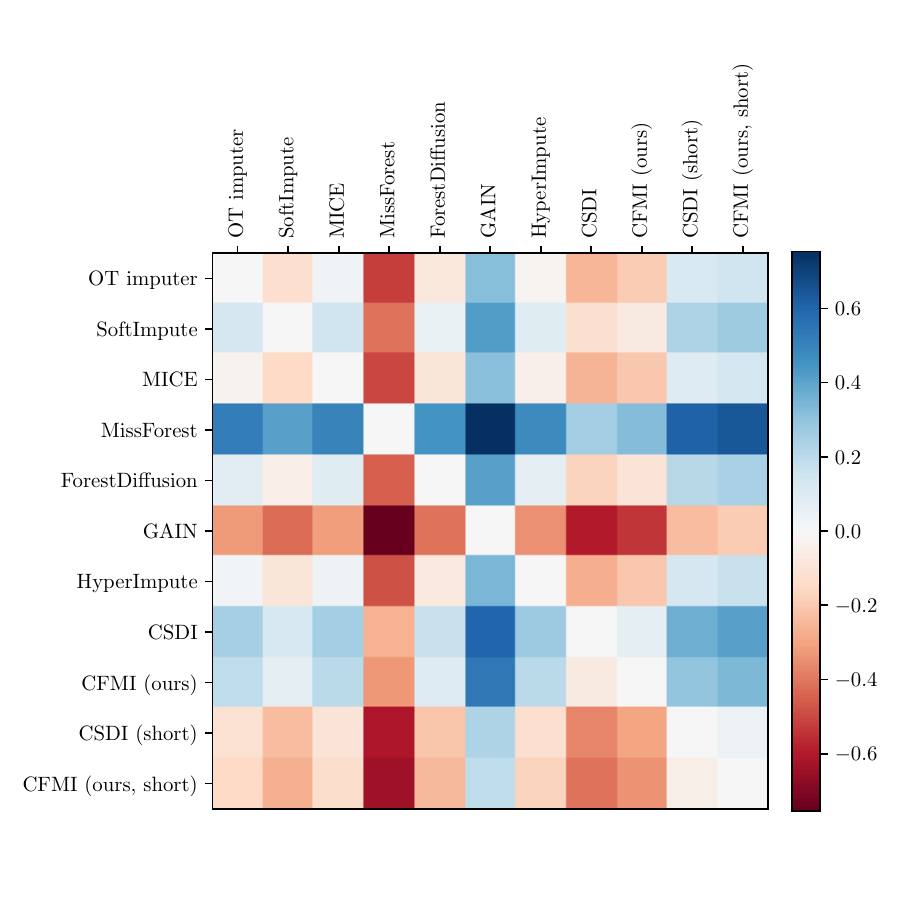}
  \caption{Regression parameter 1-CR results: average relative distance (blue means row method outperforms the column method). MCAR 50\% missingness.}
\end{figure}
\end{minipage}
\begin{minipage}{.48\textwidth}
\begin{figure}[H]
  \centering
  \includegraphics[width=1.\linewidth]{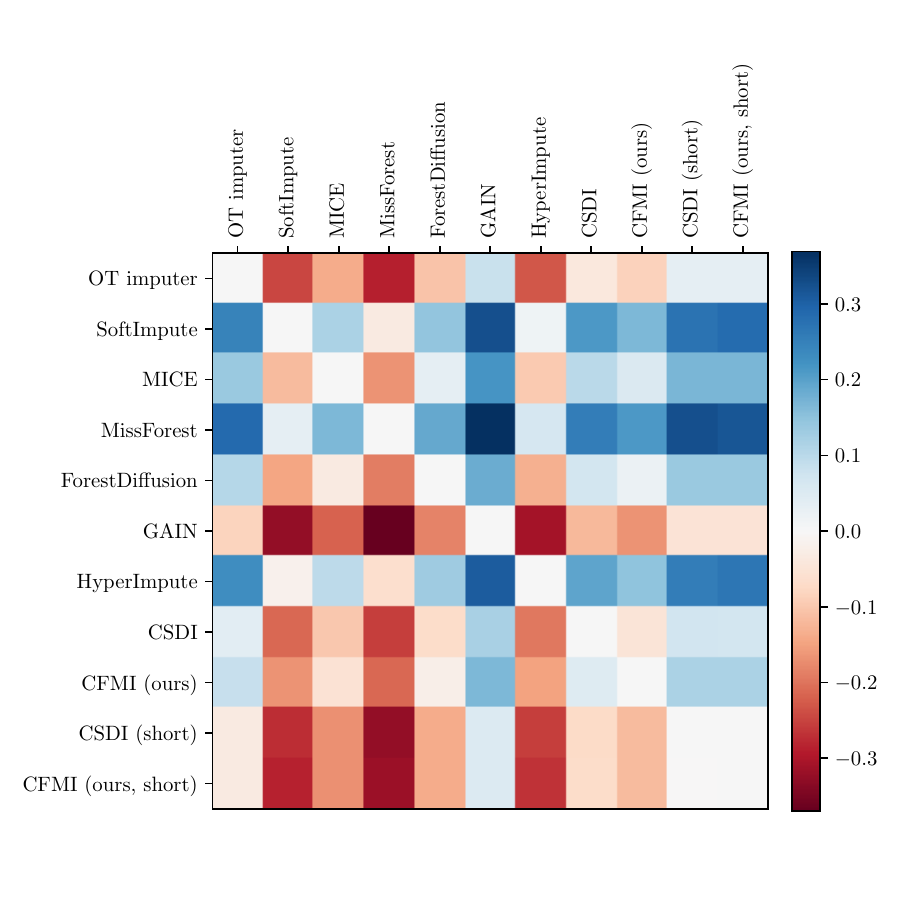}
  \caption{Regression parameter 1-CR results: average relative distance (blue means row method outperforms the column method). MCAR 75\% missingness.}
\end{figure}
\end{minipage}
\begin{minipage}{.48\textwidth}
\begin{figure}[H]
  \centering
  \includegraphics[width=1.\linewidth]{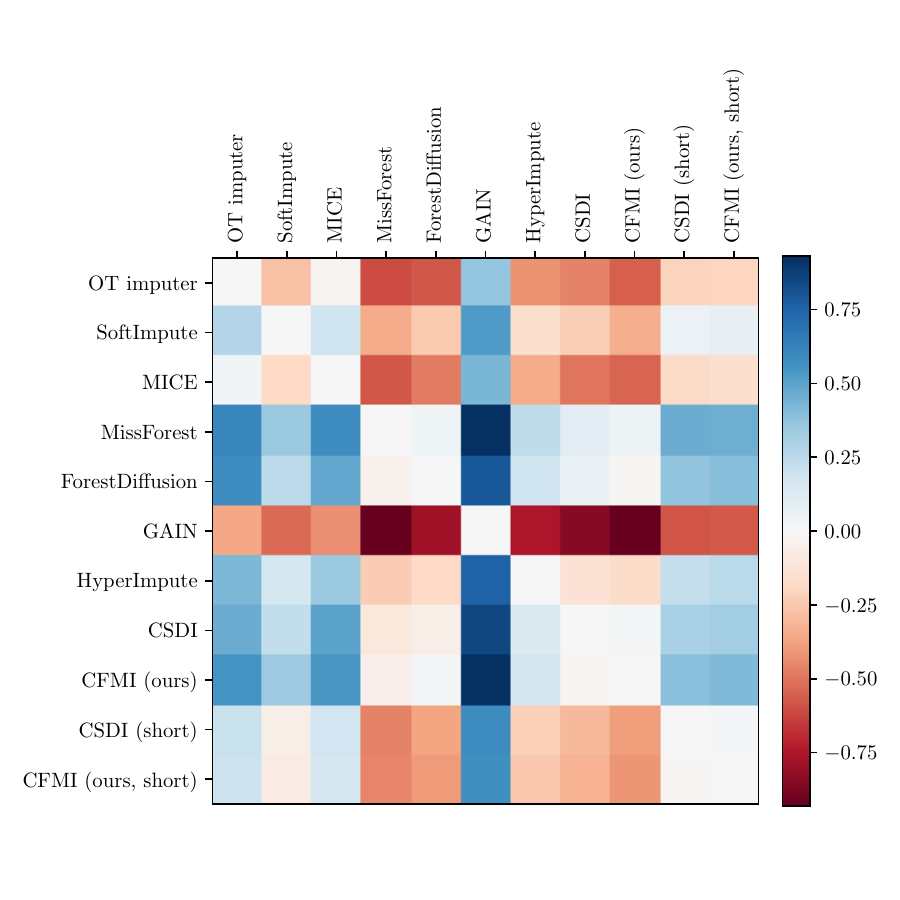}
  \caption{Regression parameter 1-CR results: average relative distance (blue means row method outperforms the column method). MAR 25\% missingness.}
\end{figure}
\end{minipage}
\begin{minipage}{.48\textwidth}
\begin{figure}[H]
  \centering
  \includegraphics[width=1.\linewidth]{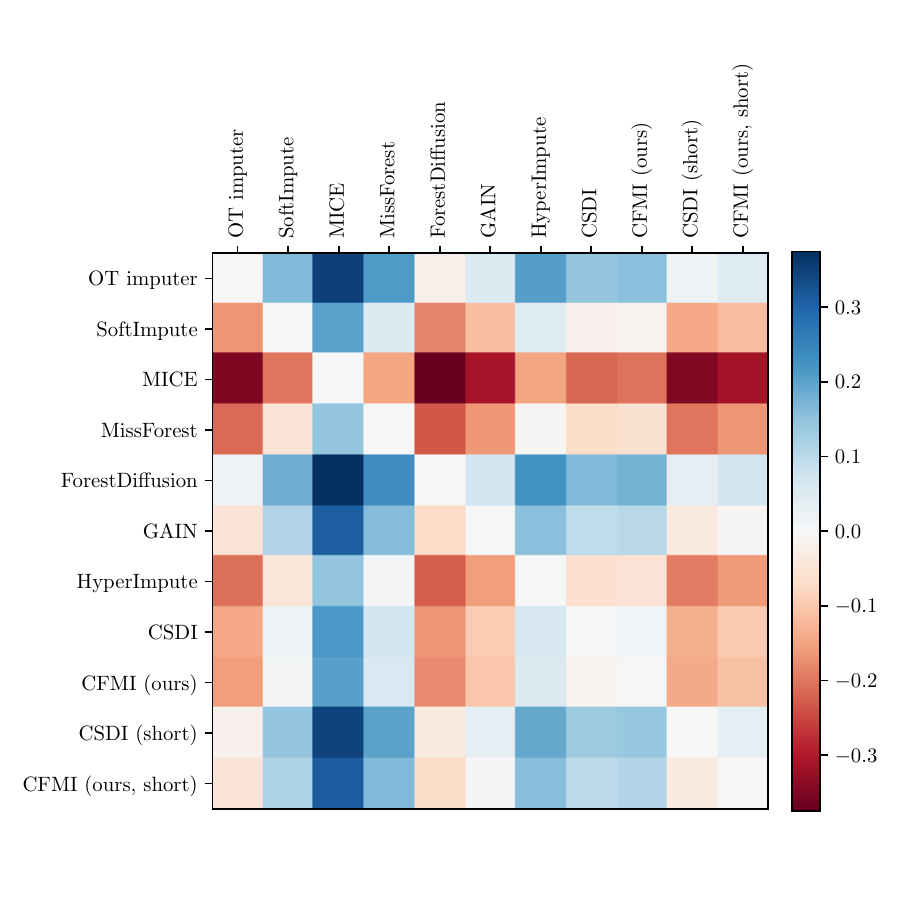}
  \caption{Regression parameter AW results: average relative distance (blue means row method outperforms the column method). MCAR 25\% missingness.}
\end{figure}
\end{minipage}
\begin{minipage}{.48\textwidth}
\begin{figure}[H]
  \centering
  \includegraphics[width=1.\linewidth]{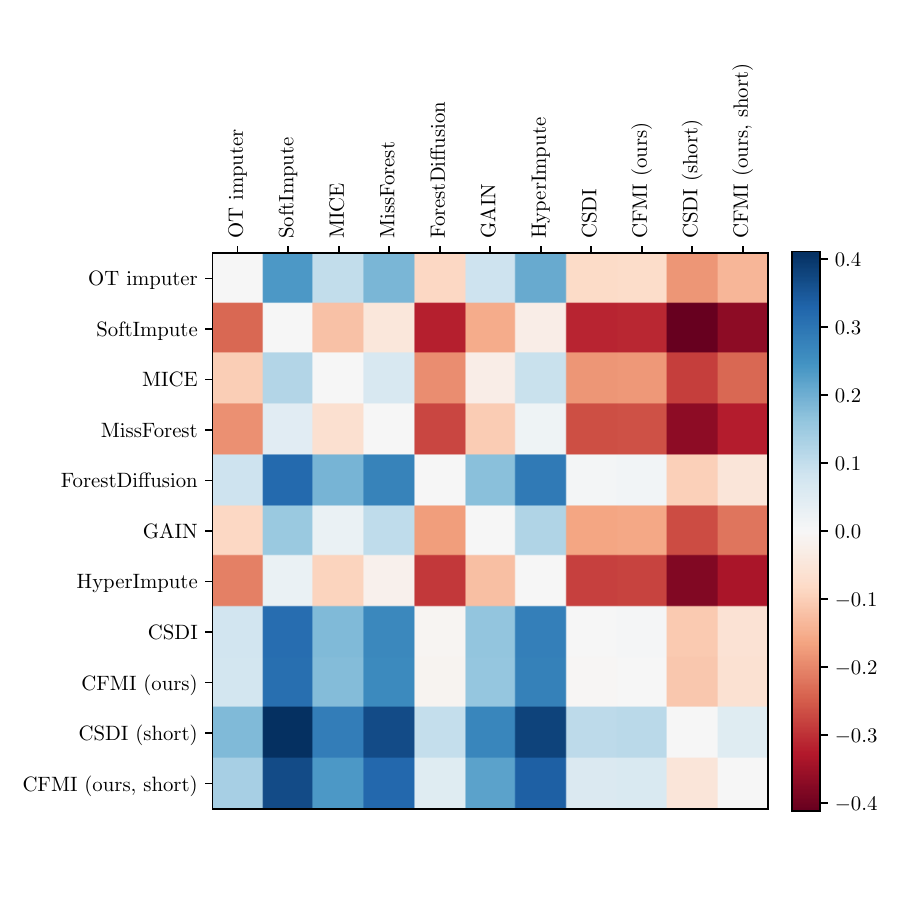}
  \caption{Regression parameter AW results: average relative distance (blue means row method outperforms the column method). MCAR 50\% missingness.}
\end{figure}
\end{minipage}
\begin{minipage}{.48\textwidth}
\begin{figure}[H]
  \centering
  \includegraphics[width=1.\linewidth]{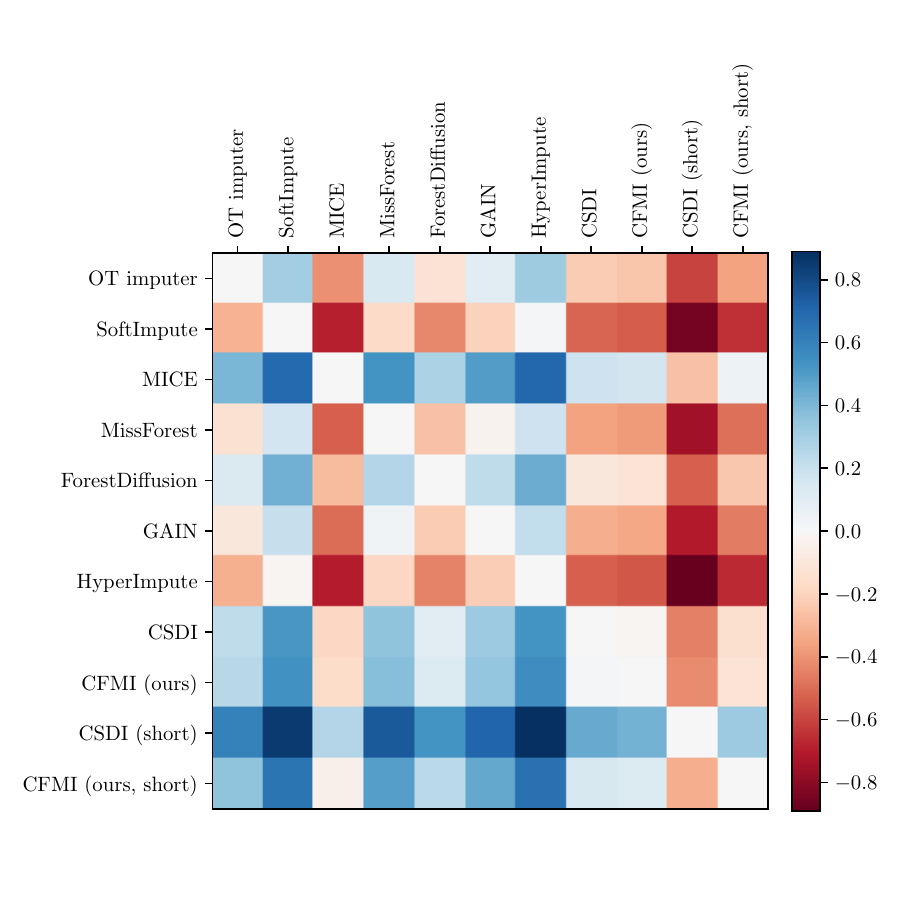}
  \caption{Regression parameter AW results: average relative distance (blue means row method outperforms the column method). MCAR 75\% missingness.}
\end{figure}
\end{minipage}
\begin{minipage}{.48\textwidth}
\begin{figure}[H]
  \centering
  \includegraphics[width=1.\linewidth]{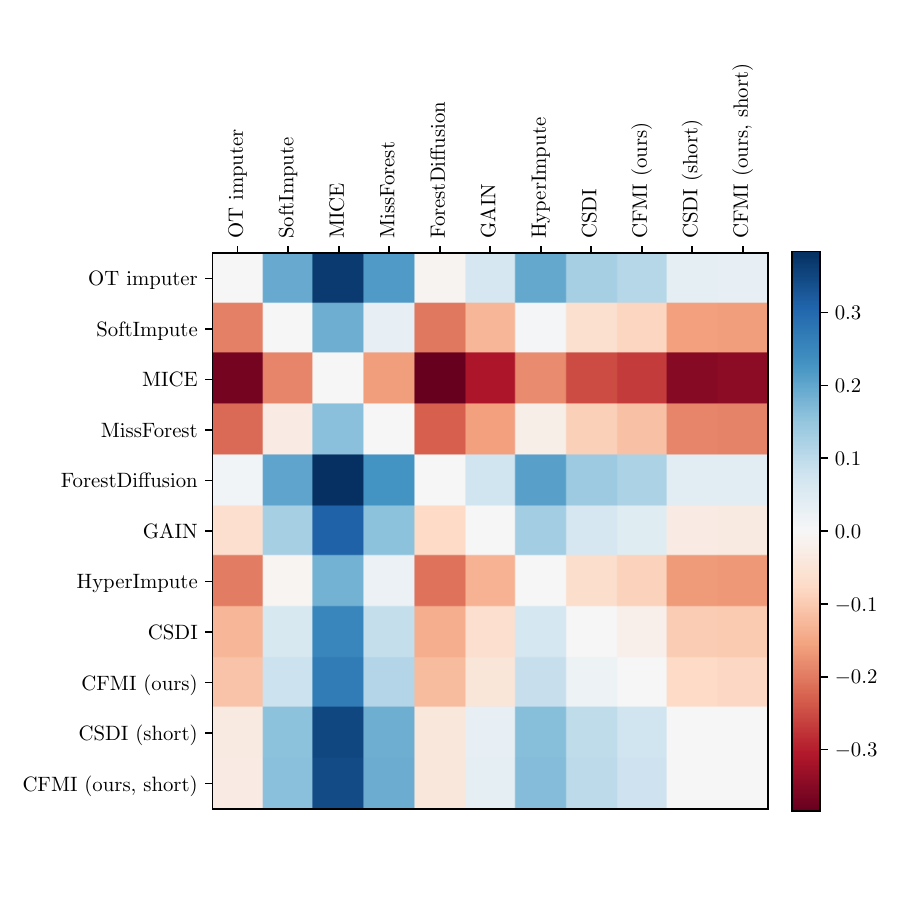}
  \caption{Regression parameter AW results: average relative distance (blue means row method outperforms the column method). MAR 25\% missingness.}
\end{figure}
\end{minipage}

\subsubsection{Average rank tables}
\label{apx:uci-avg-rank-tables}

\begin{table}[H]
\centering
\caption{Average rank of each method for all metrics, mean and standard error. Lowest mean value are highlighted in bold, as well as values that fall within lowest mean + standard error range. MCAR 25\% missingness. }
\resizebox{1.\linewidth}{!}{\input{figures/uci/25miss_rank_table}}
\end{table}

\begin{table}[H]
\centering
\caption{Average rank of each method for all metrics, mean and standard error. Lowest mean value are highlighted in bold, as well as values that fall within lowest mean + standard error range. MCAR 50\% missingness. }
\resizebox{1.\linewidth}{!}{\input{figures/uci/50miss_rank_table}}
\end{table}

\begin{table}[H]
\centering
\caption{Average rank of each method for all metrics, mean and standard error. Lowest mean value are highlighted in bold, as well as values that fall within lowest mean + standard error range. MCAR 75\% missingness. }
\resizebox{1.\linewidth}{!}{\input{figures/uci/75miss_rank_table}}
\end{table}

\begin{table}[H]
\centering
\caption{Average rank of each method for all metrics, mean and standard error. Lowest mean value are highlighted in bold, as well as values that fall within lowest mean + standard error range. MAR 25\% missingness. }
\resizebox{1.\linewidth}{!}{\input{figures/uci/25marlogmiss_rank_table}}
\end{table}

\subsubsection{Average rank vs fraction of missingness}
\label{apx:uci-avgrank-vs-missingness}

\begin{figure}[H]
  \centering
  \includegraphics[width=0.9\linewidth]{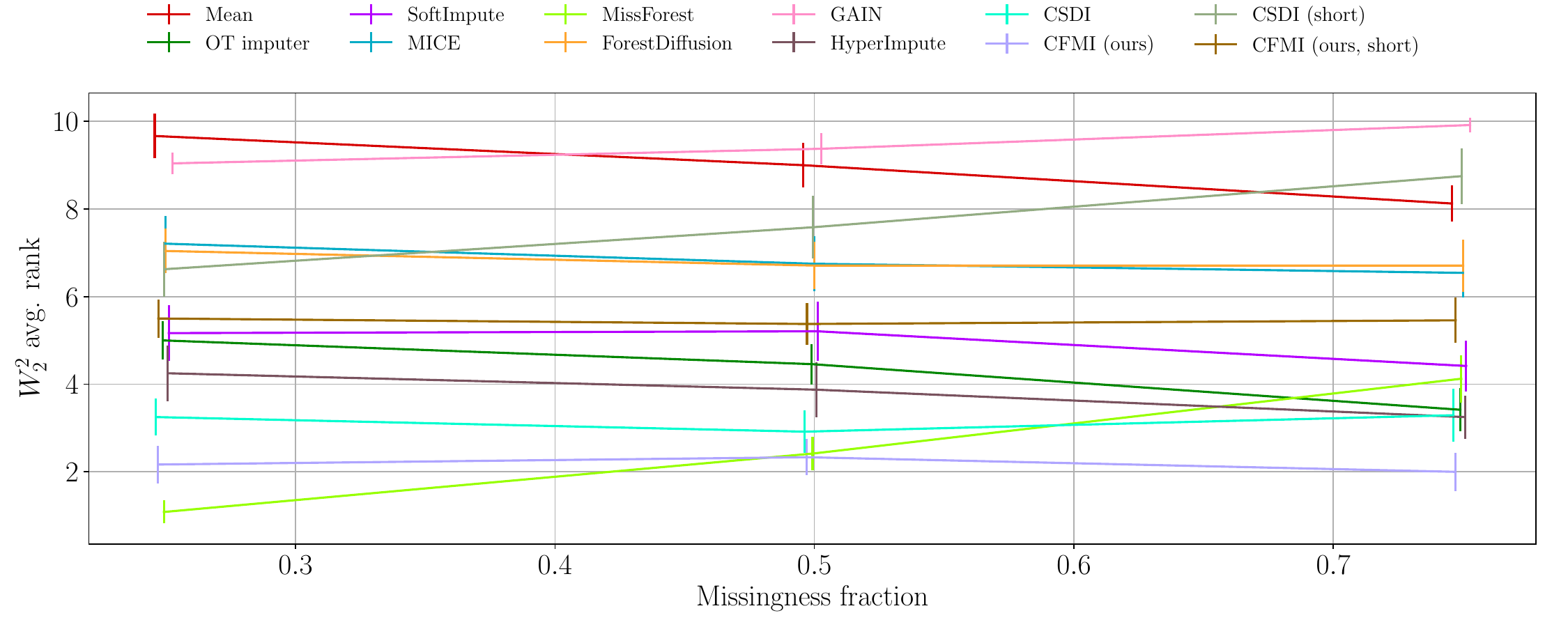}
  \caption{Average rank on Wasserstein-2 over all data sets versus missingness fraction. MCAR missingness.}
\end{figure}

\begin{figure}[H]
  \centering
  \includegraphics[width=0.9\linewidth]{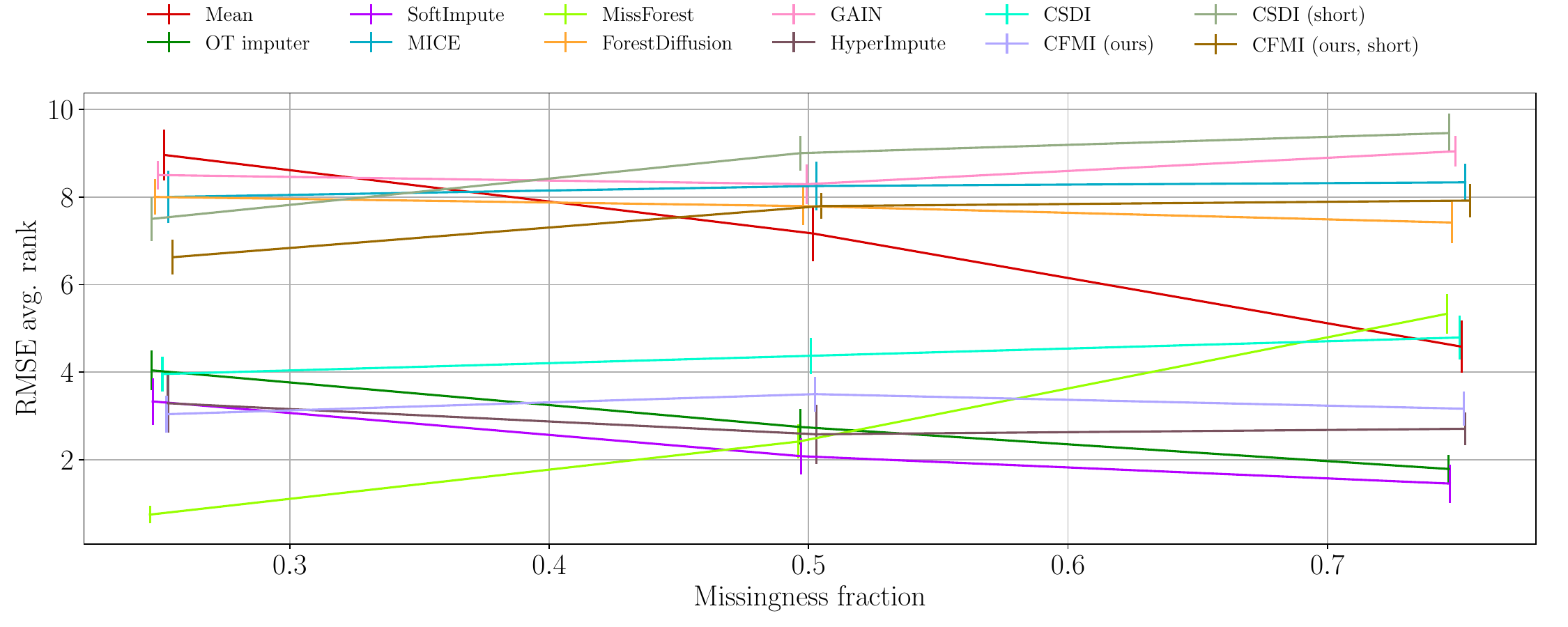}
  \caption{Average rank on average RMSE over all data sets versus missingness fraction. MCAR missingness.}
\end{figure}

\begin{figure}[H]
  \centering
  \includegraphics[width=0.9\linewidth]{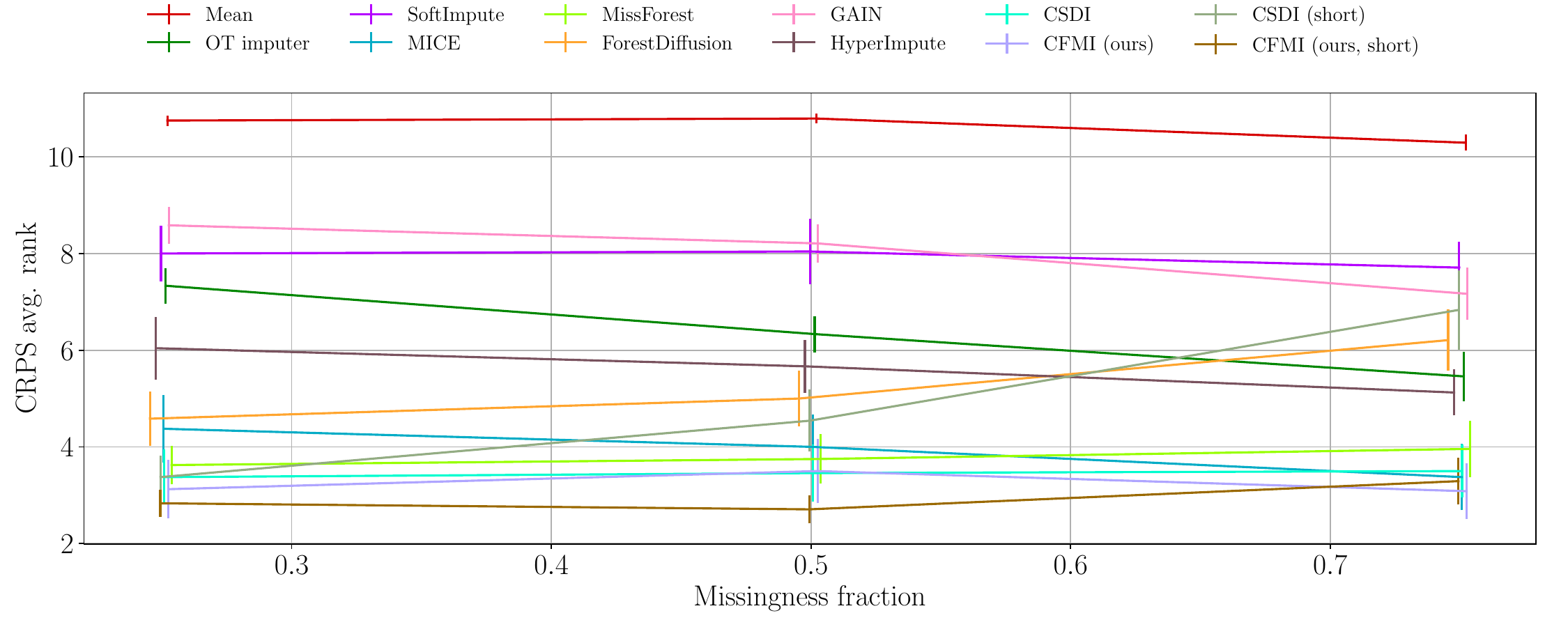}
  \caption{Average rank on CRPS over all data sets versus missingness fraction. MCAR missingness.}
\end{figure}

\begin{figure}[H]
  \centering
  \includegraphics[width=0.9\linewidth]{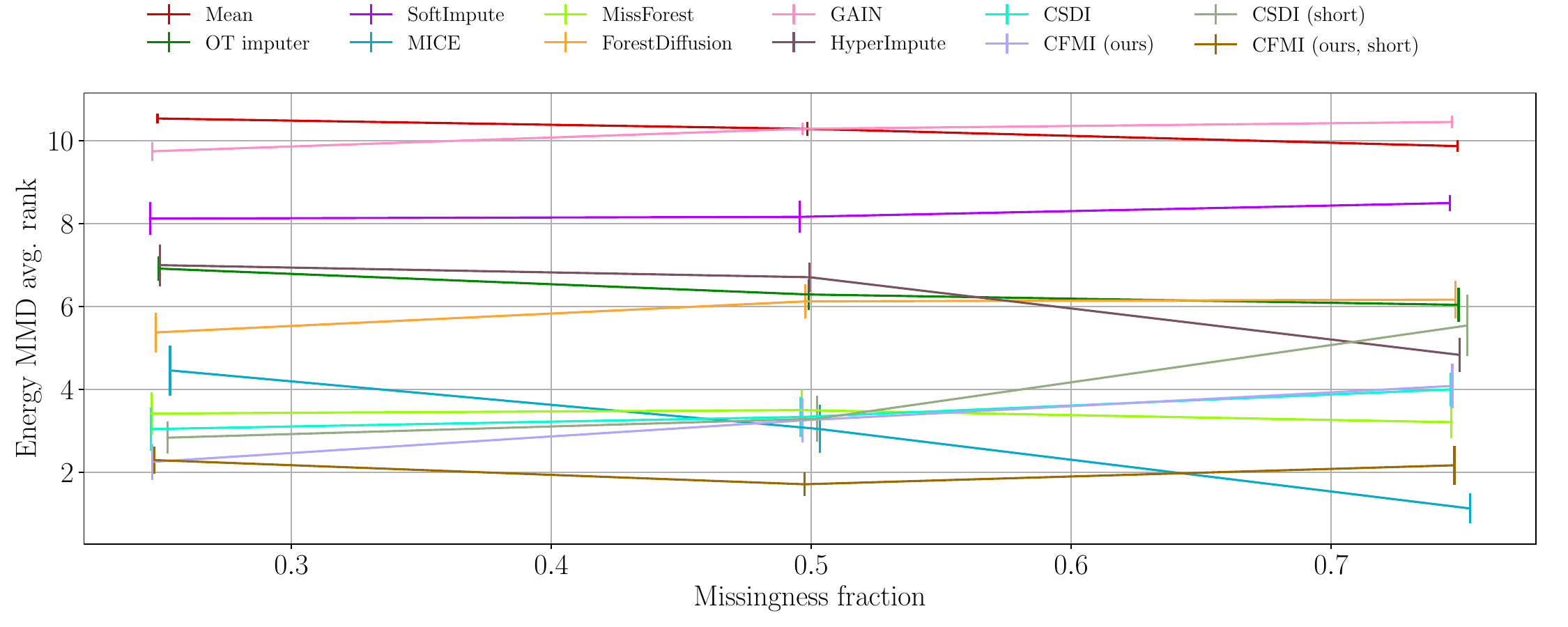}
  \caption{Average rank on energy MMD over all data sets versus missingness fraction. MCAR missingness.}
\end{figure}

\begin{figure}[H]
  \centering
  \includegraphics[width=0.9\linewidth]{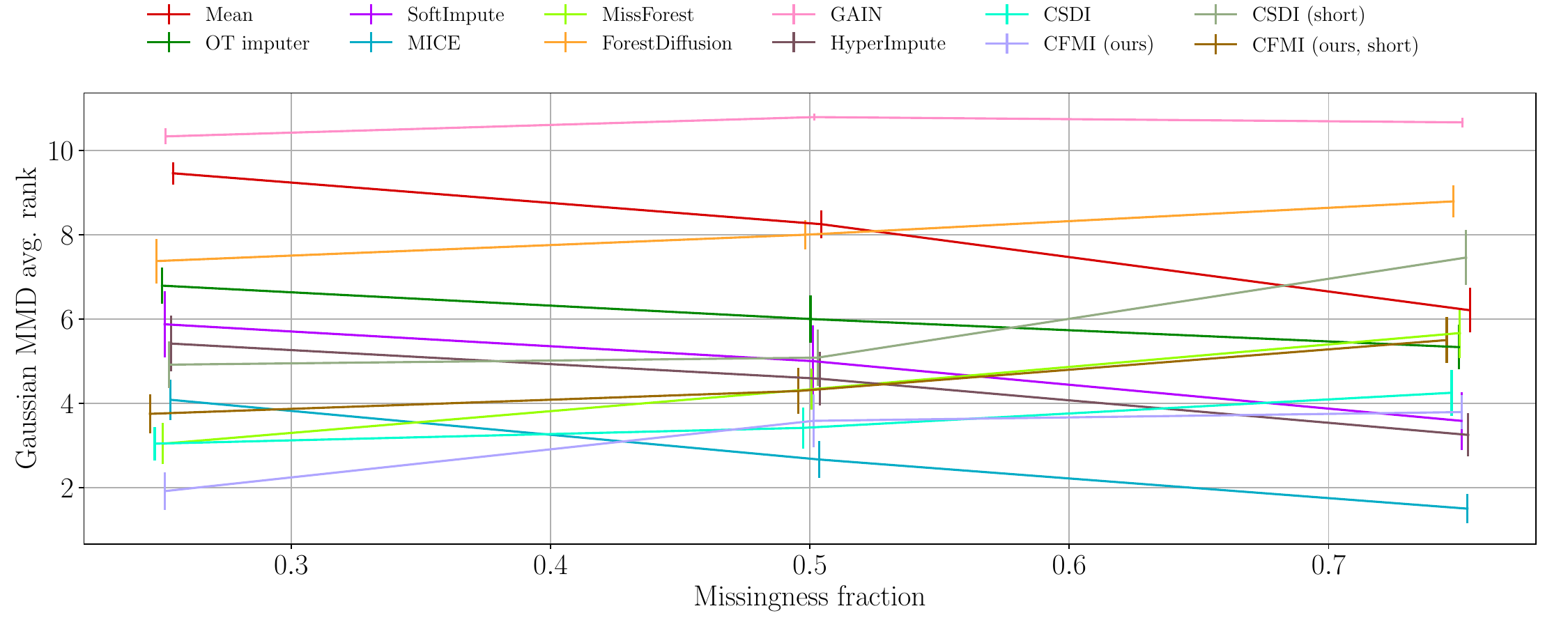}
  \caption{Average rank on Gaussian MMD over all data sets versus missingness fraction. MCAR missingness.}
\end{figure}

\begin{figure}[H]
  \centering
  \includegraphics[width=0.9\linewidth]{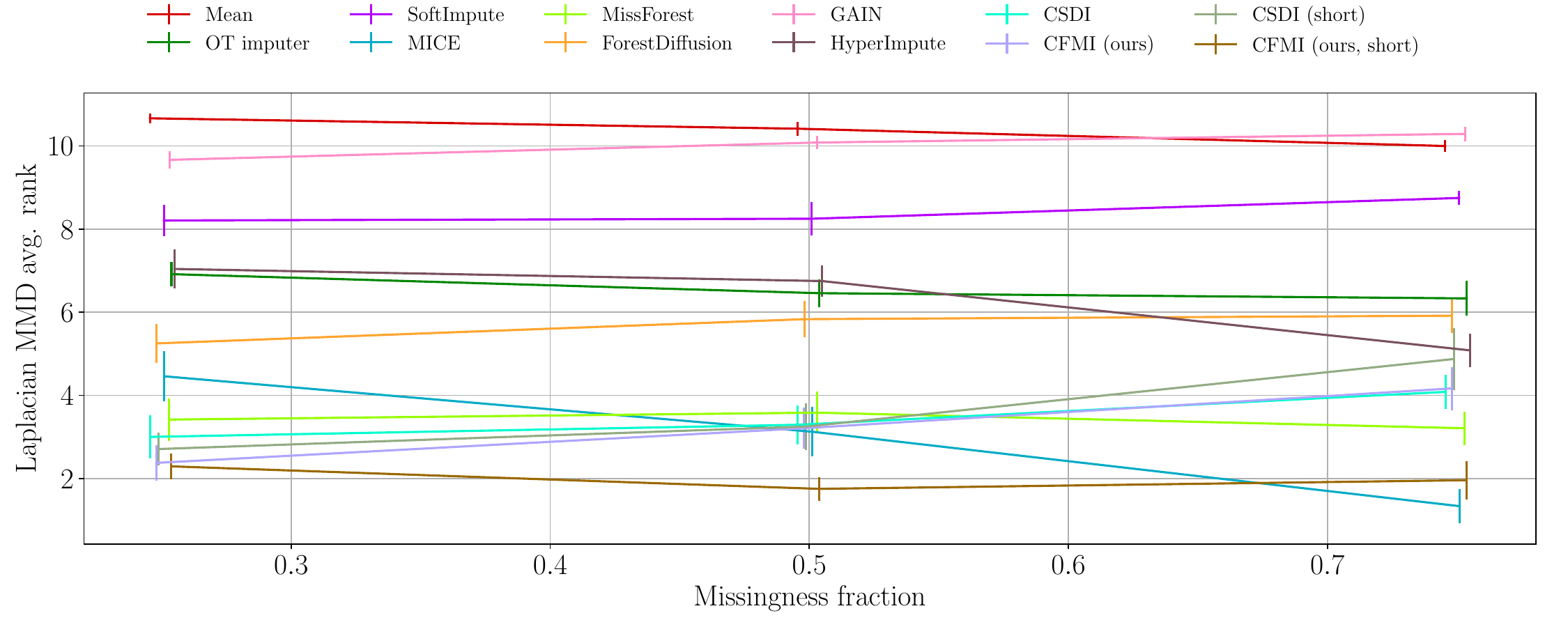}
  \caption{Average rank on Laplacian MMD over all data sets versus missingness fraction. MCAR missingness.}
\end{figure}

\begin{figure}[H]
  \centering
  \includegraphics[width=0.9\linewidth]{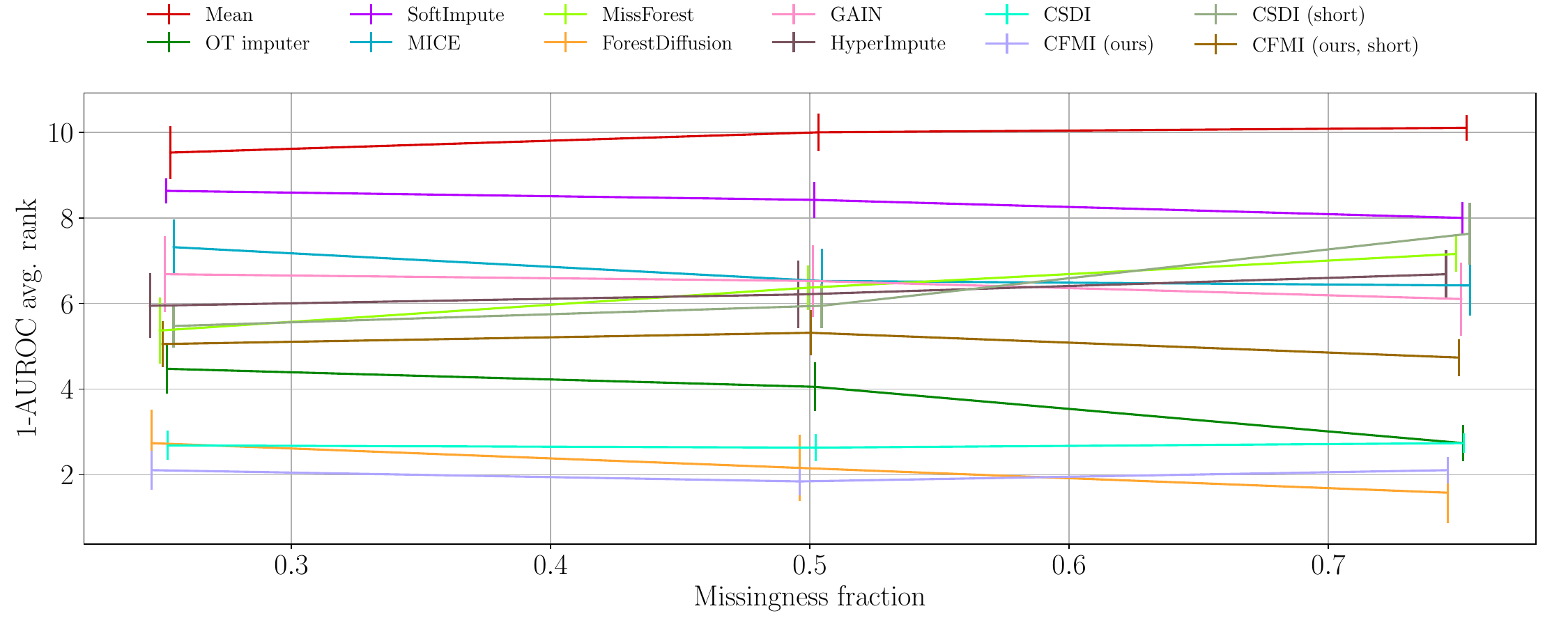}
  \caption{Average rank on classifier 1-AUROC over all data sets versus missingness fraction. MCAR missingness.}
\end{figure}

\begin{figure}[H]
  \centering
  \includegraphics[width=0.9\linewidth]{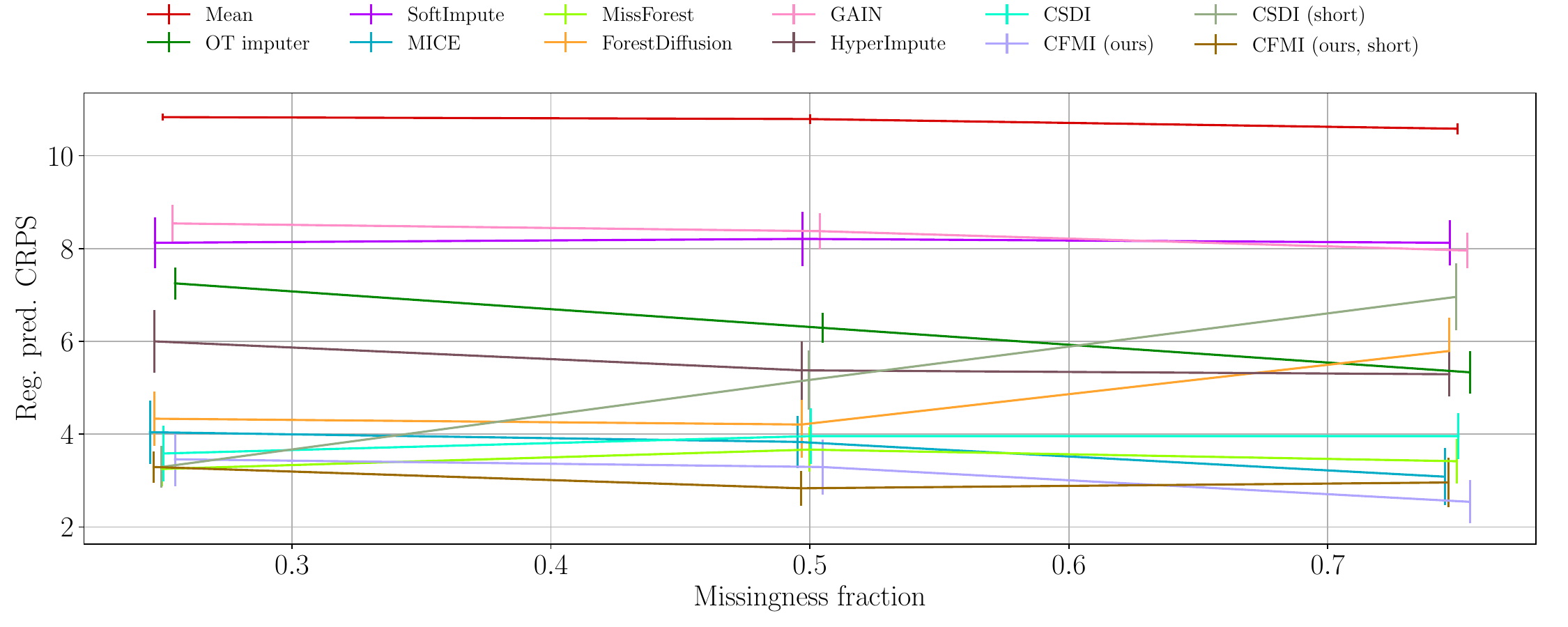}
  \caption{Average rank on regression CRPS over all data sets versus missingness fraction. MCAR missingness.}
\end{figure}

\begin{figure}[H]
  \centering
  \includegraphics[width=0.9\linewidth]{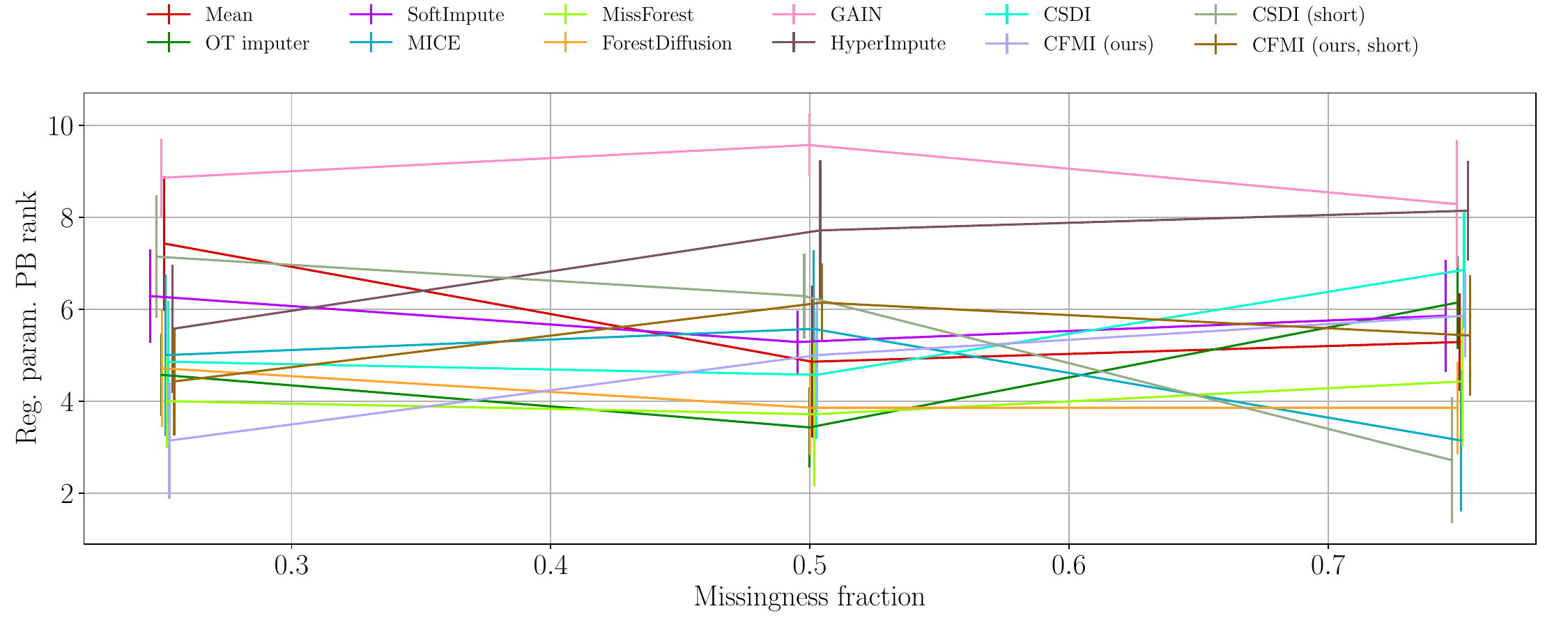}
  \caption{Average rank on regression parameter PB over all data sets versus missingness fraction. MCAR missingness.}
\end{figure}

\begin{figure}[H]
  \centering
  \includegraphics[width=0.9\linewidth]{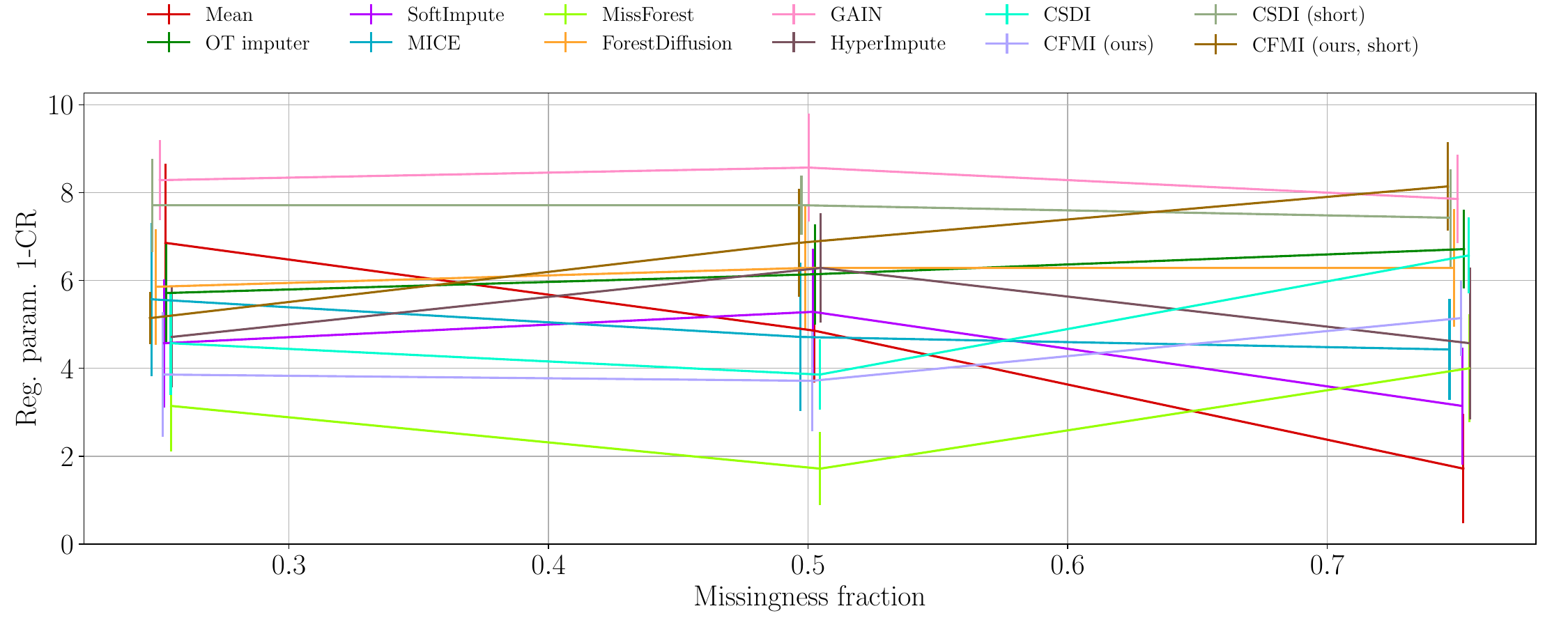}
  \caption{Average rank on regression parameter 1-CR over all data sets versus missingness fraction. MCAR missingness.}
\end{figure}

\begin{figure}[H]
  \centering
  \includegraphics[width=0.9\linewidth]{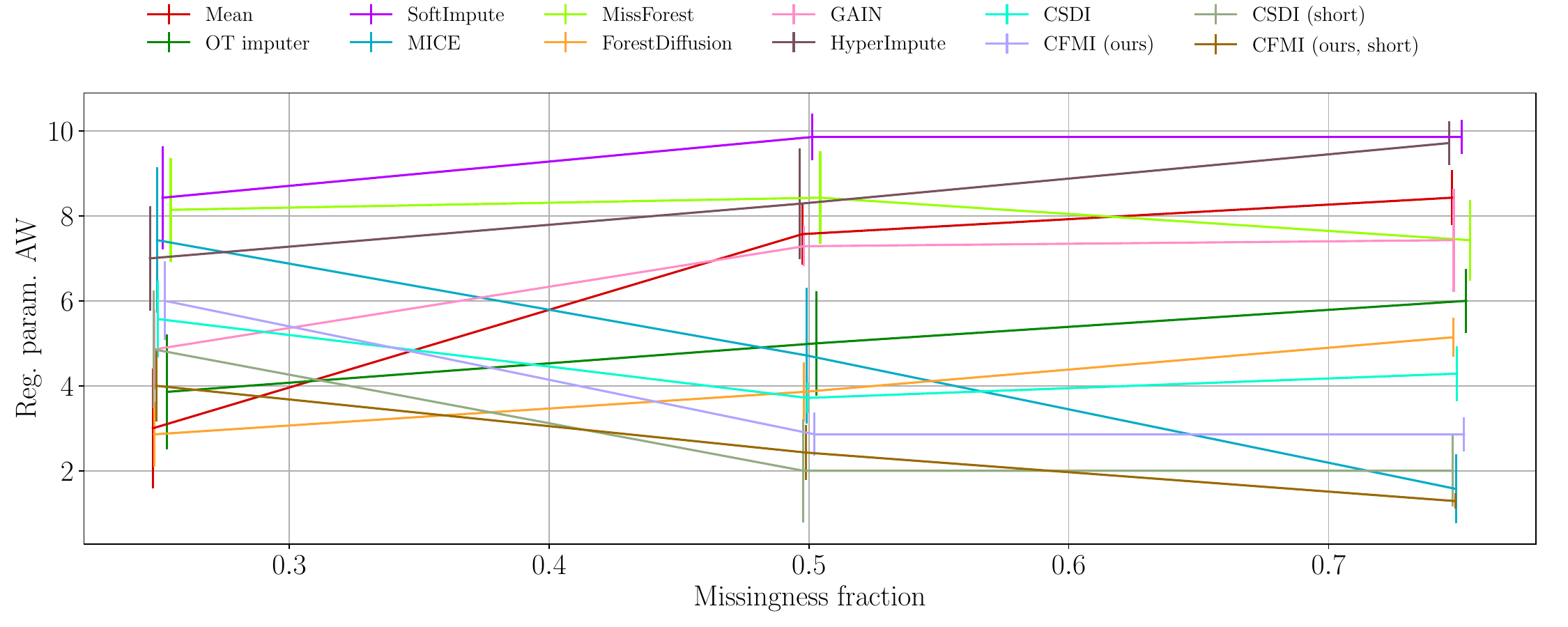}
  \caption{Average rank on regression parameter AW over all data sets versus missingness fraction. MCAR missingness.}
\end{figure}

\subsubsection{Imputation metrics vs number of training steps}
\label{apx:uci-metrics-vs-num-train-steps}

\begin{figure}[H]
  \centering
  \includegraphics[width=0.95\linewidth]{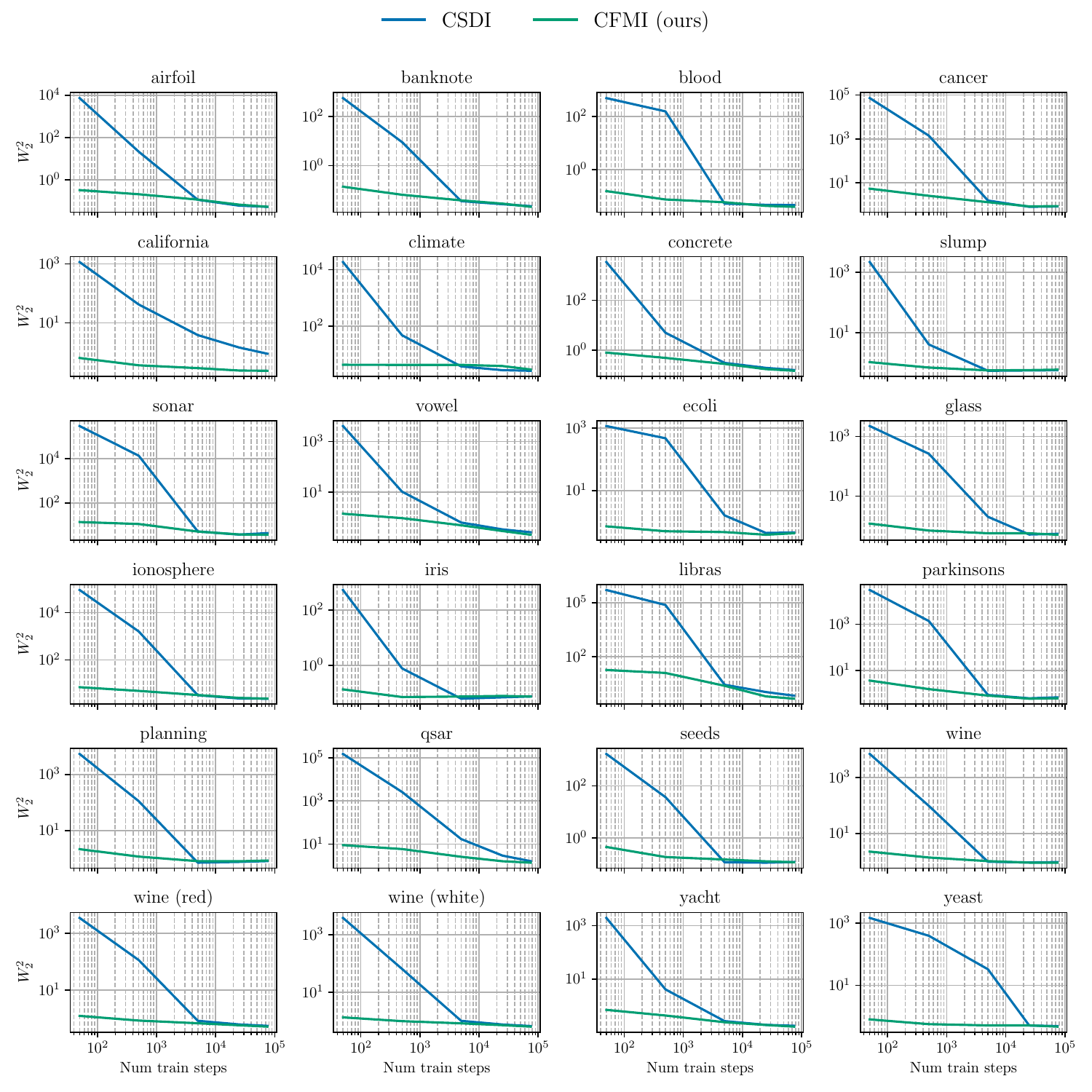}
  \caption{Wasserstein-2 metric vs number of training steps. MCAR 25\% missingness.}
\end{figure}

\begin{figure}[H]
  \centering
  \includegraphics[width=0.95\linewidth]{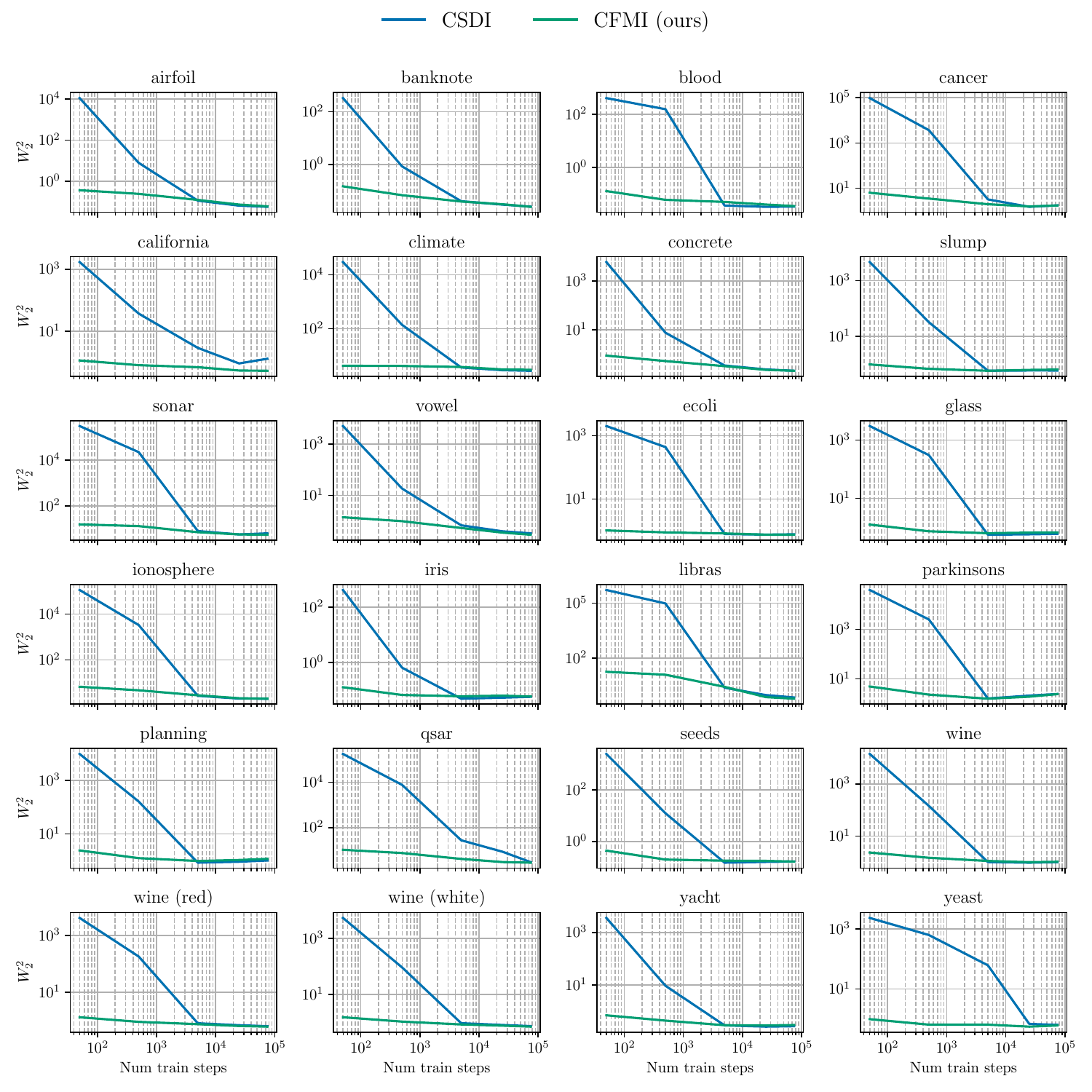}
  \caption{Wasserstein-2 metric vs number of training steps. MAR 25\% missingness.}
\end{figure}

\begin{figure}[H]
  \centering
  \includegraphics[width=0.95\linewidth]{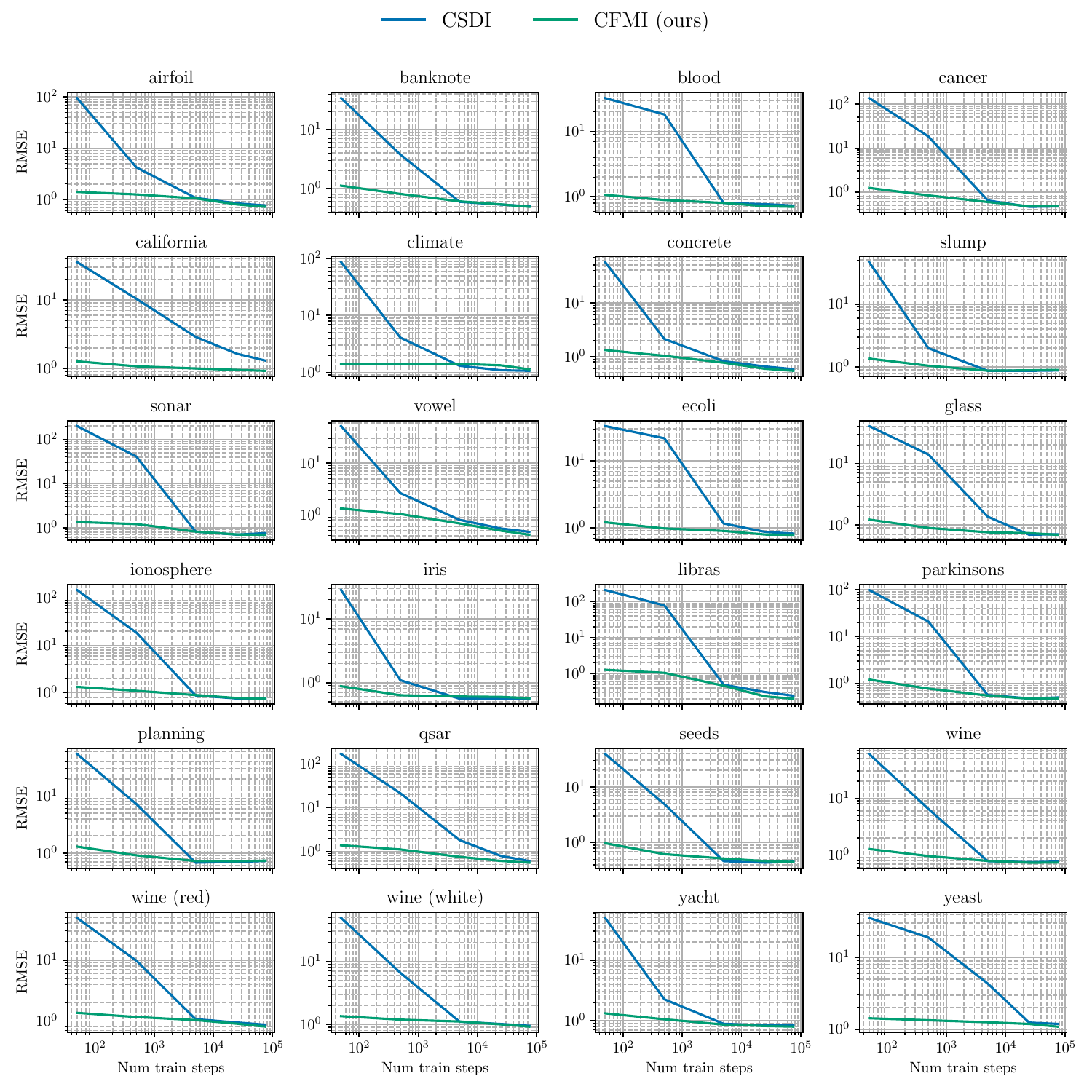}
  \caption{Average RMSE metric vs number of training steps. MCAR 25\% missingness.}
\end{figure}

\begin{figure}[H]
  \centering
  \includegraphics[width=0.95\linewidth]{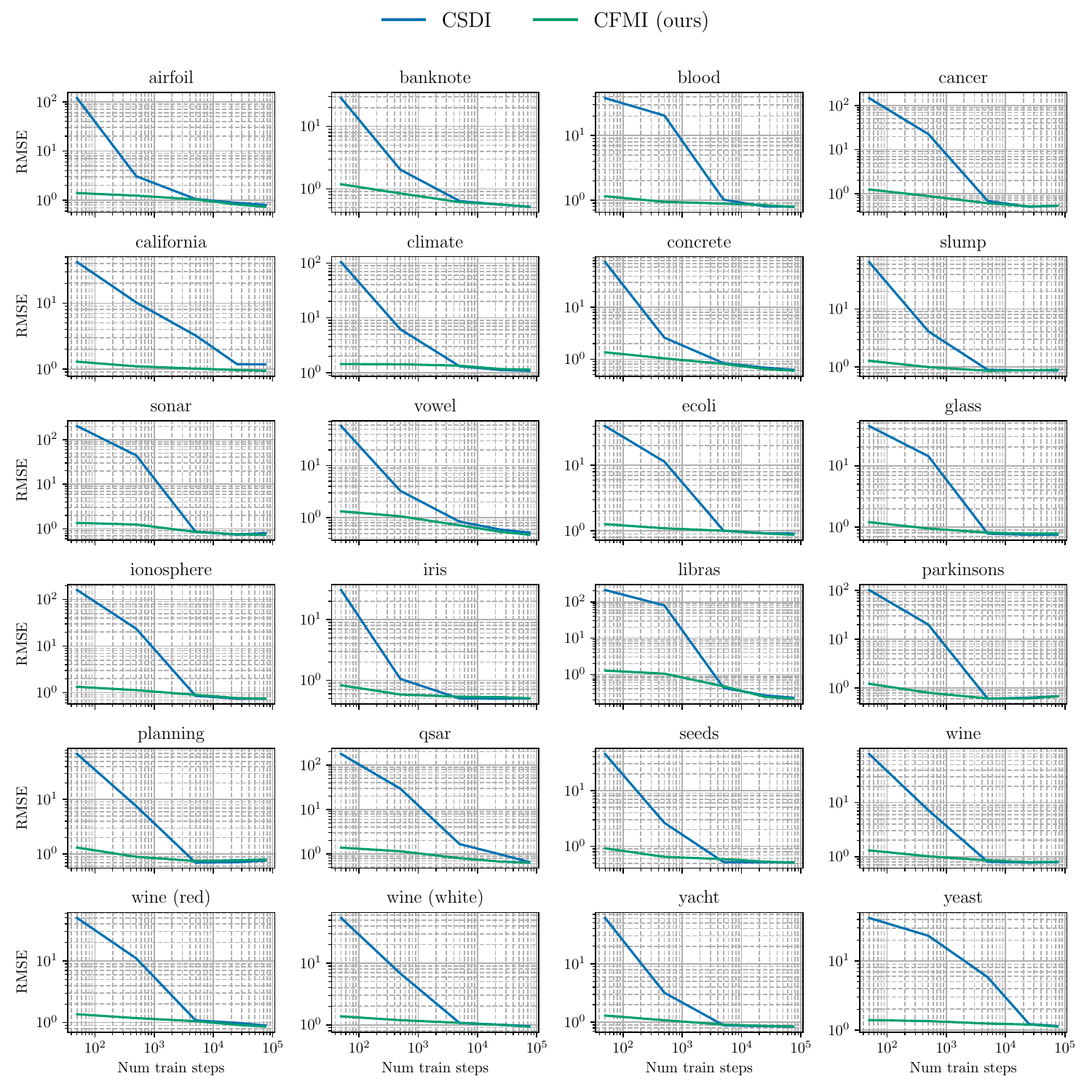}
  \caption{Average RMSE metric vs number of training steps. MAR 25\% missingness.}
\end{figure}

\begin{figure}[H]
  \centering
  \includegraphics[width=0.95\linewidth]{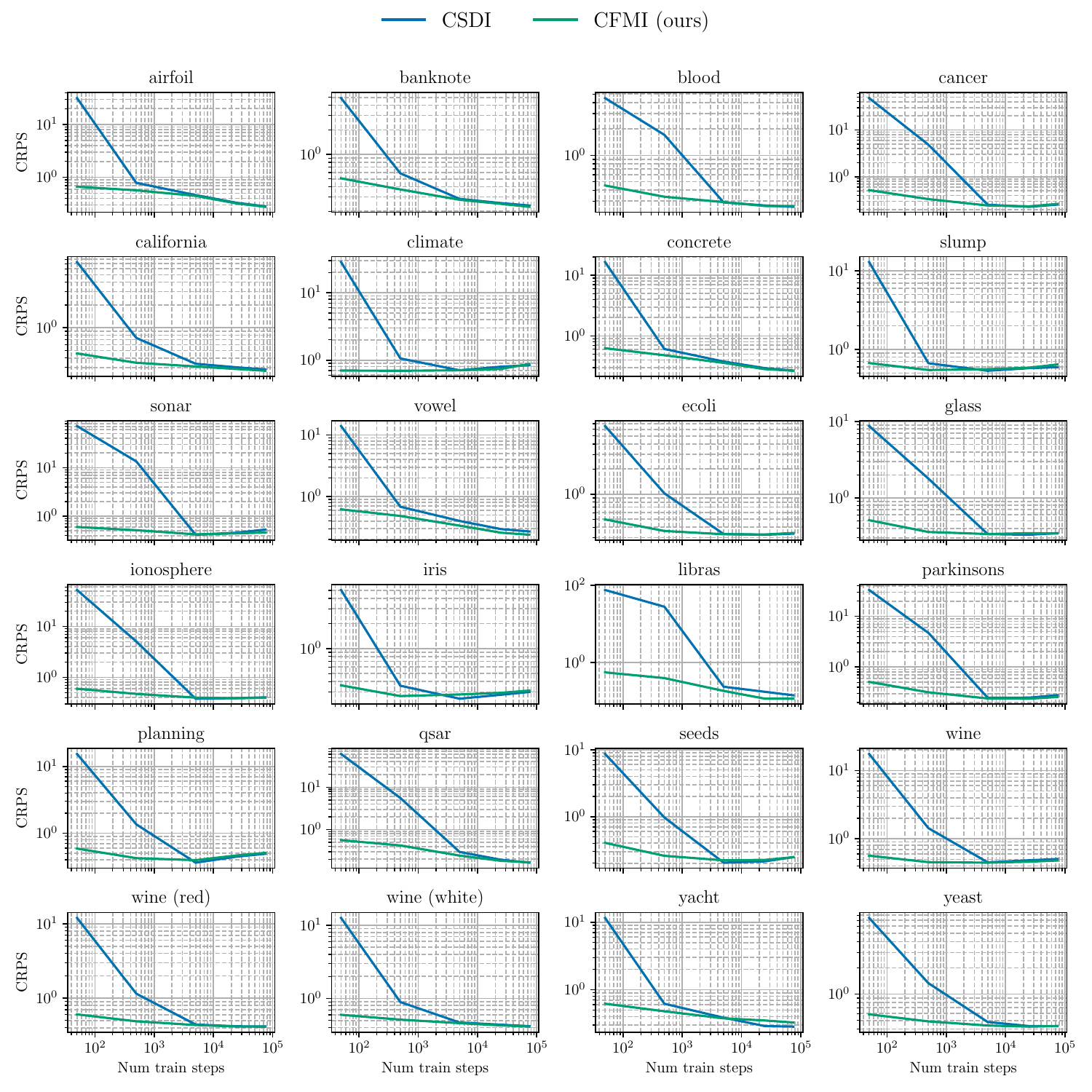}
  \caption{CRPS metric vs number of training steps. MCAR 25\% missingness.}
\end{figure}

\begin{figure}[H]
  \centering
  \includegraphics[width=0.95\linewidth]{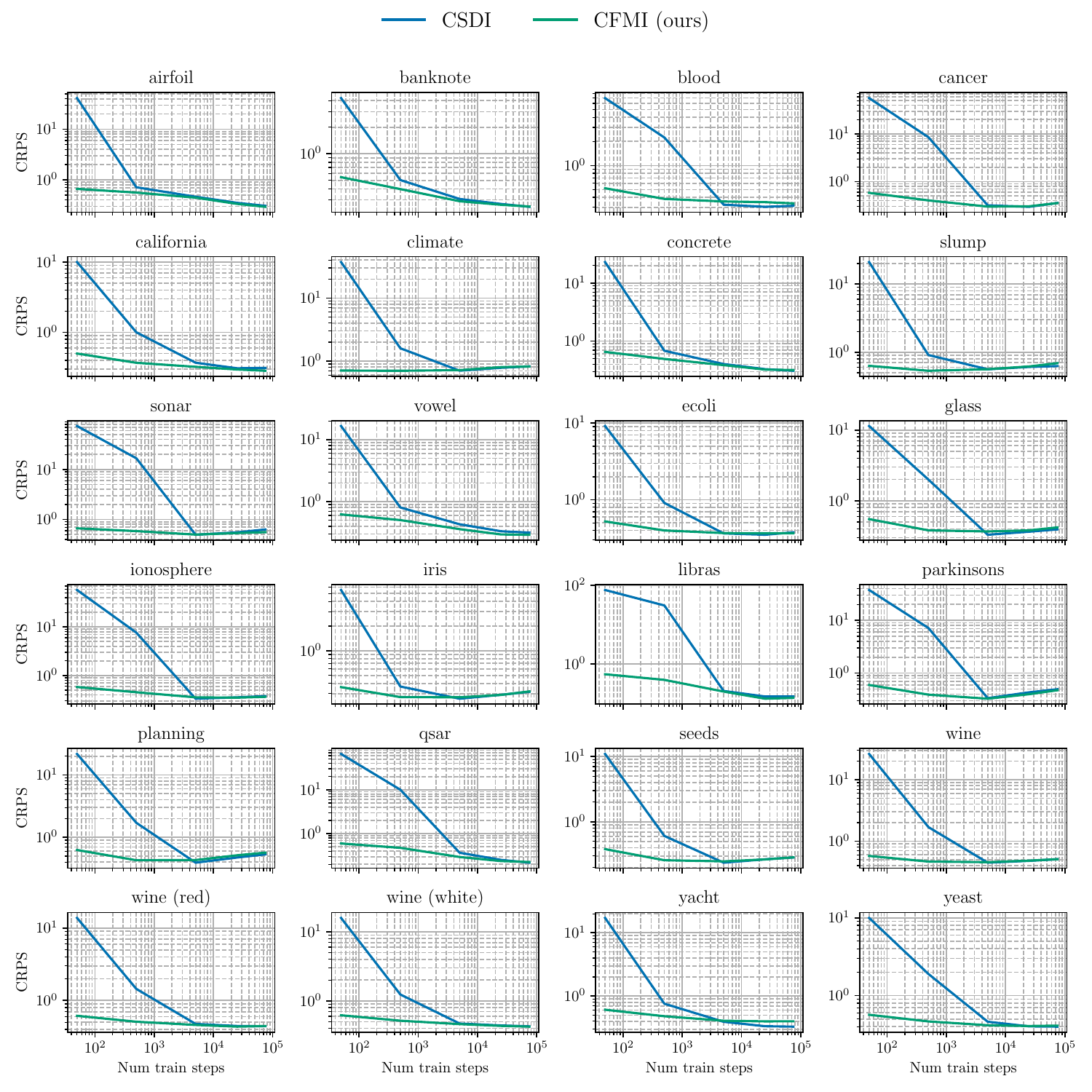}
  \caption{CRPS metric vs number of training steps. MAR 25\% missingness.}
\end{figure}

\begin{figure}[H]
  \centering
  \includegraphics[width=0.95\linewidth]{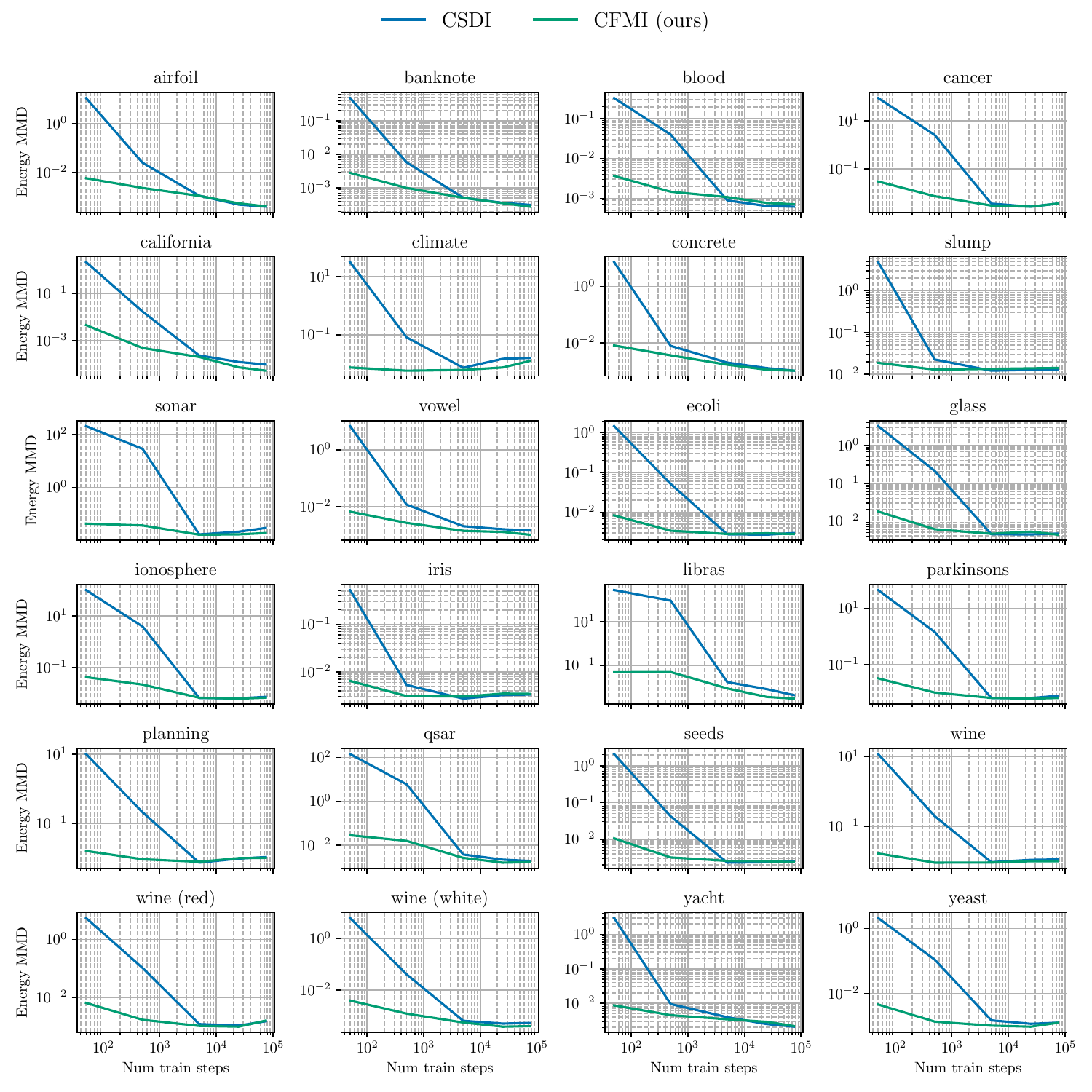}
  \caption{Energy MMD metric vs number of training steps. MCAR 25\% missingness.}
\end{figure}

\begin{figure}[H]
  \centering
  \includegraphics[width=0.95\linewidth]{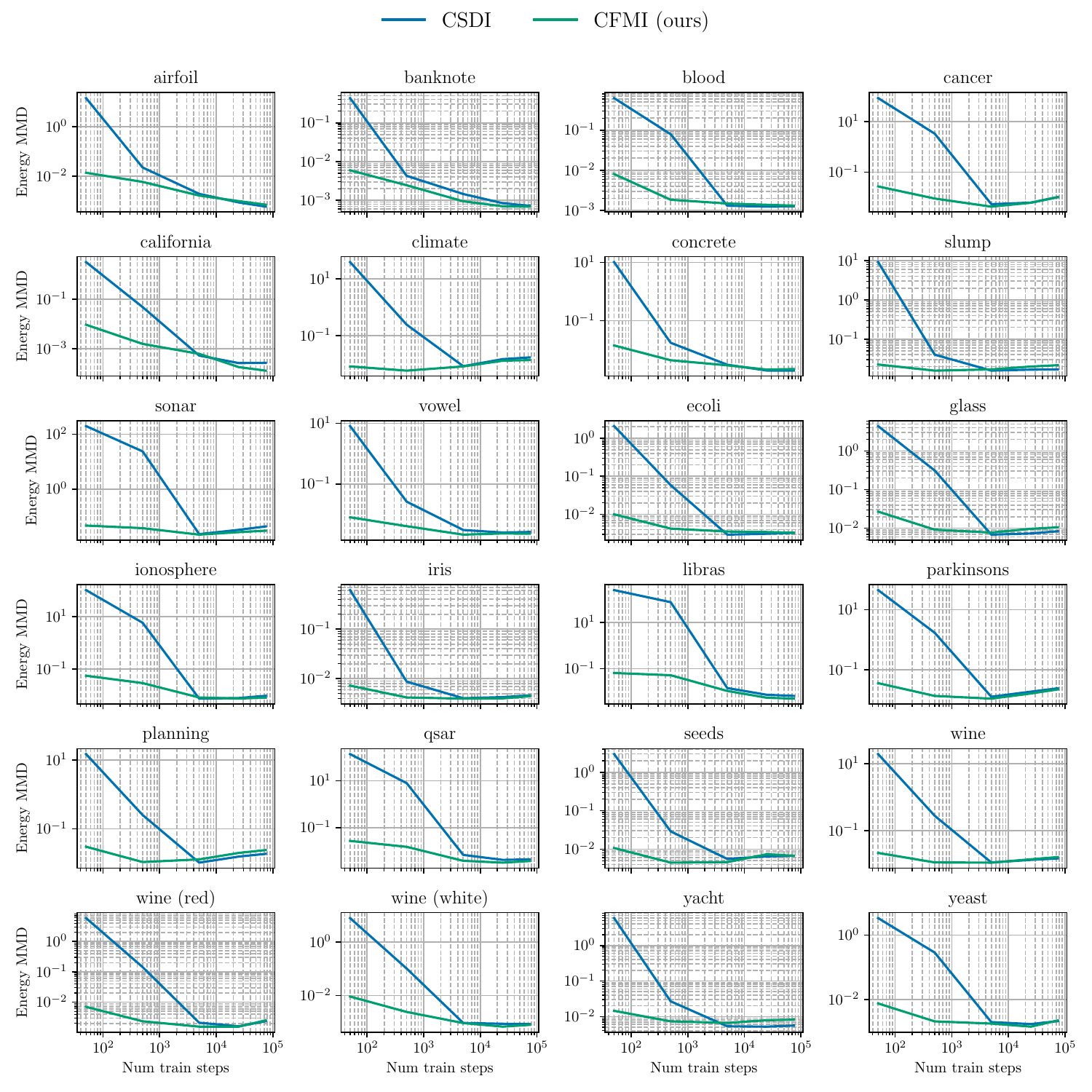}
  \caption{Energy MMC metric vs number of training steps. MAR 25\% missingness.}
\end{figure}

\begin{figure}[H]
  \centering
  \includegraphics[width=0.95\linewidth]{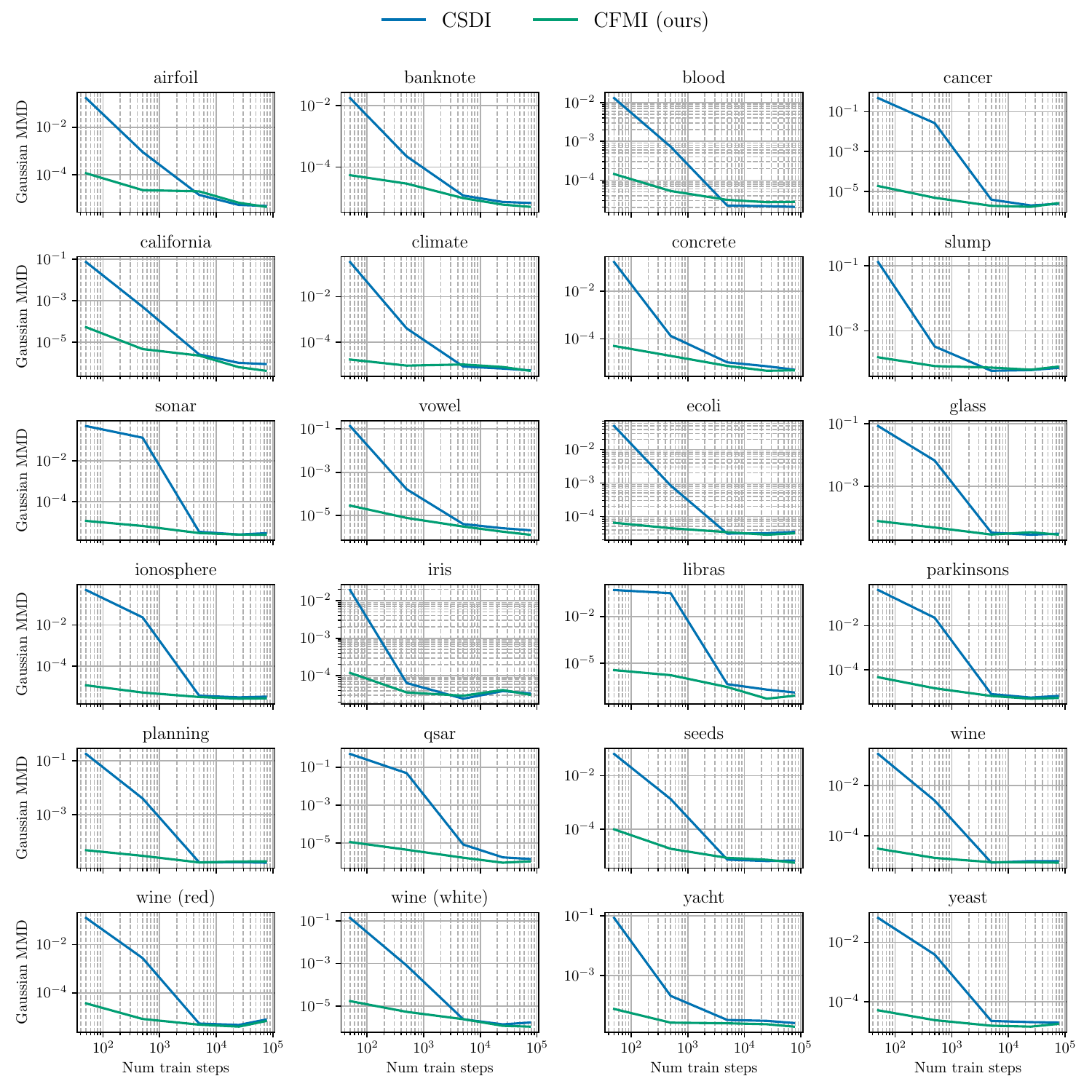}
  \caption{Gaussian MMD metric vs number of training steps. MCAR 25\% missingness.}
\end{figure}

\begin{figure}[H]
  \centering
  \includegraphics[width=0.95\linewidth]{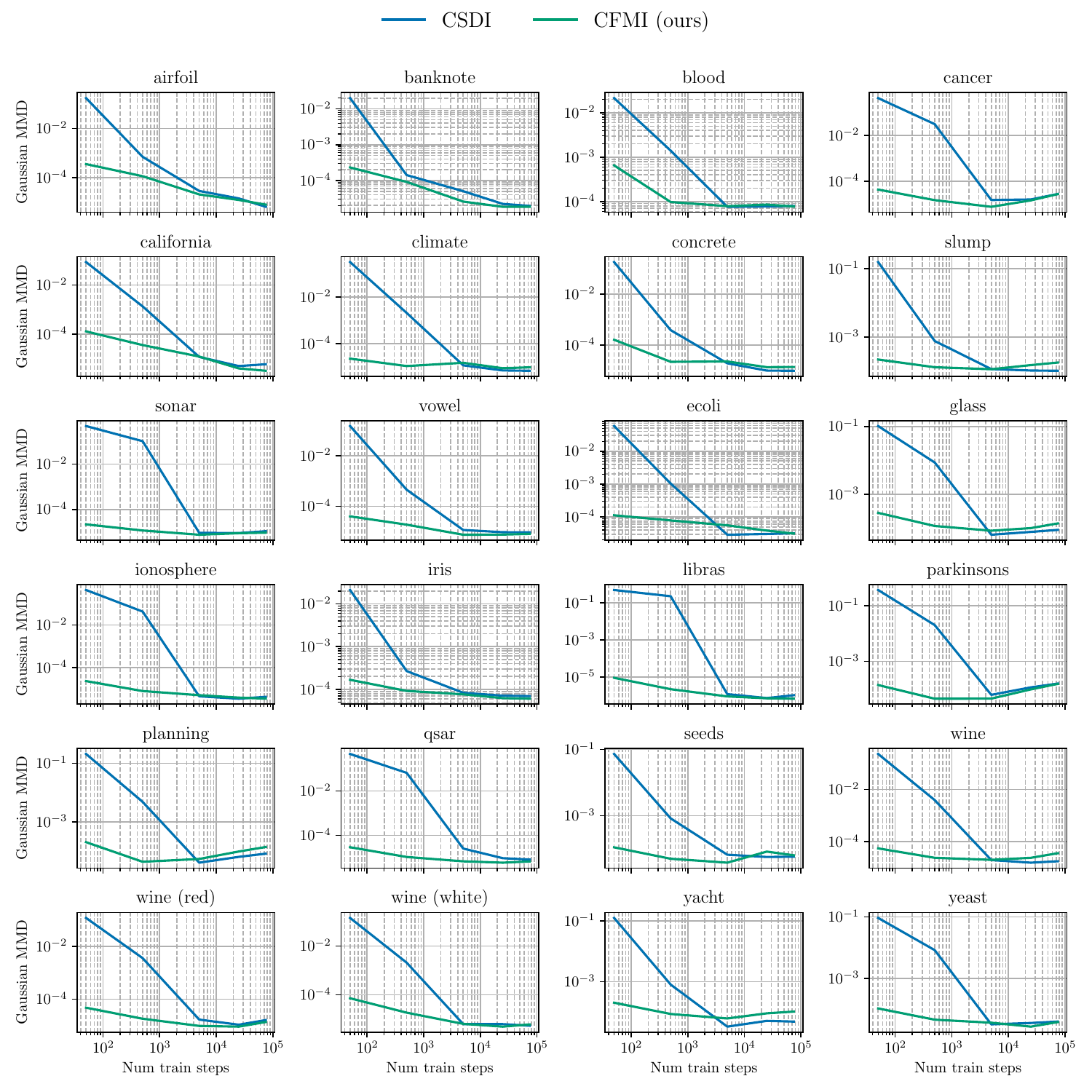}
  \caption{Gaussian MMC metrics vs number of training steps. MAR 25\% missingness.}
\end{figure}

\begin{figure}[H]
  \centering
  \includegraphics[width=0.95\linewidth]{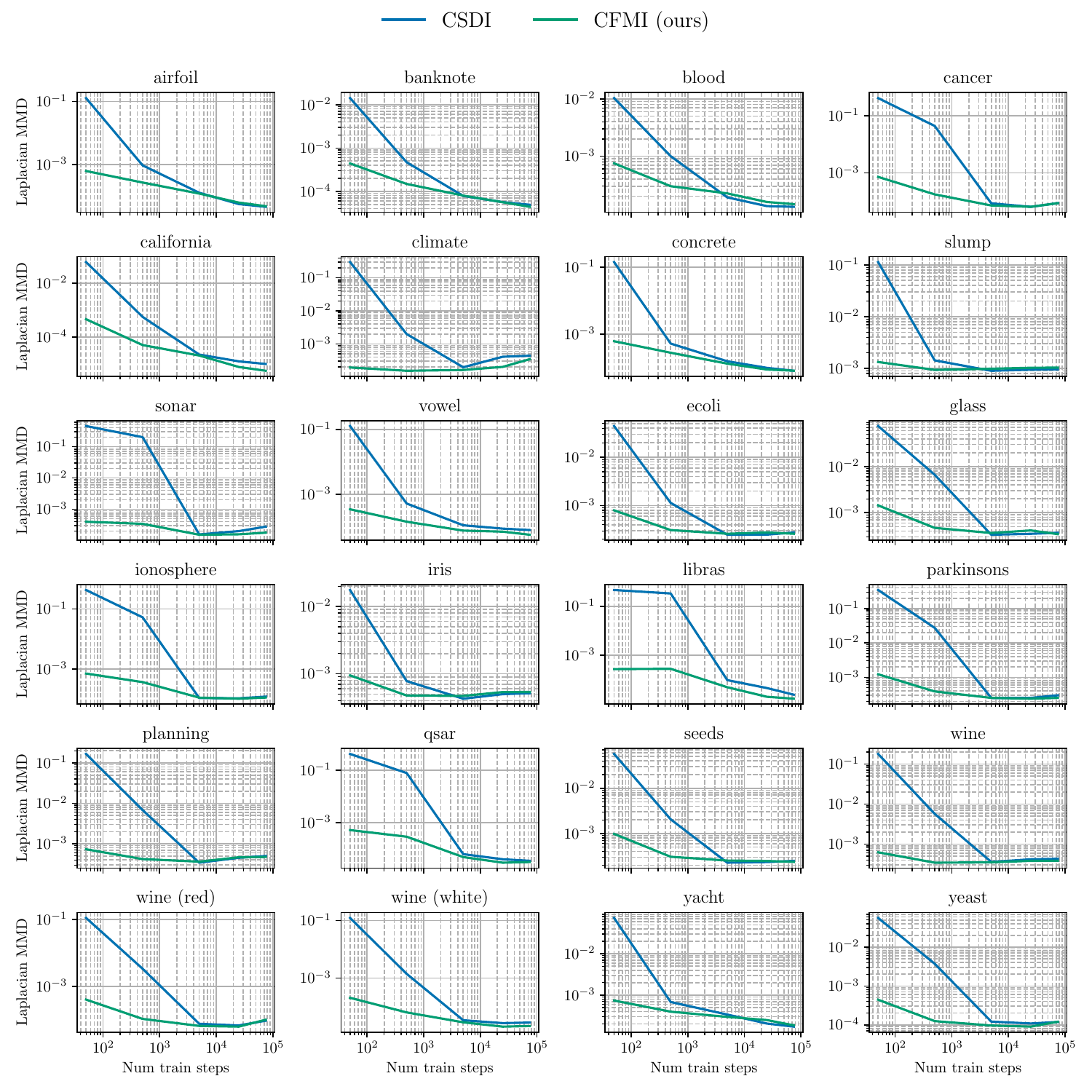}
  \caption{Laplacian MMD metric vs number of training steps. MCAR 25\% missingness.}
\end{figure}

\begin{figure}[H]
  \centering
  \includegraphics[width=0.95\linewidth]{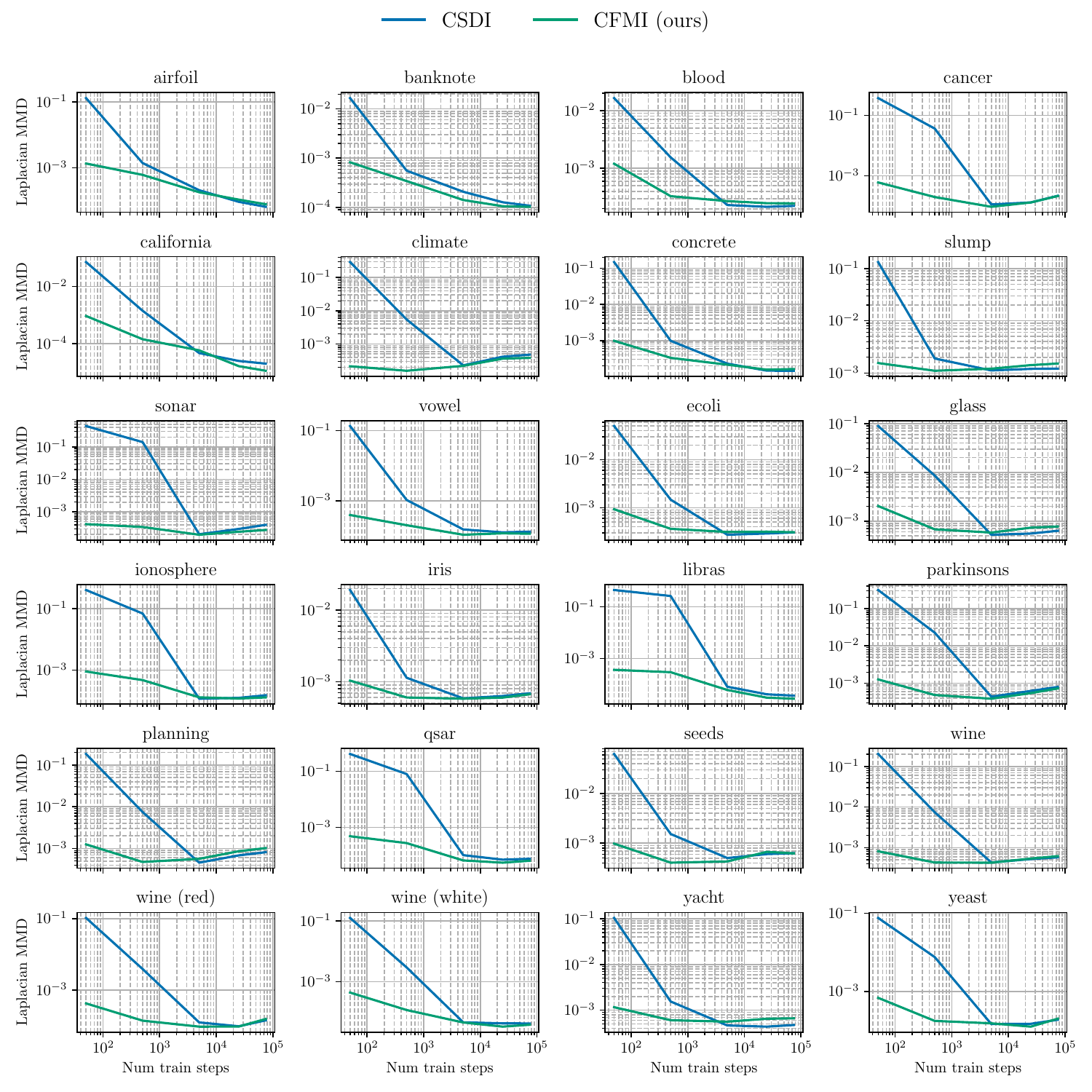}
  \caption{Laplacian MMC metric vs number of training steps. MAR 25\% missingness.}
\end{figure}

\begin{figure}[H]
  \centering
  \includegraphics[width=0.95\linewidth]{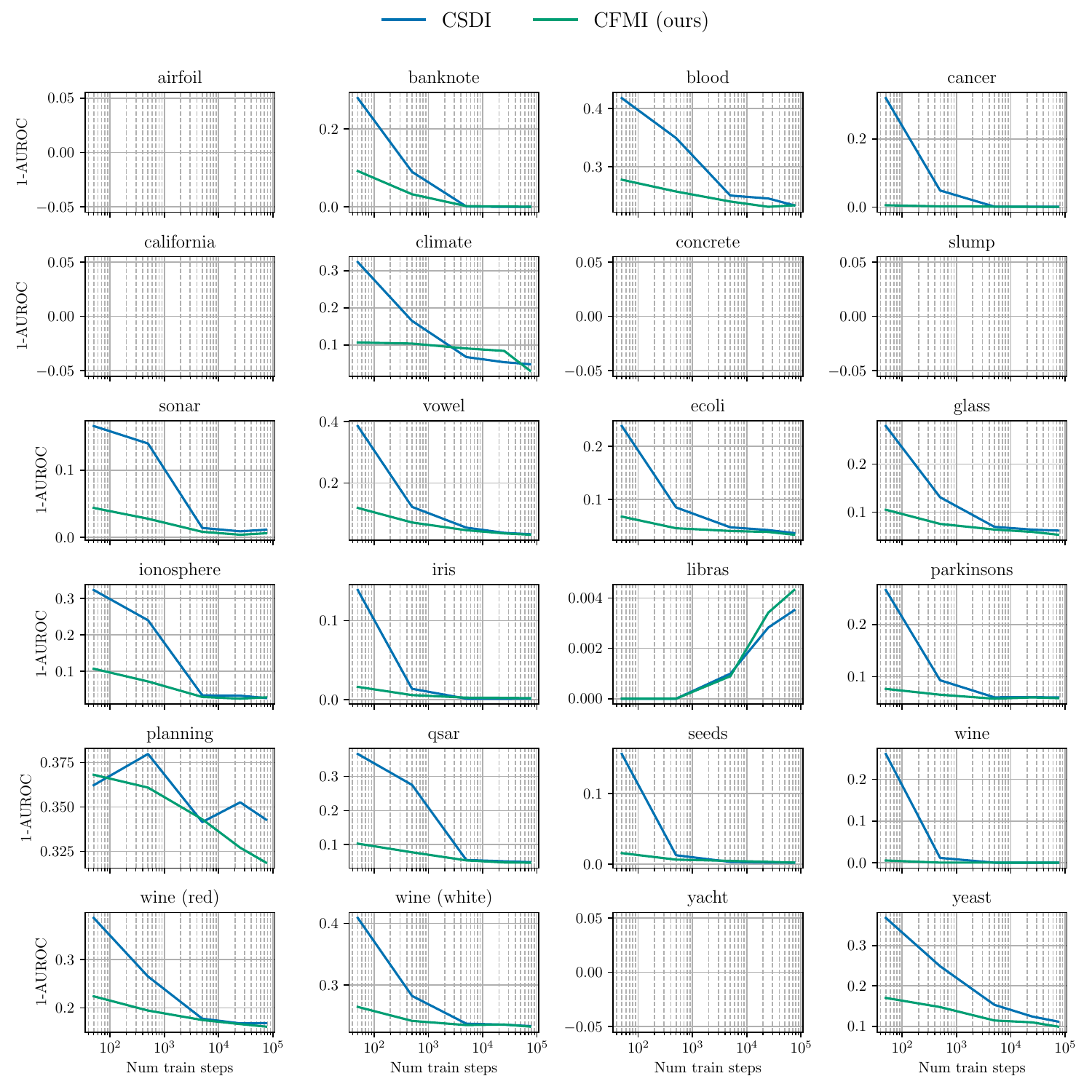}
  \caption{Classifier 1-AUROC metric vs number of training steps. MCAR 25\% missingness.}
\end{figure}

\begin{figure}[H]
  \centering
  \includegraphics[width=0.95\linewidth]{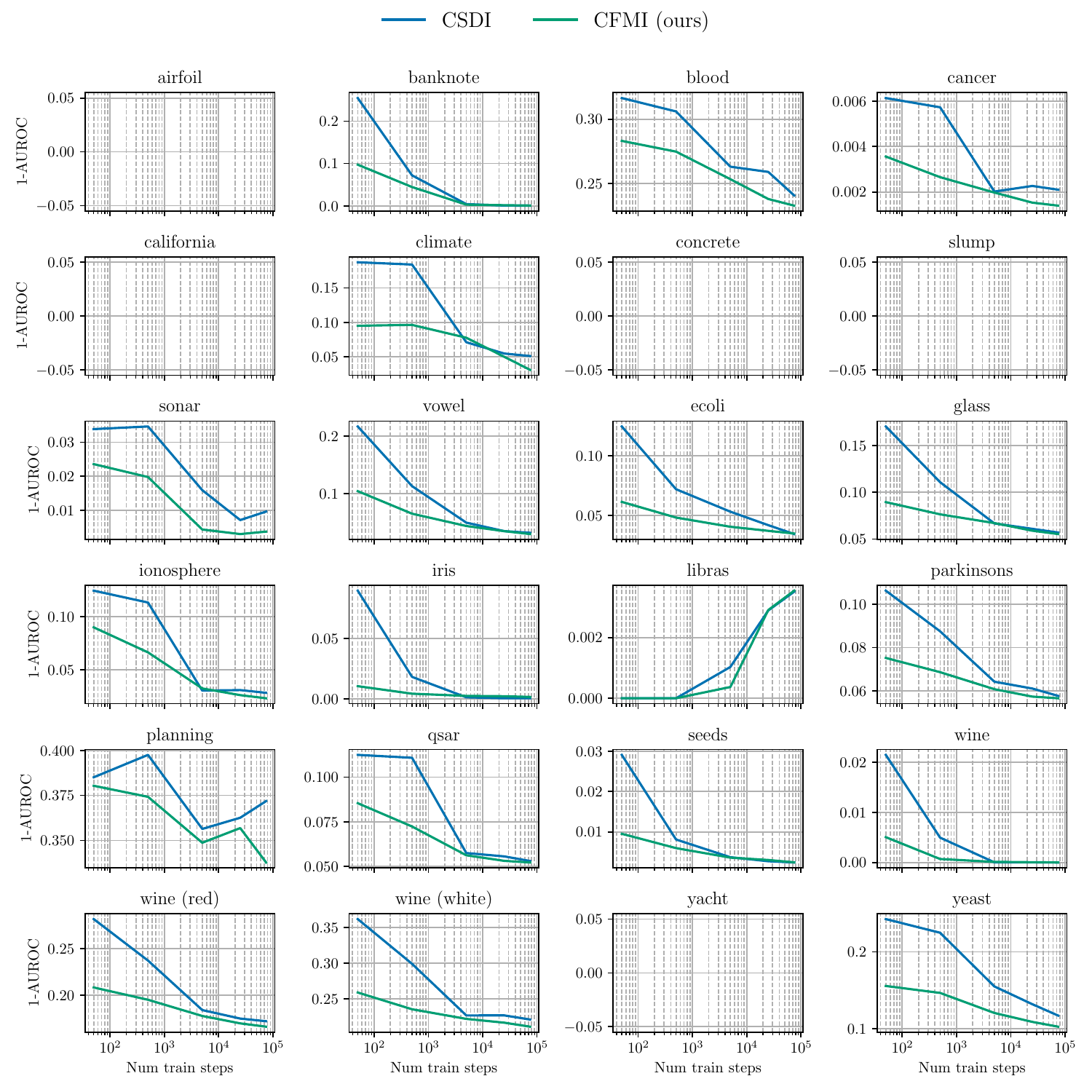}
  \caption{Classifier 1-AUROC metric vs number of training steps. MAR 25\% missingness.}
\end{figure}

\begin{figure}[H]
  \centering
  \includegraphics[width=0.95\linewidth]{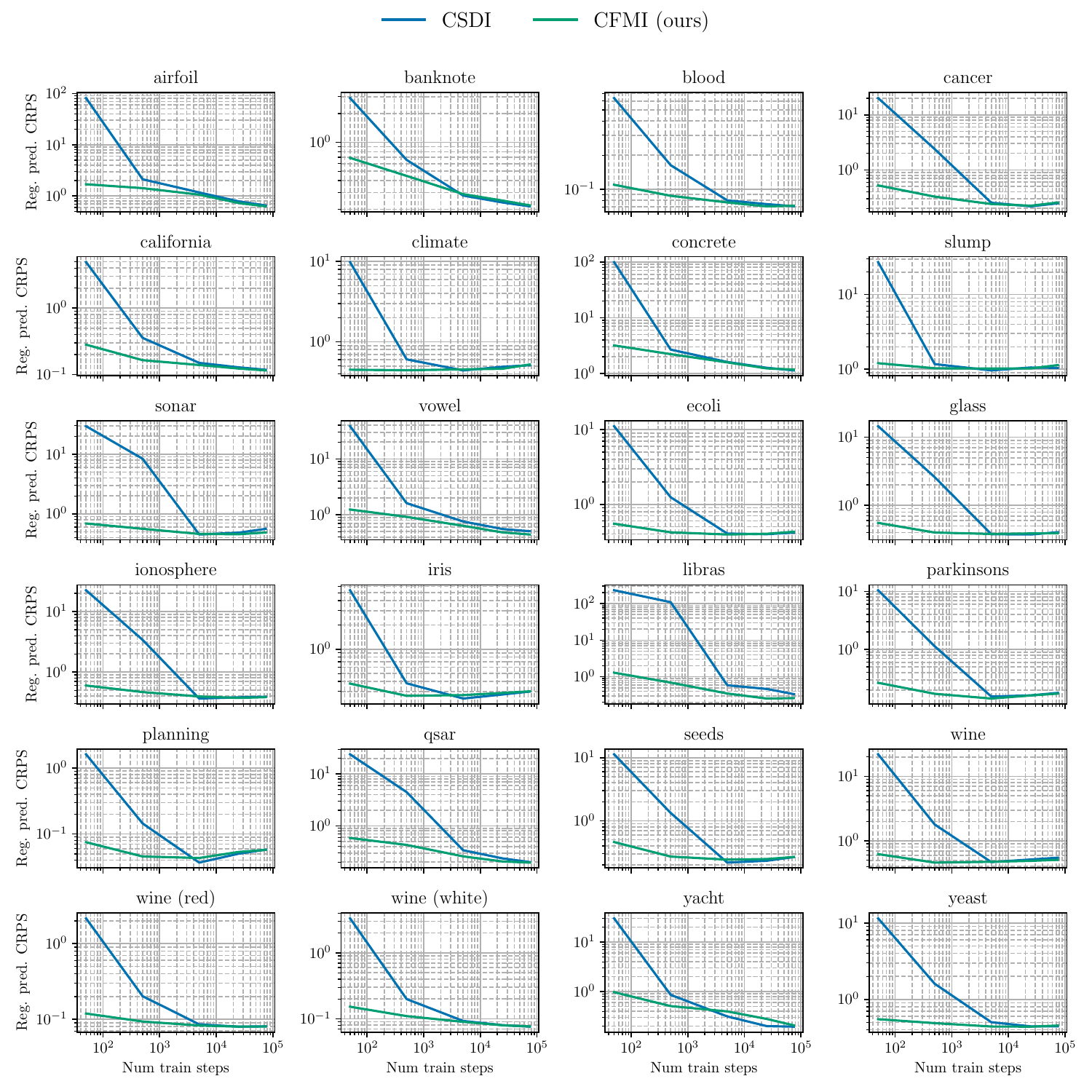}
  \caption{Regression CRPS metric vs number of training steps. MCAR 25\% missingness.}
\end{figure}

\begin{figure}[H]
  \centering
  \includegraphics[width=0.95\linewidth]{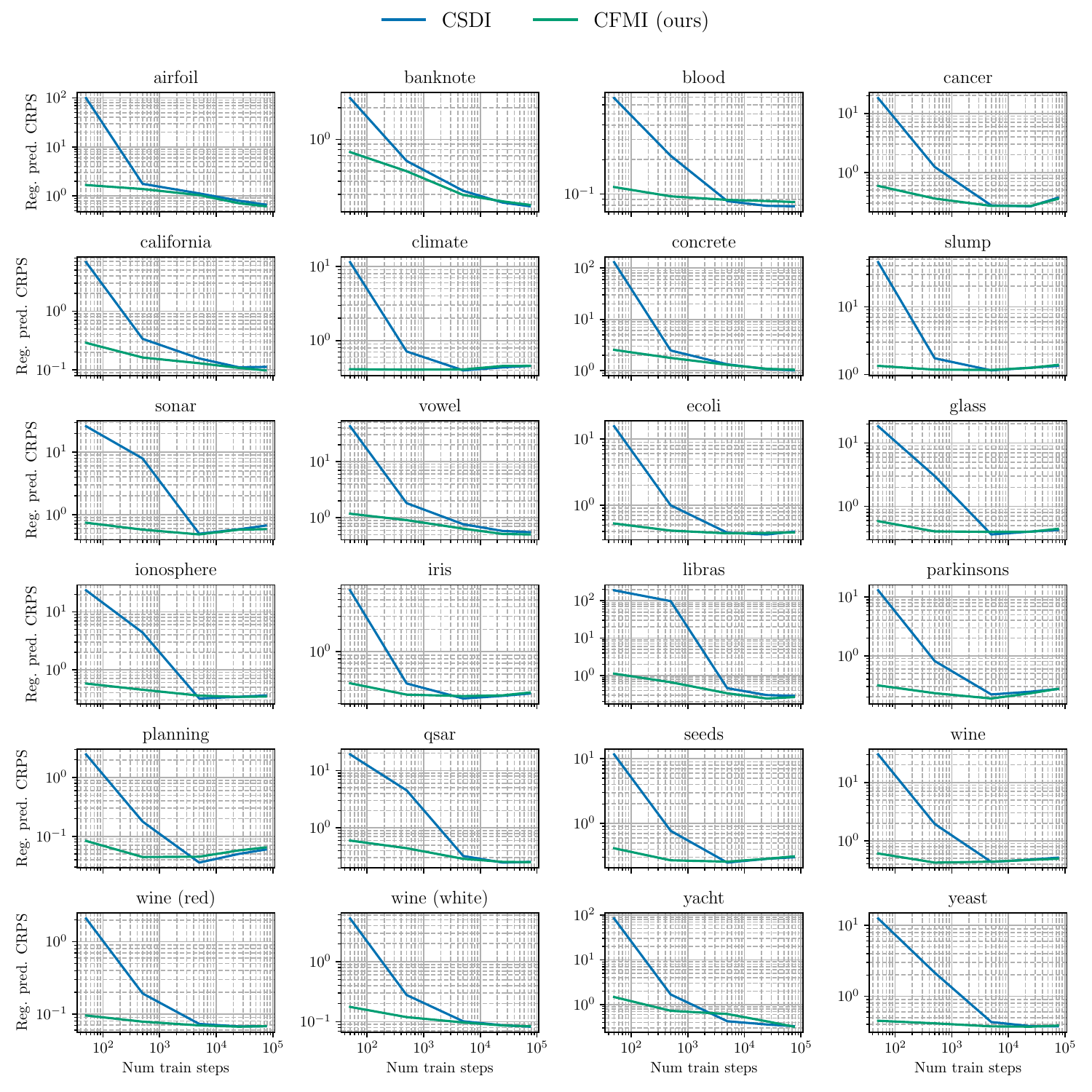}
  \caption{Regression CRPS metric vs number of training steps. MAR 25\% missingness.}
\end{figure}

\begin{figure}[H]
  \centering
  \includegraphics[width=0.95\linewidth]{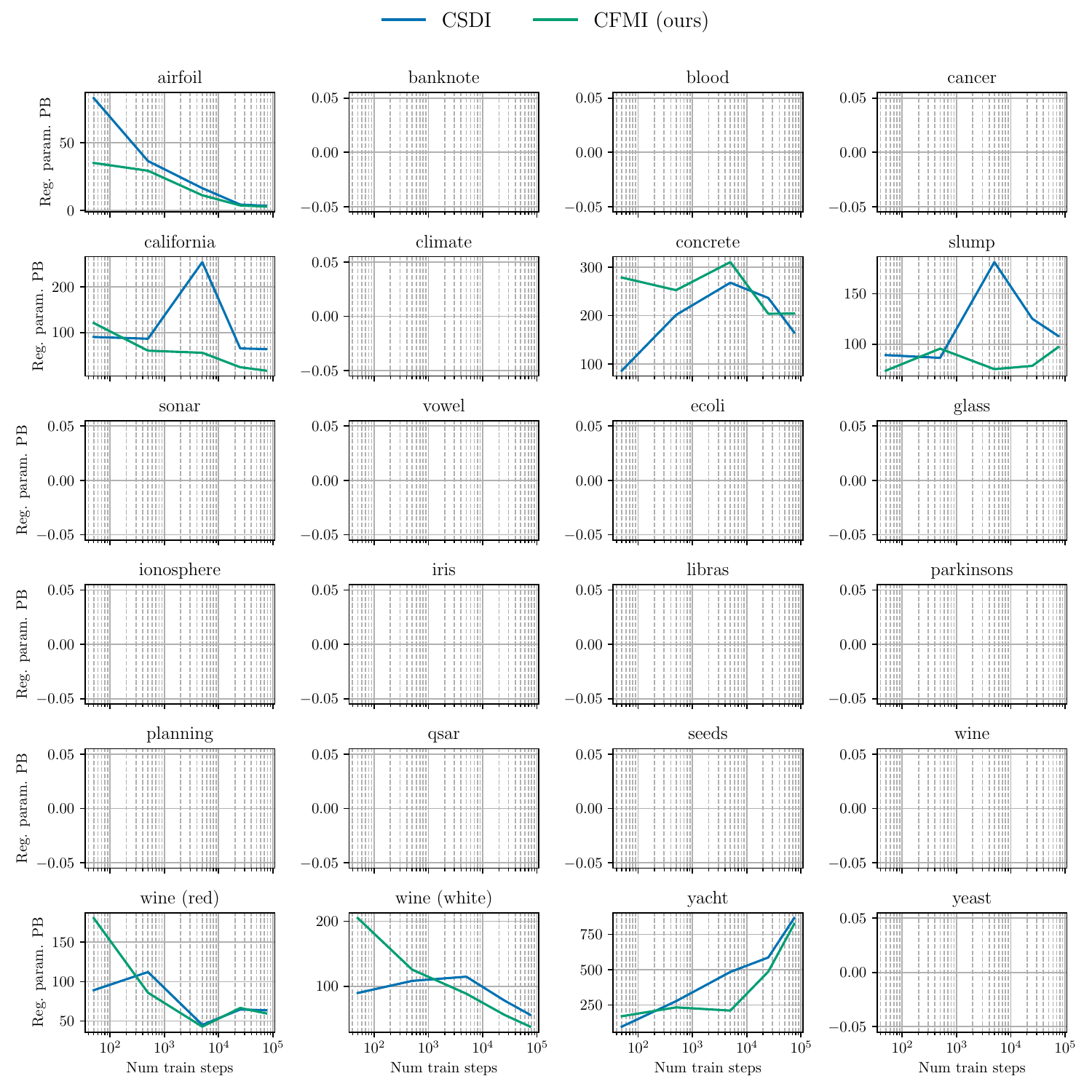}
  \caption{Regression parameter PB metric vs number of training steps. MCAR 25\% missingness.}
\end{figure}

\begin{figure}[H]
  \centering
  \includegraphics[width=0.95\linewidth]{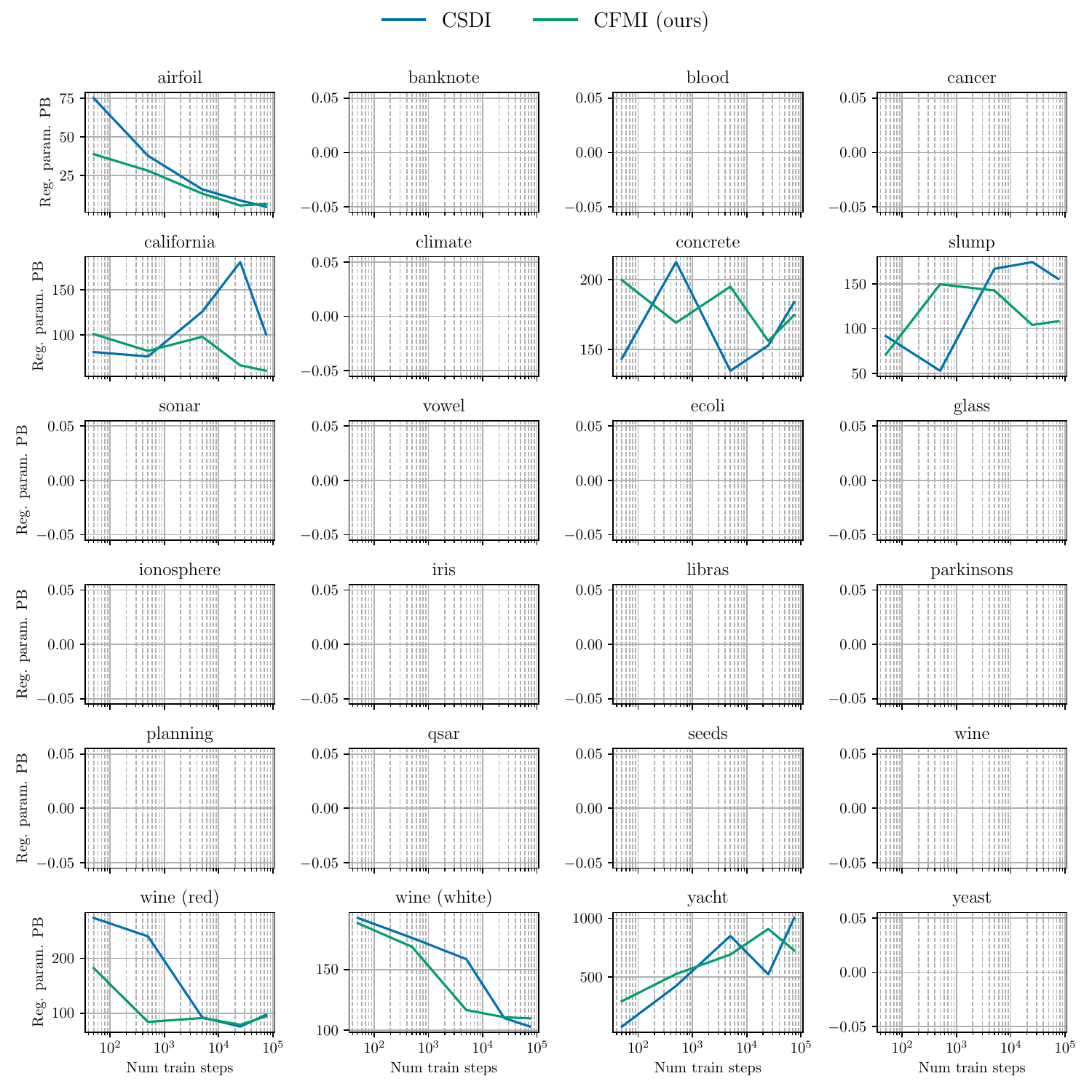}
  \caption{Regression parameter PB metric vs number of training steps. MAR 25\% missingness.}
\end{figure}

\begin{figure}[H]
  \centering
  \includegraphics[width=0.95\linewidth]{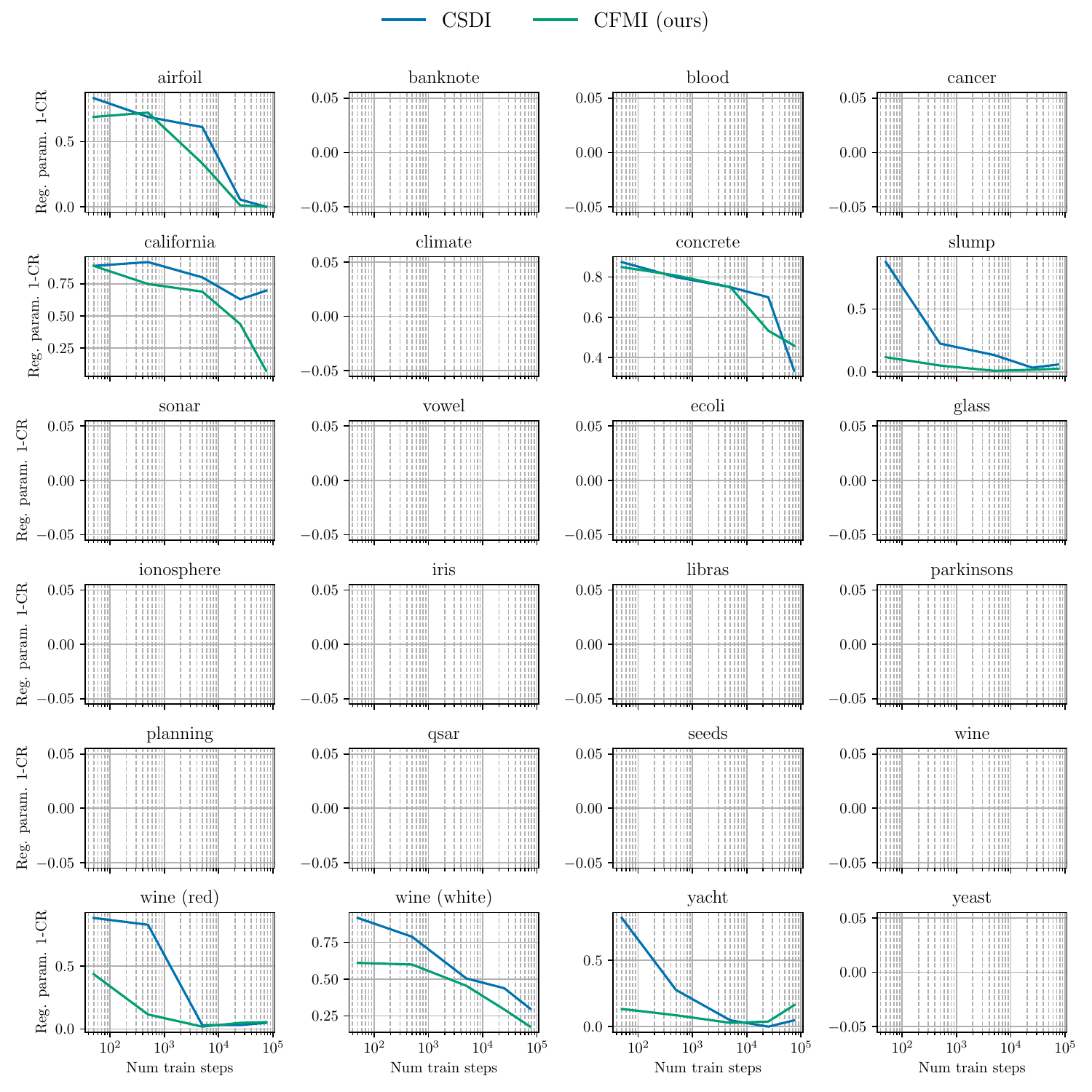}
  \caption{Regression parameter 1-CR metric vs number of training steps. MCAR 25\% missingness.}
\end{figure}

\begin{figure}[H]
  \centering
  \includegraphics[width=0.95\linewidth]{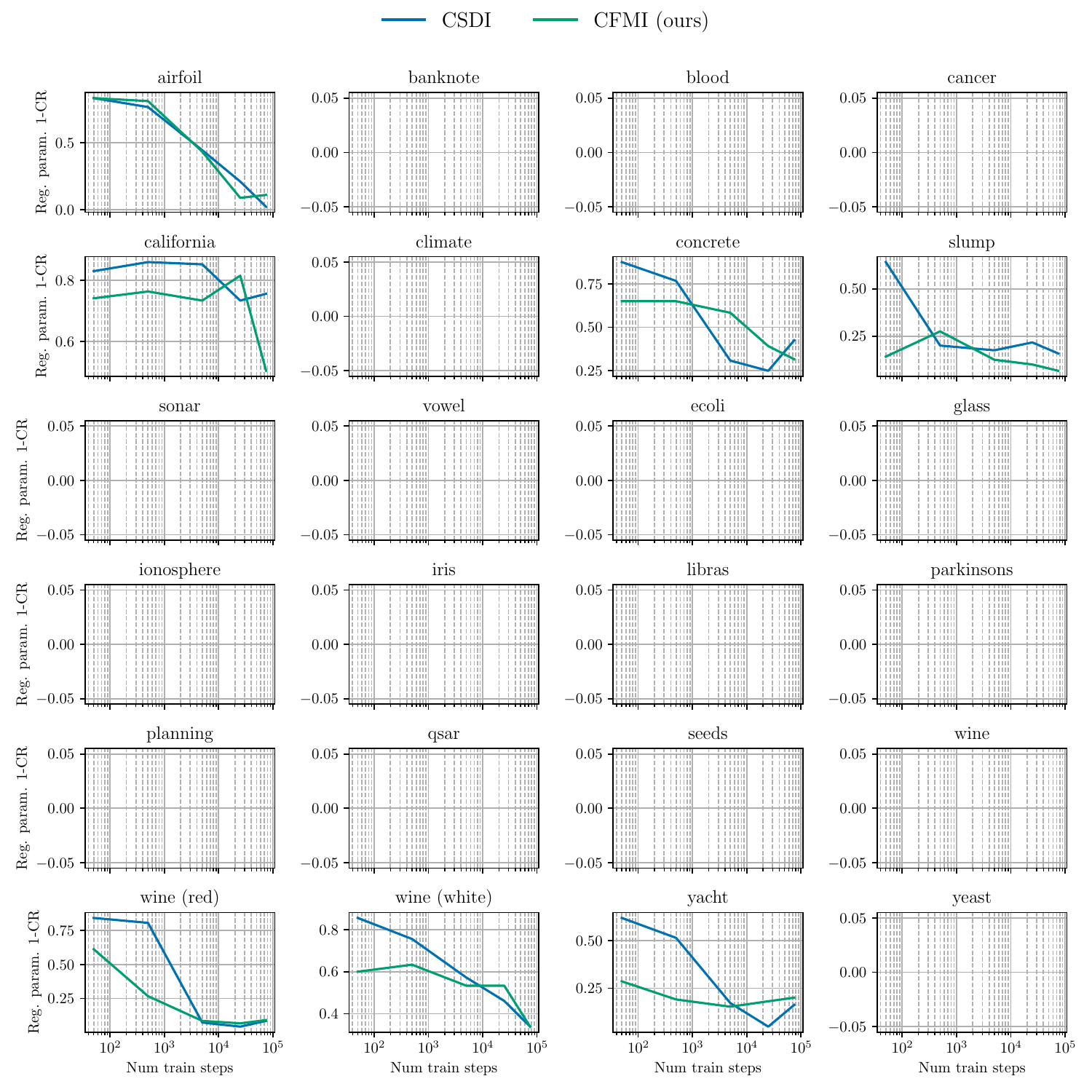}
  \caption{Regression parameter 1-CR metric vs number of training steps. MAR 25\% missingness.}
\end{figure}

\begin{figure}[H]
  \centering
  \includegraphics[width=0.95\linewidth]{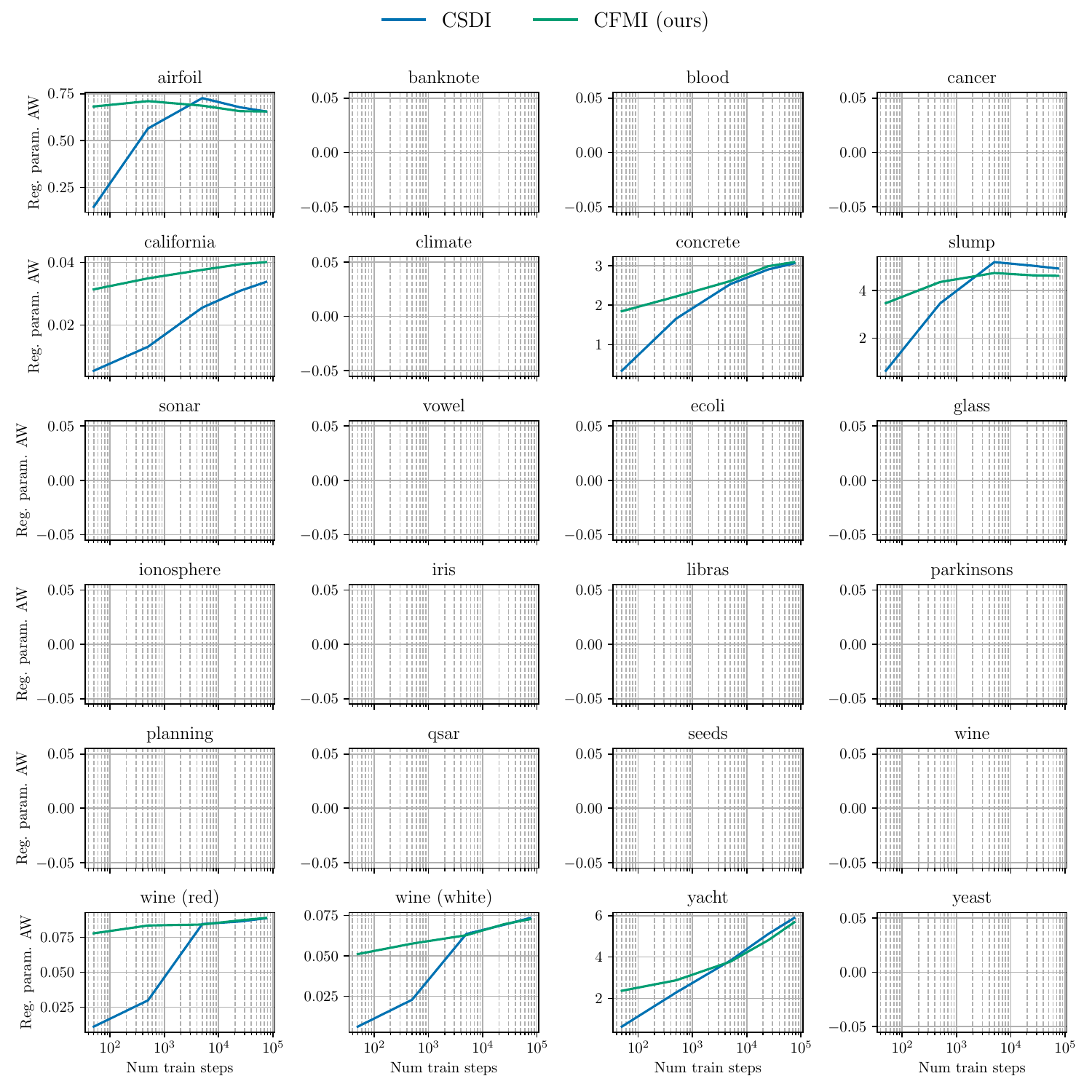}
  \caption{Regression parameter AW metric vs number of training steps. MCAR 25\% missingness.}
\end{figure}

\begin{figure}[H]
  \centering
  \includegraphics[width=0.95\linewidth]{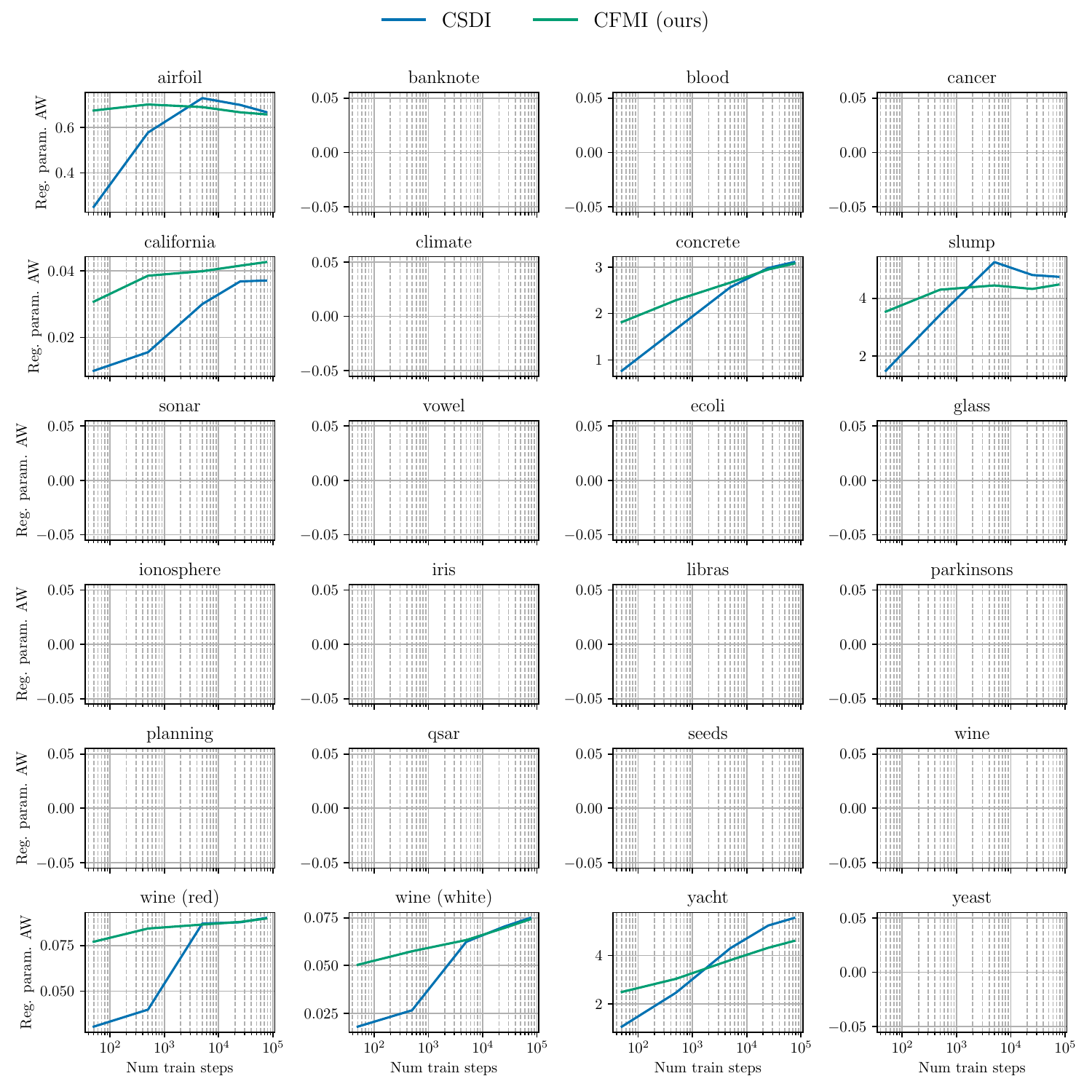}
  \caption{Regression parameter AW metric vs number of training steps. MAR 25\% missingness.}
\end{figure}

\newpage
\subsection{Time-series data sets}

\subsubsection{Raw results tables}
\label{apx:timeseries-raw-tables}

In this appendix section, we compare the performance of the proposed method, CFMI, with both our implementation of CSDI and the results reported by \citet{tashiroCSDIConditionalScorebased2021}. 
Overall, our CSDI implementation achieves results comparable to the original, with some notable differences. 
Specifically, our implementation demonstrates improved performance on the PM2.5 data set but slightly underperforms on the PhysioNet data set. 
This discrepancy is likely attributable to implementation differences, as detailed in \cref{apx:csdi-implementation}.

Despite these variations, CFMI consistently matches the performance of CSDI while demonstrating better efficiency, as discussed in \cref{sec:eval-time-series}.

\begin{table}[H]
\caption{CRPS on the time-series data. We report the mean and standard error for five runs.}
\begin{center}
\begin{tabular}{lccccc} 
\toprule
& \multicolumn{3}{c} { PhysioNet } & PM2.5 \\
\cmidrule{2-4} & $10 \%$ missing & $50 \%$ missing & $90 \%$ missing & \\
\midrule
CSDI & $0.219 (0.002)$ & $0.294 (0.005)$ & $0.446 (0.008)$ & $6.767 (0.018)$ \\    
CFMI (ours) & $0.220 (0.004)$ & $0.282 (0.010)$ & $0.410 (0.014)$ & $6.726 (0.036)$ \\
\bottomrule
\end{tabular}
\end{center}
\end{table}

\begin{table}[H]
\caption{Magnitude-normalised CRPS on the time-series data. We report the mean and standard error for five runs. Asterisks mean the results were copied from \citep{tashiroCSDIConditionalScorebased2021}.}
\begin{center}
\begin{tabular}{lcccc}
\toprule
& \multicolumn{3}{c} { PhysioNet } & PM2.5 \\
\cmidrule{2-4} & $10 \%$ missing & $50 \%$ missing& $90 \%$ missing& \\
    \midrule 
    Multitask GP~\citep{bonillaMultitaskGaussianProcess2007} (*) & $0.489(0.005)$ & $0.581(0.003)$ & $0.942(0.010)$ & $0.301(0.003)$ \\
    GP-VAE~\citep{fortuinGPVAEDeepProbabilistic2020} (*) & $0.574(0.003)$ & $0.774(0.004)$ & $0.998(0.001)$ & $0.397(0.009)$ \\
    V-RIN~\citep{mulyadiUncertaintyAwareVariationalRecurrentImputation2020} (*) & $0.808(0.008)$ & $0.831(0.005)$ & $0.922(0.003)$ & $0.526(0.025)$ \\ 
    CSDI~\citep{tashiroCSDIConditionalScorebased2021} (*) & ${0.238(0.001)}$ & ${ 0.330(0.002)}$ & ${0.522(0.002)}$ & ${0.108(0.001)}$ \\
    \midrule
    CSDI~\citep{tashiroCSDIConditionalScorebased2021} & $0.240 (0.003)$ & $0.333 (0.003)$ & $0.550 (0.006)$ & $0.099 (0.000)$ \\
    CFMI (ours) & $0.250 (0.004)$ & $0.338 (0.007)$ & $0.538 (0.019)$ & $0.098 (0.001)$ \\
\bottomrule		
\end{tabular}		
\end{center}
\end{table}

\begin{table}[H]
\caption{MAE on the time-series data. We report the mean and standard error for five runs. Asterisks mean the results were copied from \citep{tashiroCSDIConditionalScorebased2021}.}
\begin{center}
\begin{tabular}{lcccc} 
\toprule
& \multicolumn{3}{c} { PhysioNet } & PM2.5 \\
\cmidrule{2-4} & $10 \%$ missing & $50 \%$ missing& $90 \%$ missing& \\
\midrule V-RIN~\citep{mulyadiUncertaintyAwareVariationalRecurrentImputation2020} (*) & $0.271(0.001)$ & $0.365(0.002)$ & $0.606(0.006)$ & $25.4(0.62)$ \\
BRITS~\citep{caoBRITSBidirectionalRecurrent2018} (*) & $0.284(0.001)$ & $0.368(0.002)$ & $0.517(0.002)$ & $14.11(0.26)$ \\
GLIMA~\citep{suoGLIMAGlobalLocal2020} (*) & $0.265$ & $-$ & $-$ & $10.54$ \\ 
RDIS~\citep{choiRDISRandomDrop2023} (*) & $0.319(0.002)$ & $0.419(0.002)$ &	$0.631(0.002)$& $22.11(0.35)$\\ 
CSDI~\citep{tashiroCSDIConditionalScorebased2021} (*) & ${0.217(0.001)}$ &  ${0.301(0.002)}$ & ${0.481(0.003)}$ & ${9.60(0.04)}$ \\
\midrule
CSDI~\citep{tashiroCSDIConditionalScorebased2021} & $0.228 (0.002)$ & $0.319 (0.002)$ &  $0.530 (0.005)$ & $9.430 (0.021)$ \\
CFMI (ours) & $0.234 (0.002)$ & $0.319 (0.005)$ & $0.504 (0.009)$ & $9.404 (0.050)$\\
\bottomrule
\end{tabular}	
\end{center}
\end{table}

\begin{table}[H]
\caption{RMSE on the time-series data. We report the mean and standard error for five runs. Asterisks mean the results were copied from \citep{tashiroCSDIConditionalScorebased2021}.}
\begin{center}
\begin{tabular}{lccccc} 
\toprule
& \multicolumn{3}{c} { PhysioNet } & PM2.5 \\
\cmidrule{2-4} & $10 \%$ missing & $50 \%$ missing & $90 \%$ missing & \\
\midrule V-RIN~\citep{mulyadiUncertaintyAwareVariationalRecurrentImputation2020} (*) & $0.628(0.025)$ & $0.693(0.022)$ & $0.928(0.013)$ & $40.11(1.14)$ \\
 BRITS~\citep{caoBRITSBidirectionalRecurrent2018} (*) & $0.619(0.022)$ & $0.693(0.023)$ & $0.836(0.015)$ & $24.47(0.73)$ \\
 RDIS~\citep{choiRDISRandomDrop2023} (*) & $0.633(0.021)$ & $0.741(0.018)$ &	$0.934(0.013)$& $37.49(0.28)$ \\
 SSGAN~\citep{miaoGenerativeSemisupervisedLearning2021} (*) & $0.598$ & $0.762$ & $0.818$ & $-$ \\
 CSDI~\citep{tashiroCSDIConditionalScorebased2021} (*) & ${0.498(0.020)}$ & ${0.614(0.017)}$ & ${0.803(0.012)}$ & ${19.30(0.13)}$ \\
 \midrule
 CSDI~\citep{tashiroCSDIConditionalScorebased2021} & $0.582 (0.055)$ & $0.678 (0.044)$ & $0.850 (0.013)$ & $17.978 (0.089)$ \\    
 CFMI (ours) & $0.538 (0.052)$ & $0.624 (0.019)$ & $0.822 (0.019)$ & $17.701 (0.106)$ \\
\bottomrule
\end{tabular}
\end{center}
\end{table}

\subsubsection{Imputation metrics vs number of training epochs}
\label{apx:timeseries-metrics-vs-num-train-epochs}

\begin{figure}[H]
  \centering
  \includegraphics[width=0.95\linewidth]{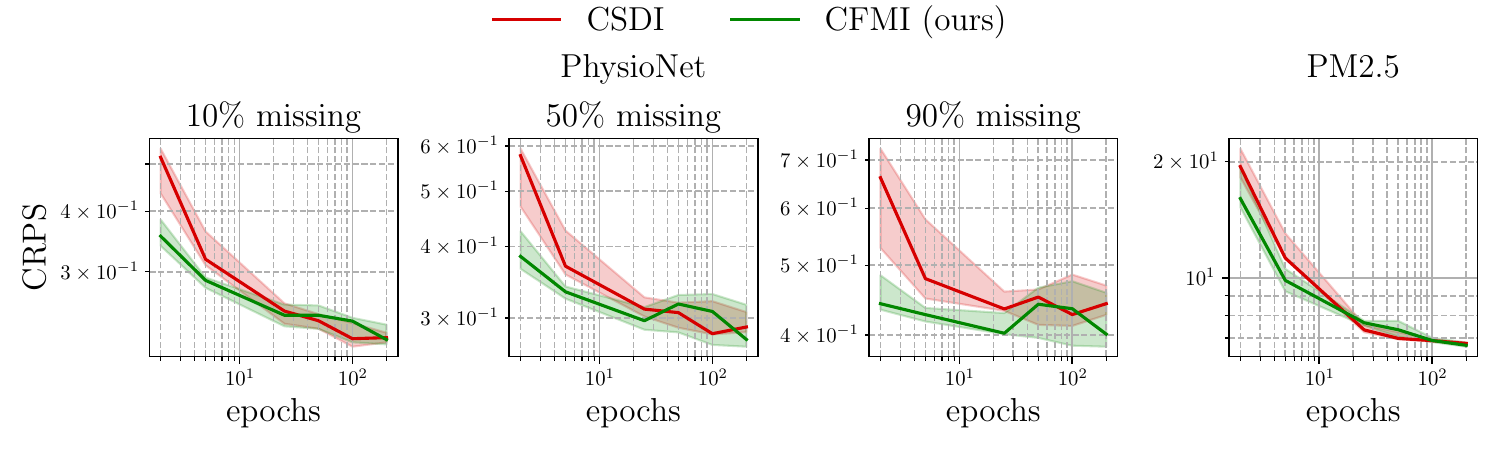}
  \caption{CRPS versus number of training epochs.}
\end{figure}

\begin{figure}[H]
  \centering
  \includegraphics[width=0.95\linewidth]{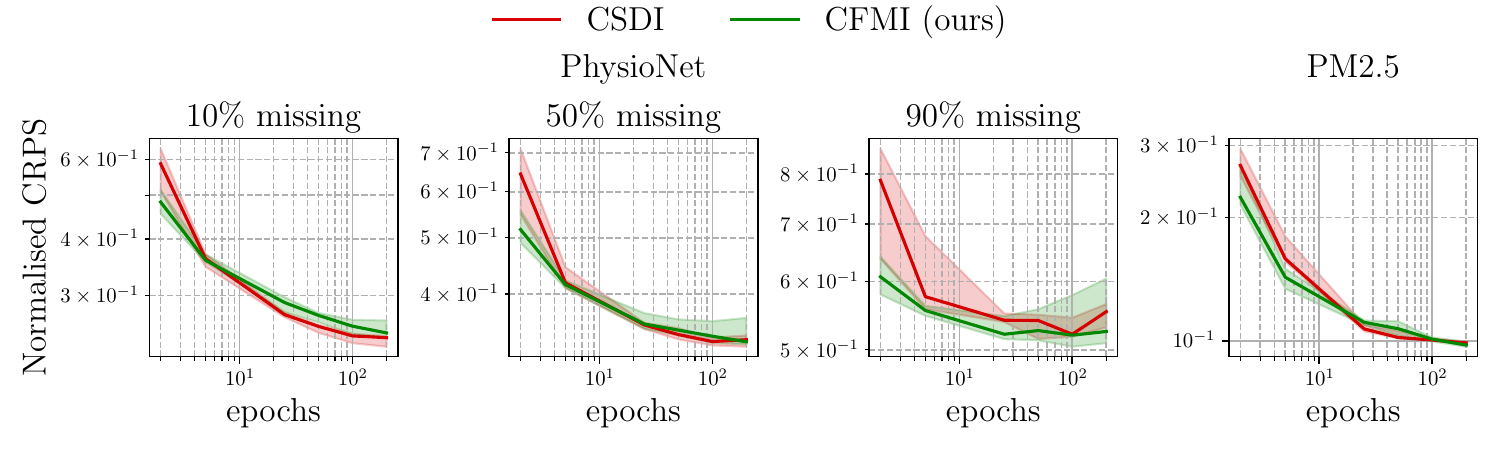}
  \caption{Magnitude-normalised CRPS versus number of training epochs.}
\end{figure}

\begin{figure}[H]
  \centering
  \includegraphics[width=0.95\linewidth]{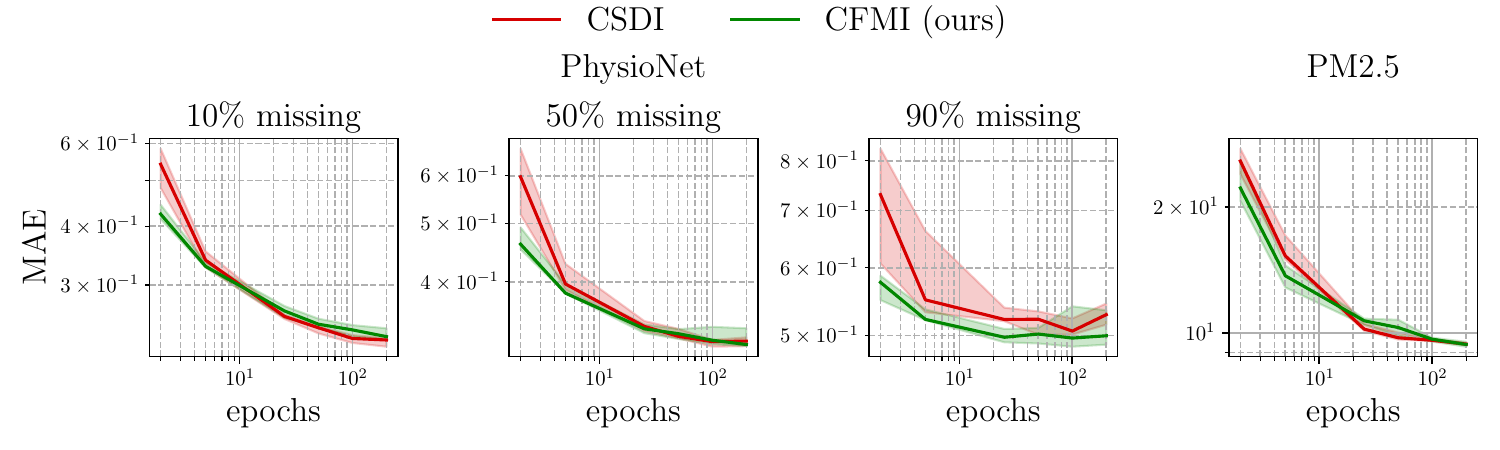}
  \caption{MAE versus number of training epochs.}
\end{figure}

\begin{figure}[H]
  \centering
  \includegraphics[width=0.95\linewidth]{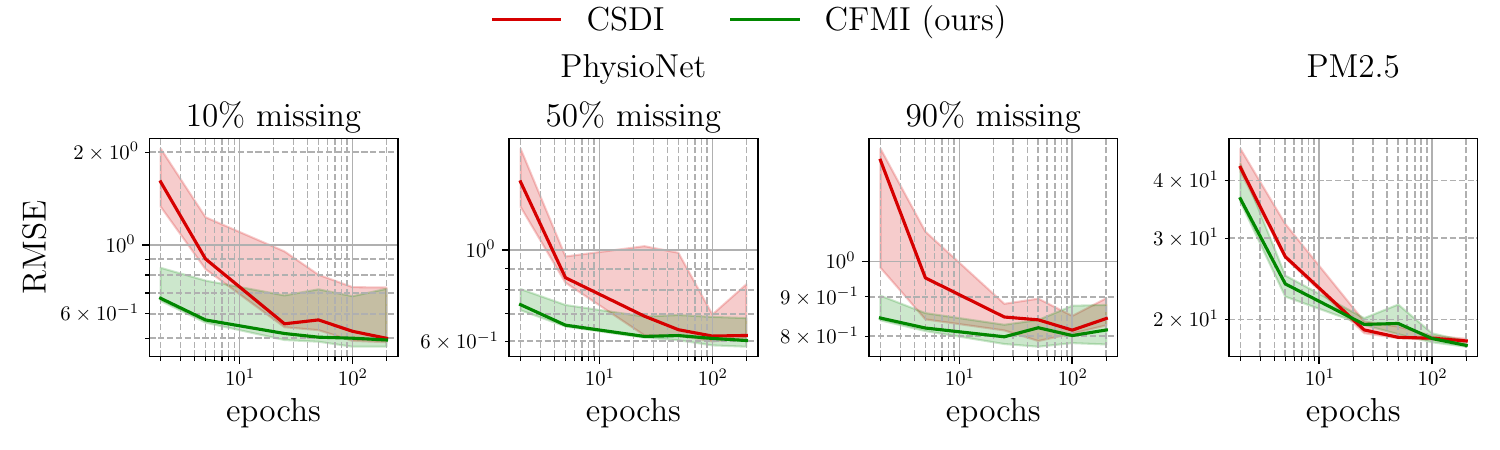}
  \caption{RMSE versus number of training epochs.}
\end{figure}

\end{document}